\def\year{2022}\relax
\documentclass[letterpaper]{article} % DO NOT CHANGE THIS
\usepackage{aaai22}  % DO NOT CHANGE THIS
\usepackage{times}  % DO NOT CHANGE THIS
\usepackage{helvet}  % DO NOT CHANGE THIS
\usepackage{courier}  % DO NOT CHANGE THIS
\usepackage[hyphens]{url}  % DO NOT CHANGE THIS
\usepackage{graphicx} % DO NOT CHANGE THIS
\usepackage{graphbox}

\usepackage{natbib}  % DO NOT CHANGE THIS AND DO NOT ADD ANY OPTIONS TO IT
\usepackage{caption} % DO NOT CHANGE THIS AND DO NOT ADD ANY OPTIONS TO IT
\DeclareCaptionStyle{ruled}{labelfont=normalfont,labelsep=colon,strut=off} % DO NOT CHANGE THIS
\usepackage[dvipsnames]{xcolor}
\usepackage{hyperref}
\usepackage{amsmath}
\usepackage{amsthm}
\usepackage{enumitem}
\usepackage{amsfonts}
\usepackage{amstext} % for \text macro
\usepackage{array}   % for \newcolumntype macro
\usepackage{subcaption}
\usepackage[capitalise]{cleveref}
\usepackage[disable]{todonotes} %Use this to disable the comments
\newtheorem{theorem}{Theorem}
\newtheorem{problem}{Problem} %[theorem]
\newtheorem{remark}{Remark} %[theorem]
\usepackage[pscoord]{eso-pic} % for adding acceptance info at top of page
\newcolumntype{C}{>{$}c<{$}} % math-mode version of "c" column type
\newcommand{\tup}[1]{\langle #1 \rangle}
\newcommand{\cC}{\mathcal{C}}

\usepackage[ruled, vlined, linesnumbered]{algorithm2e}  % algorithms
\usepackage{complexity}

\title{Conflict-Based Search for Explainable Multi-Agent Path Finding}
\author{
    Justin Kottinger\textsuperscript{\rm 1},
    Shaull Almagor\textsuperscript{\rm 2},
    Morteza Lahijanian\textsuperscript{\rm 1,3}
}
\affiliations{
    \textsuperscript{\rm 1}Department of Aerospace Engineering Sciences, University of Colorado Boulder, USA \\
    \textsuperscript{\rm 2}The Henry and Marilyn Taub Faculty of Computer Science, Technion, Israel\\
    \textsuperscript{\rm 3}Department of Computer Science, University of Colorado Boulder, USA \\
    \{justin.kottinger, morteza.lahijanian\}@colorado.edu, shaull@cs.technion.ac.il
}
\begin{document}
\maketitle
%\placetextbox{0.5}{0.95}{To appear in International Conference on Automated Planning and Scheduling (ICAPS 2022), June 2022.}%

\begin{abstract}
The goal of the Multi-Agent Path Finding (MAPF) problem is to find non-colliding paths for agents in an environment, such that each agent reaches its goal from its initial location. 
In safety-critical applications, a human supervisor may want to verify that the plan is indeed collision-free. To this end, a recent work introduces a notion of explainability for MAPF based on a visualization of the plan as a short sequence of images representing time segments, where in each time segment the trajectories of the agents are disjoint. 
Then, the problem of \emph{Explainable MAPF via Segmentation} asks for a set of non-colliding paths that admit a short-enough explanation. Explainable MAPF adds a new difficulty to MAPF, in that it is \NP-hard with respect to the size of the environment, and not just the number of agents. Thus, traditional MAPF algorithms are not equipped to directly handle Explainable MAPF. In this work, we adapt Conflict Based Search (CBS), a well-studied algorithm for MAPF, to handle Explainable MAPF. We show how to add explainability constraints on top of the standard CBS tree and its underlying $A^*$ search. We examine the usefulness of this approach and, in particular, the trade-off between planning time and explainability.

\end{abstract}
% \shtodo{General note: please replace all textit with emph}
% \shtodo{General note: please replace all EG with XG}
% \shtodo{Are you ok with section numbering? I find it annoying without, but if you prefer without that's fine.}
\section{Introduction}
\label{sec:intro}
Multi-Agent Path Finding (MAPF) is a fundamental problem in AI, in which the goal is to plan paths for several agents to reach their targets, such that paths can be taken simultaneously without the agents colliding. Applications of MAPF are ubiquitous in any area where several moving agents are involved, such as air-traffic control, UAVs, warehouse robots, autonomous cars, etc. 
While MAPF is generally intractable, the importance of this problem has generated a significant body of work over the past decade~\cite{stern2019mapf,standley2010finding,AFeln17c,AFeln16c,JSvan18b,SKoen18k,SKoen19a}, dealing with various aspects of the problem and suggesting increasingly scalable solutions. 
% In particular, a well-performing algorithm for MAPF is \emph{Conflict-Based Search (CBS)}~\cite{SHARON201540}. CBS is a decentralized algorithm, which plans for a single agent at each iteration, while placing constraints on the suggested plans, so as to avoid collisions (we elaborate on CBS in Section~\ref{subsec:CBS}).
In particular, a well-performing algorithm for MAPF is \emph{Conflict-Based Search (CBS)}~\cite{SHARON201540}, which is a decentralized approach, and has extensions with various heuristics 
% that further improve its performance~
\cite{boyarski2015icbs,li2019disjoint,li2019improved,felner2018adding}.
% \ml{add refs for variants of CBS}

% As with many problems and algorithms in AI, an emerging fundamental problem is that of \emph{explainability}: in safety-critical and heavily-regulated applications (e.g., air-traffic control, hazardous-materials warehouses), planning is not fully automatic. Indeed, automated planning has to be trusted in order to act upon, and in order to maintain legal and ethical accountability.
% In the context of MAPF, a plan is only suggested to a human supervisor, who may act upon it. 
% Then, the plan has to be presented to the supervisor in some humanly-understandable manner. In particular, the presentation should enable the supervisor to understand the paths taken by the agents, and to easily verify that the agents do not collide, as otherwise the supervisor would not necessarily trust the plan. 

% \jk{Added content here to (hopefully) stress the fact that the user may not understand (or trust) the algorithm as much as the designer.}
% \ml{it's in the next paragraph}

A major barrier in adopting such capable MAPF algorithms in \emph{safety-critical} applications, as with many algorithms in AI, is 
%that of 
\emph{trust} (or lack thereof) between the designers of such algorithms, and their potential user.  That is, in heavily-regulated applications (e.g., air-traffic control, hazardous-materials warehouses), automated planning has to be trusted before acting upon in order to maintain legal and ethical accountability. Designers can gain trust in their algorithms through studying, developing, and exhaustive testing. The same trust building methods may not be available for the user. To combat this dilemma in the context of MAPF, the current practice is to suggest a computed plan to a human supervisor, who has to verify the correctness of it to allow its execution~\cite{fines2020agent}. This poses an additional problem in MAPF -- \emph{explainability} of plans.  In other words, plans must be presented to the supervisor in some humanly-understandable manner.  In particular, the presentation (explanation) should enable the supervisor to understand the paths taken by the agents, and to easily verify that the agents do not collide, as otherwise the supervisor would not necessarily approve the plan. 
This study focuses on the problem of Explainable MAPF and aims to develop a scalable algorithm whose solutions to the MAPF problem are easily-interpretable and verifiable by humans. 

\begin{figure}[t]
     \centering
     \begin{subfigure}{0.24\linewidth}
         \centering
         \includegraphics[scale=0.13]{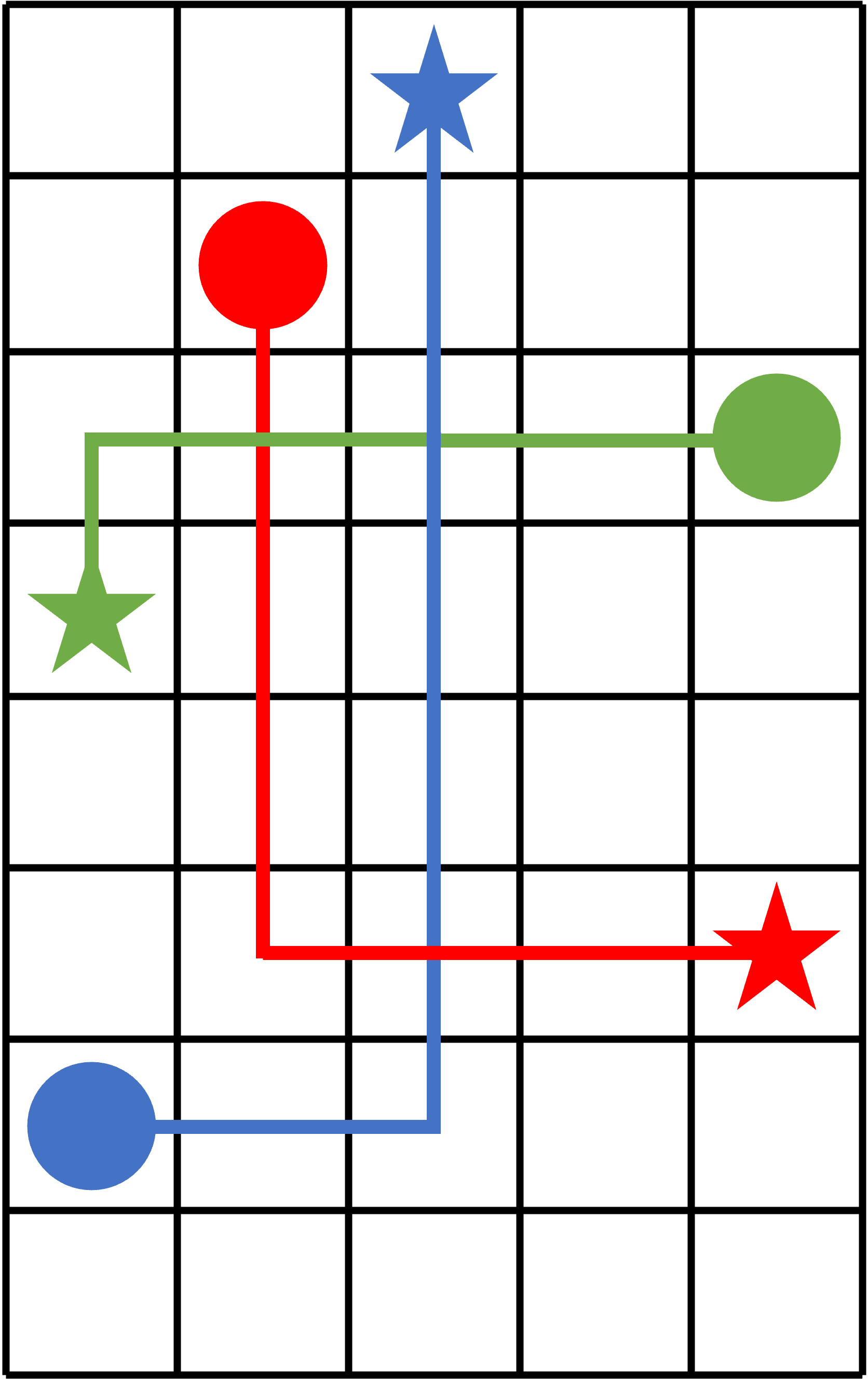}
         \caption{Full Plan}
        %  \label{fig:y equals x}
     \end{subfigure}
     \hfill
     \begin{subfigure}{0.24\linewidth}
         \centering
         \includegraphics[scale=0.13]{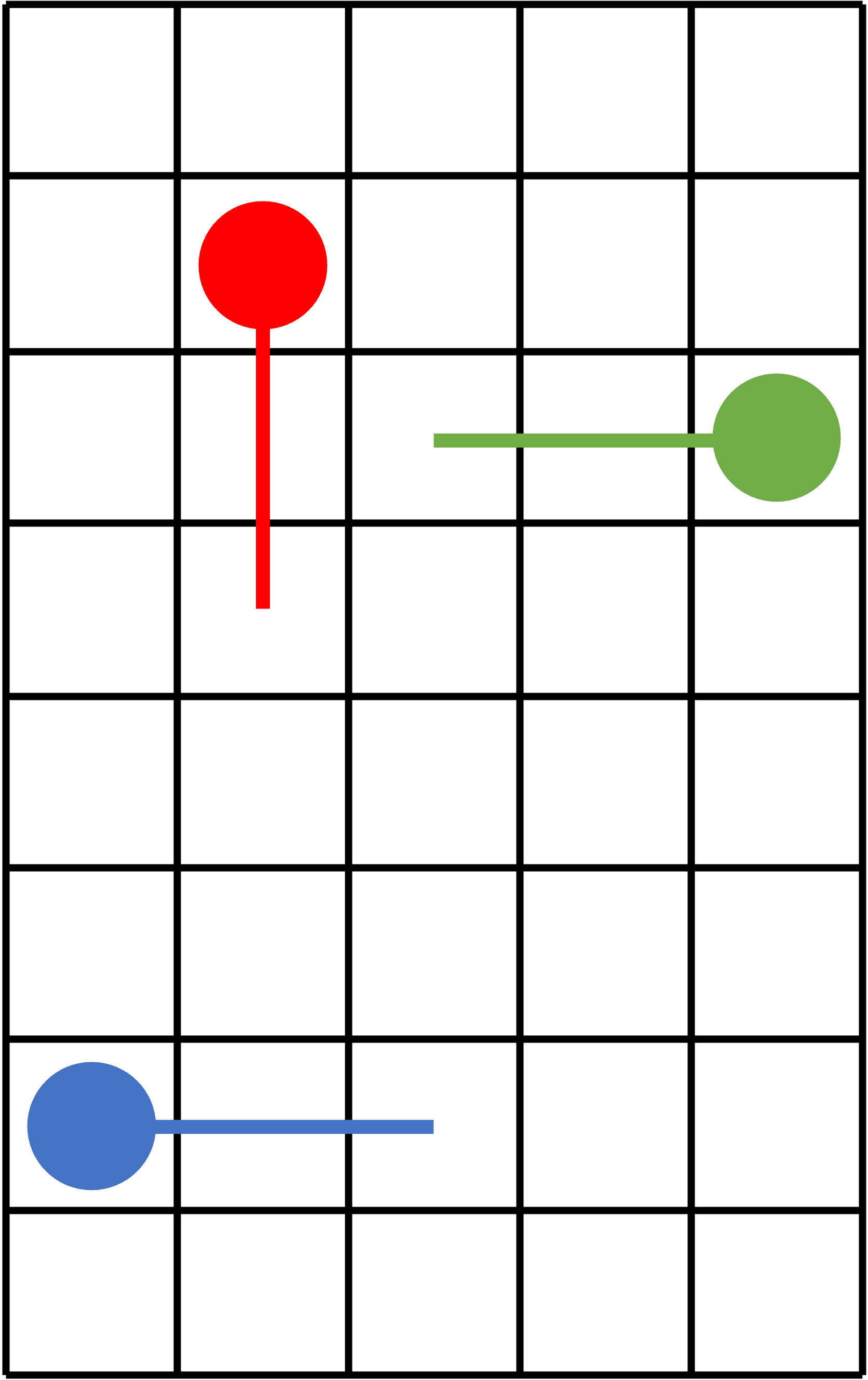}
         \caption{$k=[0,2]$}
        %  \label{fig:three sin x}
     \end{subfigure}
     \hfill
    %  \newline
     \begin{subfigure}{0.24\linewidth}
         \centering
         \includegraphics[scale=0.13]{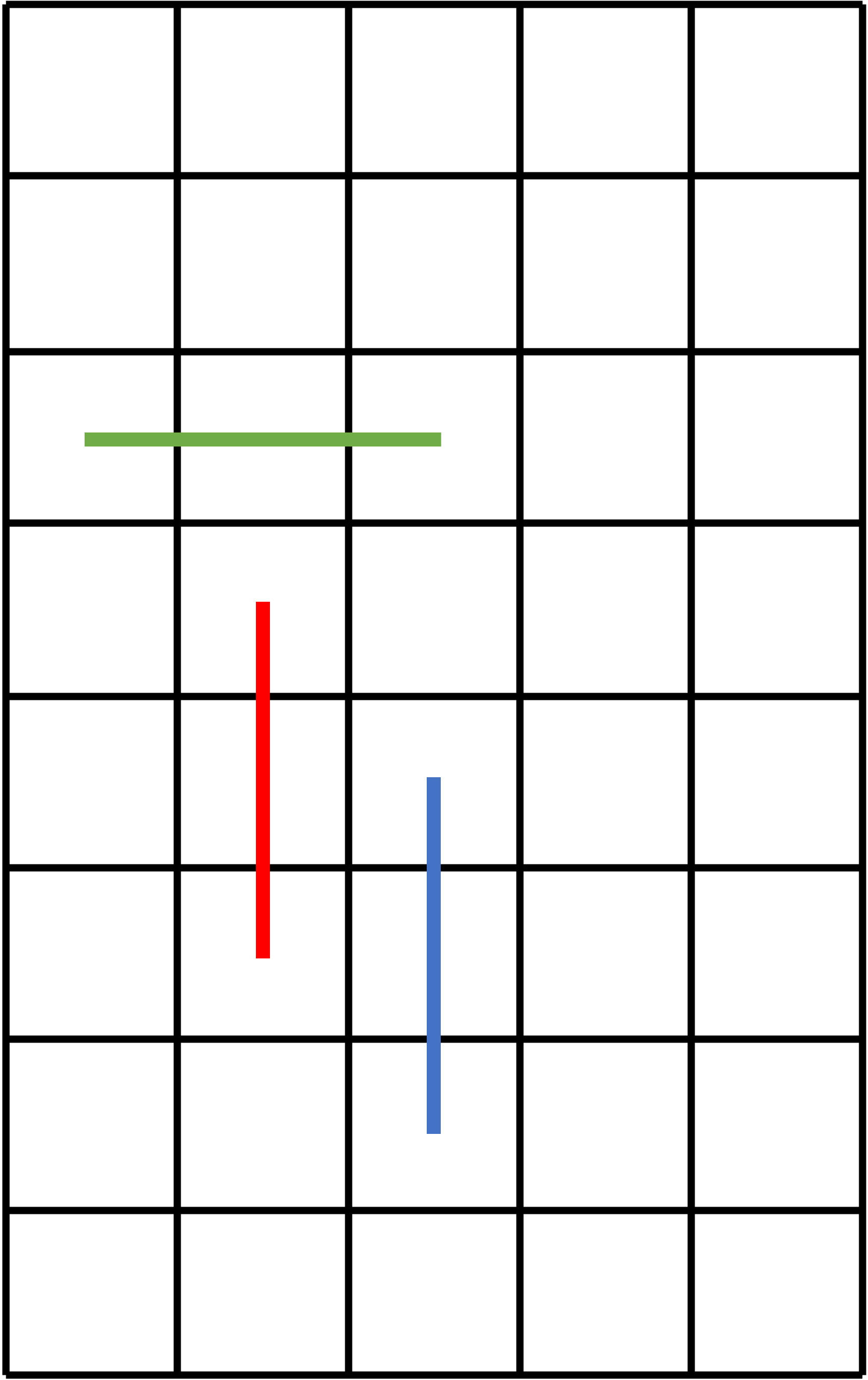}
         \caption{$k=[2,4]$}
        %  \label{fig:five over x}
     \end{subfigure}
     \hfill
     \begin{subfigure}{0.24\linewidth}
         \centering
         \includegraphics[scale=0.13]{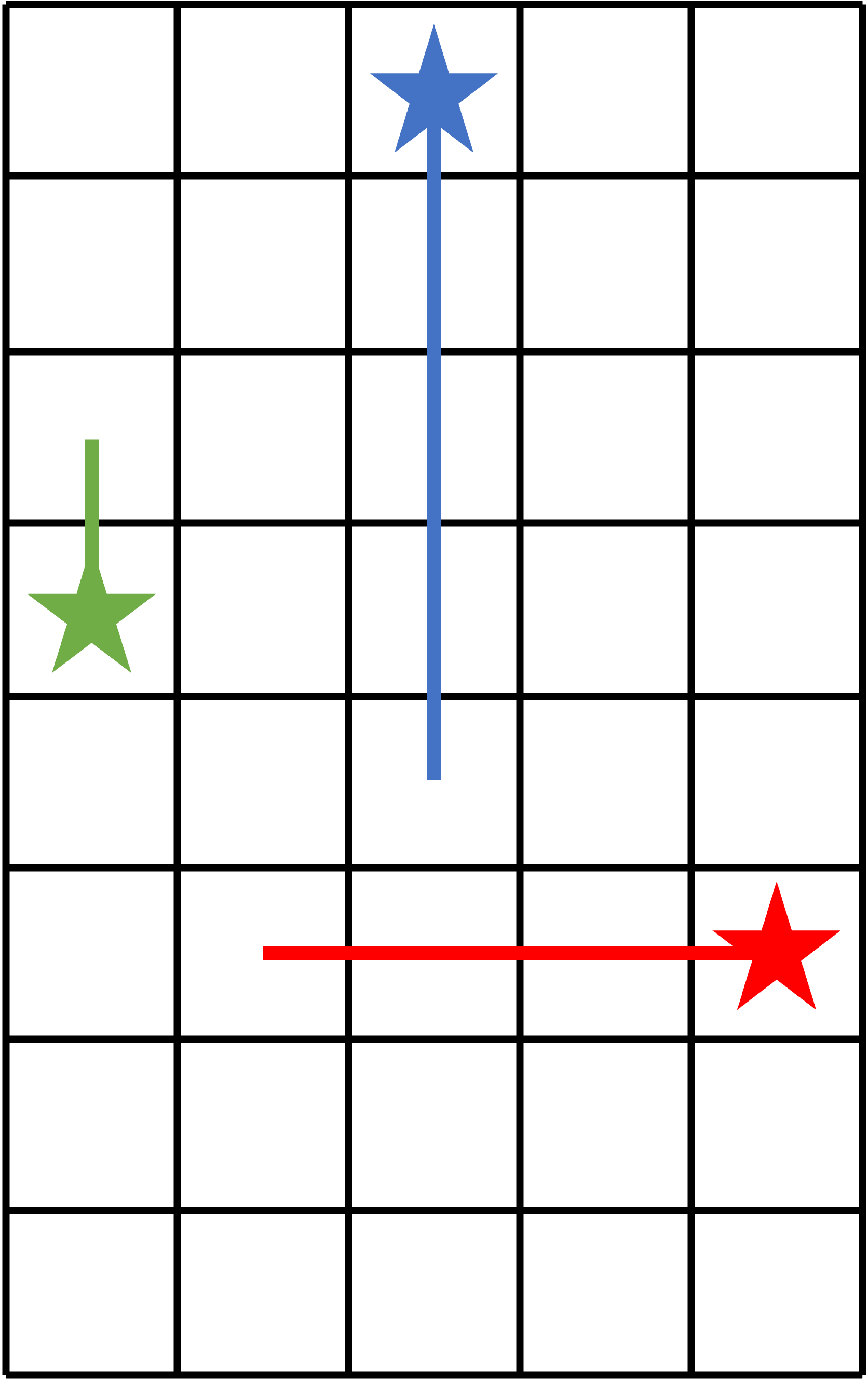}
         \caption{$k=[4,8]$}
        %  \label{fig:five over x}
     \end{subfigure}
    \caption{A plan for three agents (a), and a corresponding explanation via disjoint decomposition (b)-(d). The circles and stars mark the start and goal vertices, respectively.}
    \label{fig:intuition}
    \vspace{-4mm}
\end{figure}

% \ml{may need a paragraph on existing approaches to explanations and state why they're not suitable for MAPF.}
Explainable AI (XAI) is an active area of research in recent years.  Many studies focus on explaining decisions made by machine learning algorithms, in both categories of classification and regression \cite{arrieta2020explainable}.  In those works, various forms of explanations are explored, but visual explanations seem to be dominant for their ease of interpretability, especially for classifiers \cite{Lapuschkin_2019}.  In other cases, such as fault detection algorithms, explanations are typically in the form of witness executions \cite{mari2021explaining}.  In the planning community, explanations are mostly studied in the context of the single agent problem, and they often take a non-visual form.  For example, explanations are given based on alternative plans \cite{eifler_cashmore_hoffmann}, minimal differences between plans \cite{10.5555/3306127.3331663}, or reasoning on quantitative advantage of one plan over another \cite{fox2017explainable}.  None of these studies, 
however, focus on the MAPF problem.

Recent works \cite{Almagor:AAMAS:2020,kottinger2021mapsx} propose an explanation scheme for MAPF by means of visualization.  
There, the idea is to decompose a non-colliding plan into time segments, such that within each segment the paths of the agents are disjoint. 
Then, by depicting each segment separately (see Figure~\ref{fig:intuition}), it is easy for a human supervisor to verify that the agents do not collide. 
Indeed, recognizing line intersections takes place early in the visual cognitive process~\cite{Hubel,Tang}, making it easy to verify that the depicted lines in each segment are disjoint.
The usefulness of such explanations is also supported by the findings of the survey study \cite{brandao2021experts}.  
While decomposition can be readily used on any MAPF plan obtained by any algorithm, it is not guaranteed that doing so  results in a small number of segments. In case the number of segments is very high, this undermines the explanation scheme. 
Thus, the central problem in \textit{Explainable MAPF via Segmentation} is to find a plan for the agents that can be decomposed to a small number of segments (and hence can be explained with a small number of pictures). 

Unfortunately, Explainable MAPF is much harder than standard MAPF, in the sense that, unlike MAPF, it is \NP-hard already for two agents. In particular, the hardness of Explainable MAPF is with respect to the size of the environment (as well as the number of agents), rendering the runtime of algorithms for Explainable MAPF exponential in the size of the environment. In contrast, the complexity of classical MAPF is polynomial in size of the environment, making the problem much easier, especially with a low number of agents. 
Unsurprisingly, centralized algorithms for Explainable MAPF do not scale well, as shown in \cite{Almagor:AAMAS:2020}. 

In this work, we consider a decentralized approach to the Explainable MAPF problem.  Specifically, we adapt
CBS, a two-level algorithm that, in its low-level, plans individually for each agent, and in its high-level, identifies collisions between the agents and places constraints to resolve them in the next low-level iteration (see \cref{subsec:CBS}).
%, which in each iteration, computes paths for agents individually (using $A^*$) and then places constraints on the computed paths to avoid collisions in the next iteration.
Our \emph{main contribution} is accordingly split to two levels: at the high-level, 
%Our \emph{first contribution} is to 
we show how we can use similar constraints as those used by CBS to capture \emph{segmentation conflicts}, namely plans whose minimal decompositions have too many segments. We then discuss how to adapt CBS to compute and place these constraints during its search, thus obtaining our new algorithm, dubbed \emph{Explanation-Guided CBS (XG-CBS)}.
% We then turn our attention to the low-level planner of XG-CBS. As we discuss in Section~\ref{subsec:astar}, standard $A^*$ seems, intuitively, ill-fitted to work with XG-CBS. Indeed, minimizing the number of disjoint segments of a plan often requires lengthening the plan, which $A^*$ is reluctant to do. 

We then turn our attention to the low-level planner of XG-CBS. As we discuss in Section~\ref{subsec:astar}, standard $A^*$ seems, intuitively, ill-fitted to work with XG-CBS. Indeed, minimizing the number of disjoint segments of a plan often requires lengthening the plan, which $A^*$ is reluctant to do since it minimizes the path length.
% We then demonstrate that using standard $A^*$ as a low-level search algorithm does not perform well with XG-CBS.
% \jk{sadly, we have yet to fully prove this in experiments}
% \ml{well... it does not work well in absolute terms. Our proposed XG-A* improves it for small envs but suffers in the larger envs}
% \jk{Either way, I think that the statement contradicts our results as it is currently stated. What about the following: ``We then demonstrate that using standard $A^*$ as a low-level search algorithm is not well suited for transforming system behavior around explanations." I get that it might not be as strong as a statement. But it is more truthful w.r.t the results we see in the table.}
% To mitigate this, our \emph{second contribution} is a new search space and heuristic for $A^*$ that incorporates information on the segmentation of the plan, in order to direct XG-CBS toward plans with small decompositions.  We show that this approach significantly improves the performance of XG-CBS in small environments even with large number of agents, as well as in scenarios where solutions are scarce. In large environments though, where there are many solutions that do not require large deviations from shortest paths, this method is at disadvantage due to its exhaustive exploration for plans with low number of segments, and using classical $A^*$ as a low-level search results in smaller computation times.
Thus, at the low level, our contribution focuses on developing appropriate search algorithms, that are guided towards plans with small decompositions, 
%Our \emph{second contribution} is a new search space and heuristic for $A^*$ that guides the search towards plans with small decompositions, 
which are appropriate for XG-CBS.  
To this end, we propose three low-level search algorithms.
The first algorithm, dubbed XG-$A^*$ (\cref{subsec:XG-Astar}), guides the search toward a plan with minimum number of disjoint segments, while maintaining the completeness of XG-CBS. Moreover, it can 
%and exhaustive 
%and can be even 
be combined with standard $A^*$ to improve performance through a meta-parameter, resulting in the algorithm WXG-$A^*$ (\cref{subsec:weighted}).  

We discuss how the optimal value for this parameter is highly dependant on the instance of the problem, and hence, difficult to choose \textit{a priori}. 
Also, both XG-$A^*$ and WXG-$A^*$, due to their completeness, are subject to the inherent difficulty of Explainable MAPF with respect to the environment size. This is manifested by the need to track the history of paths within the search space.
To address these problems, 
%our \emph{third contribution} is a parameter-free 
we propose another low-level algorithm, SR-$A^*$ (\cref{subsec:sr}) that uses the segmentation information in a coarse way such that it does not need to track history, yet obtains solutions with small number of decomposition. Theoretically, SR-$A^*$ sacrifices completeness, 
but our experimental results (\cref{subsec:performance}) show that with SR-$A^*$, XG-CBS has comparable computation time to \emph{vanilla} CBS (and even outperforms it), while obtaining plans with much smaller decompositions.  This is despite solving a much harder problem.
Thus, our overall contribution is 
a decentralized algorithm for the Explainable MAPF problem.  To the best of our knowledge, this is the first
%\shtodo{Despite the double blind, anyone reading this already guessed it's by the same authors, so I'm not sure if priding ourselves on being the first is in place.} 
algorithm of its class that scales,
%can scale up to 10 \ml{12?} agents, 
significantly outperforming previous algorithms.
%the state of the art.  
We show properties of our algorithm and further evaluate it on many benchmarks, comprising examples that demonstrate specific intricacies of Explainable MAPF, as well as standard MAPF benchmarks.  
Overall, this work illustrates the unique computational challenges faced in Explainable MAPF and paves the way for further algorithmic exploration of this problem.
%Furthermore, the proposed techniques to develop XG-CBS from CBS are general and can be employed to extend other MAPF algorithms to Explainable MAPF.
%\ml{need a better transition to the last sentence}
%\shtodo{Why is the last sentence true? I have no idea how to transfer e.g., SAT-based algorithms, or even PBS (priority based search) to the explainable setting. It's certainly not by plugging the same ideas.}

\section{Problem Statement}
\label{sec:problem}
%\shtodo{Made many rephrases and changes in this section. Please read and comment.}\jk{This section reads well. I just corrected a couple very minor typos.}
Consider $n\in \mathbb{N}$ agents, acting in a directed graph $G=\tup{V, E}$ where each agent $i\in \{1,\ldots,n\}$ has a source $s_i \in V$ and a goal $g_i \in V$. A \emph{path} in $G$ is a sequence of vertices $\pi=v_1v_2\ldots v_m$ 
such that $(v_k, v_{k+1})\in E$ for all $1 \leq k < m$. 

Given paths $\pi_1=v_1v_2 \ldots v_{m}$ and $\pi_2=u_1 u_2 \ldots u_{m}$ in $G$ for some $m > 1$, we say that $\pi_1$ and $\pi_2$ are \emph{non-colliding} if the following conditions are satisfied 
for all $1\le k< m$:
\begin{enumerate} [label={(\roman*)}, leftmargin=0.5in]
    \item $v_k \neq u_k$ (i.e., no vertex collisions),
    \label{condition:collision} 
    \item $(v_k, v_{k+1}) \neq (u_{k+1}, u_k)$ (i.e., no edge collisions).
    \label{condition:edgeSwap}
\end{enumerate}
We extend the definition to paths of different lengths by truncating the longer path. Intuitively, this means that once a path ends, the respective agent ``disappears''\footnote{Changing this to have the agents remain at the target location does not impact our results in any significant way.}.

%Intuitively, since we only consider indices up to $\min{(m_1, m_2)}$, once the goal is reached, the agent ``disappears".
% Requiring these constraints to hold for up to $\min{(m_1, m_2)}$ steps removes the need to check against paths that have already been completed \ml{this sentence is unclear!}. 
%The conditions described above are extended to an arbitrary set of paths by requiring that they are pairwise non-colliding.

Given $n$ agents on a graph $G$ and two lists $s_1, \ldots, s_n$ 
% \ml{let's use `,' for lists, not paths}
and $g_1, \ldots, g_n$ of source and goal vertices, respectively, a \emph{plan} $P=\{\pi_1, \ldots, \pi_n\}$ is a set of non-colliding paths (i.e., $\pi_i$ and $\pi_j$ are non-colliding for all 
% $1\le i,j\le n$) 
$i,j \in \{1,\ldots,n\}$ and $i \neq j$) 
such that $\pi_i$ drives agent $i$ from $s_i$ to $g_i$ for every $i\in \{1, \ldots, n\}$. The \emph{length} of the plan is the maximal length of a path in $P$.
The classical \emph{Multi-Agent Path Finding (MAPF)} problem is to find a plan\footnote{Typically, the plan is required to be optimal with respect to some cost function, e.g., makespan or sum-of-costs.} $P$ on $G$.

%In this paper, we consider a variation of this problem, dubbed \emph{explainable MAPF}.
% Thus, the classical \emph{Multi-Agent Path Finding (MAPF)} problem immediately follows.
% \ml{no need to define MAPF formally. Just say something like ... classical MAPF seeks to find a $P$ on $G$.}
% \begin{problem}{MAPF:}
% Given a graph $G=\langle V, E\rangle$, and lists $s_1, \ldots, s_n$, and $g_1, \ldots, g_n$ of source and goal vertices, where $s_i, g_i\in V$ for $i\in \{1, \ldots, n\}$, find a plan $P$ for all the agents, or answer that no such plan exists. 
% \end{problem}

% \shtodo{I think we should start with an intuitive explanation of disjoint segmentation, and only then give the formal details (possibly shortened).}
We now turn to recap the definitions of Explainable MAPF via Segmentation from \cite{Almagor:AAMAS:2020}. Consider a path $\pi=v_1\ldots v_m$ and $t_1\le t_2$. We define $\pi[t_1,t_2]=v_{t_1}\ldots v_{t_2}$ to be the segment of $\pi$ between $t_1$ and $t_2$. If either $t_1$ or $t_2$ are not within the range $\{1,\ldots, m\}$, we simply disregard the out-of-bounds vertices.
%Further, given two time points $1 \leq t_a \leq t_b \leq m$, we define $\pi[t_a, t_b]$ 
%to be a segment of $\pi$ between vertices $v_{t_a}$ and $v_{t_b}$. \shtodo{note to self: move segments to later}
% \ml{can we say $1\leq t_1 \leq t_2 \leq m$ and remove the last sentence?}

A set of paths (better thought of as path segments) $\{\tau_1, \ldots, \tau_n\}$ where each $\tau_i=v_{i1} \ldots v_{ik_i}$ is \emph{vertex disjoint} if for all $i\neq j$ we have $\{v_{i1}, \ldots, v_{ik_i}\} \cap \{v_{j1}, \ldots, v_{jk_j}\} = \emptyset$.
Next, consider a plan $P=\{\pi_1,\ldots, \pi_n\}$ as above and let $K=\max_{i}{m_i}$ be its length, where $m_i$ is the length of $\pi_i \in P$. A \emph{vertex-disjoint decomposition} of $P$ is an ordered list of natural numbers $1=t_0 < t_1 < \ldots < t_r=K+1$ 
such that for every $1 \leq k\leq r$, the path segments $\{\pi_j[t_{k-1}, t_k-1]\}_{j=1}^{n}$ are vertex-disjoint. We refer to $r$ as the \emph{index} of the decomposition. The minimal index of a  vertex-disjoint decomposition of $P$ is referred to as the \emph{index} of $P$. As shown in~\cite{Almagor:AAMAS:2020} (and we recap in Section~\ref{subsec:XGCBS}), computing a minimal-index decomposition can be done in polynomial time using a greedy algorithm, hence, we only consider minimal index decompositions here.
We now present the formal definition of the Explainable MAPF via Segmentation problem.

\begin{problem}[Explainable MAPF via Segmentation]
\label{problem:expMAPF}
Given a graph $G=\langle V,E \rangle$ with lists $s_1, \ldots, s_n$ and $g_1, \ldots, g_n$ of source and goal vertices, respectively, and bound $r \in\mathbb{N}$,
find a plan $P$ for the agents with index of at most $r$ or answer that the instance is \emph{unsolvable} -- no such plan exists.
\end{problem}
\citet{Almagor:AAMAS:2020} proved that (the decision version of) Problem~\ref{problem:expMAPF} is \NP-complete, even for 2 agents (unlike MAPF, which is in $\P$ for a fixed number of agents). They propose a centralized algorithm for the problem, but demonstrate that it does not scale. To this end, the goal of this paper is to develop a \emph{decentralized} algorithm that is capable of solving Problem~\ref{problem:expMAPF} and scaling to a large number of agents.
% \ml{can we say that, in small envs, our XG-CBS is scalable wrt the number agents with XG-A* scalable wrt the size of the env with classical A*?}
% \shtodo{I don't think so. First, there is no formal justification of scalability, it's not like CBS performs well on all cases. \\
% Second, even if you fix the size of the environment, the tradeoff between XG-$A*$ and classical $A^*$, in the context of XG-CBS, is how much time is spent in the low-level v.s. how large the constraint tree becomes. But we don't know whether this tradeoff always leans towards one or the other.}
% \shtodo{Move these to Sec 5 or the discussion.}
\section{Explanation-Guided CBS} % "Explanation-Guided CBS for Explainable MAPF" is too long
\label{sec:XGCBS}
% remind reader what CBS is

% Our solution to Problem~\ref{problem:expMAPF} extends from a decentralized MAPF algorithm known as \emph{Conflict-Based Search (CBS)}~\cite{SHARON201540}. Here, we first review the CBS algorithm and then present our extensions to it to obtain Explanation-Guided CBS (XG-CBS).
Our solution to Problem~\ref{problem:expMAPF} extends from CBS~\cite{SHARON201540}, a decentralized MAPF algorithm. 
Here, we first review this algorithm and then present our extensions to it to obtain Explanation-Guided CBS (XG-CBS). 
\subsection{CBS for MAPF}
\label{subsec:CBS}
%\jk{get to two paragraphs}
CBS is a two-level search on the space of possible plans, consisting of a high-level conflict-tree search and a low-level graph search. 
At the high-level, CBS keeps track of a \emph{constraint-tree}, in which each node represents a suggested plan, which might have collisions, referred to as \emph{conflicts}. Initially, a root node is obtained by using a low-level graph search algorithm, typically $A^*$ with a shortest-path heuristic, to find a path for each agent from start to goal, ignoring the other agents (hence the decentralized nature of the method).

At each iteration, CBS picks an unexplored node from the tree, based on some heuristic. Then, the conflicts (namely collisions) in the plan corresponding to that node are identified. CBS attempts to resolve the conflicts by creating child nodes based on the conflicts, as follows: if Agents $i$ and $j$ collide at time $t$ in vertex $v$, then two children are created for the node, one with the constraint that Agent $i$ cannot be in vertex $v$ at time $t$, and the other dually for Agent $j$. Then, in each child node, a low-level search is used to replan a path for the newly-constrained agent, given the set of constraints obtained thus far along the branch of the constraint tree.
% %The idea is to plan for each agent individually using low-level graph search, while the high-level constraint tree tracks conflicts (collisions) between the agents and defines constraints to resolve them. 
This process repeats until either a non-colliding plan is found, or no new nodes are created in the constraint tree, 
at which point CBS returns that there is no solution.

\begin{figure}
    \centering
    \includegraphics[scale=0.13]{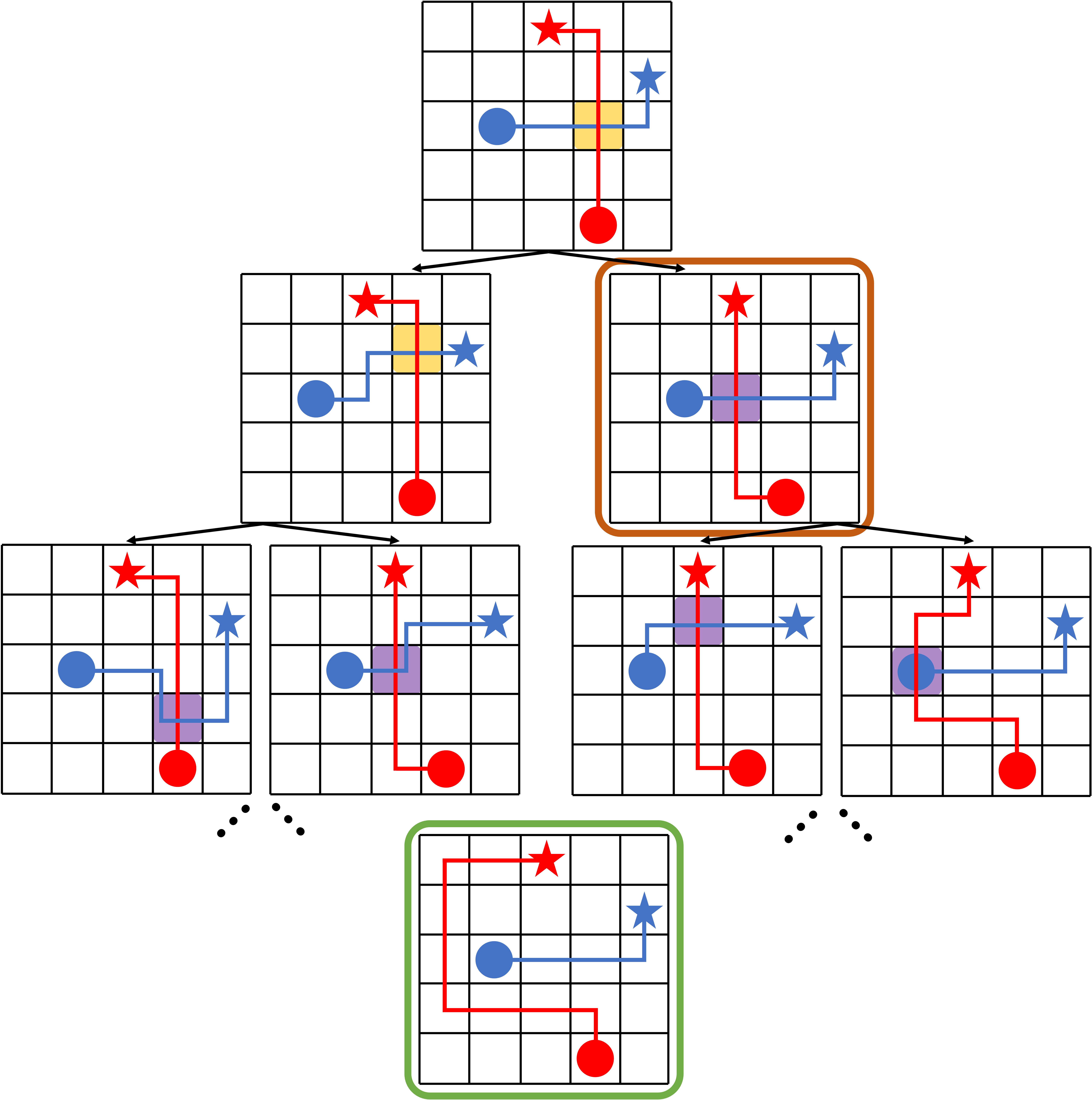}
    \caption{Illustration of XG-CBS with $A^*$ as the low-level planner. Yellow and purple colors indicate collision and segmentation conflicts, respectively.}
    \label{fig:expConflictRes}
\end{figure}

We (partially) demonstrate CBS in Figure~\ref{fig:expConflictRes}.  In this example, the root node has a colliding plan
% for the agents, 
and hence a constraint is placed on the yellow vertex at time 2, so two children are created with new plans for each of the two colliding agents.
% The priority queue places solutions with \emph{lower} cost at \emph{higher} priority. Next, the high-level algorithm evaluates the highest-priority node for new conflicts. This process repeats until either a non-colliding plan is found, or the priority queue becomes empty, at which point CBS returns no solution. 
% \ml{just say that the nodes of the constraint tree are searched via a priority queue that is based on the solution cost at each node.  Often, the solution cost is chosen to be the sum of the path lengths of the agents.}

CBS performs well for standard MAPF queries. However, it is ill-suited for solving Problem~\ref{problem:expMAPF} due to its lack of regard for vertex-disjoint decomposition of the proposed solutions. More precisely, CBS is guided toward short plans, whereas minimizing the index typically incurs a tradeoff with plan length. 
Below, we build upon CBS to plan for explainability, in order to address Problem~\ref{problem:expMAPF}. 
%%%% end sub-section %%%%

\subsection{CBS for Explainable MAPF}
\label{subsec:XG-CBS}
% \shtodo{why aren't we using CBS extensions}
% \jk{addressed Shaull's comment here}
We modify CBS both at the high-level (constraint tree) and low-level (graph search), to obtain a new algorithm dubbed \emph{Explanation-Guided CBS (XG-CBS)}. 
% We design XG-CBS algorithm to solve Problem~\ref{problem:expMAPF} by modifying CBS both at the high-level (constraint tree) and low-level (graph search). 
To this end, we first introduce \emph{segmentation conflicts} to the constraint tree. These are conflicts that occur when the plan is non-colliding, but whose index is greater than the bound $r$. These conflicts are resolved by placing appropriate constraints, as we detail below. 
% We handle the low-level search in Section~\ref{subsec:astar}.
We elaborate on the low-level search in Section~\ref{subsec:astar}.

We remark that we focus on the classical CBS algorithm, as opposed to improvements thereof (e.g., ICBS~\cite{boyarski2015icbs}), as our goal is to study the efficacy of the well-understood  constraint-tree method to explanations. 
%For the low-level search, we observe that a shortest-path guided $A^*$ does not work well for resolving segmentation conflicts, especially in congested environment.
%We propose modifications to the heuristic for $A^*$ to be more oriented toward these constraints in Section~\ref{subsec:astar}.

\label{subsec:XGCBS}

\subsubsection{Segmentation Conflicts}
% \jk{Clarify the segmentation conflicts produce constraints for \emph{all} agents not to cross paths }
Recall that in CBS, whenever a plan has collisions, constraints are placed on the colliding agents to force one of them away from the collision point. We keep these constraints in XG-CBS, and introduce additional constraints to handle segmentation. In order to define the new constraints, we recall how vertex-disjoint decompositions are computed. 

Consider a plan $P=\{\pi_1,\ldots, \pi_n\}$. In \cite{Almagor:AAMAS:2020}, it is shown that a minimal decomposition of $P$ can be found greedily by lengthening the current interval as long as the paths are disjoint, and starting a new segment once an intersection occurs. More precisely, we set $t_0=t_1=0$ and check $\{\pi_1[t_0,t_1],\ldots,\pi_n[t_0,t_1]\}$ for disjointedness. If it is disjoint, then $t_1$ is incremented by one. The process continues until the segment is not disjoint, at which point we add $t_1-1$ as a segmentation point, set $t_0=t_1$, and start the process again. This continues until the entire plan is segmented.

We use this greedy characterization to define segmentation constraints as follows. For a non-colliding plan $P=\{\pi_1,\ldots, \pi_n\}$ of length $K$, let $1=t_0 < t_1 < \ldots < t_r=K+1$ be a vertex-disjoint decomposition found as above. It follows that for every $1\le \ell\le r$, we cannot extend the disjoint segment $[t_{\ell-1},t_\ell-1]$ to time $t_\ell$. That is, there exist agents $i\neq j$ with $\pi_i[t_{\ell},t_{\ell}]\in \pi_j[t_{\ell-1},t_{\ell}]$, where $\pi_i[t_{\ell},t_{\ell}]$ is a single vertex. With each such pair of agents $i,j$, we associate the vertex $v=\pi_i[t_\ell,t_\ell]$ and the times $T_i=t_\ell$ and $T_j$ to be a time such that $\pi_j[T_j,T_j]=v$. Intuitively, $T_i$ and $T_j$ are the times when Agents $i$ and $j$, respectively, visit $v$ in the segment $[t_{\ell-1},t_\ell]$. 
Then, for a node with plan $P$ in the constraint tree of XG-CBS, we add two children with the following constraints: one child prevents Agent $i$ from visiting $v$ at time $T_i$, and the other prevents Agent $j$ from visiting $v$ at time $T_j$. 
% \jk{I clarified that segmentation conflicts occur for every agent that intersects in this paragraph.}
% \ml{modified the text below}
Note that, for a node with multiple segmentation conflicts, several such pairs of child nodes are added, one pair per conflict.
% Note that, in general, a node with a segmentation conflict may have several such pairs of child nodes because there may be multiple segmentation conflicts for a single plan $P$.

In Figure~\ref{fig:expConflictRes}, we depict segmentation conflicts as purple squares. For example, in the orange node of the tree, the plan requires two segments, due to the path intersection in the purple node, visited by the blue agent at time 1 and by the red agent at time 3. The two children of this node prevent each of these visits, and replan for the corresponding agent.

% \paragraph{XG-CBS}
\subsubsection{XG-CBS}
We are now ready to describe the operation of XG-CBS, with the caveat that we do not explicitly state the implementation of the low-level graph search algorithm. We leave this detail to Section~\ref{subsec:astar} and only assume that the low-level search algorithm is sound and complete, e.g., $A^*$.

XG-CBS algorithm proceeds as follows. Initially, the low-level algorithm is called for each agent separately to 
%XG-CBS begins by calling XG-$A^*$ for every individual agent to 
obtain an initial plan. If the initial solution does not have any conflicts (collision nor segmentation), the plan and its decomposition are returned as the solution. If conflicts exist, they are resolved by extending the tree according to the constraints as above. Once a new node is created with a constraint on Agent $i$, the low-level algorithm is called to replan for Agent $i$. Each new node is assigned a cost (as we discuss below) and added to a priority queue. At the next iteration, the minimum cost plan is popped from the queue, and gets evaluated for conflicts. This process repeats until either a satisfactory plan is found or the search is exhausted. 
We refer the reader to the Appendix (Section~\ref{sec:xg-cbs-alg}) for the pseudocode presentation of the XG-CBS algorithm.

An important remark is that during the low-level planning, an upper bound is set on the length of the path. The bound originates from the proof of membership in $\NP$ of
% the Explainable-MAPF problem
Problem~\ref{problem:expMAPF}, and serves to bound the constraint tree.

A crucial aspect of XG-CBS is the cost function on the tree nodes.
Recall that a common cost function for the high-level CBS is the combined length of all the paths (a.k.a. sum-of-costs). %Just as the case of the low-level planner, path length metrics alone could conflict with optimizing for explanations. 
This approach, however, tends to conflict with optimizing for explainability.
Thus, XG-CBS utilizes the index of the plan to define a cost function. Specifically, 
the primary cost of a proposed plan is the index of the plan, with a small tweak.  Recall that plans in the constraint tree may contain collisions, in which case the index is undefined. We circumvent this by viewing collisions as an end of a segment. Then, the combined length of paths is only used as a tie-breaker. 
This cost function enables XG-CBS to prioritize plans with a lower number of segments. 

Figure~\ref{fig:expConflictRes} demonstrates a run of XG-CBS where the low-level planner is standard $A^*$, and the index bound $r = 1$.
We conclude this section by showing that XG-CBS is complete. %(regardless of the low-level algorithm).

\begin{theorem}[XG-CBS Completeness]
\label{thm:XGCBS_complete}
XG-CBS always terminates, and given a solvable instance of Explainable MAPF via Segmentation, XG-CBS will terminate with a valid solution.
\end{theorem}

We refer the reader to the Appendix (Section~\ref{sec:proof}) for the proof of Theorem~\ref{thm:XGCBS_complete}.

\section{Low-Level Search}
\label{subsec:astar}

The low-level search has a twofold impact on the behavior of CBS.  First, it determines the concrete paths obtained after placing constraints. Second, since it is run for every node, it has a significant impact on the runtime.
In this section, we study four low-level search algorithms for XG-CBS. We start with an overview of our approaches.

In classical CBS, the goal is to find the shortest plan, making $A^*$ (with Hamming distance heuristic) a reasonable choice. For XG-CBS, however, 
%Theorem~\ref{thm:XGCBS_complete} shows that XG-CBS remains complete with standard $A^*$. However, in cases where low-index plans are scarce, 
the typical behavior of $A^*$ is ill-fitting. Intuitively, this is because $A^*$ tends to make very local changes in plans. Then, a segmentation conflict, which occurs on an intersection of paths, is likely resolved in a way that still intersects the same path in a nearby location or time. To illustrate this, consider the orange node in Figure~\ref{fig:expConflictRes}, and observe that the segmentation conflict for the red agent is resolved by going through the blue agent's origin, creating another segmentation conflict. 
Hence, many segmentation conflicts are typically required to be able to reduce the index of the plan. Despite this, $A^*$ is very fast, and thus allows a rapid exploration of the constraint tree. Thus, $A^*$ can be seen as one extreme, where speed is preferred over explanation-oriented paths.

At the other extreme, in order to orient XG-CBS toward a minimal-index plan, we propose a  low-level search called \emph{Explanation-Guided $A^*$ (XG-$A^*$)} that uses $A^*$ with 
%modify the low-level search. More precisely, the low-level search remains $A^*$, but with 
a novel, segmentation-based heuristic. Intuitively, XG-$A^*$ guides the search by minimizing the number of segments, as opposed to minizing the length.
As we discuss in~\cref{subsec:XG-Astar}, XG-$A^*$ is highly guided towards minimal explanations but is slow 
due to keeping track of the path \emph{history}. 
% The latter is due to keeping track of the \emph{history} of the path, and essentially performing segmentation at every step.
Our next approach is to get the best of both worlds, by combining XG-$A^*$ and $A^*$ in a weighted manner. We elaborate on this in~\cref{subsec:weighted}.  

The three approaches above maintain the completeness of XG-CBS. Our final low-level planner, discussed in~\cref{subsec:sr}, sacrifices completeness in favor of circumventing the need to keep track of the path's history in XG-$A^*$, thus obtaining a fast, explanation-oriented search (Section~\ref{subsec:sr}).

\subsection{XG-\texorpdfstring{$A^*$}{A*} -- Explanation Guided $A^*$}
\label{subsec:XG-Astar}
Recall that in CBS, the low-level search $A^*$ %does not utilize 
ignores 
the existing explanation of other agents when replanning for a certain agent. Thus, standard $A^*$ takes as input the graph $G=\tup{V,E}$, start and goal vertices $s,g\in V$ for an agent, and the set of constraints $\cC$ in the current node. 
% \ml{How does XG-A* account for existing explanations? Need to give the intuition first and then get to the algorithm}
In contrast, XG-$A^*$ accounts for existing segments, and hence, also receives as input the set of paths of the other agents, denoted by  $P_{-1}$, and a bound $B$ on the maximal allowed path length for the agent. We remark that the bound $B$ is only used to terminate the search if the plan becomes too long. This assures progress so that completeness is retained
(c.f., Theorem~\ref{thm:XGCBS_complete}). 

% In contrast, XG-$A^*$ also receives as input the set of paths of the other agents, denoted by  $P_{-1}$, and a bound $B$ on the maximal allowed path length for the agent. We remark that the bound $B$ is only used to terminate the search if the plan becomes too long. This assures
%maintains 
% progress so that completeness is retained
%and is used to retain completeness 
% (c.f., Theorem~\ref{thm:XGCBS_complete}). 

For brevity, in the following, we assume XG-$A^*$ plans for Agent 1, and the paths for the other agents are $P_{-1}=\{\pi_2,\ldots, \pi_n\}$.  Intuitively, XG-$A^*$ searches for a path for Agent 1 from $s$ to $g$ (that does not violate the constraints in $\cC$), while maintaining that the index of the decomposition of $P_{-1}$ combined with the planned path so far remains minimal. We demonstrate this before giving the precise details.

Consider the root node of Figure~\ref{fig:expConflictRes} 
with the colliding paths of the two agents.
% . Given the agents' paths in this node, first a collision conflict is identified, 
Once the collision constraint is identified, two children are generated with the respective constraints. Now consider XG-$A^*$ planning for the red agent given the path of the blue agent. XG-$A^*$ initially attempts to keep the index at 1, i.e., to keep the paths of the agents disjoint. To this end, XG-$A^*$  arrives at the plan in the green node (bottom of Figure~\ref{fig:expConflictRes}) before even suggesting the plan in the orange node, which the standard $A^*$ does. Indeed, the orange node has index 2, and therefore is not explored until all index 1 plans are exhausted.
This example demonstrates how XG-$A^*$ directs XG-CBS toward a minimal-index plan.

We now turn to the details of XG-$A^*$.
% (see Algorithm~\ref{alg:eg-Astar} in Section~\ref{sec:appendix}
%\ml{supplementary material?}
%). 
The search space of XG-$A^*$ consists of nodes of the form $(v,t,H,i)$ where $v\in V$ is the vertex, $t\in \mathbb{N}$ is the timestamp, $H$ is a sequence of vertices, representing the history of the path from the last segmentation time, and $i$ represents the plan index up to time $t$.
%The pseudocode is shown in Algorithm~\ref{alg:eg-Astar}. 
% XG-$A^*$ performs a search on the graph $G$ from the start node $(s,0,\emptyset,1)$ guided toward any node corresponding to the goal vertex $g$ as long as $i\le r$ (recall that $r$ is the allowed bound on the index of the plan). 
XG-$A^*$ performs a search on the graph $G$ from the start node $(s,0,\emptyset,1)$ guided toward any node corresponding to the goal vertex $g$ as long as $i\le \bar{r}$, where $\bar{r}$ is the index of $P_{-1}$.
% \jk{This is technically incorrect. XG-A* has no idea what r is. It only uses the minimal disjoint decomposition of $P_{-1}$. I think it should be something like ``XG-$A^*$ performs a search on the graph $G$ from the start node $(s,0,\emptyset,1)$ guided toward any node corresponding to the goal vertex $g$ as long as $i\le w$ where $w$ is the index of $P_{-1}$." We can use a different variable if you do not like $w$.}
% \ml{well... in that case, we run into a problem in with our heuristic 2, no?}
The central element is the heuristic guiding the search. A node $(v,t,H,i)$ is assigned two values: the current index $i$, which is the primary heuristic value, and the shortest-path metric from $v$ to the goal $g$ in the graph $G$ itself, which is used as a tie-breaker.
In order to expand a node, a neighbor of $v$ is selected on the graph, and $t$ is increased by 1. At this point the new vertex and time are checked against the constraints $\cC$, and if they are not constrained, $H$ and $i$ are computed as per the greedy approach described in Section~\ref{subsec:XGCBS}. 
Thus, XG-$A^*$ starts by exploring all 1-segment plans, and only once these are exhausted, moves on to 2-segments, etc. 
We refer the reader to the Appendix (Section~\ref{sec:xg-astar-alg}) for the pseudocode of XG-$A^*$ and two methods of speeding it up.

We now make two important observations regarding the behavior of XG-$A^*$. 

\begin{figure}
     \centering
     \begin{subfigure}{0.49\linewidth}
         \centering
         \includegraphics[width=0.95\linewidth]{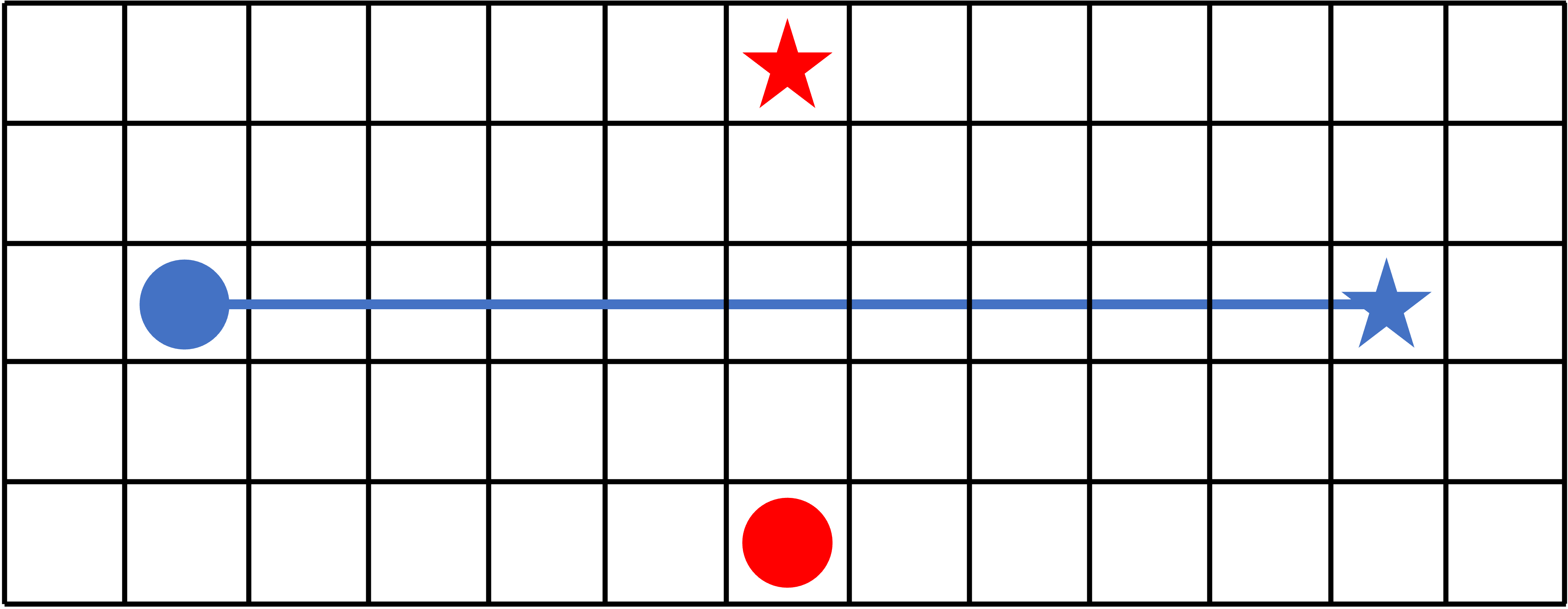}
         \caption{$r=1$ plan exists.}
         \label{fig:egAstar_easy}
     \end{subfigure}
     \hfill
     \begin{subfigure}{0.49\linewidth}
         \centering
         \includegraphics[width=0.95\linewidth]{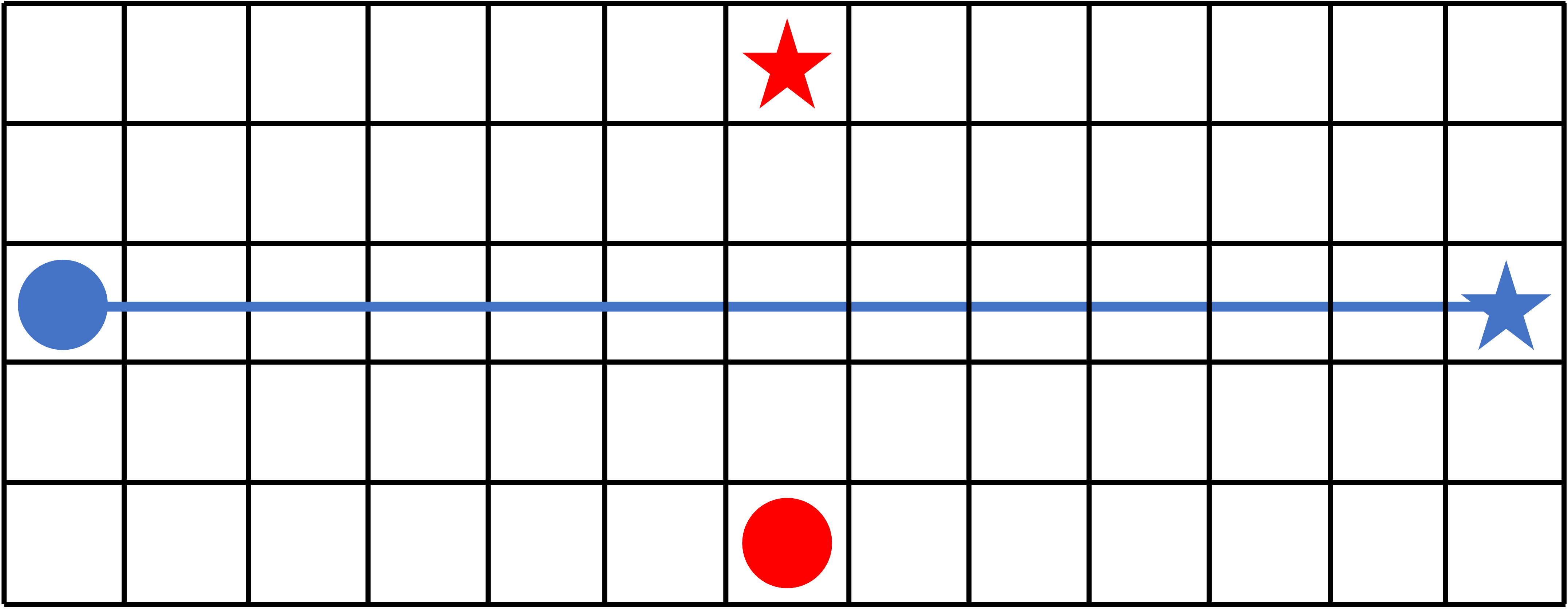}
         \caption{$r=1$ plan does not exist.}
         \label{fig:egAstar_hard}
     \end{subfigure} 
    \caption{XG-$A^*$ drawback of Remark~\ref{rmk:XGAstar_cycles}.}
    \label{fig:egAstar_challenge}
\end{figure}

\begin{remark}
\label{rmk:XGAstar_sync}
Observe that the index of a node depends not only on the plan for Agent 1, but also on the decomposition of the plan $P_{-1}$. Therefore, if $P_{-1}$ alone causes segmentation, XG-$A^*$ also increases $i$ in the current node. 
This causes XG-$A^*$ to ``synchronize'' segmentations. That is, if a path intersection in $P_{-1}$ induces a new segment, the XG-$A^*$ attempts to make Agent 1 intersect another path at that exact time, in order to avoid creating a new segment, which ultimately leads to a lower index.
\end{remark}
% \jk{I think $P$ should be $P_{-1}$ in this remark}
% \ml{good catch! yes!}
\begin{remark}
\label{rmk:XGAstar_cycles}
Since the primary heuristic is not guided toward the goal $g$, XG-$A^*$ spends a lot of time covering.  For instance, it exhausts all index-1 plans before incrementing the index, even if it is impossible to reach $g$ in 1 segment. 
This is demonstrated in Figure~\ref{fig:egAstar_challenge}, where XG-$A^*$ is used to compute a path for the red agent given the existing path of the blue agent.  In Figure~\ref{fig:egAstar_easy}, an index-1 plan exists, and XG-$A^*$ finds it relatively quickly, as it is guided toward the goal within the space of index-1 plans (by going around the blue agent). 
In Figure~\ref{fig:egAstar_hard}, it is clear that no index-1 plan exists. However, XG-$A^*$ first has to exhaust all index 1 plans, before attempting index 2 (the shortest path).
This severe drawback
means XG-$A^*$ is slow with respect to the size of the graph $G$, rather than the number of agents. As we mention in Section~\ref{sec:problem}, this difficulty is inherent to Explainable MAPF via Segmentation.
\end{remark}

Since XG-$A^*$ eventually exhausts the space of possible plans, ordered by index, and since this space is bounded using the bound $B$, we obtain the following.
\begin{theorem}[XG-$A^*$ Completeness]
\label{thm:xgAstar_complete}
    Given a set of paths $\{\pi_2, \ldots, \pi_n\}$, source and goal vertices $s_1$ and $g_1$, respectively, a set of constraints $\mathcal{C}$, and a bound $B$, if there exists a path from $s_i$ to $g_i$ of length at most $B$ that does not violate the constraints in $\mathcal{C}$, then XG-$A^*$ will terminate with such a path $\pi_i$ that minimizes the index of $\{\pi_1, \ldots, \pi_n\}$.
\end{theorem}

\subsection{WXG-\texorpdfstring{$A^*$}{A*} -- Weighted Explanation Guided $A^*$}
\label{subsec:weighted}
As demonstrated in Remark~\ref{rmk:XGAstar_cycles}, XG-$A^*$ spends a lot of time exhausting the plans of a certain index before making any progress towards the goal. 
% This occurs because the cost of node $q = (v,t,H,i)$ within XG-$A^*$ solely depends on the index $i$ and ignores path length from source $q_s = (s,0,\emptyset,1)$ to $q$.
This occurs because the cost function of XG-$A^*$ 
% attempts to minimize the plan index in exhaustive way.
is the plan index and uses path length only as a tie-breaker.
% from source $q_s = (s,0,\emptyset,1)$ to $q$.
% Conversely, the cost of $q$ in standard $A^*$ is the sum of the path length from $q_s$ to $q$ and heuristic value from $q$ to goal.
% Standard $A^*$ ignores the index that $q$ creates in relation to $P_{-1}$.
% Conversely, in standard $A^*$, the cost of $q$ is the path length from $q_s$ to $q$ plus the heuristic value from $q$ to goal,
Conversely, standard $A^*$ uses path length as the cost function and becomes efficient with a heuristic (estimate of path length to goal),
% Standard $A^*$ ignores the index that $q$ creates in relation to $P_{-1}$.
% completely ignoring the index that $q$ creates in relation to $P_{-1}$.
completely ignoring the plan index.
These algorithms are two extremities of explanation-guided graph search. 
% An intermediate approach is to combine these algorithms by means of weighting.
To get the best of both worlds, we design a general algorithm
called \emph{weighted} XG-$A^*$ (WXG-$A^*$)
that combines 
the two search methods.
% the intuitions behind XG-$A^*$ and $A^*$.
% We now turn to combining the intuitions behind XG-$A^*$ and $A^*$ into a single algorithm. 
% We refer to this approach as \emph{weighted} XG-$A^*$ (WXG-$A^*$). 
The premise behind WXG-$A^*$ is to simultaneously inherit the index-minimization property of XG-$A^*$ and the efficient search property of $A^*$.  
% To this end, WXG-$A^*$ combines the cost functions of XG-$A^*$ and $A^*$.

Let $f_x$ and $f_a$ denote the cost functions of XG-$A^*$ and $A^*$, respectively. We define the cost function of WXG-$A^*$ to be a linear combination of $f_x$ and $f_a$, i.e, for node $q$, 
% a linear combination of $f_x(n)$ and $f_a(n)$. Formally, 
$$ f_w(q)=wf_x(q) + (1-w) f_a(q), $$
where $w \in (0, 1)$.   
% As demonstrated in Remark~\ref{rmk:XGAstar_cycles}, XG-$A^*$ may spend a lot of time exhausting the plans of a certain index before making any progress towards goal. This occurs because the cost of node $n$ within XG-$A^*$ is equivalent to the index $i$ of $n$ in relation to $P_{-1}$. We denote the cost function for XG-$A^*$ as $f_x(n)$. Conversely, the cost function used in standard $A^*$ (denoted as $f_a(n)$) is equivalent to path length of from $s$ to $n$ plus an admissible heuristic that estimates the distance from $n$ to $g$. 
% \ml{why do you introduce new notations here when you don't use them here.  You're still describing the high-level ideas.  The details come in the next paragraph.  That's where you should introduce new notations}
% We treat both algorithms as the two extremes for explanation guided graph search. That is, XG-$A^*$ is \emph{solely} guided by the explanation and \emph{blind} to path length while $A^*$ is \emph{blind} to explanation and \emph{solely} guided by path length. We now present an alternative approach to explanation guided graph search that combines both approaches.
% By combining these two cost terms, 
% Function $f_w(q)$ is
% \ml{not really a middle point... more like in-between}
% between the two extreme forms of explanation-guided planning by encouraging 
The function $f_w(q)$ encourages
\emph{both} index 
minimization and efficient graph search. The amount that $f_w$ tends toward either type of graph-search depends on weight $w$.  As $w \to 1$, $f_w$ biases more towards minimal-index paths, and hence, the search becomes exhaustive (slower). Conversely, as $w \to 0$, 
% the index of the plan has a lesser effect on the path. 
the search tends more towards shortest path length (hence faster).
% \ml{give intuition on the choice of $w_x$ and $w_a$.  What happens as you increase $w_x$ or $w_a$? Explain what this function $f_w$ really represents}
Algorithmically, WXG-$A^*$ is simply XG-$A^*$ guided by $f_w$ rather than $f_x$.
% \ml{OK, now you have $f_w$, then what?  Do you do A* or XG-A* with this cost function?}
% \ml{still can give more intuition... discuss how $f_a$ is often dominant unless the number of segments is large.  Because of this dominance, $f_x$ acts more like a tie breaker / slight preference toward small segment-paths among paths with similar path lengths when $w_a$ is similar to $w_x$ in value.}

We note that careful consideration is needed in choosing a value for $w$. An intuition is that $f_a$ (path length) is typically much greater than $f_x$ (number of segments).  Unless $w$ is very large, $f_a$ is dominant and $f_x$ acts more like a tie-breaker.
In Section~\ref{sec:experiments}, we empirically show how varying $w$ changes the behavior of XG-CBS.
% \ml{say that WXG-A* is also complete because ...}
Finally, note that WXG-$A^*$ exhausts the same search space as XG-$A^*$, differing only in the order of the search. Therefore, Theorem~\ref{thm:xgAstar_complete} still holds for WXG-$A^*$, i.e.,  WXG-$A^*$ is complete.
% We show how varying $w_x$ and $w_a$ changes the behavior of XG-CBS in Section~\ref{sec:experiments}.
% \ml{explain why this would work better.. e.g., by combining these two cost terms, we encourage exploration of bluh bluh... hence, the search is biased ... bluh bluh ... As a results, bluh bluh bluh}
\subsection{SR-\texorpdfstring{$A^*$}{A*} -- Segmentation Respecting $A^*$}
\label{subsec:sr}

% Note that it is difficult, or even impossible, to determine whether more weight should be given to $f_x(n)$ or $f_a(n)$ for a given problem \textit{a priori}. Therefore, an improved, problem-agnostic, graph search is needed.

While WXG-$A^*$ can theoretically provide a good balance (trade-off) between efficiency and index minimization, it suffers from two drawbacks. First, it is difficult to choose an appropriate weight $w$ \textit{a priori} to achieve a good balance, since it is highly instance dependent. Second, WXG-$A^*$ needs to maintain the history of the path (as in XG-$A^*$) in order to perform segmentation, resulting in a slow search algorithm.
%Therefore, an improved, problem-agnostic, graph search is needed. Here, we introduce such an algorithm by sacrificing completeness.
We propose a new low-level algorithm that does not keep track of history, thus obtaining a significant speedup.

Recall from Remark~\ref{rmk:XGAstar_sync}
% \ml{it doesn't claim... it states}
that XG-$A^*$ computes paths that fit within the existing segmentation of $P_{-1}$ by
keeping track of the index of $P_{-1}$ combined with the new path, which requires keeping the history of the path from the last segmentation point.
A coarse way of eliminating the need to keep the history is to make sure the planned path completely avoids all paths in $P_{-1}$, and so does not contribute to segmentation. This, however, likely results in no plans being found, as it amounts to keeping the agents disjoint. 
Our proposed algorithm, dubbed \emph{segmentation-respecting $A^*$} (SR-$A^*$), refines this idea, by making sure that the planned path is disjoint from all paths \emph{within the current segment}. 
%reasoning over the index of paths.  We now present an $A^*$ search that mimics the behavior of XG-$A^*$ without the need for the index cost function $f_x$. We call the algorithm segment-respecting $A^*$ (SR-$A^*$). 
Intuitively, SR-$A^*$ treats every disjoint segment within $P_{-1}$ as time dependent obstacles. That is, existing paths within a segment become obstacles only for the time window of the segment. 
% That is, the paths within every segment are treated as obstacles in the time window of the corresponding segment.
% \ml{then, what happens? explain the behavior...}
The resulting behavior is an efficient graph search algorithm that is dedicated to fitting within an existing segmentation. 
% We now present a more coherent way 
% \ml{is it more coherent?}
% to combine the intuitions behind XG-$A^*$ and $A^*$. We call our method \emph{Segment Respecting}-$A^*$ (SR-$A^*$). Intuitively, SR-$A^*$ performs standard $A^*$ graph search while treating the existing explanation for $P_{-1}$ as time dependent obstacles. 
% \ml{First you need to explain what the shortcomings/difficulties of the above methods are. Then, state an observation, based on which you can design a fast algorithm.  Then explain the new algorithm}

Formally, consider a plan $P_{-1}$ with a disjoint decomposition $t_0<t_1<\ldots<t_r$, and a planning query for Agent $1$. The search space is now modified by adding a ``timed obstacle'' at vertex $v$ at time $t$ as follows. Let $1\le i\le r$ be the segment such that $t_i\le t\le t_{i+1}$, then we add a timed obstacle if there is a path of $P_{-1}$ that visits vertex $v$ at the interval $[t_i,t_{i+1}]$. For example, if $P_{-1}$ contains the segment $v_1,v_2,v_3$ at times $3,4,5$, respectively, then vertices $v_1,v_2$ and $v_3$ are all obstacles at times $[3,5]$.

%SR-$A^*$ begins by iterating through every path segment $\{\tau_1^s, \ldots, \tau_{n-1}^s\}$ for $s=\{1, \ldots, p\}$. For the first segment, SR-$A^*$ examines every $\tau_i^1=v_{i1} \ldots v_{ik_i}$ for $i\in \{1,\ldots,n-1\}$ and creates a timed obstacle at every $v_{ij}$ that exists for time range $\Delta t\in\left[1,k_i\right]$. Outside of $\Delta t$, the obstacles do not exist. This process repeats for all $p$ segments of $P_{-1}$. Once all timed obstacles have been created, standard $A^*$ is used to generate a path that respects physical obstacles, timed obstacles, and constraints. 
% Intuitively, the first segment of $P_{-1}$ is treated as obstacles for the same amount of time that the segment exists. Then, those obstacles vanish and new obstacles appear for the second segment, and so on. 

Observe that crucially, if Agent $1$ does not intersect with any timed obstacle, then it also does not create new segments, and hence ``respects'' the segmentation of $P_{-1}$. In particular, SR-$A^*$ breaks the completeness of XG-CBS. Indeed, the restriction of the search space means that some paths are never explored. 
From an efficiency perspective, however, SR-$A^*$ both limits the search space, and eliminates the tracking of history, rendering this search comparable to $A^*$. 
In \cref{sec:experiments}, we demonstrate that SR-$A^*$ performs exceedingly well, both in terms of efficiency and plan index.

\section{Case Studies}
\label{sec:experiments}

We evaluate the performance of XG-CBS %Our case studies are 
on a combination of self-designed problems and standard MAPF benchmark problems available in \cite{mapfBenchmarks}. The self-designed spaces %are used to 
exhibit interesting 
%explanation 
behaviors that are unique to Problem~\ref{problem:expMAPF}. The results of our benchmarks
%are used to 
show the advantages and disadvantages of the proposed algorithms in various environments and scenarios.  
% shown in Table~\ref{tab:table_benchmark}.
% Our C++ implementation is available on GitHub~\cite{sourceCode} 
% \ml{can't have GitHub link since it's double blind review.  Maybe we can submit our code as supplementary material?}\jk{Yes, I have been thinking about working with John to set up a Docker image for my code. This would make it easy for reviewers to look at the code and reproduce the results}\ml{can you submit anything other than pdf as supplementary material?}. 
All experiments were performed on a machine with an AMD Ryzen 7 3.9GHz CPU and 64 GB of RAM. Our implementation is available on GitHub~\cite{sourceCode}.

\subsection{Illustrative Examples}
\begin{figure}[t]
     \centering
     \begin{subfigure}{0.49\linewidth}
         \centering
         \includegraphics[scale=0.2]{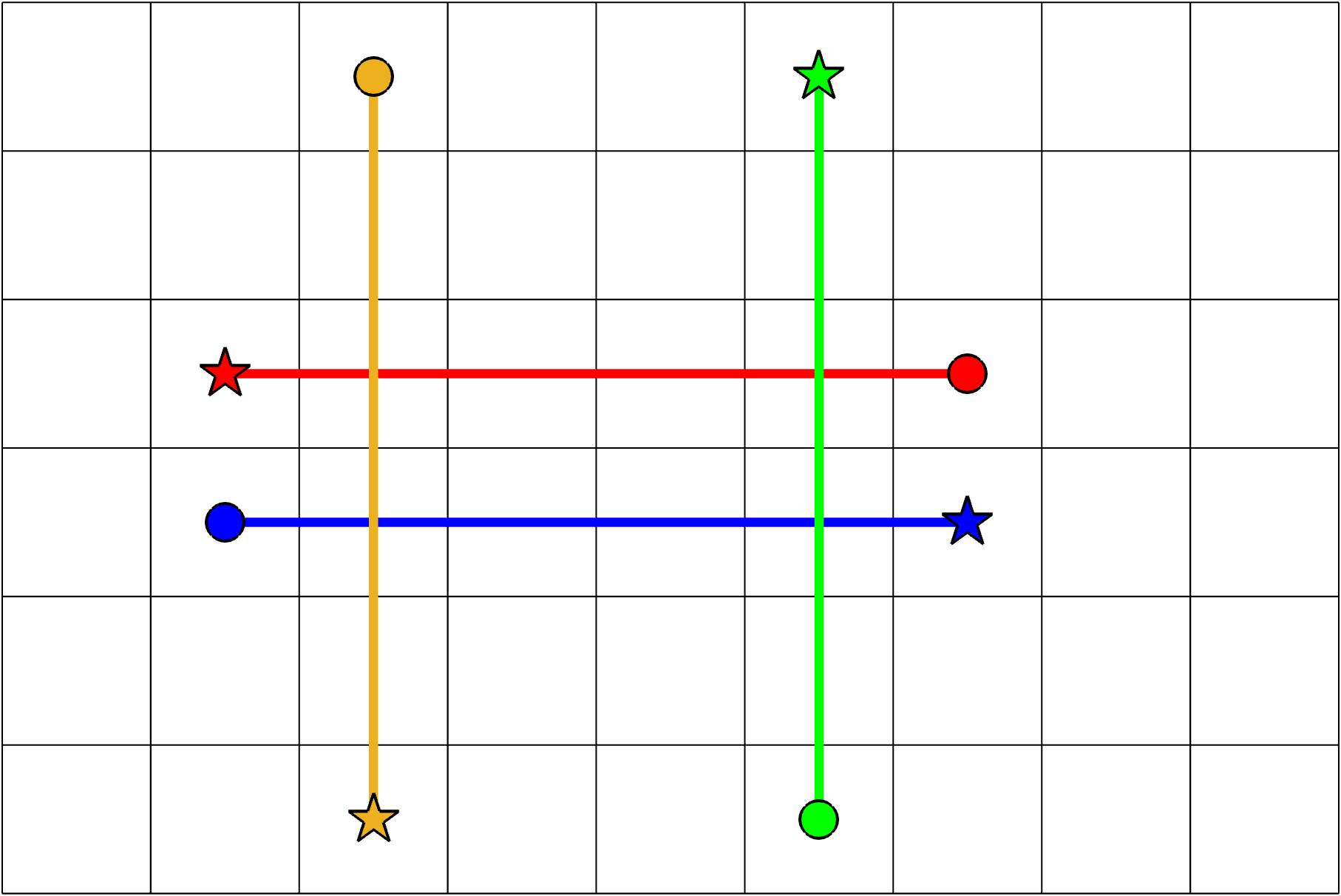}
         \caption{CBS}
         \label{fig:HT7_cbs_full_plan}
     \end{subfigure}
     \hfill
     \begin{subfigure}{0.49\linewidth}
         \centering
         \includegraphics[scale=0.2]{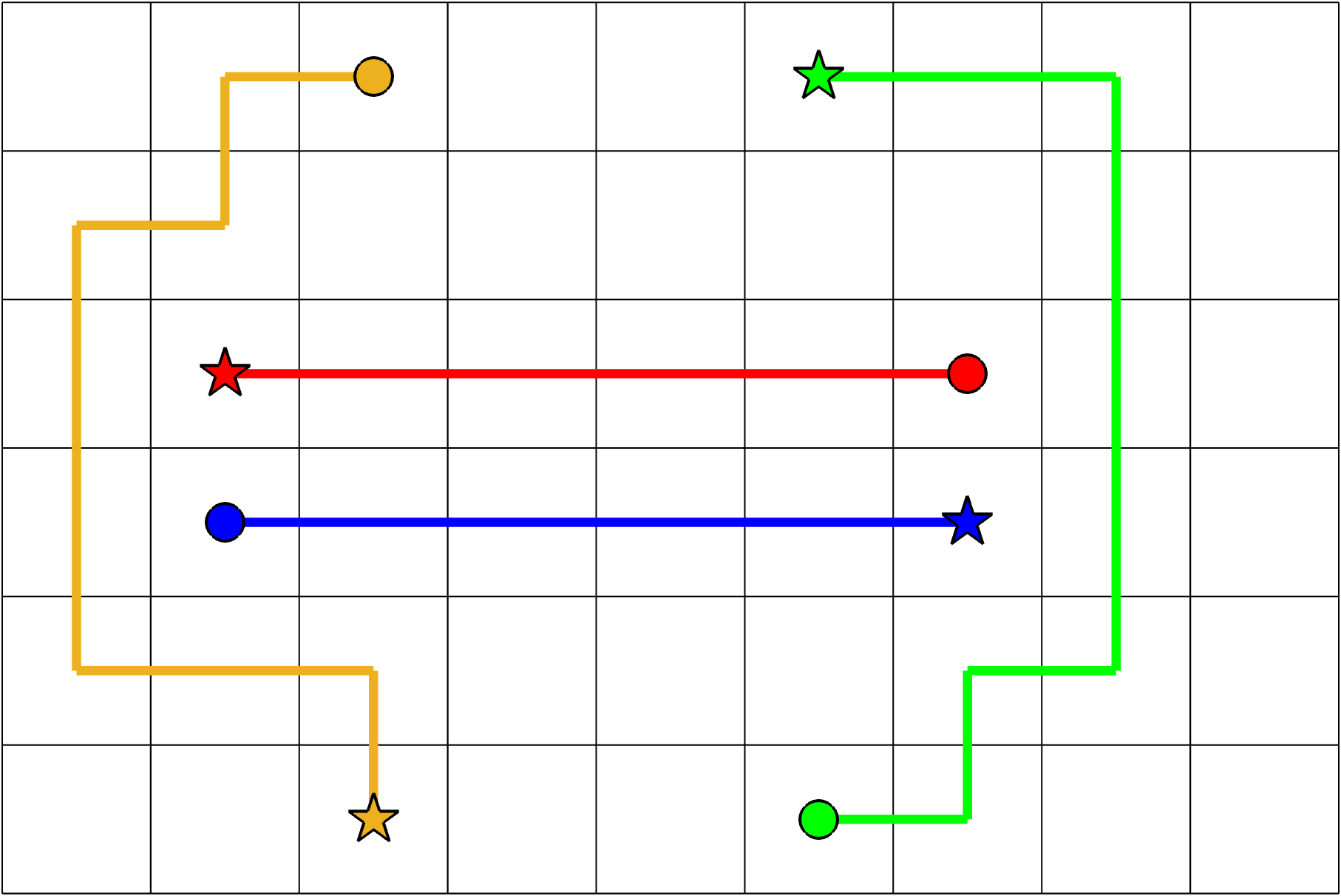}
         \caption{XG-CBS, $r=1$}
         \label{fig:HT7_egcbs_full_plan}
     \end{subfigure}
     \hfill
     \newline
     \begin{subfigure}{0.49\linewidth}
         \centering
         \includegraphics[scale=0.2]{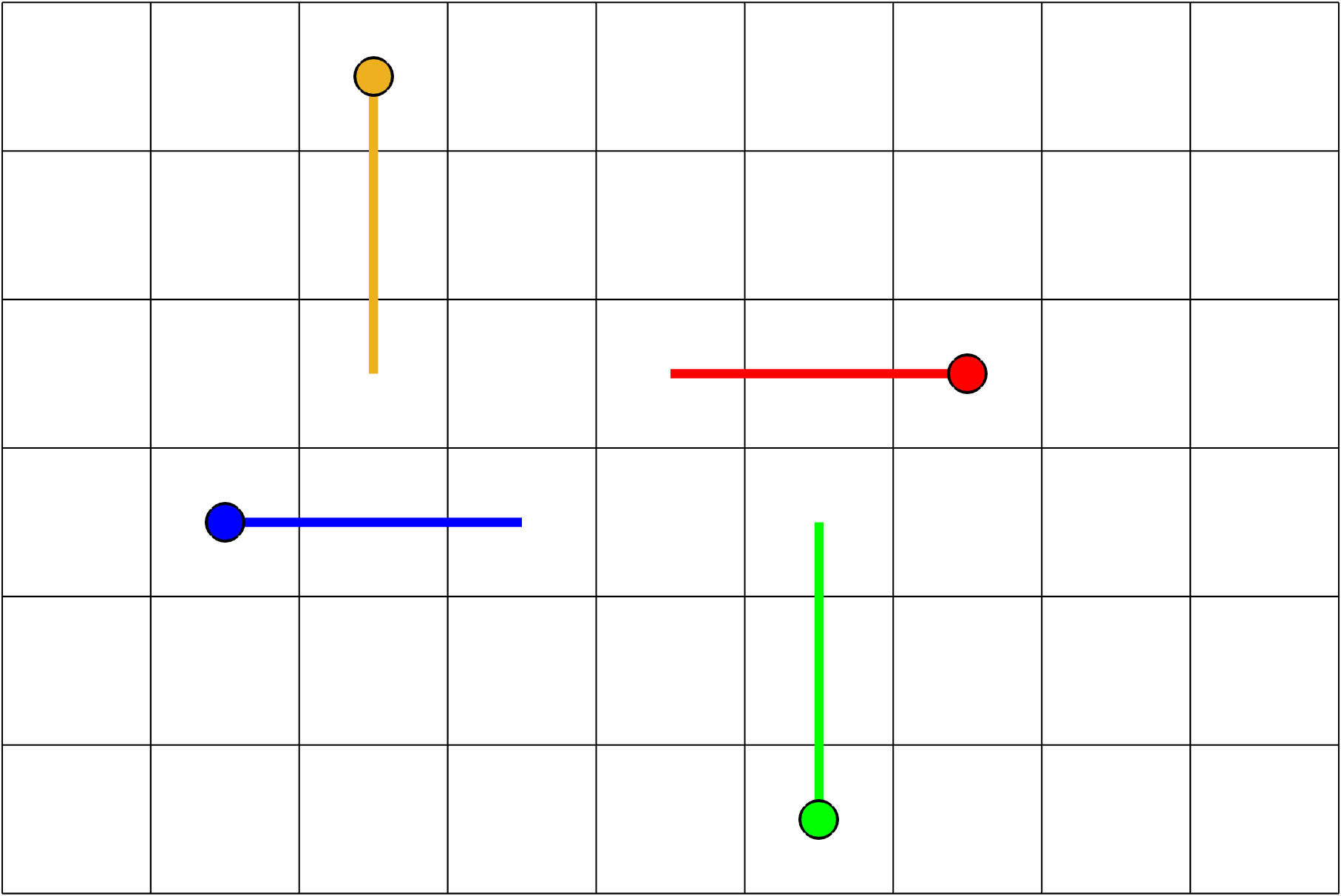}
         \caption{CBS $\Delta k=[0,2]$}
         \label{fig:HT7_cbs_s1}
     \end{subfigure}
     \begin{subfigure}{0.49\linewidth}
         \centering
         \includegraphics[scale=0.2]{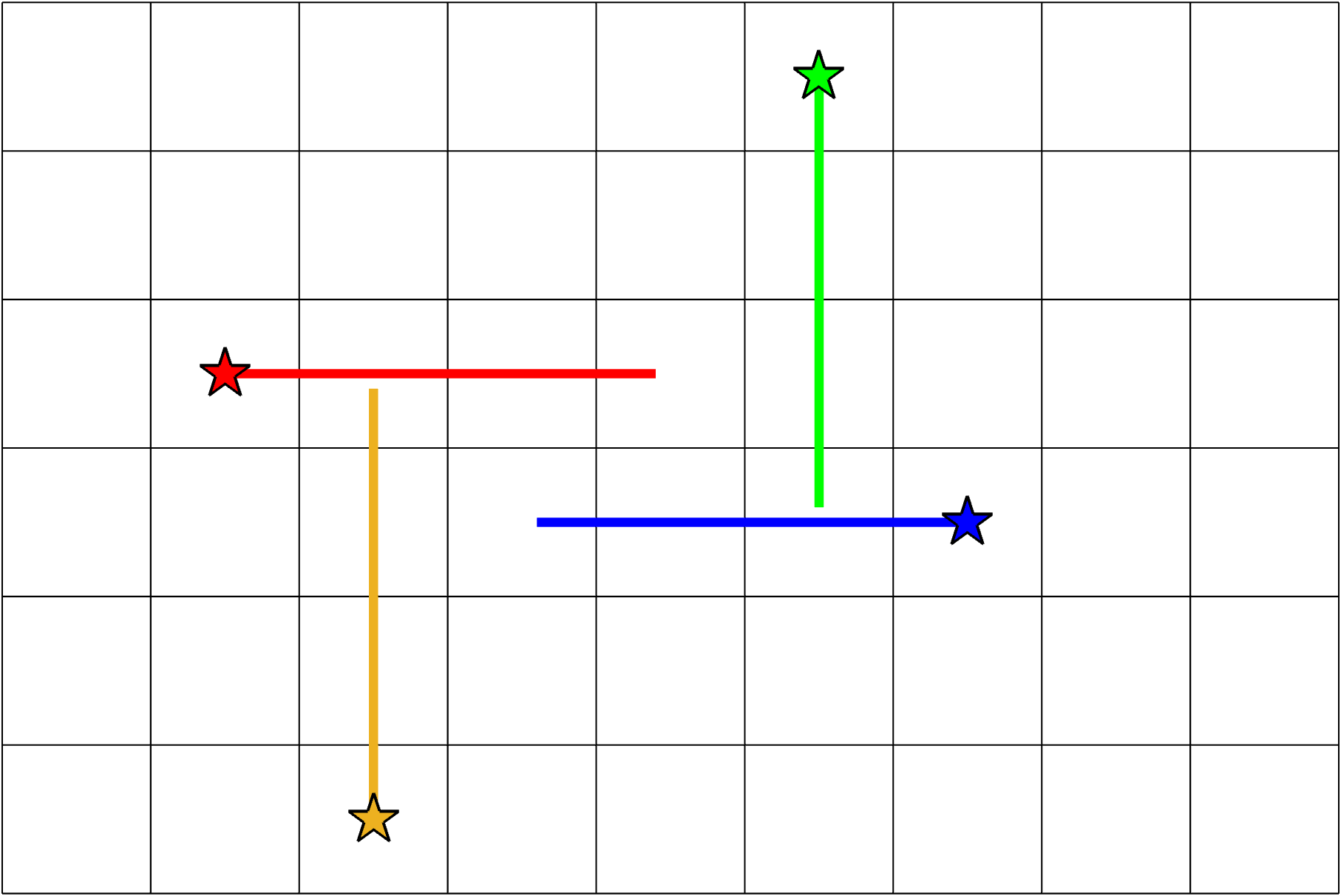}
         \caption{CBS $\Delta k=[2,5]$}
         \label{fig:HT7_cbs_s2}
     \end{subfigure}
     \vspace{-1mm}
    \caption{Road crossing: solutions via CBS and XG-CBS}
    \label{fig:expConcept}
    \vspace{-1mm}
\end{figure}
\begin{figure}[t]
     \centering
     \begin{subfigure}{0.32\linewidth}
         \centering
         \includegraphics[scale=0.18]{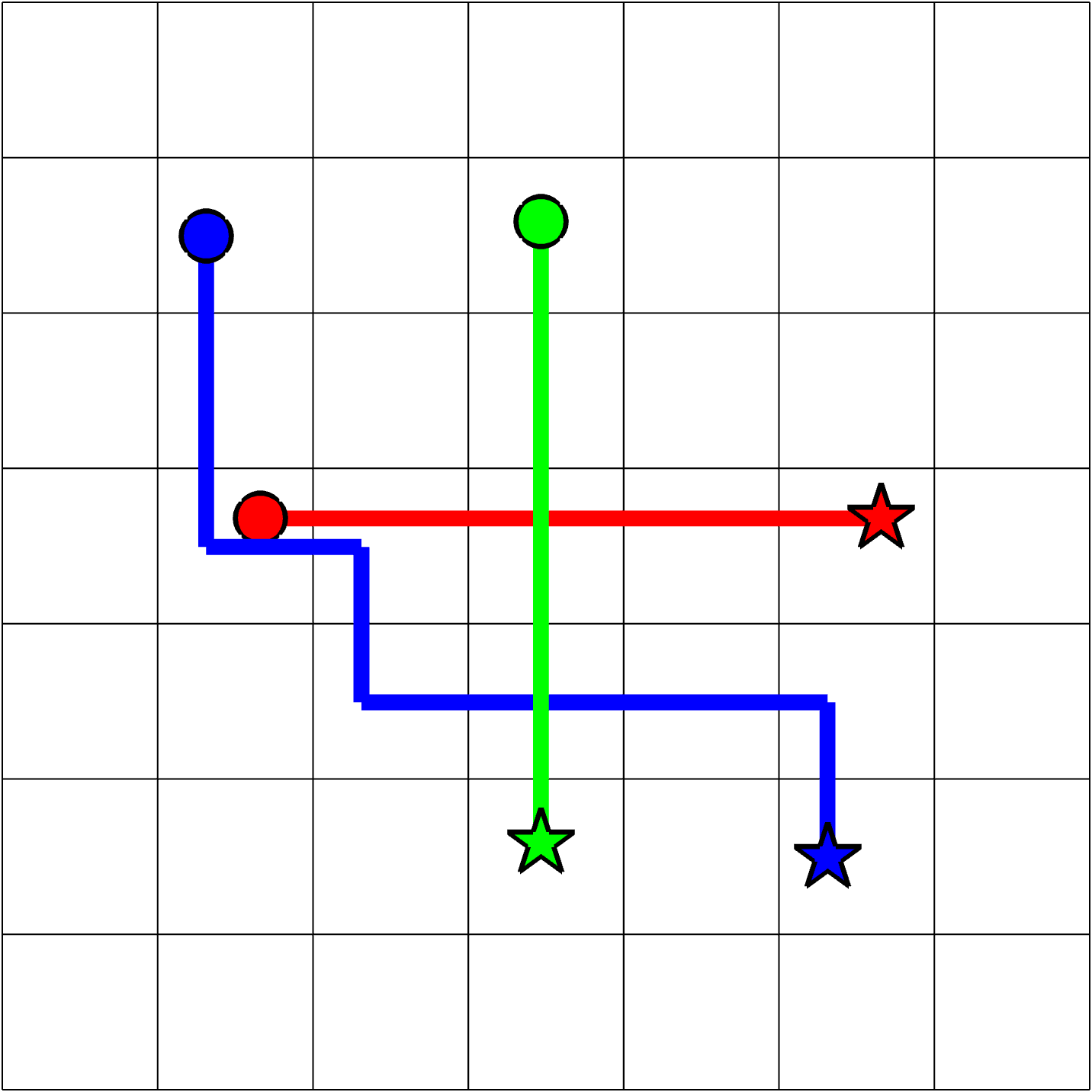}
         \caption{CBS}
         \label{fig:ht3_cbs_full}
     \end{subfigure}
     \hfill
     \begin{subfigure}{0.32\linewidth}
         \centering
         \includegraphics[scale=0.18]{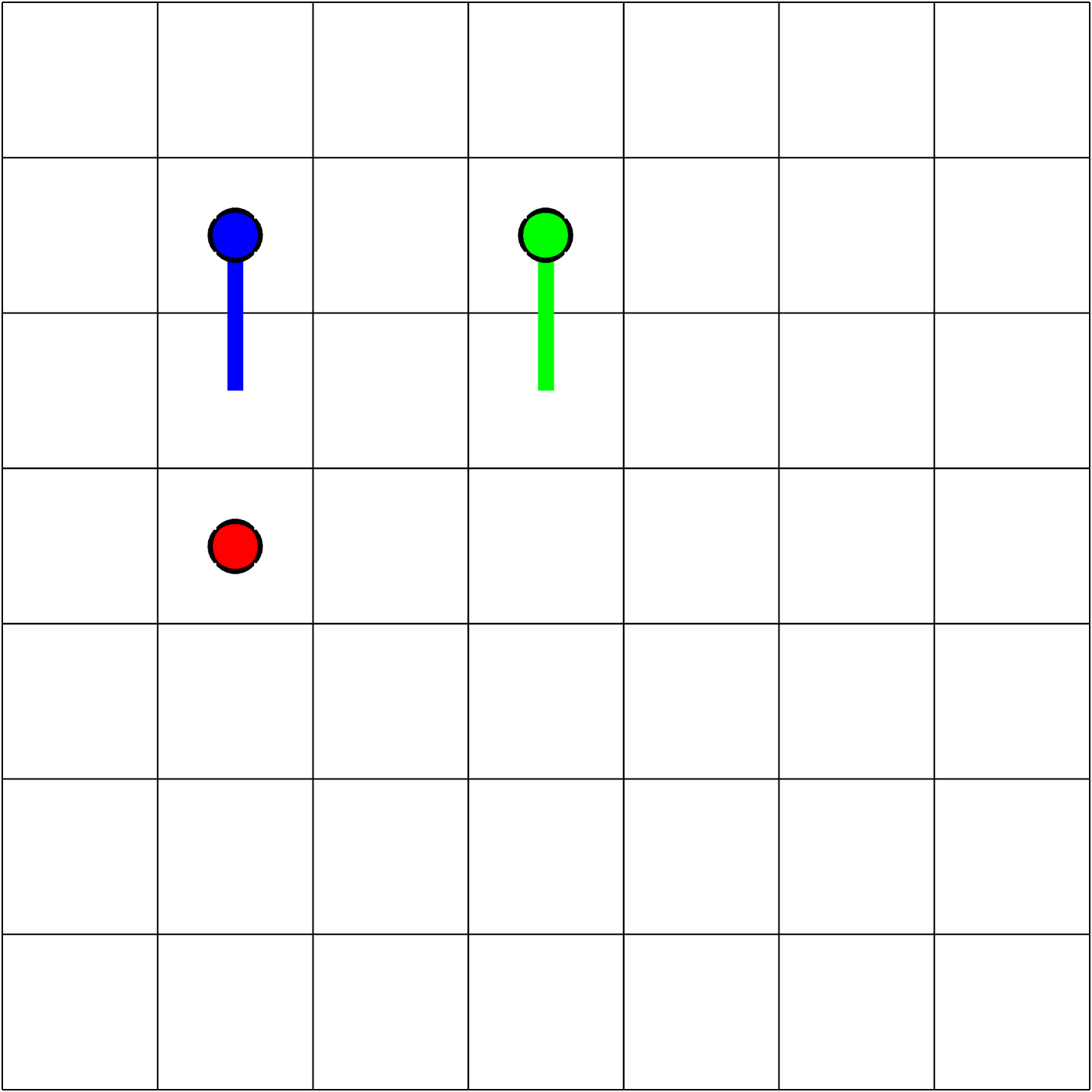}
         \caption{CBS $\Delta k=[0,1]$}
         \label{fig:ht3_cbs_s1}
     \end{subfigure}
     \hfill
     \begin{subfigure}{0.32\linewidth}
         \centering
         \includegraphics[scale=0.18]{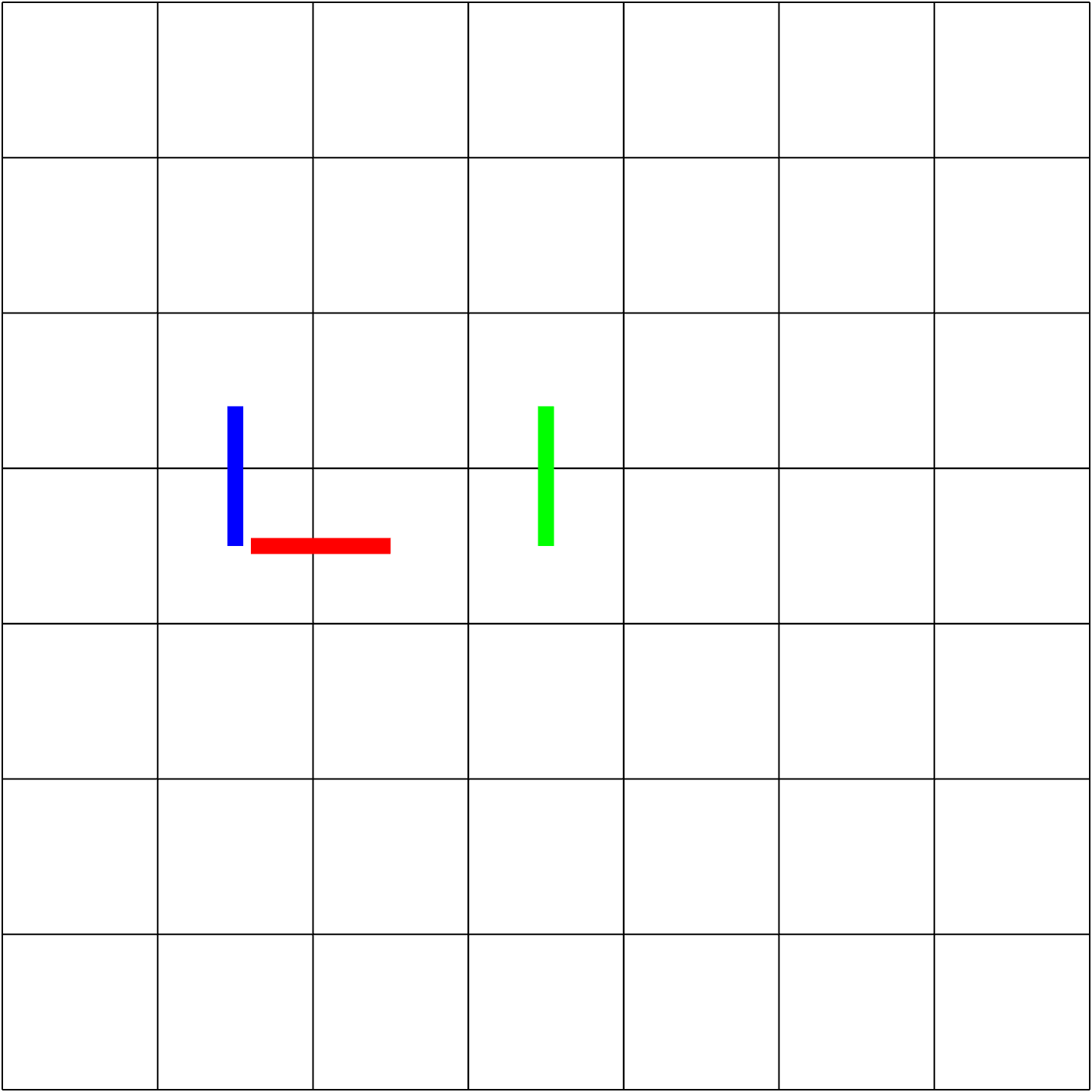}
         \caption{CBS $\Delta k=[1,2]$}
         \label{fig:ht3_cbs_s2}
     \end{subfigure}
     \hfill
     \newline
     \begin{subfigure}{0.32\linewidth}
         \centering
         \includegraphics[scale=0.18]{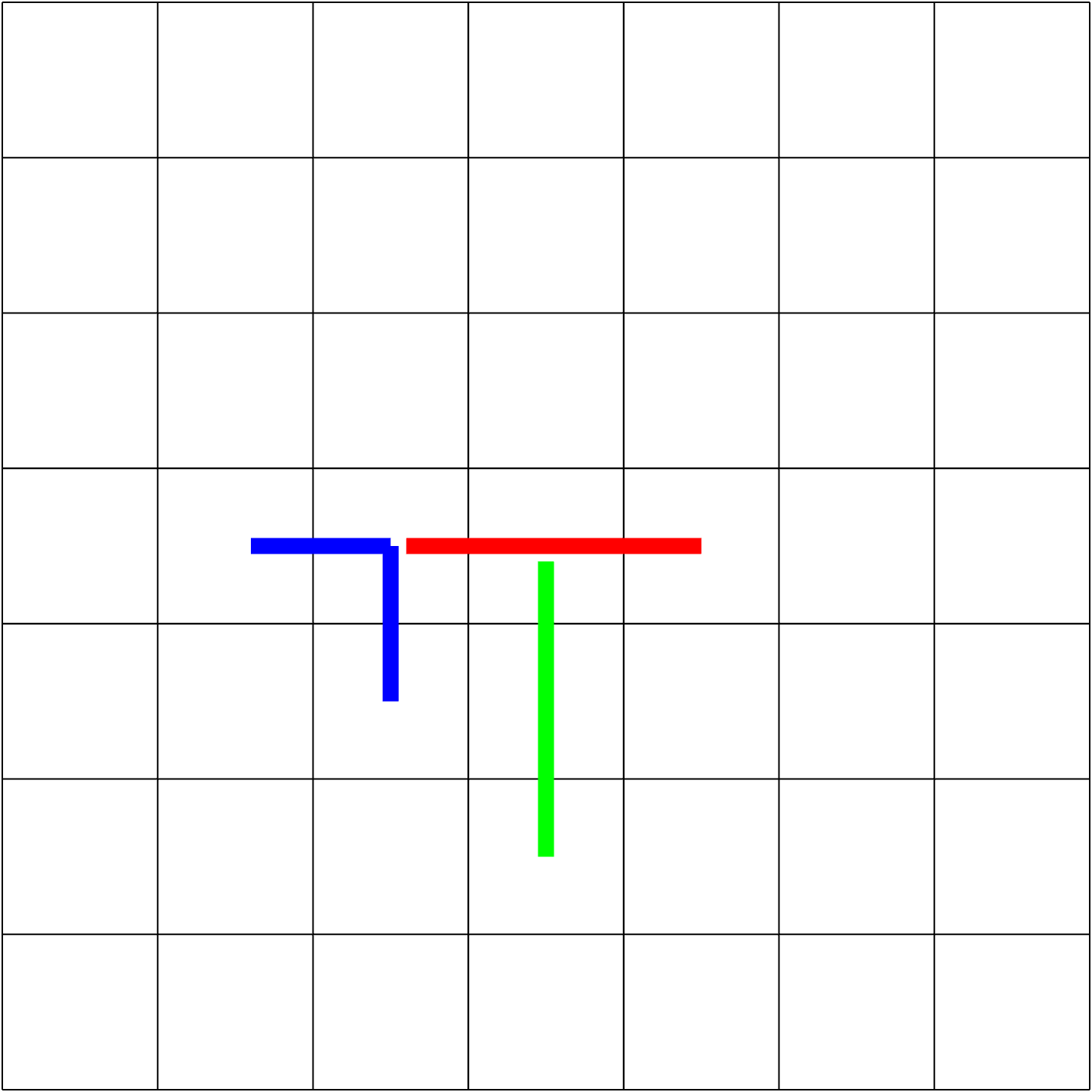}
         \caption{CBS $\Delta k=[2,4]$}
         \label{fig:ht3_cbs_s3}
     \end{subfigure}
     \hfill
     \begin{subfigure}{0.32\linewidth}
         \centering
         \includegraphics[scale=0.18]{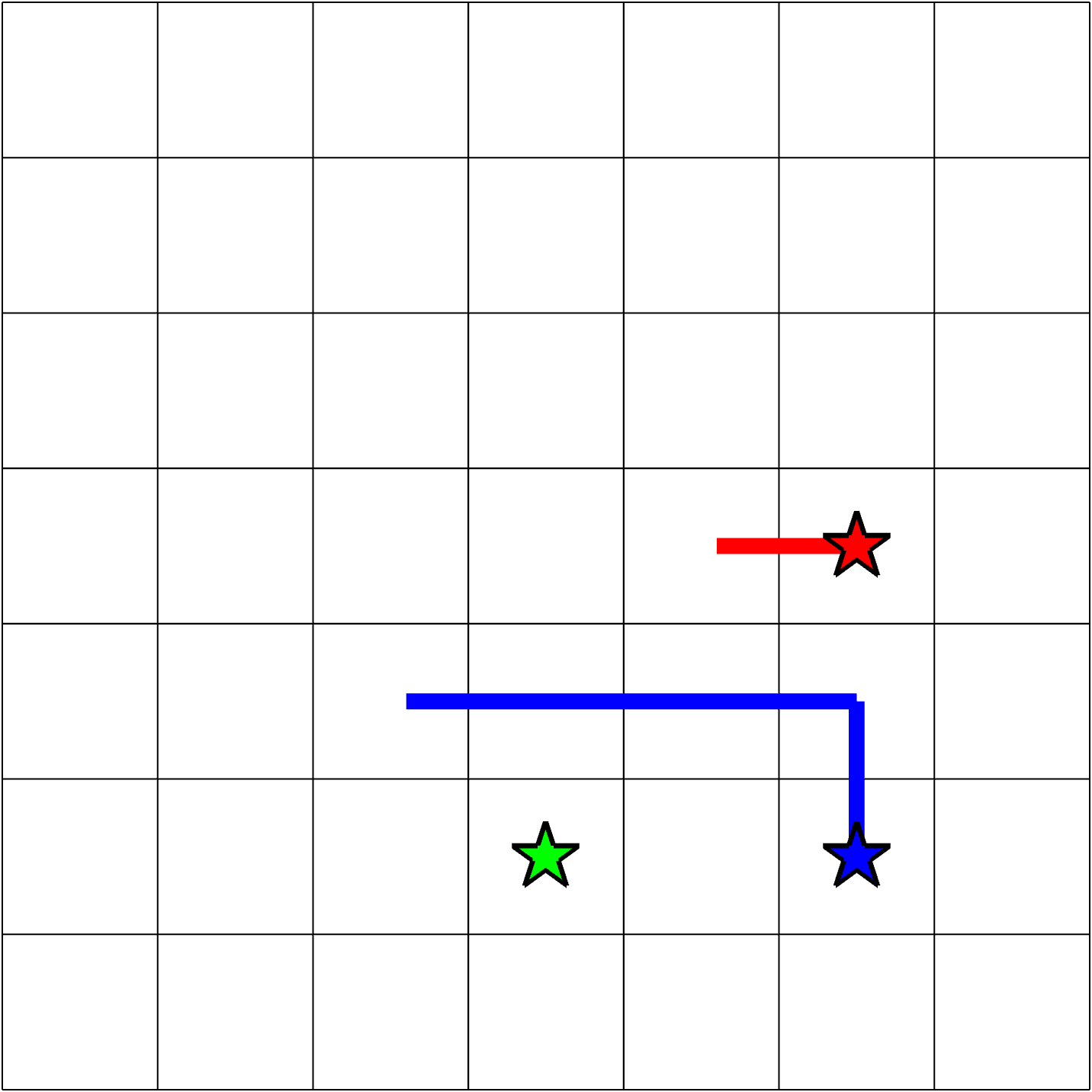}
         \caption{CBS $\Delta k=[4,8]$}
         \label{fig:ht3_cbs_s4}
     \end{subfigure}
     \begin{subfigure}{0.32\linewidth}
         \centering
         \includegraphics[scale=0.18]{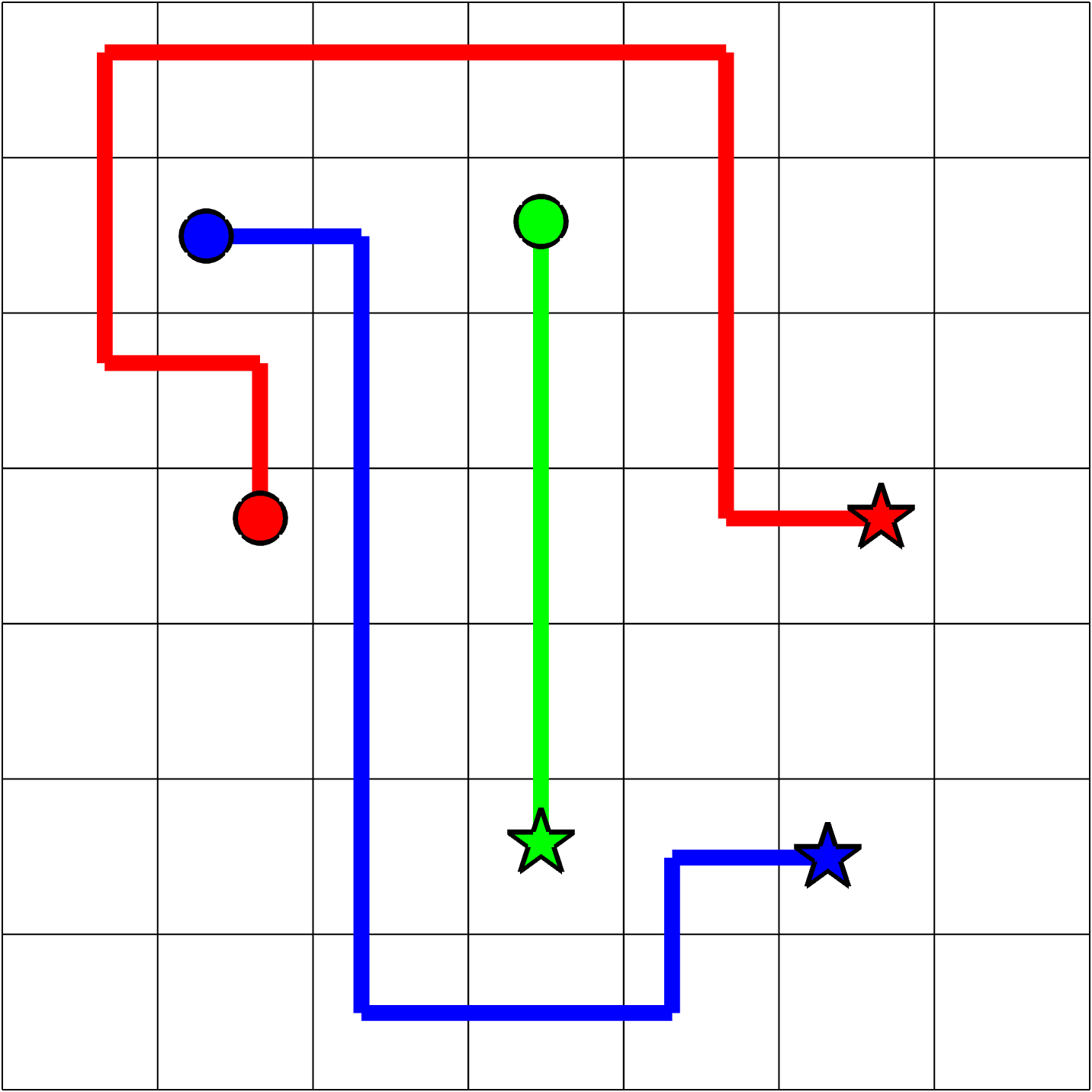}
         \caption{XG-CBS, $r=1$}
         \label{fig:ht3_egcbs_full}
     \end{subfigure}
     \vspace{-1mm}
    \caption{Apparent collision in short plan vs. optimal index}
    %\caption{Warehouse: Shortest Plan vs. Optimal Explanation}
    \label{fig:ht3}
    \vspace{-3mm}
\end{figure}

To gain insight into XG-CBS, we showcase it on
%We begin by showcasing XG-CBS in %simple 
settings that present unique explanation challenges. %, to gain insight into its properties.
Figure~\ref{fig:HT7_cbs_full_plan} shows a CBS solution of MAPF, where four agents need to cross an intersection. 
Visually verifying that the plan is collision free is difficult.
% , as
%Simply examining the entire plan makes it difficult for a human to validate that the plan is collision free since 
% it requires careful tracking of the agents over time. 
It becomes easy using the explanation scheme, which decomposes the plan into two disjoint segments in Fig.~\ref{fig:HT7_cbs_s1} and~\ref{fig:HT7_cbs_s2}. 
Using XG-CBS, we obtain a plan with index 1, as depicted in Fig.~\ref{fig:HT7_egcbs_full_plan}, which is much easier to verify.
This demonstrates the trade-off between plan length and explanations: the shortest plan requires index 2, while index 1 can be achieved with a longer plan.

% This setting demonstrates the 
% trade-off between plan length and explanations: the shortest plan requires index 2, while a longer plan, found using XG-CBS, can attain index 1, as depicted in Fig.~\ref{fig:HT7_egcbs_full_plan}.

Performance-wise, XG-CBS with $A^*$, as proposed in Section~\ref{subsec:XGCBS}, timed out after a 15 minute threshold, whereas XG-CBS with XG-$A^*$ arrived at an index-1 solution in $0.05$ seconds. This difference can be attributed to the facts that the set of index-1 plans is comparatively sparse in the set of plans, and that index-1 plans 
%We begin examining the most intuitive solution: XG-CBS using $A^*$ as the low-level planner, as proposed in Section~\ref{subsec:XGCBS}. The algorithm times-out without a solution after the threshold of $15$ minutes of planning time. There are two reasons for the decreased efficiency of the approach. 
%First, the set of plans that return the optimal explanation are comparatively sparse in relation to the full set of possibly plans. 
%Secondly, plans that produce the optimal explanation 
greatly deviate from the shortest plan. As we discuss in Section~\ref{subsec:performance}, these factors have a significant effect on the efficacy of each algorithm.

%These challenges mean that XG-CBS must iterate through many constraints before $A^*$ can realize the goal of the high-level algorithm. In contrast, replacing the low-level planner with XG-$A^*$ returns the optimal solution shown in Figure~\ref{fig:HT7_egcbs_full_plan} almost immediately ($0.05$ seconds). The difference is that XG-$A^*$ is equipped to prioritize paths that result in optimal explanation schemes, enabling it to direct its search towards more explainable solutions. 
% \jk{do we need to specify that we are using the heuristics here? If so, can we put it in the beginning of the section since we do not present non-heuristic approach at all?}
% \ml{we should definitely mention it in 5.2}

Our next use case %concerns  
is depicted in 
%We now focus on a use case of our approach. Consider the plan in 
Figure~\ref{fig:ht3_cbs_full}. %, for a set of hazardous material warehouse robots. 
A human examining the plan may notice a possible collision between the red and green agents. % in the middle of the space. 
However, it becomes clear in the explanation (Figures~\ref{fig:ht3_cbs_s1}-\ref{fig:ht3_cbs_s4}) that the red agent does, in fact, wait at the first vertex, %for the green agent to cross in front, 
thus avoiding collision. An improved explanation can be obtained using XG-CBS with XG-$A^*$ as shown in Fig.~\ref{fig:ht3_egcbs_full}.
This solution was obtained in $0.5$ seconds, whereas XG-CBS with $A^*$ again timed out.
For more case studies, we refer the reader to the supplementary material.

\subsection{Benchmark Evaluation}
\label{subsec:performance}
We now evaluate XG-CBS with the different low-level algorithms on a large set of MAPF benchmarks from~\cite{mapfBenchmarks}. Our comparison of the algorithms is along three axes: computation time, segmentation index, and plan length (average cost, i.e., sum-of-costs divided by number of agents). We also evaluate CBS as a baseline. 

\begin{figure*}[t]
    \centering
    \begin{subfigure}{\textwidth}
        \centering
        \includegraphics[width=0.858\textwidth]{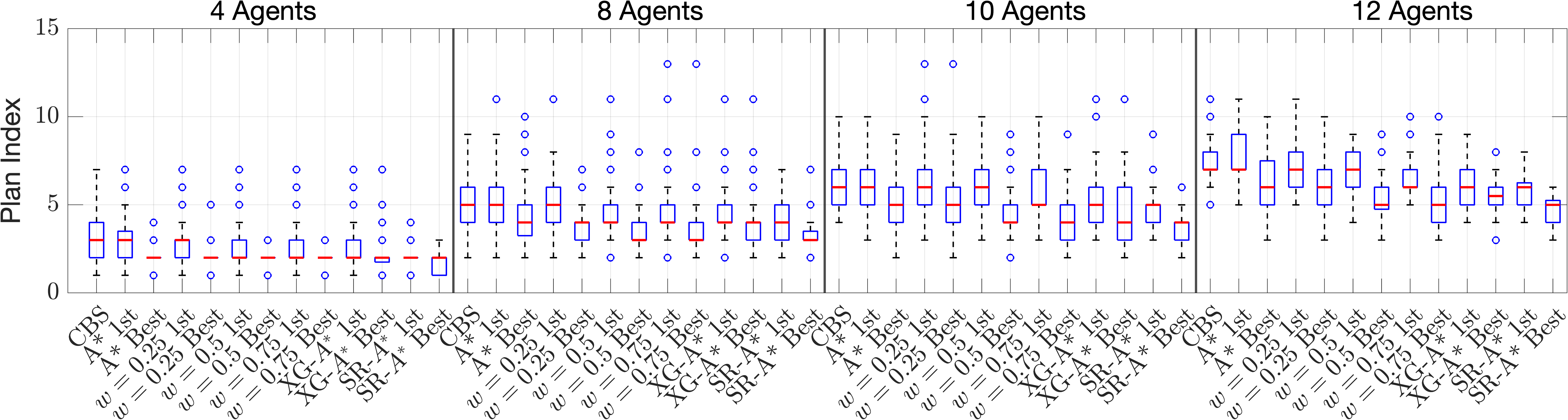}
        \caption{Plan index for $9\times 9$ environments.}
        \label{fig:plan_index_9x9}
    \end{subfigure}
    \hfill
    \begin{subfigure}{\textwidth}
        \centering
        \includegraphics[width=0.858\textwidth]{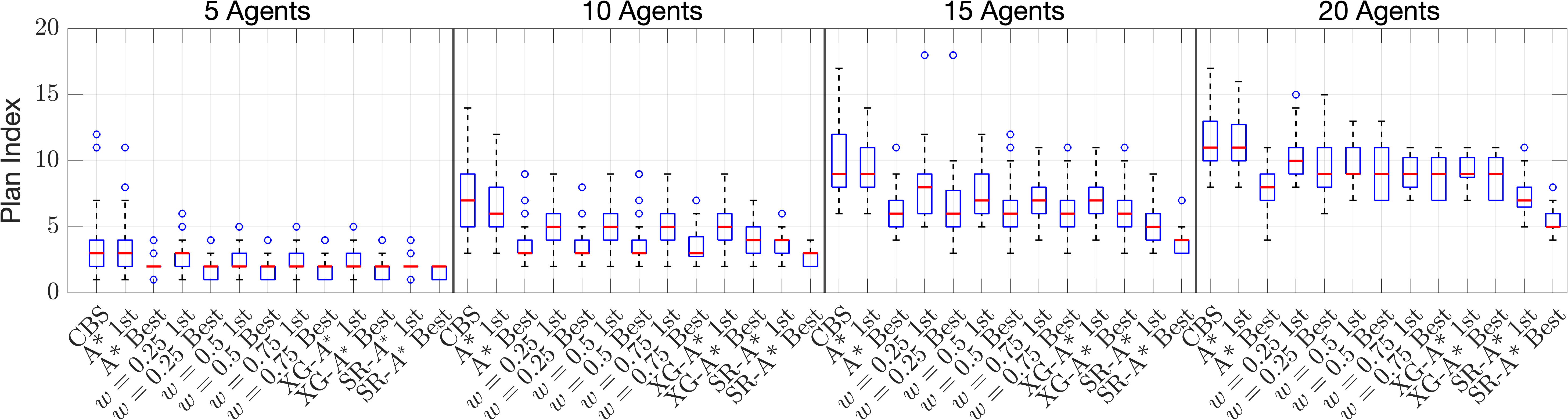}
        \caption{Plan index for $16\times 16$ environments.}
        \label{fig:plan_index_15x15}
    \end{subfigure}
    \hfill
    \begin{subfigure}{0.49\textwidth}
        \centering
        \includegraphics[width=\textwidth]{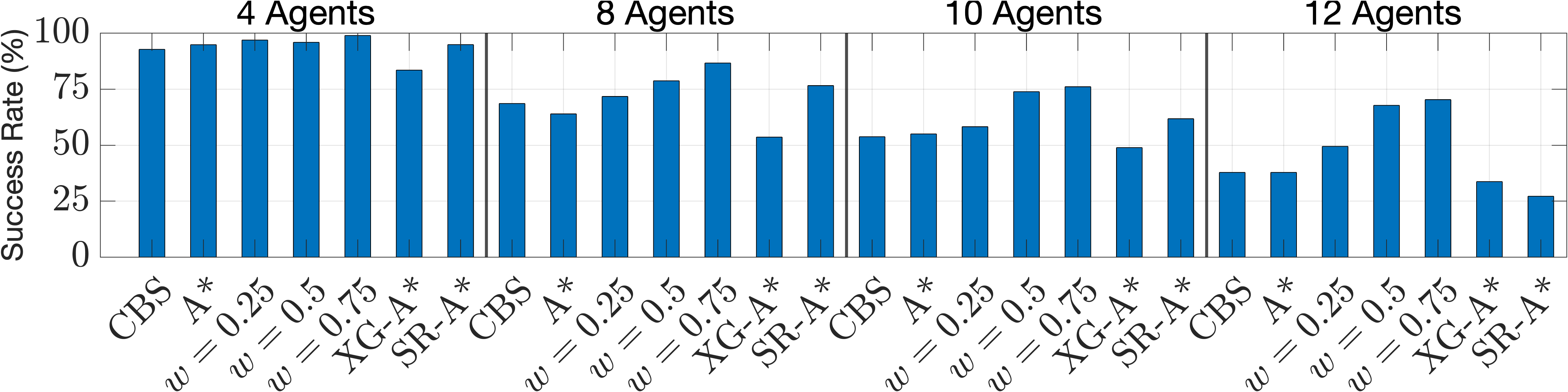}
        \caption{Success Rate for $9\times 9$ environments.}
        \label{fig:suc_rate_9x9}
    \end{subfigure}
    \hfill
    \begin{subfigure}{0.49\textwidth}
        \centering
        \includegraphics[width=\textwidth]{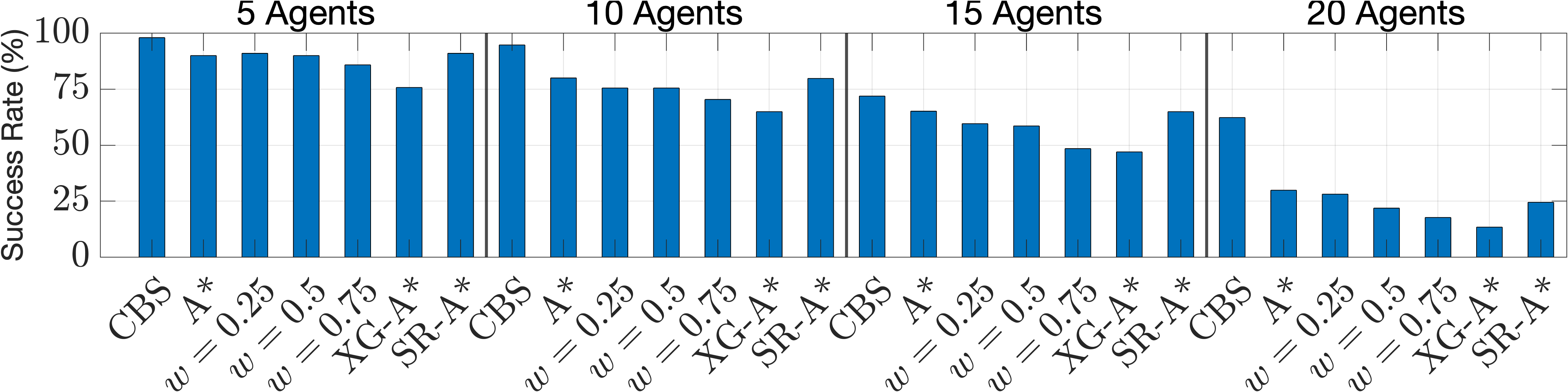}
        \caption{Success Rate for $16\times 16$ environments.}
        \label{fig:suc_rate_15x15x}
    \end{subfigure}
    \caption{Benchmark results for $9\times9$ and $16\times 16$ environments.}
    \label{fig:benchmark_8_15}
    % \vspace*{1mm}
\end{figure*}

Our experiments are run as follows. For each benchmark, we run CBS. If CBS finds a plan, we segment it and use the index as an upper bound for XG-CBS. We then repeatedly lower the bound in XG-CBS, until it times out. We refer to the former result as \emph{first} and to the latter as \emph{best}. In case CBS does not terminate, we run XG-CBS with an initial bound of $\infty$. We remark that whenever an algorithm times out without a solution, we do not include this in the computation time.
Our benchmarks were on grid worlds with the following sizes and number of agents:
$9\times 9$ with $4$, $8$, $10$, and $12$ agents, $16\times 16$ with $5$, $10$, $15$, and $20$ agents, and $33\times 33$ with $10$, $20$, and $30$ agents. For each grid size and agent number combination, we ran $100$ unique experiments. The timeout for a single algorithm on a single benchmark was 5 minutes (while this may seem like a high threshold, recall that Problem~\ref{problem:expMAPF} is computationally harder than MAPF).
The results are partially presented in Figures \ref{fig:benchmark_8_15} and \ref{fig:benchmark_33}. We refer to Section~\ref{sec:append} for the full set of benchmark results. For $33\times 33$ environments, XG-$A^*$ and WXG-$A^*$ nearly always time out, and hence not evaluated.

For the most part, the results match our expectations: vanilla CBS offers the best tradeoff between plan length and computation time, but invariably outputs plans with high index. Of the two extremities $A^*$ and XG-$A^*$, the speed of $A^*$ allows it to eventually find smaller index plans than XG-$A^*$, with comparable path length. However, XG-$A^*$, being guided towards minimal index plans, often outputs a lower index plan initially (c.f., \emph{first} column). Moreover, as the environment becomes smaller and more congested ($9 \times 9$, 12 agents), XG-$A^*$ outperforms $A^*$. 
% In such environments, WXG-$A^*$ has the best performance both in success rate and plan index.
Unfortunately, the history-dependence of XG-$A^*$ means that it does not scale to larger environments, and times out. 
In particular, this rules out the use of WXG-$A^*$, which is also history-dependent, for larger environments.
\begin{figure}[t]
    \centering
    \includegraphics[width=0.49\linewidth, align=t]{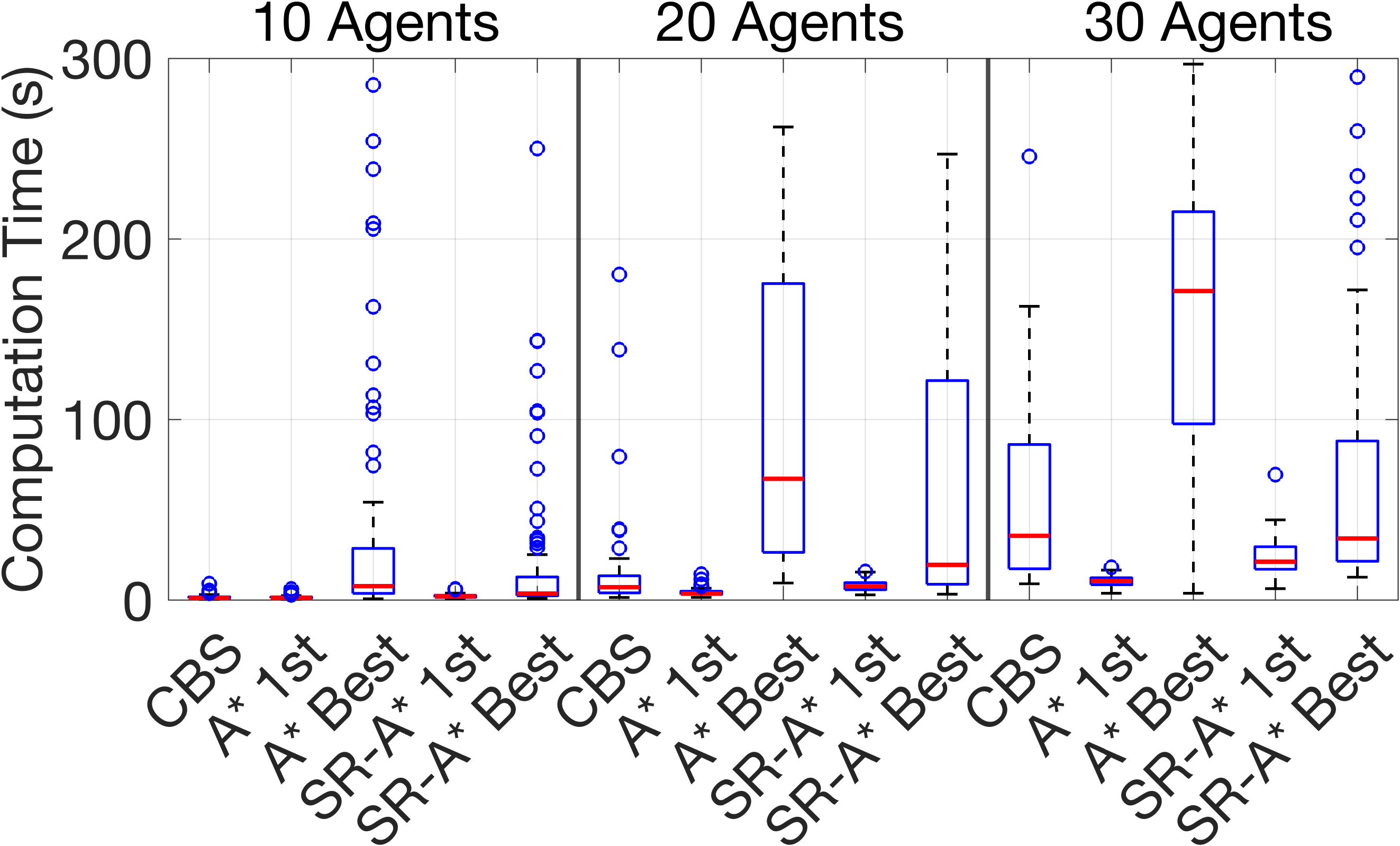}
    \hfill
    \includegraphics[width=0.49\linewidth, align=t]{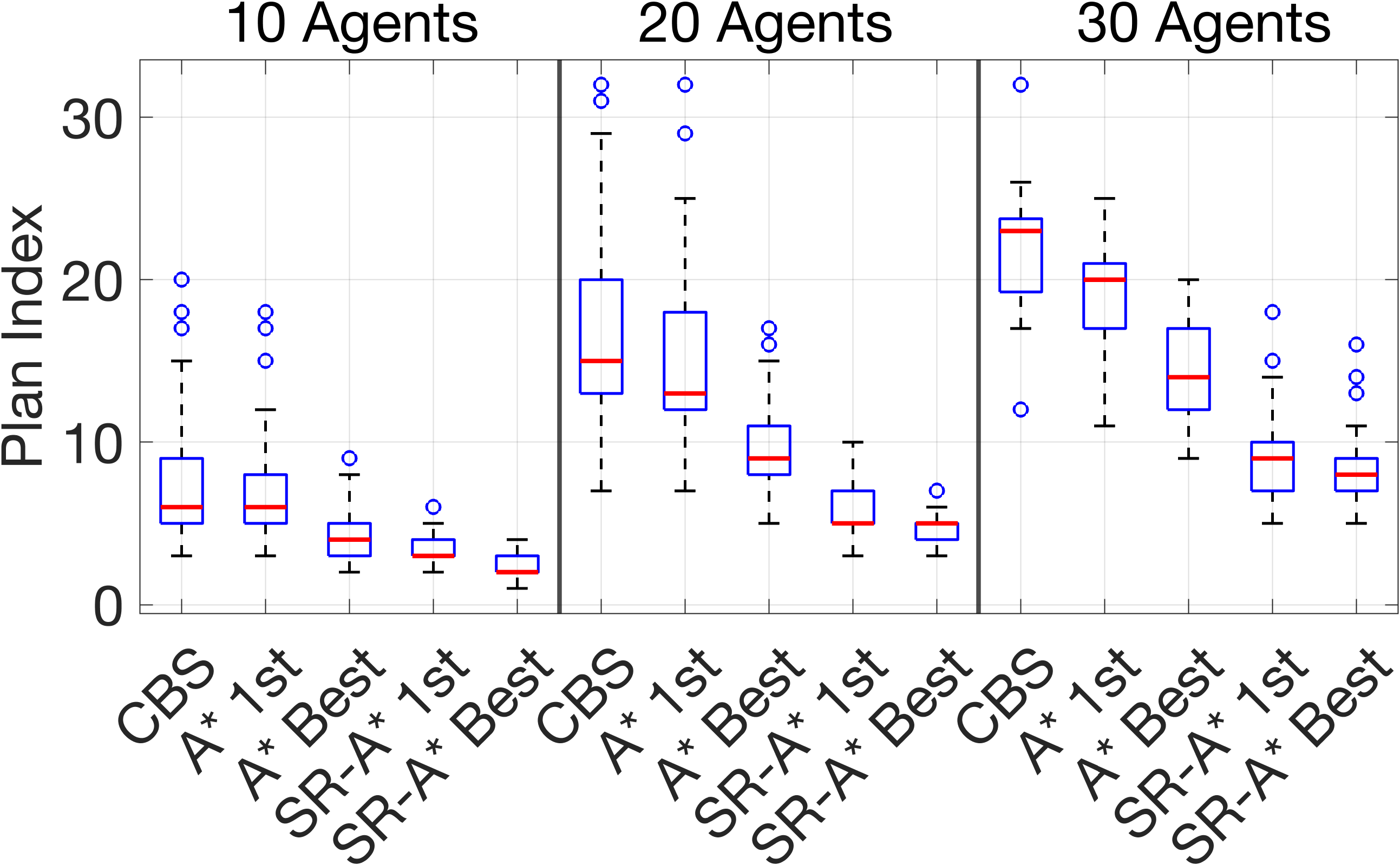}\vspace{1mm} \\
    \includegraphics[width=0.49\linewidth, align=t]{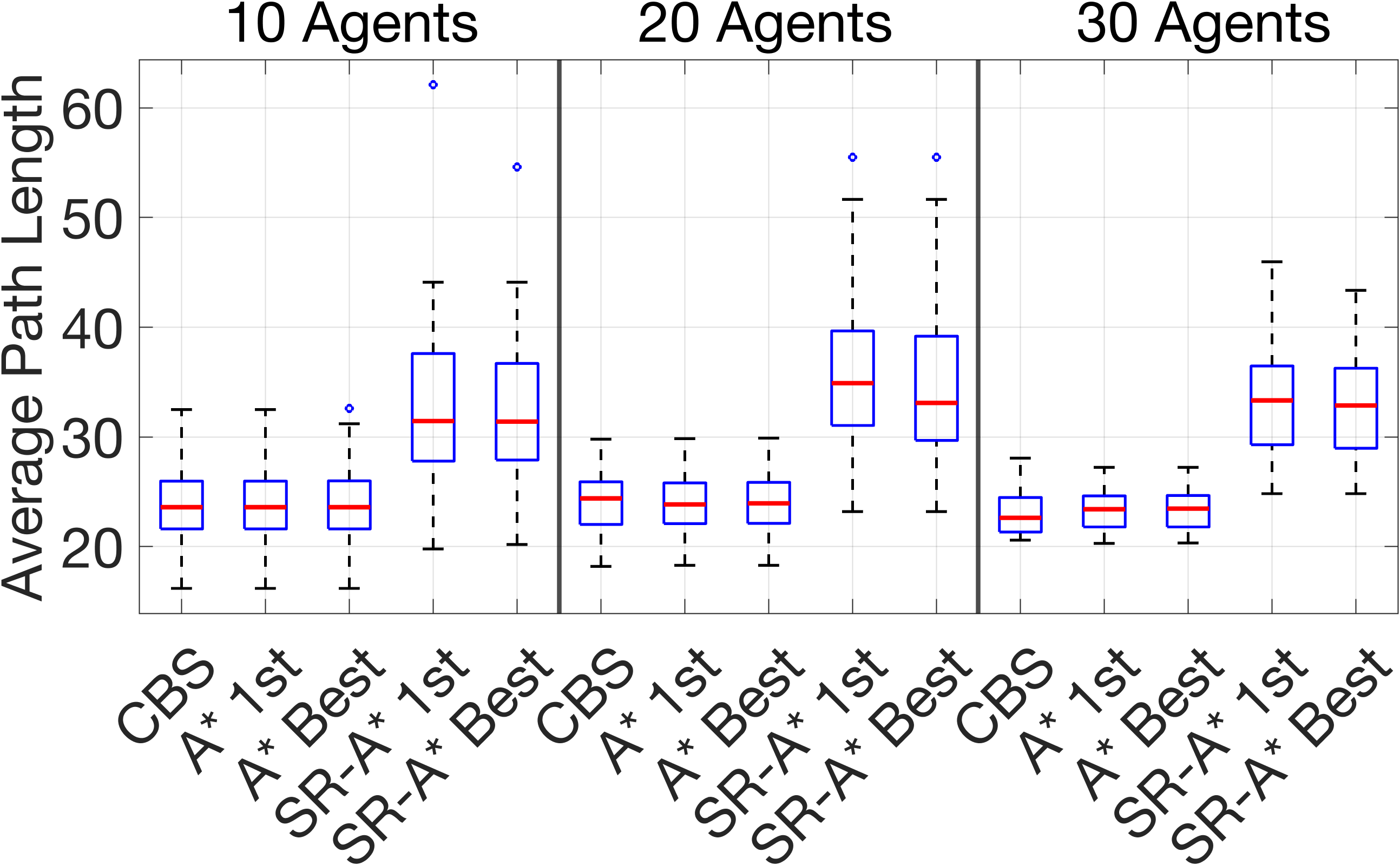}
    \hfill
    \includegraphics[width=0.49\linewidth, align=t]{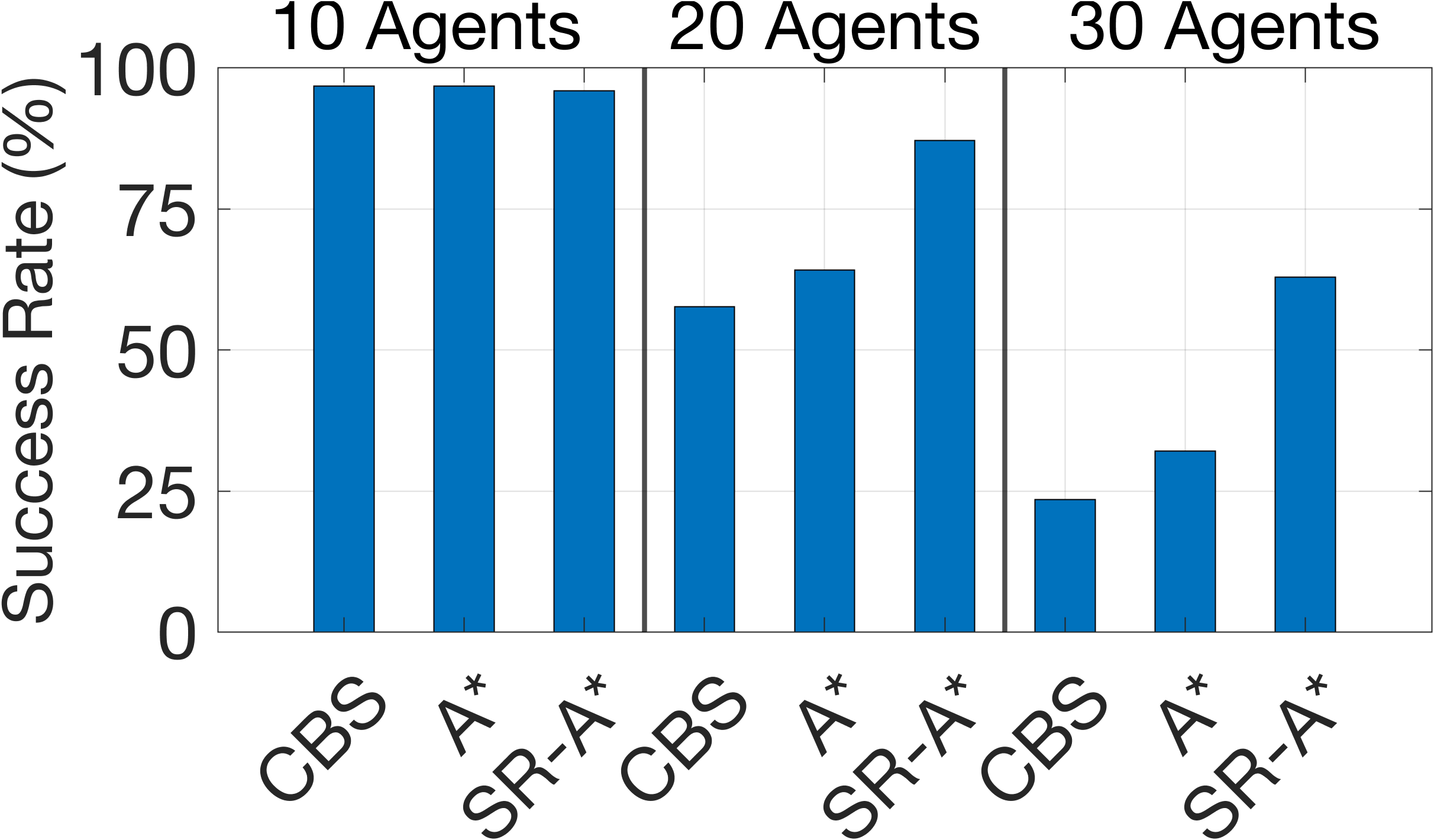}
    \vspace{-1mm}
    \caption{Benchmark results for $33 \times 33$ environment.}
    \label{fig:benchmark_33}
    % \vspace{-4mm}
\end{figure}
% \begin{figure}
%     \centering
%     \includegraphics[width=0.95\linewidth]{Figures/Plots_8x8_combined_3_1/Plan Index.png}
%     \includegraphics[width=0.95\linewidth]{Figures/Plots_15x15_combined_3_1/Success Rate.png}
%     \caption{Benchmark results for  $9\times 9$ environments.}
%     \label{fig:plan_index_9x9}
% \end{figure}
% \begin{figure}
%     \centering
%     \includegraphics[width=0.95\linewidth]{Figures/Plots_15x15_combined_3_1/Plan Index.png}
%     \includegraphics[width=0.95\linewidth]{Figures/Plots_8x8_combined_3_1/Success Rate.png}
%     % \begin{subfigure}{1\linewidth}
%     %      \centering
%     %      \includegraphics[width=1\linewidth]{Figures/Plots_8x8_combined_3_1/Plan Index.png}
%     %      \caption{$9\times 9$}
%     %      \label{fig:plan_index_9x9}
%     %  \end{subfigure}
%     %  \begin{subfigure}{1\linewidth}
%     %      \centering
%     %      \includegraphics[width=1\linewidth]{Figures/Plots_8x8_combined_3_1/Success Rate.png}
%     %      \caption{$16\times 16$}
%     %      \label{fig:Plan_index_16x16}
%     %  \end{subfigure}
%     \caption{Benchmark results for  $16\times 16$ environments.}
%     \label{fig:plan_index_15x15}
% \end{figure}

The surprising results come from SR-$A^*$. Despite being theoretically incomplete, in practice it offers an excellent success-rate (matching CBS), and invariably reduces the index of the plan (compared to CBS) by a significant amount (e.g., for $33 \times 33$, 30 agents, the reduction is from roughly 24 segments to 6 segments!). Moreover, it's limited search space allowes it to match CBS in computation time, and sometimes even outperform it. The tradeoff, naturally, comes in the path length, which increases.

Another pleasant surprise comes from $A^*$, which despite the expected increase in computation time, does manage to give some decrease in the index, even on larger environments. Moreover, since $A^*$ uses the distance to the goal as a  heuristic, the plans found by $A^*$ are typically shorter than those of SR-$A^*$ (but usually have a higher cost, since lowering the cost eventually causes it to time out).

On smaller environments, WXG-$A^*$ sometimes finds smaller index plans than XG-$A^*$ on the first try. In addition, it has a higher success rate (often higher than all other versions, including CBS) due to being guided in part by $A^*$. However, since finding the best weight parameter is instance-dependent (c.f., \cref{subsec:weighted}), it is not clear whether batch experiments capture the performance of WXG-$A^*$. 

To summarize the results, on larger environments, if one wishes to optimize explainability, then SR-$A^*$ is a clear winner. On smaller environments, $A^*$ usually works fairly well, but XG-$A^*$ can offer smaller index on congested environments. Finally, by carefully tuning a combined weight (e.g., by trying different options), one can obtain better explanations in small environments using WXG-$A^*$.

\section{Discussion and Future Work}
\label{sec:discussion}
In this paper, we introduced a CBS-based decentralized algorithm for Explainable MAPF via Segmentation. %, based on an adaptation of CBS.
%We have taken the first steps toward a scalable and explainable MAPF implementation, by which a human supervisor can gain insight and trust in automated plans.
Our technical contribution is twofold: first, we describe the extension XG-CBS, which can be readily implemented on top of existing CBS implementations. Second, we describe new low-level search algorithms, namely XG-$A^*$, WXG-$A^*$ and SR-$A^*$ oriented toward low index plans. While the former two yield a complete algorithm, we study their efficacy and show they do not scale well. The latter, despite not being complete, scales well and is often as efficient as CBS, while yielding easily explainable plans.

%We also study the efficacy of using XG-$A^*$ versus standard $A^*$ in XG-CBS. 
%Our experimental results show that for large environments, using $A^*$ 
%as a low-level planner 
%is often preferable. Nonetheless, on some instances, we do get better plans using XG-$A^*$. 
%In future research, we aim at characterizing the instances where one planner is preferable to the other. 
In future research, we will adapt other MAPF algorithms to the explainable settings, such as Priority-Based Search~\cite{ma2019searching}, and SAT-based solutions, as well as extensions and improvements of ``vanilla'' CBS.
%\jk{AAAI reviewer claimed tethered UAVs were not a good application for this work.}
%\shtodo{Right, but that reviewer was an idiot.}
Finally, we remark that Explainable MAPF via Segmentation has potential applications beyond gaining trust. Indeed, disjoint decompositions can be used during the actual execution of the plan, in case the agents' paths must not cross. 
%For example,  consider robots \ml{these robots are called tethered robots} hooked to e.g., electrical sockets. If we want to keep their cables from entwining, we could occasionally disconnect and plug them to a closer socket. However, given a disjoint segmentation, we know the optimal times to do so. 
%\ml{how about this:  
For example in tethered robots (e.g., tethered UAVs), we wish to 
%avoid 
minimize tangling of the tethers as they constrain robot motion. This essentially means we desire plans with a small index, which XG-CBS enables us to achieve. Similarly, in applications of MAPF to 3D pipe routing~\cite{belov2020multi}, small segmentations may allow for simpler routing.
%}
Other applications can be found in multi-layered circuit board design. 
In particular, such applications show that even the reduction of one segment from the index may have beneficial financial applications, which may be significant.

%\section{Acknowledgments}
\bibliography{main} %bibliography,

\begin{thebibliography}{28}
\providecommand{\natexlab}[1]{#1}

\bibitem[{Almagor and Lahijanian(2020)}]{Almagor:AAMAS:2020}
Almagor, S.; and Lahijanian, M. 2020.
\newblock Explainable Multi Agent Path Finding.
\newblock In \emph{Proceedings of the 19th International Conference on
  Autonomous Agents and MultiAgent Systems}, AAMAS '20, 34--42. Richland, SC:
  International Foundation for Autonomous Agents and Multiagent Systems.
\newblock ISBN 9781450375184.

\bibitem[{Arrieta et~al.(2020)Arrieta, D{\'\i}az-Rodr{\'\i}guez, Del~Ser,
  Bennetot, Tabik, Barbado, Garc{\'\i}a, Gil-L{\'o}pez, Molina, Benjamins
  et~al.}]{arrieta2020explainable}
Arrieta, A.~B.; D{\'\i}az-Rodr{\'\i}guez, N.; Del~Ser, J.; Bennetot, A.; Tabik,
  S.; Barbado, A.; Garc{\'\i}a, S.; Gil-L{\'o}pez, S.; Molina, D.; Benjamins,
  R.; et~al. 2020.
\newblock Explainable Artificial Intelligence (XAI): Concepts, taxonomies,
  opportunities and challenges toward responsible AI.
\newblock \emph{Information Fusion}, 58: 82--115.

\bibitem[{Bartak, Svancara, and Vlk(2018)}]{JSvan18b}
Bartak, R.; Svancara, J.; and Vlk, M. 2018.
\newblock A Scheduling-Based Approach to Multi-Agent Path Finding with Weighted
  and Capacitated Arcs.
\newblock In \emph{Proceedings of the International Joint Conference on
  Autonomous Agents and Multiagent Systems (AAMAS)}, 748--756.

\bibitem[{Belov et~al.(2020)Belov, Du, De~La~Banda, Harabor, Koenig, and
  Wei}]{belov2020multi}
Belov, G.; Du, W.; De~La~Banda, M.~G.; Harabor, D.; Koenig, S.; and Wei, X.
  2020.
\newblock From multi-agent pathfinding to 3D pipe routing.
\newblock In \emph{Thirteenth Annual Symposium on Combinatorial Search}.

\bibitem[{Bose and Markelov(2019)}]{mapfBenchmarks}
Bose, A.; and Markelov, I. 2019.
\newblock Multi-Agent Path Planning in Python.
\newblock \url{https://github.com/atb033/multi_agent_path_planning}.

\bibitem[{Boyarski et~al.(2015)Boyarski, Felner, Stern, Sharon, Tolpin,
  Betzalel, and Shimony}]{boyarski2015icbs}
Boyarski, E.; Felner, A.; Stern, R.; Sharon, G.; Tolpin, D.; Betzalel, O.; and
  Shimony, E. 2015.
\newblock ICBS: Improved conflict-based search algorithm for multi-agent
  pathfinding.
\newblock In \emph{Twenty-Fourth International Joint Conference on Artificial
  Intelligence}.

\bibitem[{Brandao et~al.(2021)Brandao, Canal, Krivi{\'c}, Luff, and
  Coles}]{brandao2021experts}
Brandao, M.; Canal, G.; Krivi{\'c}, S.; Luff, P.; and Coles, A. 2021.
\newblock How experts explain motion planner output: a preliminary user-study
  to inform the design of explainable planners.
\newblock In \emph{2021 30th IEEE International Conference on Robot \& Human
  Interactive Communication (RO-MAN)}, 299--306. IEEE.

\bibitem[{Cohen et~al.(2018)Cohen, Koenig, Kumar, Wagner, Choset, Chan, and
  Sturtevant}]{SKoen18k}
Cohen, L.; Koenig, S.; Kumar, S.; Wagner, G.; Choset, H.; Chan, D.; and
  Sturtevant, N. 2018.
\newblock Rapid Randomized Restarts for Multi-Agent Path Finding: Preliminary
  Results.
\newblock In \emph{Proceedings of the International Joint Conference on
  Autonomous Agents and Multiagent Systems (AAMAS)}, 1909--1911.

\bibitem[{Eifler et~al.(2019)Eifler, Cashmore, Jorg, Magazzeni, and
  Steinmetz}]{eifler_cashmore_hoffmann}
Eifler, R.; Cashmore, M.; Jorg, H.; Magazzeni, D.; and Steinmetz, M. 2019.
\newblock Explaining the Space of Plans through Plan-Property Dependencies.
\newblock \emph{Proceedings of the 2nd Workshop on Explainable Planning
  (XAIP)}.

\bibitem[{Felner et~al.(2018)Felner, Li, Boyarski, Ma, Cohen, Kumar, and
  Koenig}]{felner2018adding}
Felner, A.; Li, J.; Boyarski, E.; Ma, H.; Cohen, L.; Kumar, T.~S.; and Koenig,
  S. 2018.
\newblock Adding heuristics to conflict-based search for multi-agent path
  finding.
\newblock In \emph{Proceedings of the International Conference on Automated
  Planning and Scheduling}, volume~28.

\bibitem[{Felner et~al.(2017)Felner, Stern, Shimony, Goldenberg, Sharon,
  Sturtevant, Wagner, and Surynek}]{AFeln17c}
Felner, A.; Stern, R.; Shimony, E.; Goldenberg, M.; Sharon, G.; Sturtevant, N.;
  Wagner, G.; and Surynek, P. 2017.
\newblock Search-Based Optimal Solvers for the Multi-Agent Pathfinding Problem:
  Summary and Challenges.
\newblock In \emph{Proceedings of the Symposium on Combinatorial Search
  (SoCS)}, 28--37.

\bibitem[{Fines, Sharpanskykh, and Vert(2020)}]{fines2020agent}
Fines, K.; Sharpanskykh, A.; and Vert, M. 2020.
\newblock Agent-based distributed planning and coordination for resilient
  airport surface movement operations.
\newblock \emph{Aerospace}, 7(4): 48.

\bibitem[{Fox, Long, and Magazzeni(2017)}]{fox2017explainable}
Fox, M.; Long, D.; and Magazzeni, D. 2017.
\newblock Explainable Planning.
\newblock arXiv:1709.10256.

\bibitem[{Hubel and Wiesel(1959)}]{Hubel}
Hubel, D.~H.; and Wiesel, T.~N. 1959.
\newblock Receptive fields of single neurones in the cat's striate cortex.
\newblock \emph{The Journal of Physiology}, 148(3).

\bibitem[{Kambhampati(2019)}]{10.5555/3306127.3331663}
Kambhampati, S. 2019.
\newblock Synthesizing Explainable Behavior for Human-AI Collaboration.
\newblock In \emph{Proceedings of the 18th International Conference on
  Autonomous Agents and MultiAgent Systems}, 1–2. Richland, SC: International
  Foundation for Autonomous Agents and Multiagent Systems.
\newblock ISBN 9781450363099.

\bibitem[{Kottinger(2021)}]{sourceCode}
Kottinger, J. 2021.
\newblock Explanation-Guided Conflict-Based Search for Explainable MAPF.
\newblock \url{https://github.com/aria-systems-group/Explanation-Guided-CBS}.

\bibitem[{Kottinger, Almagor, and Lahijanian(2021)}]{kottinger2021mapsx}
Kottinger, J.; Almagor, S.; and Lahijanian, M. 2021.
\newblock MAPS-X: Explainable Multi-Robot Motion Planning via Segmentation.
\newblock In \emph{Proceedings of the IEEE International Conference on Robotics
  and Automation}. Xi'an, China: IEEE.

\bibitem[{Lapuschkin et~al.(2019)Lapuschkin, Wäldchen, Binder, Montavon,
  Samek, and Müller}]{Lapuschkin_2019}
Lapuschkin, S.; Wäldchen, S.; Binder, A.; Montavon, G.; Samek, W.; and
  Müller, K.-R. 2019.
\newblock Unmasking Clever Hans predictors and assessing what machines really
  learn.
\newblock \emph{Nature Communications}, 10(1).

\bibitem[{Li et~al.(2019{\natexlab{a}})Li, Felner, Boyarski, Ma, and
  Koenig}]{li2019improved}
Li, J.; Felner, A.; Boyarski, E.; Ma, H.; and Koenig, S. 2019{\natexlab{a}}.
\newblock Improved Heuristics for Multi-Agent Path Finding with Conflict-Based
  Search.
\newblock In \emph{IJCAI}, volume 2019, 442--449.

\bibitem[{Li et~al.(2019{\natexlab{b}})Li, Harabor, Stuckey, Felner, Ma, and
  Koenig}]{li2019disjoint}
Li, J.; Harabor, D.; Stuckey, P.~J.; Felner, A.; Ma, H.; and Koenig, S.
  2019{\natexlab{b}}.
\newblock Disjoint splitting for multi-agent path finding with conflict-based
  search.
\newblock In \emph{Proceedings of the International Conference on Automated
  Planning and Scheduling}, volume~29, 279--283.

\bibitem[{Ma et~al.(2019{\natexlab{a}})Ma, Harabor, Stuckey, Li, and
  Koenig}]{SKoen19a}
Ma, H.; Harabor, D.; Stuckey, P.; Li, J.; and Koenig, S. 2019{\natexlab{a}}.
\newblock Searching with Consistent Prioritization for Multi-Agent Path
  Finding.
\newblock In \emph{Proceedings of the AAAI Conference on Artificial
  Intelligence (AAAI)}, (in print).

\bibitem[{Ma et~al.(2019{\natexlab{b}})Ma, Harabor, Stuckey, Li, and
  Koenig}]{ma2019searching}
Ma, H.; Harabor, D.; Stuckey, P.~J.; Li, J.; and Koenig, S. 2019{\natexlab{b}}.
\newblock Searching with consistent prioritization for multi-agent path
  finding.
\newblock In \emph{Proceedings of the AAAI Conference on Artificial
  Intelligence}, 7643--7650.

\bibitem[{Mari, Dang, and G{\"o}ssler(2021)}]{mari2021explaining}
Mari, T.; Dang, T.; and G{\"o}ssler, G. 2021.
\newblock Explaining Safety Violations in Real-Time Systems.
\newblock In \emph{International Conference on Formal Modeling and Analysis of
  Timed Systems}, 100--116. Springer.

\bibitem[{Sharon et~al.(2015)Sharon, Stern, Felner, and
  Sturtevant}]{SHARON201540}
Sharon, G.; Stern, R.; Felner, A.; and Sturtevant, N.~R. 2015.
\newblock Conflict-based search for optimal multi-agent pathfinding.
\newblock \emph{Artificial Intelligence}, 219: 40--66.

\bibitem[{Standley(2010)}]{standley2010finding}
Standley, T.~S. 2010.
\newblock Finding optimal solutions to cooperative pathfinding problems.
\newblock In \emph{Twenty-Fourth AAAI Conference on Artificial Intelligence}.

\bibitem[{Stern et~al.(2019)Stern, Sturtevant, Atzmon, Walker, Li, Cohen, Ma,
  Kumar, Felner, and Koenig}]{stern2019mapf}
Stern, R.; Sturtevant, N.~R.; Atzmon, D.; Walker, T.; Li, J.; Cohen, L.; Ma,
  H.; Kumar, T. K.~S.; Felner, A.; and Koenig, S. 2019.
\newblock Multi-Agent Pathfinding: Definitions, Variants, and Benchmarks.
\newblock \emph{Symposium on Combinatorial Search (SoCS)}, 151--158.

\bibitem[{Surynek et~al.(2016)Surynek, Felner, Stern, and Boyarski}]{AFeln16c}
Surynek, P.; Felner, A.; Stern, R.; and Boyarski, E. 2016.
\newblock An Empirical Comparison of the Hardness of Multi-Agent Path Finding
  under the Makespan and the Sum of Costs Objectives.
\newblock In \emph{Proceedings of the Symposium on Combinatorial Search
  (SoCS)}, 145--147.

\bibitem[{Tang et~al.(2018)Tang, Lee, Li, Zhang, Xu, Liu, Teo, and
  Jiang}]{Tang}
Tang, S.; Lee, T.~S.; Li, M.; Zhang, Y.; Xu, Y.; Liu, F.; Teo, B.; and Jiang,
  H. 2018.
\newblock Complex pattern selectivity in macaque primary visual cortex revealed
  by large-scale two-photon imaging.

\end{thebibliography}
% \appendix
\section{Appendix}
\label{sec:append}
We now present supplementary material that provides additional insights to the behavior of our proposed algorithms presented in the paper. We begin by outlining the algorithm for the high-level of XG-CBS. Then, we present a more extensive examples section to further prove the efficacy of our proposed explanation scheme and algorithms. We conclude with an extensive table that shows all the experiments performed by our team, further validating the claims made in the paper. 
\subsection{XG-CBS Algorithm}
\label{sec:xg-cbs-alg}
Given an Explainable MAPF instance consisting of a graph $G$, a list of source $(s_i)_{i=1}^n=(s_1, \ldots, s_n)$, a list of goal vertices $(g_i)_{i=1}^n=(g_1, \ldots, g_n)$,
% \ml{I've never seen this notation before. I've seen $\{s_i\}_{i=1}^n$, which is a set, not a list.}\jk{I changed it to generic symbols that are different enough from $G$.}
% \shtodo{I think Morteza means that you need either curly brackets (for a set) or round ones, for an ordered list, which is more suitable in this case.}
% \jk{The problem is that simply using $(s_1, \ldots, s_n)$ and $(g_1, \ldots, g_n)$ does not fit in the algorithm caption. I need a shorthand variable so that it fits into the algorithm. I added round brackets to make it an ordered list. However, I think $\mathcal{S}$ and $\mathcal{G}$ should stay for the algorithm.}
an index bound $r$, and a path length bound $B$, XG-CBS proceeds as follows. 

First, a root node $R$ is initialized with an empty set of constraints $\cC$. Then, the low-level planner is called to find a path for each agent. If the graph search fails to generate a root plan $P_r$ then XG-CBS returns no solution. If, however, a full plan is found, then it is saved in $R$ along with all other important information and and added to the priority queue $Q$. 

While the queue is not empty, XG-CBS selects the highest priority node $N$, removes it from the queue, and evaluates its plan $N.plan$ for a conflict $(a_i, a_j, v_i, v_j, T_i, T_j)$ between Agent $i$ at $(v_i, T_i)$ and Agent $j$ at $(v_j, T_j)$. Note that segmentation conflicts are included in this definition by letting $v_i=v_j$.

If no conflicts exist for a given nodes plan, then it is returned as the solution. Otherwise, for every agent in the conflict, a new node $K$ is added with a new constraint $(a_i, v_i, T_i)$, and the newly constrained agent is re-planned for using low level graph search. If successful, the new plan and all its information is added to K before it is added $Q$, where it will eventually be evaluated for conflicts.

\begin{algorithm}[t]
    
    R.$\cC$, Q, $P_r \leftarrow$ $\emptyset$\;
    \For{every agent }
    {
        $P_r$.add( graphSearch($G$, $s_i$, $g_i$, R.$\cC$, $P_r$, $B$) )
    }
    \If{$P_r$.size() $< n$}
    {\KwRet{no solution}}
    R.plan, R.index, R.cost $\leftarrow P_r$\;
    Q.add(R)\;
    \While{Q not empty}
    {
    N $\leftarrow$ Q.highestPriority()\;
    Q.pop(N); c $\leftarrow$ conflictCheck(N.plan, r)\;
    \If{$c$ is empty}
    {\KwRet{N.plan}}
    \For{every agent $a_i \in c$}
    {
    K.$\cC \leftarrow $ N.$\cC \cup (a_i, v, T_i)$\;
    $P_{-i} \leftarrow K.plan \setminus \pi_i$ \;
    $\pi_i \leftarrow$ graphSearch($G$, $s_i$, $g_i$, K.$\cC$, $P_{-i}$, $B$)\;
    \If{$\pi_i$ exists}
    {
    $P_{new} \leftarrow P_{-i} \cup \pi_i$\;
    K.plan, K.index, K.cost $\leftarrow P_{new}$\;
    Q.add(K)\;
    }
    }
    }
    \caption{XG-CBS$(G, (s_i)_{i=1}^n, (g_i)_{i=1}^n, r, B)$}
    \label{alg:xg-cbs}
\end{algorithm}

Algorithm~\ref{alg:xg-cbs} outlines the pseudocode for XG-CBS. Note that the \emph{graphSearch($\cdot$)} procedure only utilizes the initial plan $P_r$ or existing $P_{-i}$ as the chosen low level planner specifies. For example, XG-$A^*$ uses the existing plan as outlined in the paper. However, using $A^*$ only segments the plan after $\pi_i$ is found but before \emph{graphSearch($\cdot$)} returns it.
\subsection{Completeness of XG-CBS}
\label{sec:proof}
%%%%%%% Begin Proof %%%%%%
We now present the proof of Theorem~\ref{thm:XGCBS_complete}.
\begin{proof}
% \shtodo{If we're short on space, let's move the proof to the appendix. It's not exciting, and suspicious reviewers are welcome to read it.}
The proof is a small variation on the completeness proof of standard CBS. Consider a solvable instance of Explainable MAPF, namely $G=\tup{V,E}$, lists $s_1,\ldots,s_n$ and $g_1,\ldots,g_n$, and a bound $r$. Since this instance is solvable, there exists a plan $P=\{\pi_1,\ldots,\pi_n\}$ with index at most $r$.

We obtain from the plan $P$ a maximal set $\cC_{\max}$ of constraints for all the agents, by adding, for each agent $i$, every constraint the prevents agent $i$ from being at vertex $v$ at time $t$, for every $v,t$ such that $\pi_i[t,t]\neq v$. Intuitively, the only paths allowed under $\cC_{\max}$ prescribe $P$ exactly. 

We claim that as long as no solution is found by XG-CBS, there exists a branch in the constraint tree whose set of constraints is a subset of $\cC_{\max}$, and that this path has an unexplored node. 
This is easily proved by induction on the constraint tree: the root node does not have any constraints, and $\emptyset\subseteq \cC_{\max}$. If the root is not a solution, then it has children obtained by conflicts. This completes the base case. For the induction step, consider the aforementioned branch, and consider the unexplored leaf node. If the plan represented in the leaf is not a solution, then it has children obtained by new constraints. We claim that at least one of these children adds a constraint from $\cC_{\max}$. Indeed, $P$ does not have any conflicts, and so any conflict must produce at least two constraints, one of which is not in $\cC_{\max}$ (otherwise, $\cC_{\max}$ allows a conflict, which is a contradiction). So we are done.%We conclude the induction.

Therefore, unless XG-CBS terminates with a solution earlier, at least one branch will be expanded toward $P$. To complete the argument, we observe that the constraint tree of XG-CBS is bounded, since the lengths of the plans are bounded. It follows that every branch will eventually be explored. In particular, $P$ will be reached.

Finally, if the instance is not solvable, then eventually every possible constraint is placed, and the constraint tree is no longer updated, thus terminating the search.
\end{proof}
%%%%%%% End Proof %%%%%%
%%%%%%%%%%%%%%%%%%%%%%%%%%%%%%%%%%%
\subsection{XG-$A^*$ Algorithm and Speedups.}
\label{sec:xg-astar-alg}
% \subsubsection{Algorithms}
The algorithm for XG-$A^*$ is shown in Algorithm~\ref{alg:eg-Astar}. Below, we present some basic observations that help mitigate the limitation presented in Remark~4.
\begin{algorithm}[t]
    Q $\leftarrow \{s_i\}$\; %Closed $\leftarrow \{\}$\;
    \While{Q not empty}
    {
        c $\leftarrow$ Q.highestPriority()\;
        \eIf{c.loc = $g_i$} 
        {
            \KwRet{$\pi_i \leftarrow$ c.Path()}
        }
        {
            Q.pop(c); N $\leftarrow$ expand(c, $G$, $\cC$, $B$)\;
            \For{every n $\in$ N}
            {
                n.index $\leftarrow$ Segment(n.Path(), $P_{-i}$)\;
                \eIf{n.index $\leq$ n.parent.index}
                {
                    n.gScore $\leftarrow$ n.parent.gScore$+1$\;
                    Q.add(n)\;
                }
                {
                    \If{n.parent.gScore$+1\leq$n.gScore}
                    {
                        n.gScore $\leftarrow$ n.parent.gScore$+1$\;
                        Q.add(n)\;
                    }
                }
            }
        }
    }
    \caption{XG-$A^*(G, s_i, g_i, \cC, P_{-i}, B)$}
    \label{alg:eg-Astar}
\end{algorithm}
\subsubsection{Speeding Up XG-\texorpdfstring{$A^*$}{A*}}
\label{subsec:heuristics}
% \shtodo{Move to appendix, add instead section about new incomplete method, and the weighting of $A*$ and $XG-A^*$.}
As demonstrated in Remark~4, XG-$A^*$ may spend a lot of time exhausting the plans of a certain index before making any progress. As we now show, we can alleviate some of the computational cost, using simple observations.

\paragraph{Eliminating Cycles}
Assume %for the following 
that the graph $G$ allows agents to stay in place, i.e., has self-loops. This is typical in, e.g., warehouse robots. Now, consider a plan $P=\{\pi_1,\ldots,\pi_n\}$ such that in its vertex-disjoint decomposition, Agent $i$ makes a cycle that is contained entirely within a certain segment. Since paths within a segment are disjoint, we can eliminate this cycle and replace it with Agent $i$ waiting in place for the duration of the cycle. Moreover, we can shift this waiting by a wait in the initial vertex of the segment.

Thus, we observe that any plan can be put in a ``normal form'' 
%replaced with a similar plan 
where within each segment, no agent makes a cycle, and staying in place is allowed only on the initial vertex. We use this to speed up XG-$A^*$ by 
limiting the search space 
%enforcing the searched plans 
to comply with this condition: it is easy to check whether an agent has a cycle within a segment (except for looping in the initial vertex),
%That is, the agent cannot revisit the same vertex within a segment (which is easily checked, 
as $H$, the history of the segment, is part of the information of each node.%), except for the initial vertex, where the agent can initially wait before starting to move.

\paragraph{Shortest Path after Segment Bound}
Recall that XG-$A^*$ exhausts all the plans with index $i$ before moving to index $i+1$. We propose a speed up, whereby when the index reaches the bound $\bar{r}$ (index of $P_{-1}$), the remaining search is performed using standard $A^*$, i.e., searching for the shortest path to the target, rather than exhausting the remaining plans. 
% \jk{should also be $w$ or whatever symbol we use above}
% \ml{then, it wouldn't be a very effective heuristic if $w$ is large and $r$ is small}
Technically, this modification retains the completeness of the algorithm, and hence Theorem~5 is still valid. Intuitively, this offers a speed up  since if we already reached an index beyond the given bound, it is unlikely that planning for the current agent helps to reduce segmentation. Therefore, we terminate the search as quickly as possible and allow for further exploration of the conflict tree. %While this heuristic is not supported by any theoretical guarantees, 
We find that empirically, this heuristic speeds up the algorithm. %, especially for small $r$.
\subsection{Extended Case Studies}

\subsubsection{Illustrative Examples (extended)}
We now further showcase XG-CBS in many settings. We begin by showing how XG-CBS XG-$A^*$ outperforms XG-CBS with $A^*$ for examples where the optimal explanation requires careful tuning of the plan. Then, we incrementally increase the space size, agent number, and solution complexity to show the capabilities of XG-CBS.

\begin{figure}[t]
     \centering
     \begin{subfigure}{0.49\linewidth}
         \centering
         \includegraphics[scale=0.3]{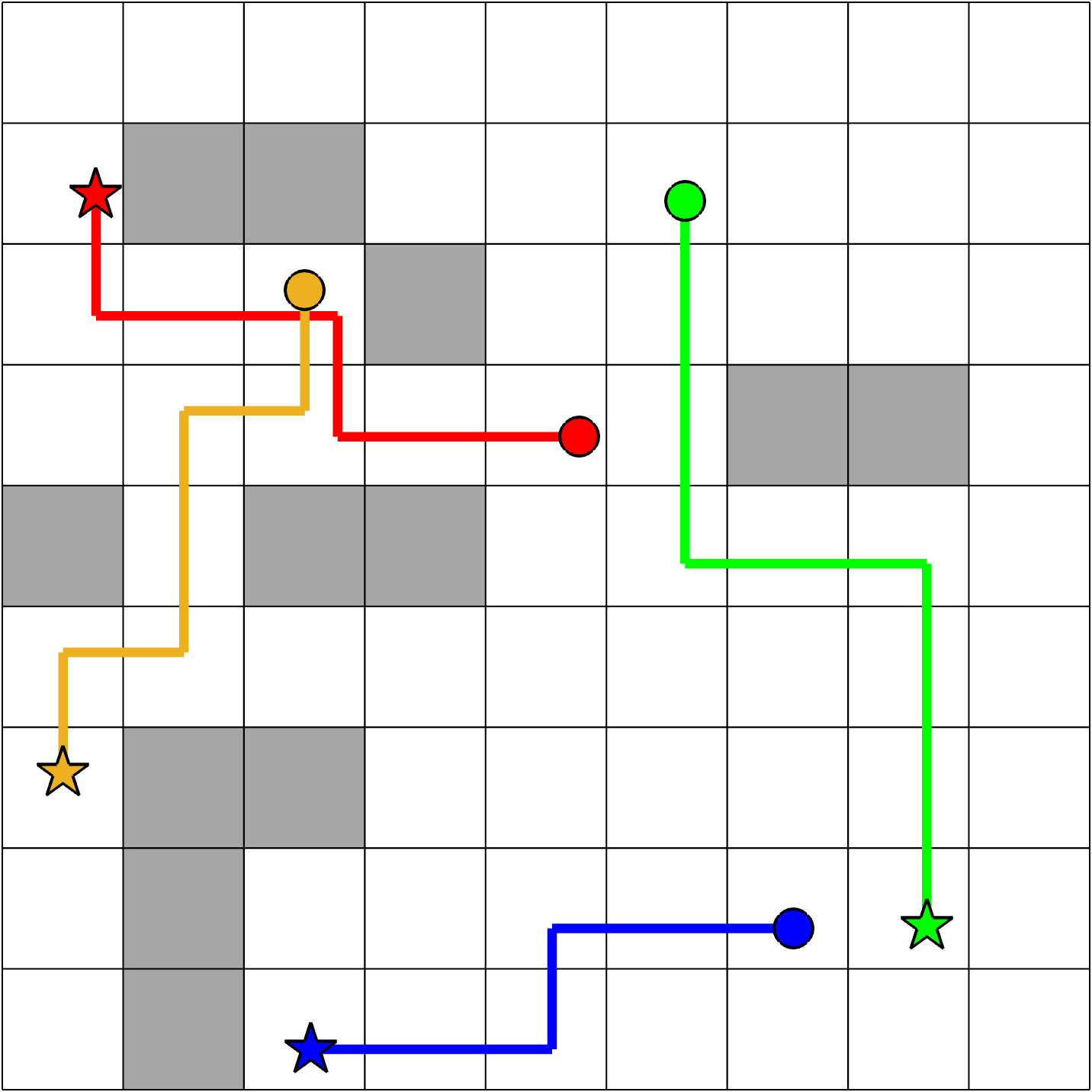}
         \caption{CBS}
         \label{fig:ln3_cbs_full}
     \end{subfigure}
     \hfill
     \begin{subfigure}{0.49\linewidth}
         \centering
         \includegraphics[scale=0.3]{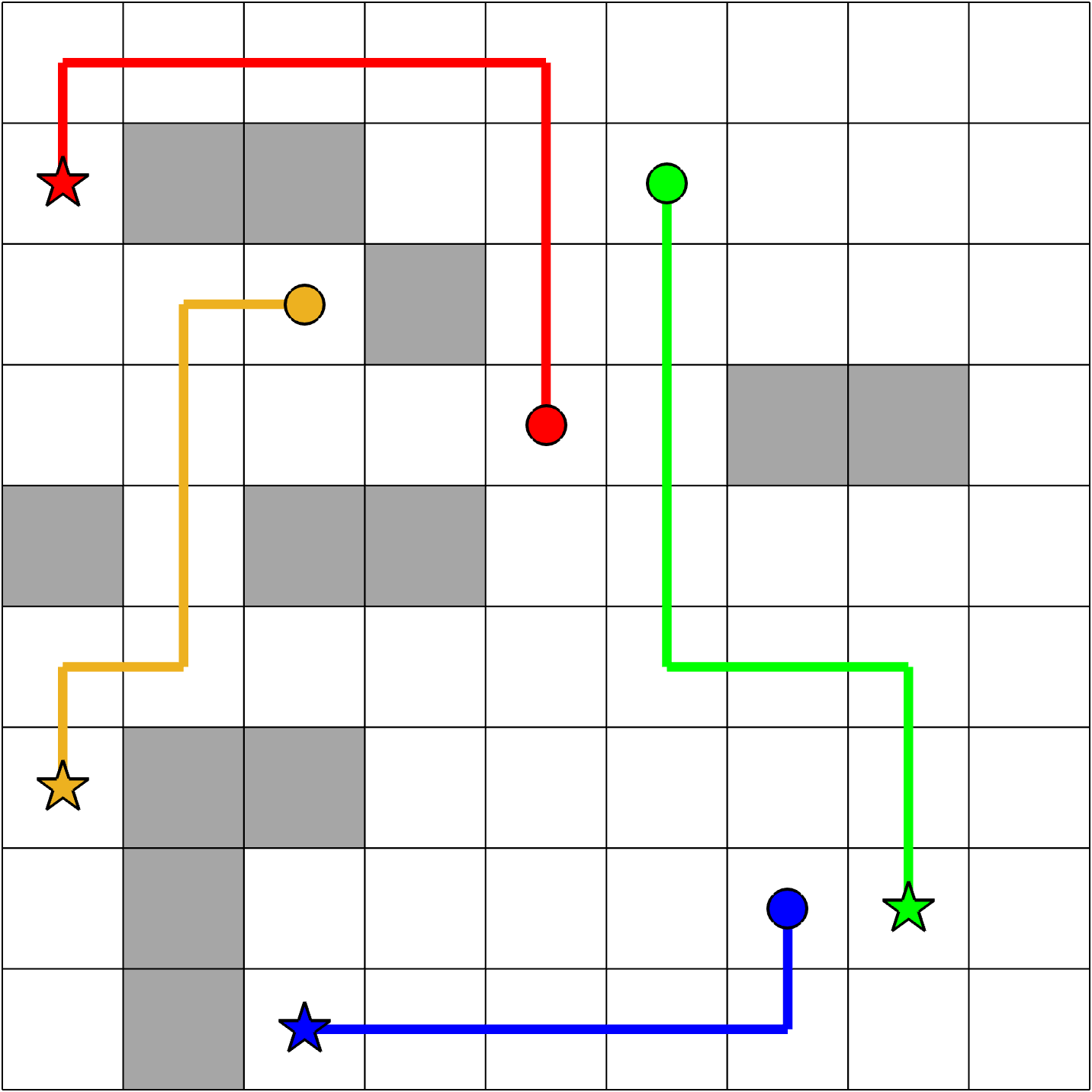}
         \caption{XG-CBS, $r=1$}
         \label{fig:ln3_xgcbs_full}
     \end{subfigure}
     \hfill
     \newline
     \begin{subfigure}{0.49\linewidth}
         \centering
         \includegraphics[scale=0.3]{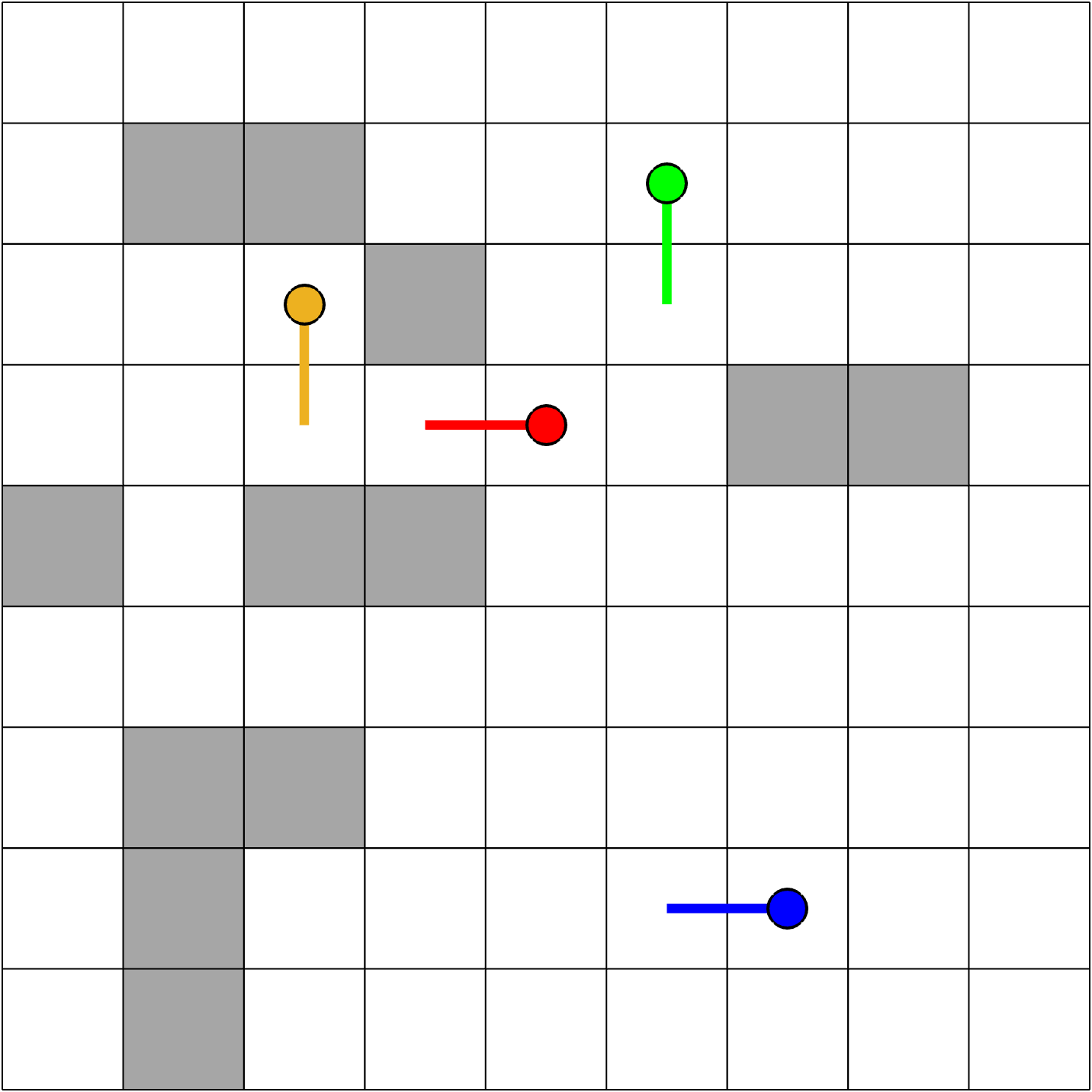}
         \caption{CBS $\Delta k=[0,1]$}
         \label{fig:ln3_cbs_seg1}
     \end{subfigure}
     \begin{subfigure}{0.49\linewidth}
         \centering
         \includegraphics[scale=0.3]{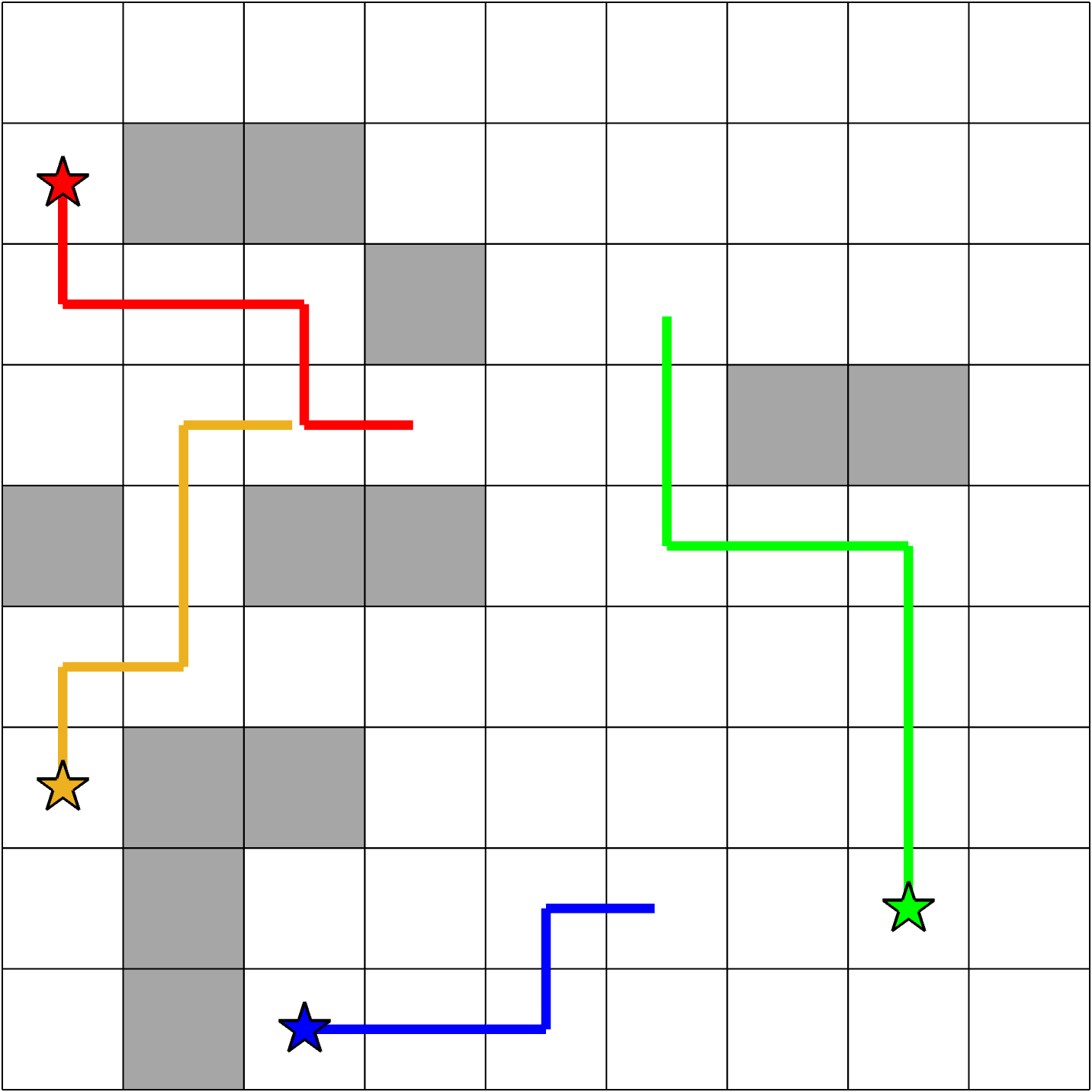}
         \caption{CBS $\Delta k=[1,8]$}
         \label{fig:ln3_cbs_seg2}
     \end{subfigure}
    % \caption{Line $16$ of Table~\ref{tab:final_benchmark}}
    \caption{Example of 1-segment solution with XG-CBS}
    \label{fig:ln_3} % numbers in labels represent row of Table 1 -- numbers in document should be from Table 2
\end{figure}

Figure~\ref{fig:ln3_cbs_full} shows an example of four agents in a $9\times 9$ grid world. Notice that the shortest plan results in a sub-optimal explanation. XG-CBS with XG-$A^*$ easily returns the optimal explanation shown in Figure~\ref{fig:ln3_xgcbs_full} in 0.04 seconds. Note that XG-CBS with $A^*$ finds a $2$-segment solution almost immediately but fails to return the preferred $1$-segment solution given a 15 minute planning time threshold. This is due to the increase in path length when attempting to untangle the red and yellow agents. 

\begin{figure}[t]
     \centering
     \begin{subfigure}{0.49\linewidth}
         \centering
         \includegraphics[scale=0.2]{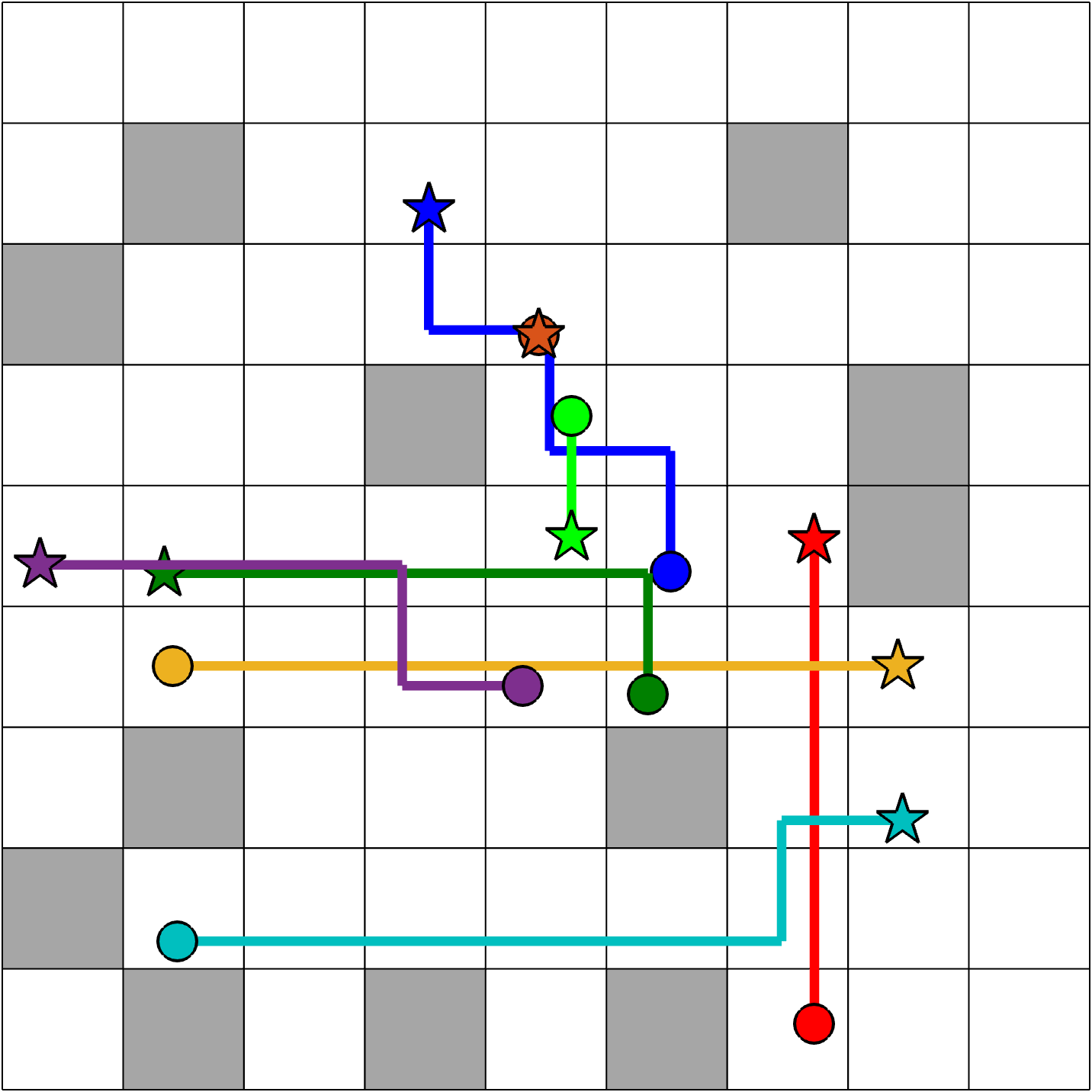}
         \caption{CBS}
         \label{fig:ln7_cbs_full}
     \end{subfigure}
     \hfill
     \begin{subfigure}{0.49\linewidth}
         \centering
         \includegraphics[scale=0.2]{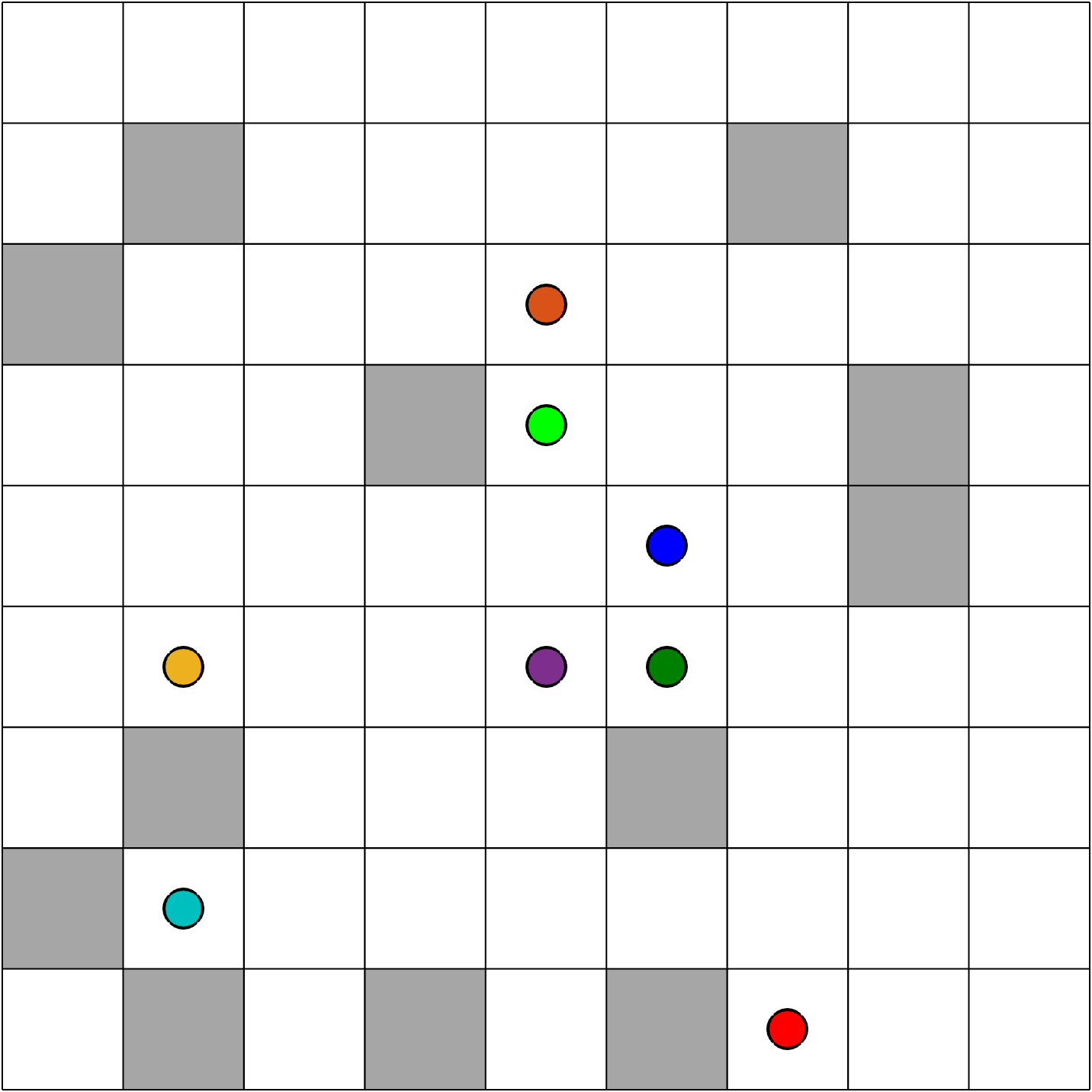}
         \caption{CBS, $\Delta k=[0,0]$}
         \label{fig:ln7_cbs_seg1}
     \end{subfigure}
     \hfill
     \newline
     \begin{subfigure}{0.49\linewidth}
         \centering
         \includegraphics[scale=0.2]{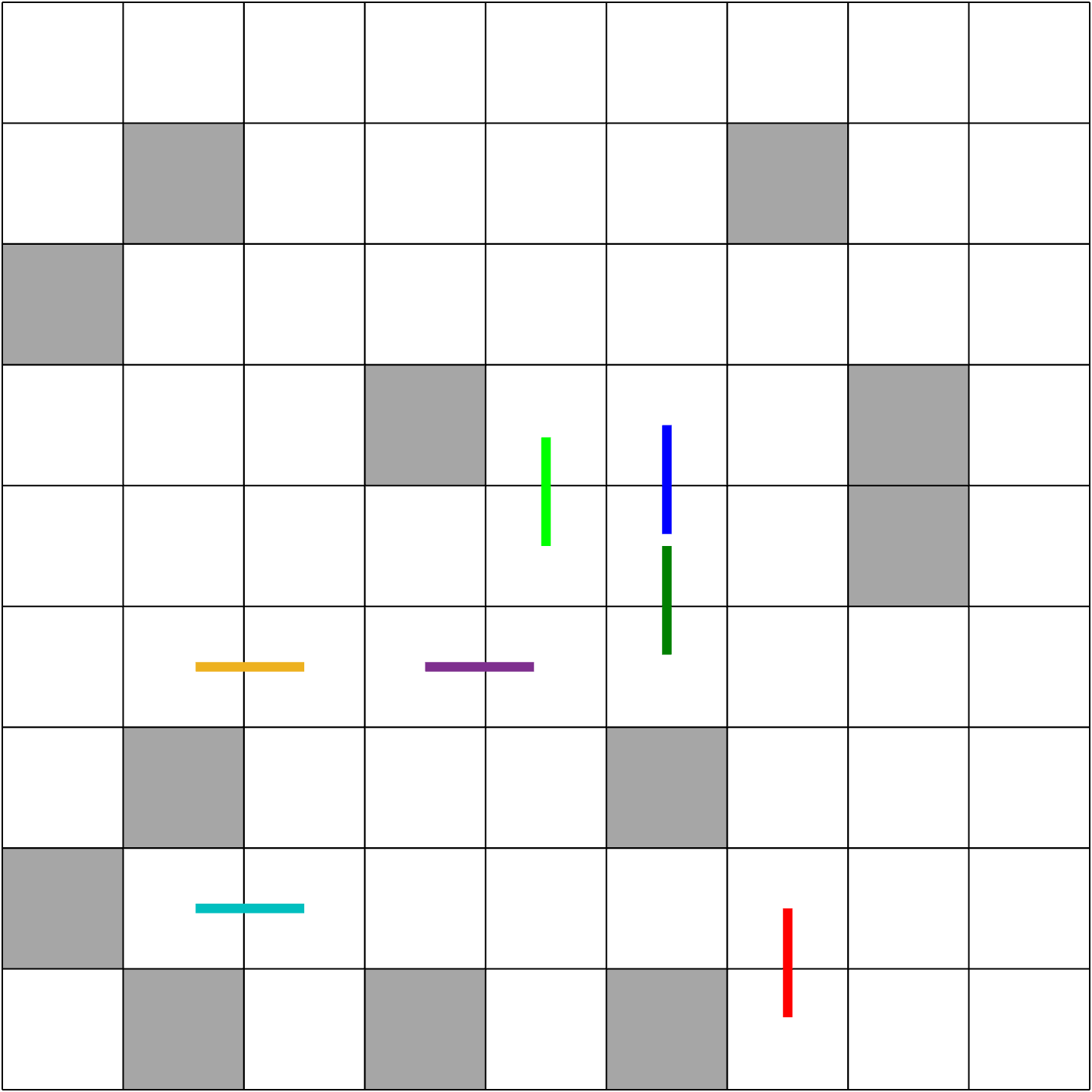}
         \caption{CBS, $\Delta k=[0,1]$}
         \label{fig:ln7_cbs_seg2}
     \end{subfigure}
     \hfill
     \begin{subfigure}{0.49\linewidth}
         \centering
         \includegraphics[scale=0.2]{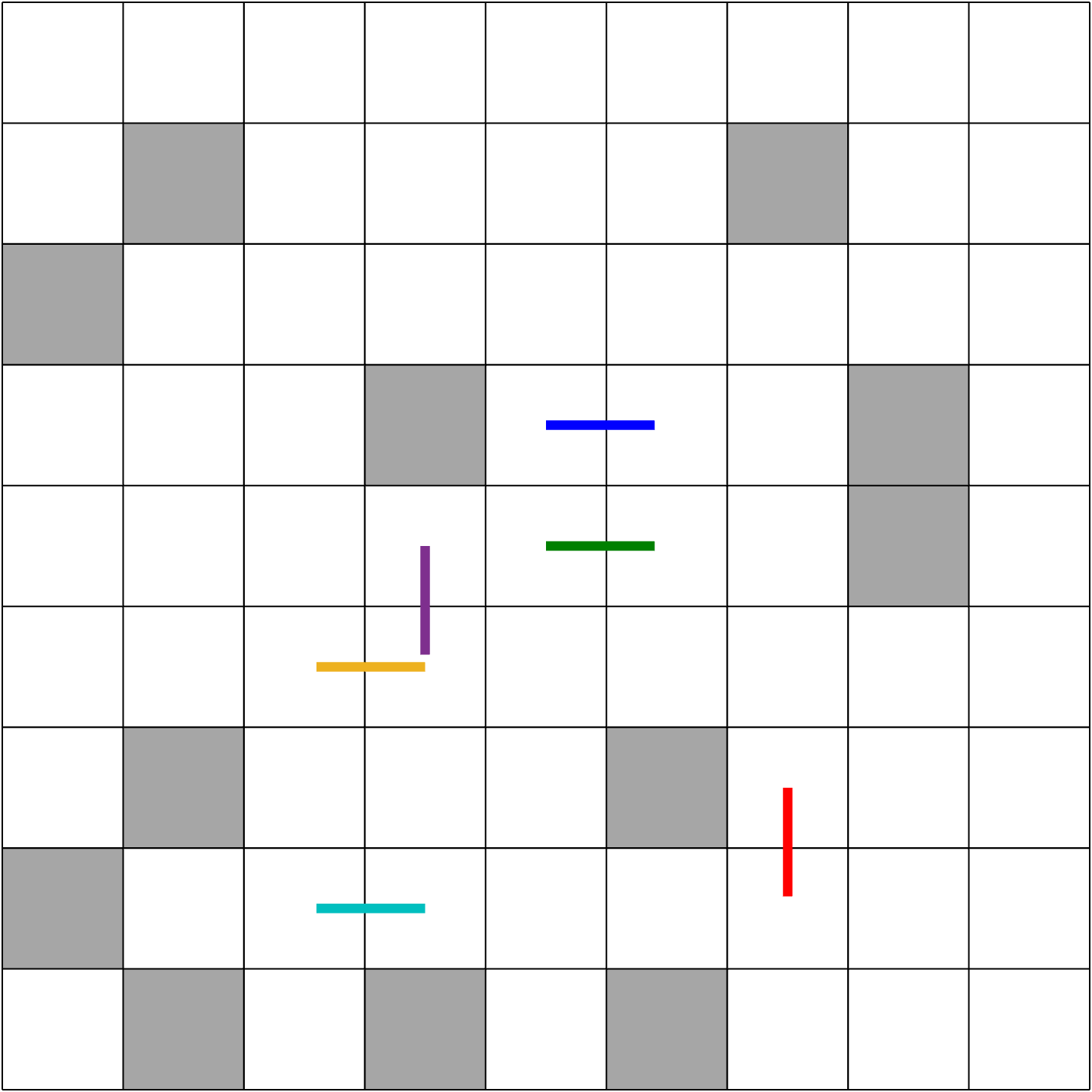}
         \caption{CBS, $\Delta k=[1,2]$}
         \label{fig:ln7_cbs_seg3}
     \end{subfigure}
     \hfill
     \newline
     \begin{subfigure}{0.49\linewidth}
         \centering
         \includegraphics[scale=0.2]{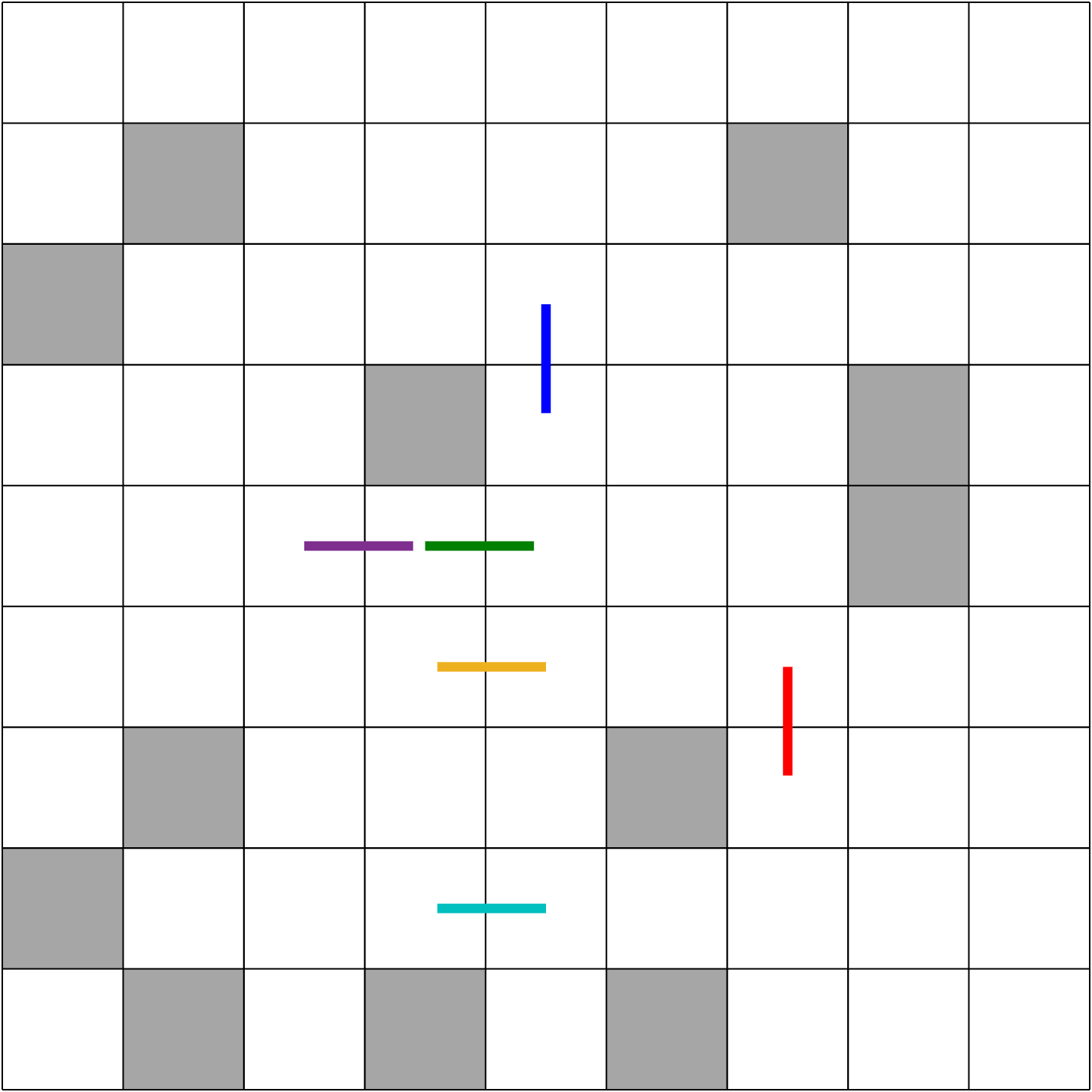}
         \caption{CBS, $\Delta k=[2,3]$}
         \label{fig:ln7_cbs_seg4}
     \end{subfigure}
     \hfill
     \begin{subfigure}{0.49\linewidth}
         \centering
         \includegraphics[scale=0.2]{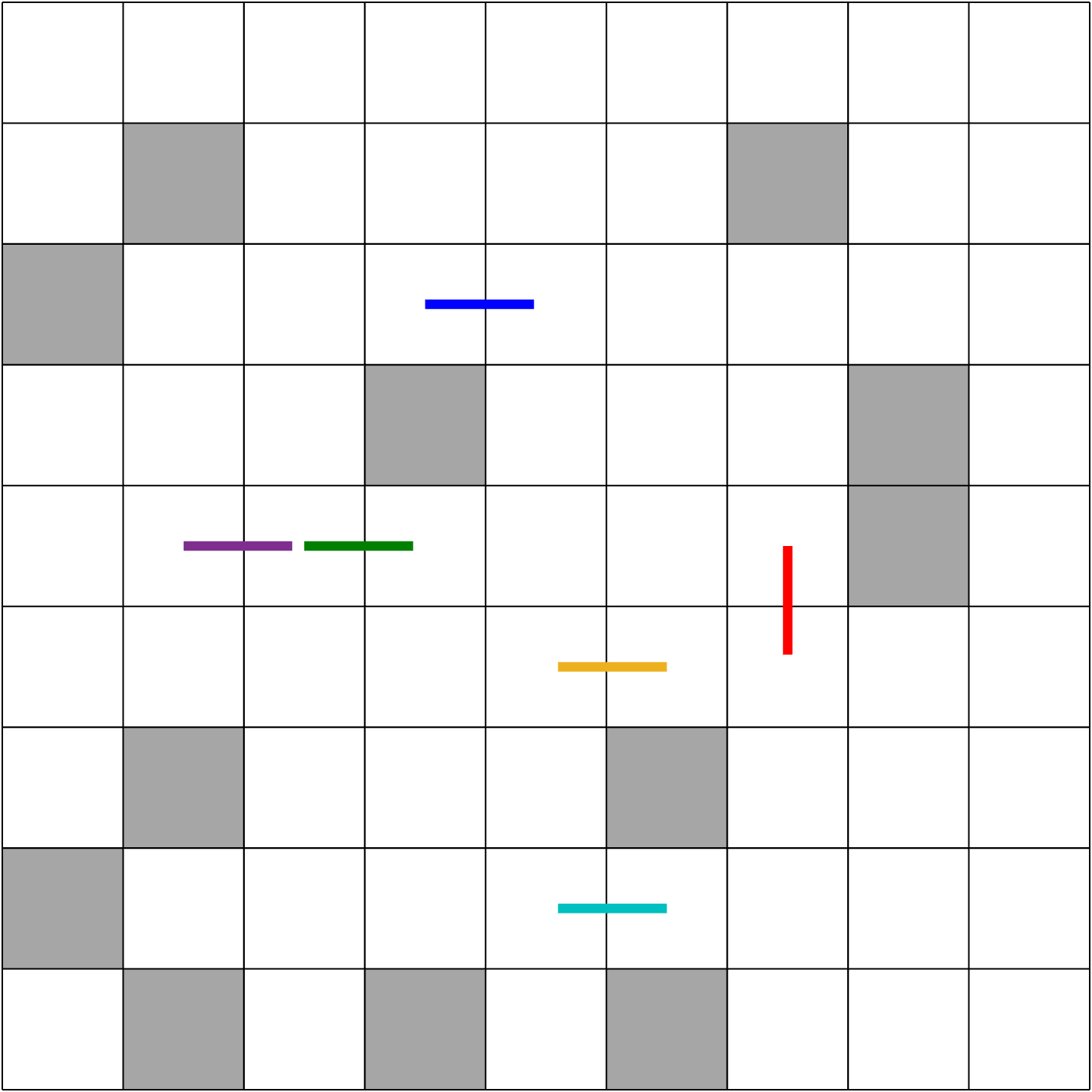}
         \caption{CBS, $\Delta k=[3,4]$}
         \label{fig:ln7_cbs_seg5}
     \end{subfigure}
     \hfill
     \newline
     \begin{subfigure}{0.49\linewidth}
         \centering
         \includegraphics[scale=0.2]{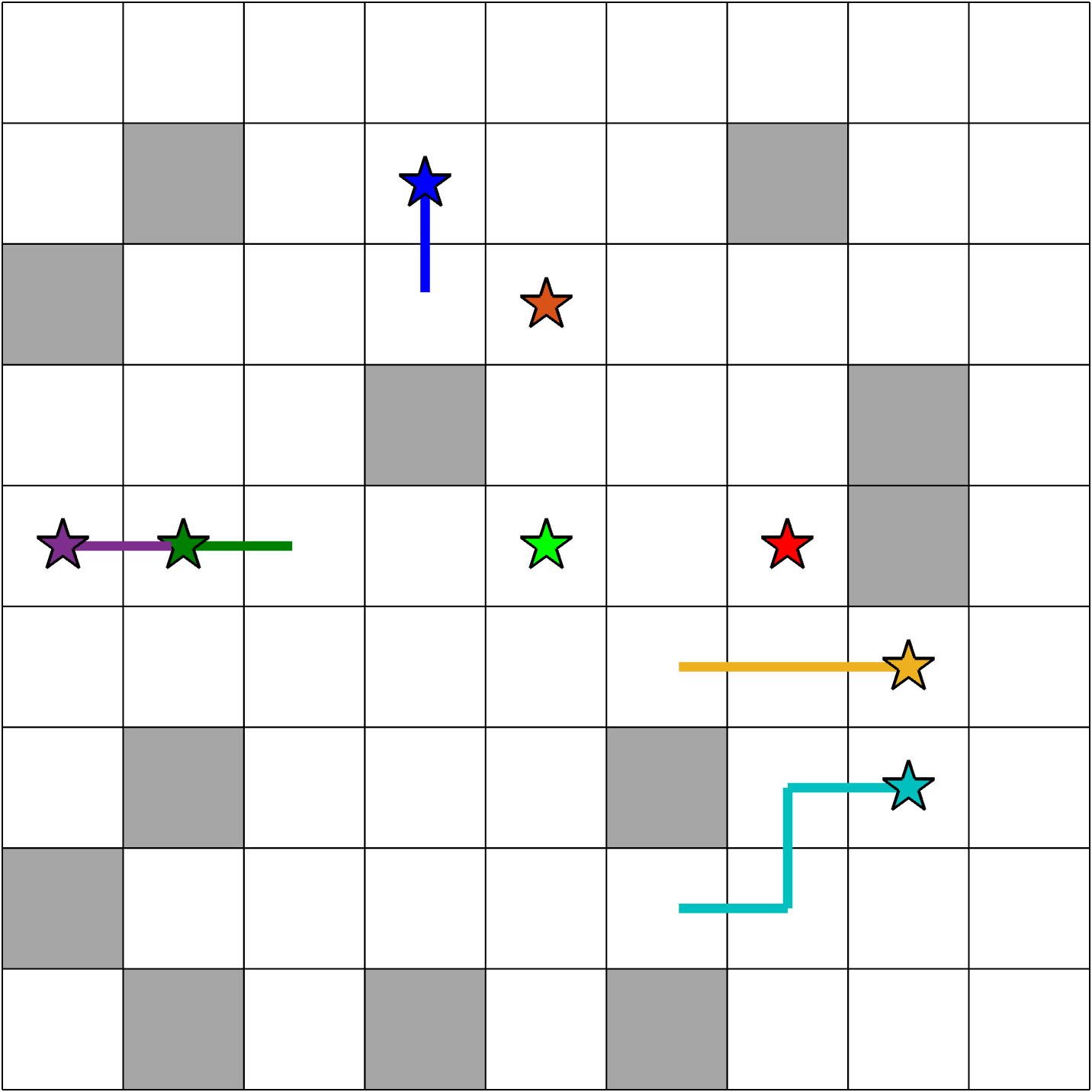}
         \caption{CBS, $\Delta k=[4,7]$}
         \label{fig:ln7_cbs_seg6}
     \end{subfigure}
     \hfill
     \begin{subfigure}{0.49\linewidth}
         \centering
         \includegraphics[scale=0.2]{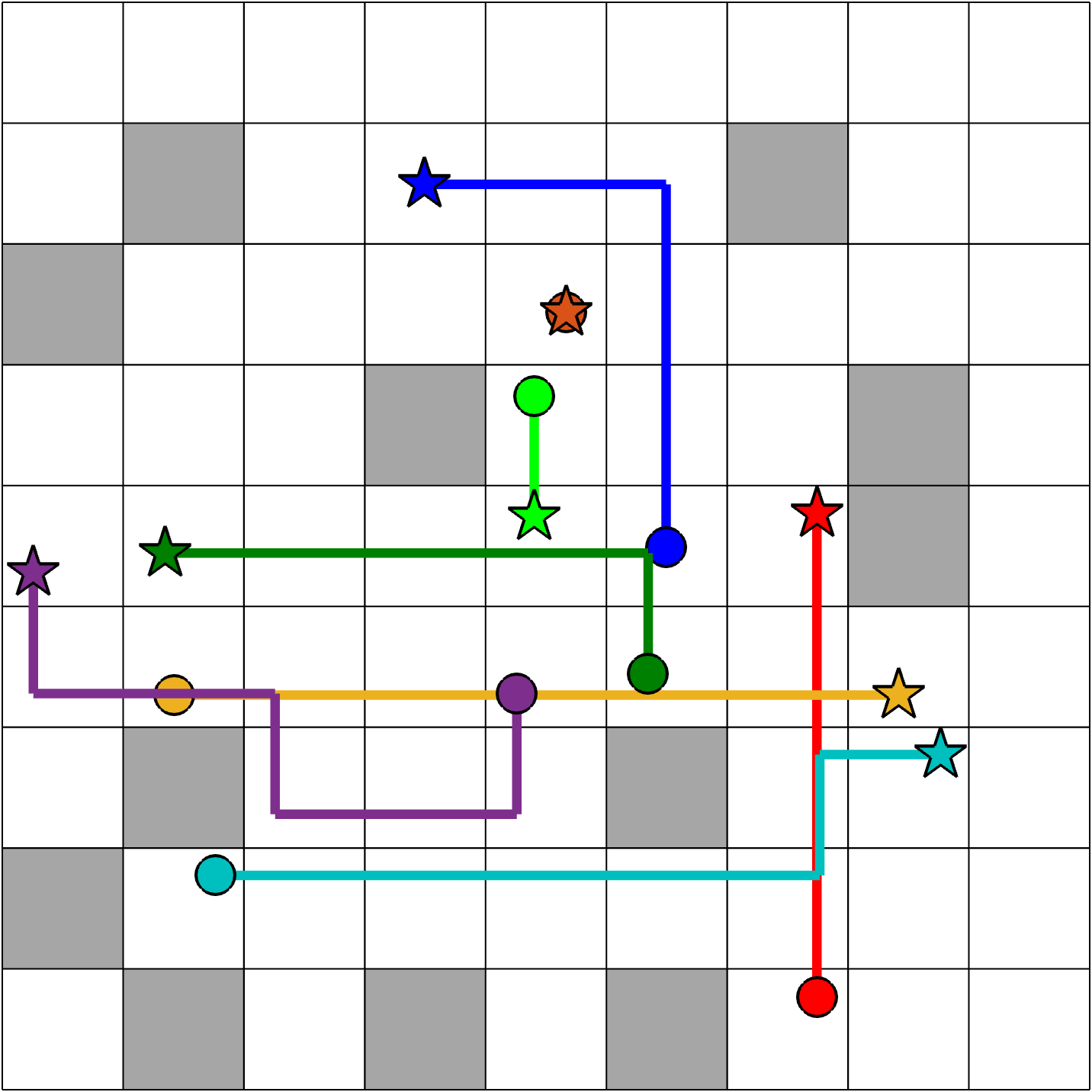}
         \caption{XG-CBS, $r=2$}
         \label{fig:ln7_xgcbs_full}
     \end{subfigure}
     \newline
     \begin{subfigure}{0.49\linewidth}
         \centering
         \includegraphics[scale=0.2]{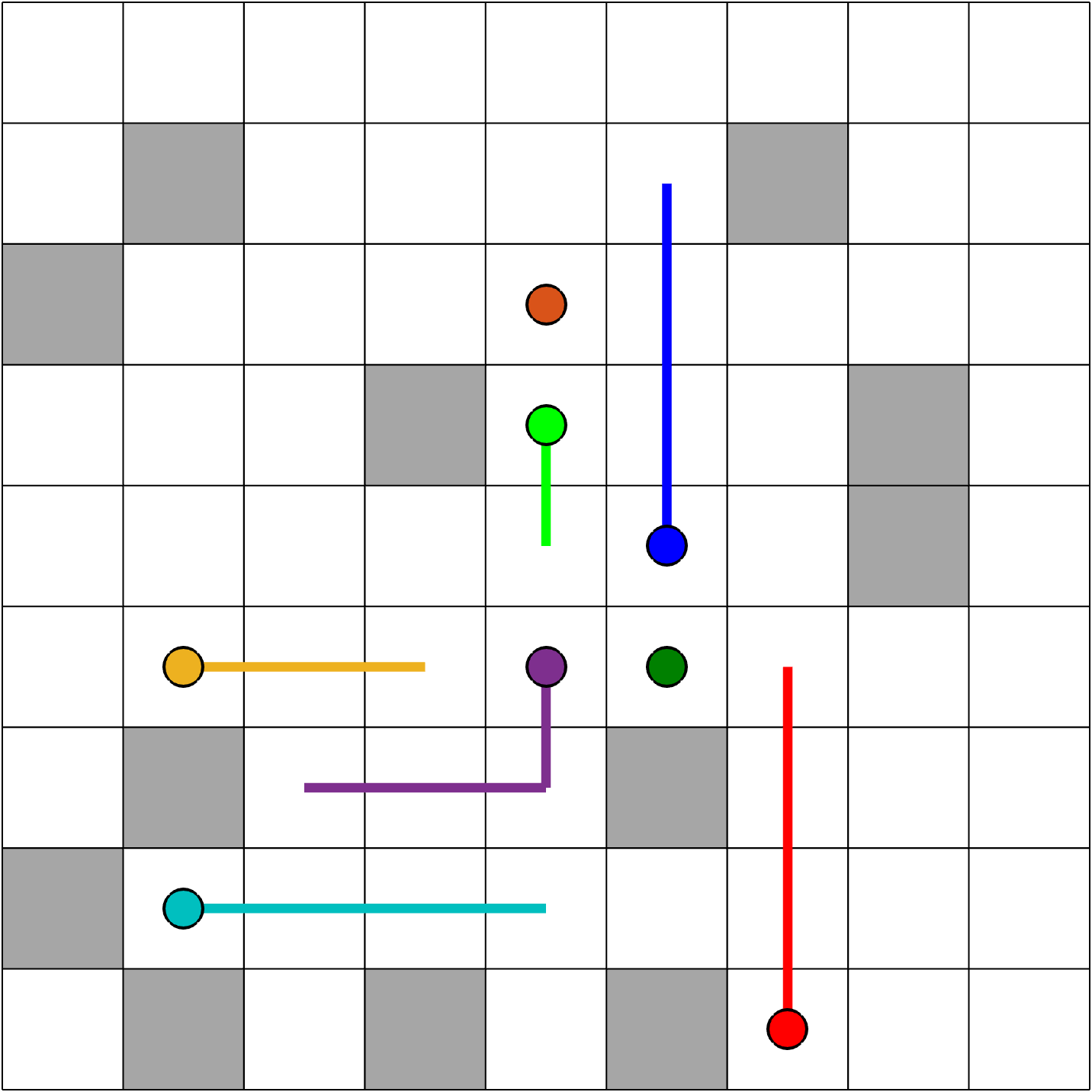}
         \caption{XG-CBS, $\Delta k=[0,3]$}
         \label{fig:ln7_xgcbs_seg1}
     \end{subfigure}
     \hfill
     \begin{subfigure}{0.49\linewidth}
         \centering
         \includegraphics[scale=0.2]{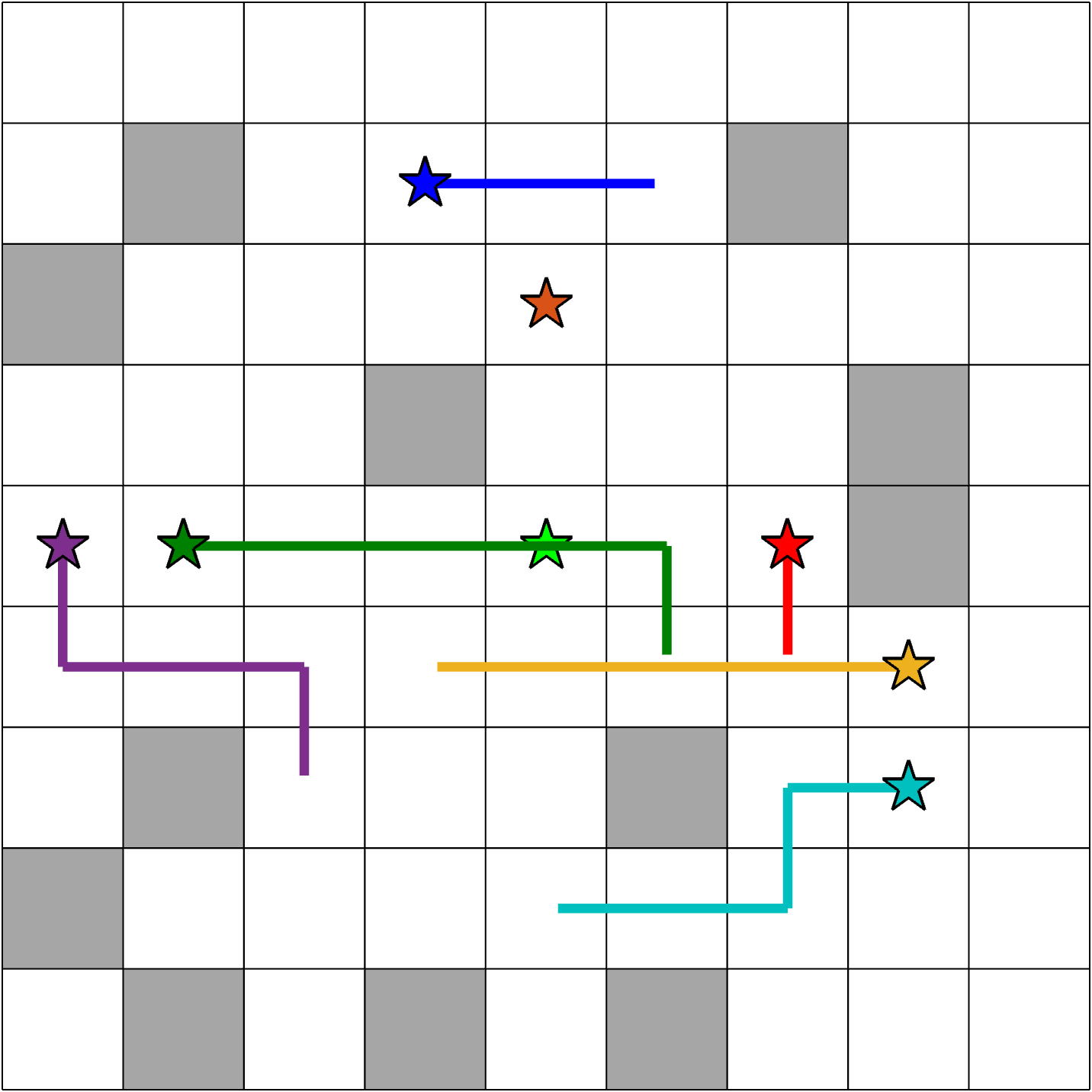}
         \caption{XG-CBS, $\Delta k=[3,8]$}
         \label{fig:ln7_xgcbs_seg2}
     \end{subfigure}
    % \caption{Line $20$ of Table~\ref{tab:final_benchmark}}
    \caption{Example of 2-segment solution with XG-CBS}
    \label{fig:ln_7} % numbers in labels represent row of Table 1 -- numbers in document should be from Table 2
\end{figure}

We now turn to a more interesting example, shown in Figure~\ref{fig:ln_7}. The shortest plan produces a 6-segment solution due to the natural leader-follower behavior that appears in the shortest plan. Inspecting the full plan in Figure~\ref{fig:ln7_cbs_full} makes it difficult to validate the plan is collision free. The explanation (Figures~\ref{fig:ln7_cbs_seg1}-\ref{fig:ln7_cbs_seg6}) clearly shows a collision free plan. However, there are many segments compared to the small example. XG-CBS is capable of producing a much more explainable plan, shown in Figure~\ref{fig:ln7_xgcbs_full}. Forcing the green agent to wait and the purple agent to take a longer path enables a 2-segment plan (Figures~\ref{fig:ln7_xgcbs_seg1}-\ref{fig:ln7_xgcbs_seg2}) that is much easier to explain. We note here that XG-CBS with XG-$A^*$ found the 2-segment plan in $128.1$ seconds while the XG-CBS using classical $A^*$ once again failed to return a 2-segment solution after a maximum of 15 minutes of planning. 

\begin{figure*}
     \centering
     \begin{subfigure}{0.3\linewidth}
     \end{subfigure}
     \hfill
     \begin{subfigure}{0.3\linewidth}
         \centering
         \includegraphics[scale=0.4]{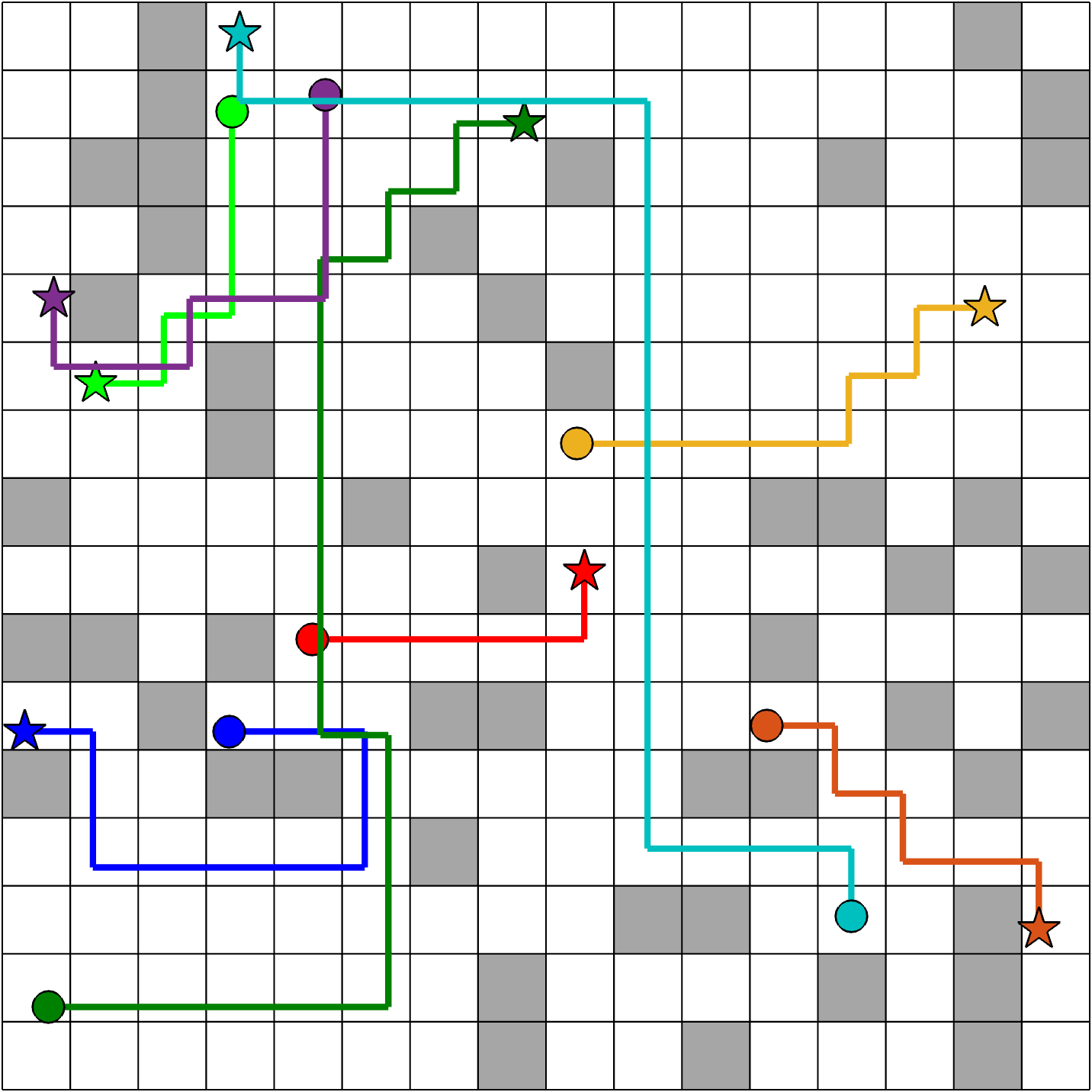}
         \caption{CBS}
         \label{fig:ln10_cbs_full}
     \end{subfigure}
     \hfill
     \begin{subfigure}{0.3\linewidth}
     \end{subfigure}
     \hfill
     \newline
     \begin{subfigure}{0.3\linewidth}
         \centering
         \includegraphics[scale=0.4]{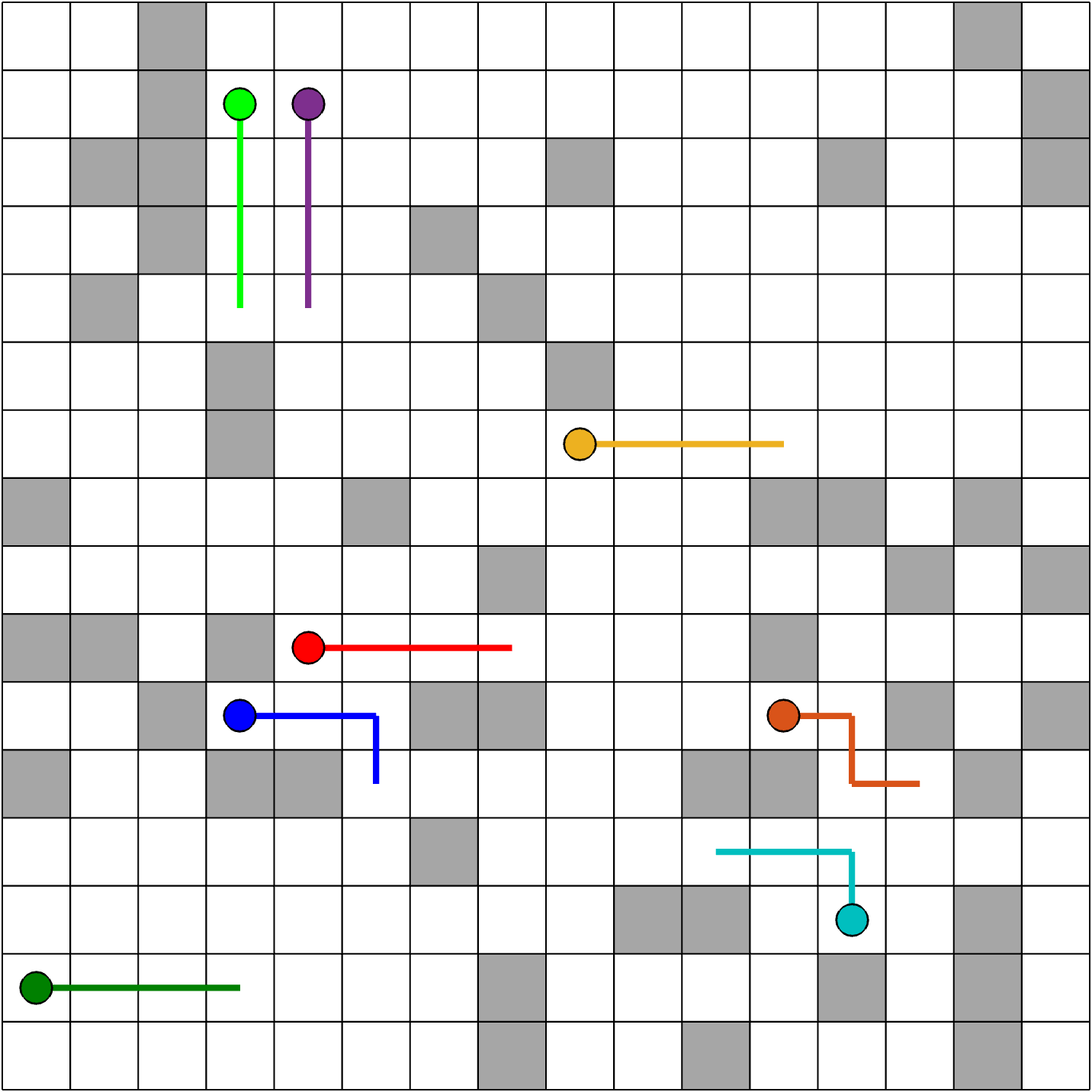}
         \caption{CBS, $\Delta k=[0,3]$}
         \label{fig:ln10_cbs_seg1}
     \end{subfigure}
     \hfill
     \begin{subfigure}{0.3\linewidth}
         \centering
         \includegraphics[scale=0.4]{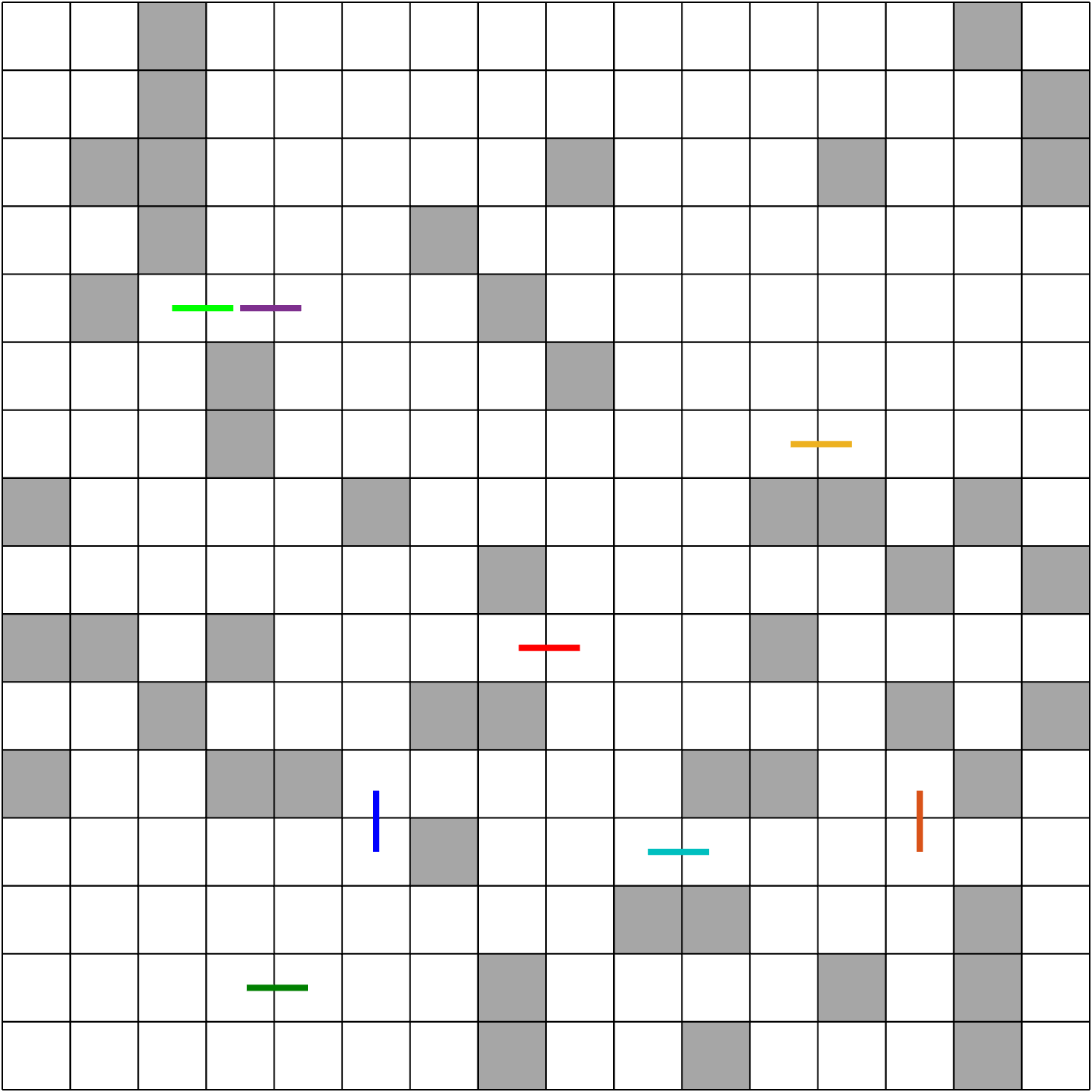}
         \caption{CBS, $\Delta k=[3,4]$}
         \label{fig:ln10_cbs_seg2}
     \end{subfigure}
     \hfill
     \begin{subfigure}{0.3\linewidth}
         \centering
         \includegraphics[scale=0.4]{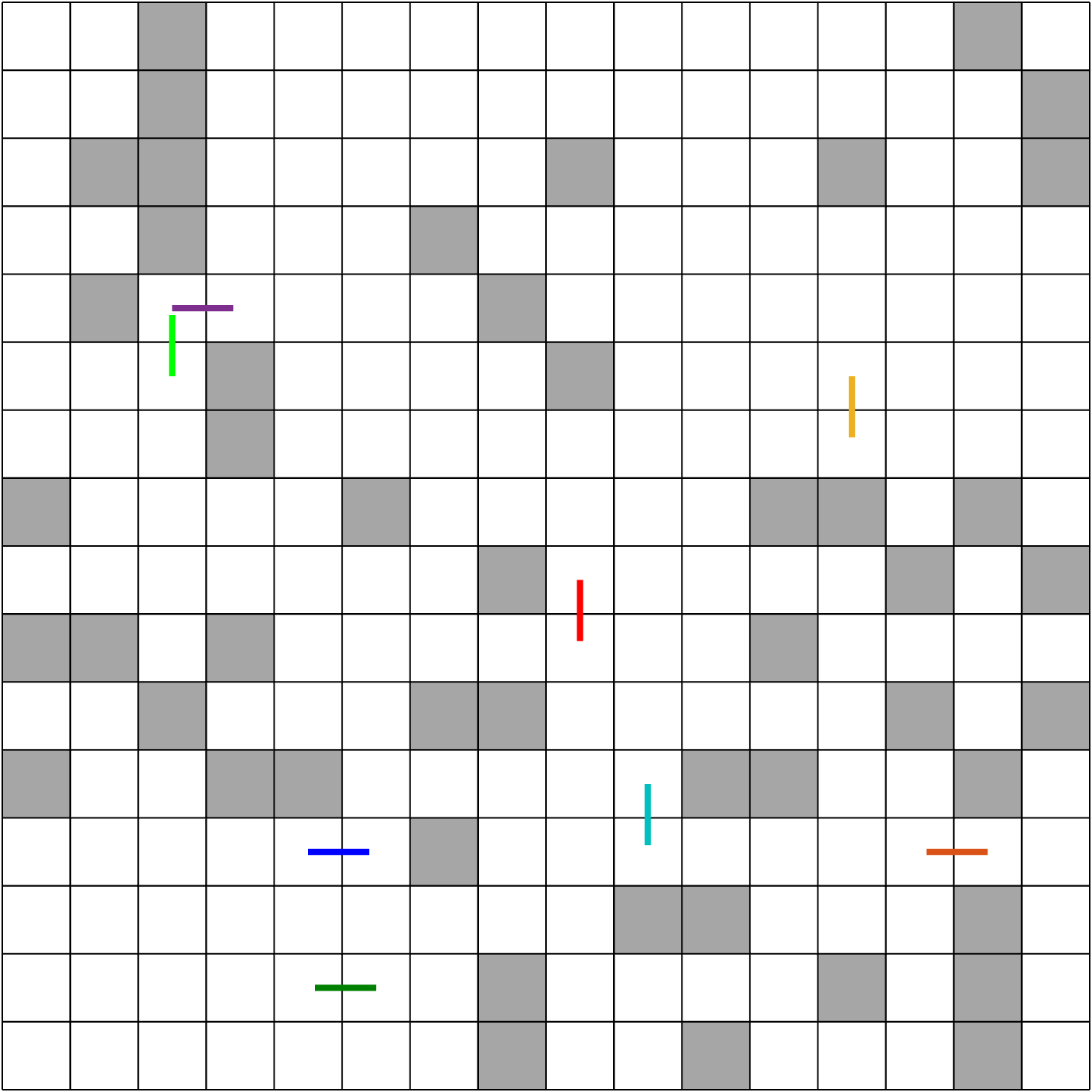}
         \caption{CBS, $\Delta k=[4,5]$}
         \label{fig:ln10_cbs_seg3}
     \end{subfigure}
     \hfill
     \newline
     \begin{subfigure}{0.3\linewidth}
         \centering
         \includegraphics[scale=0.4]{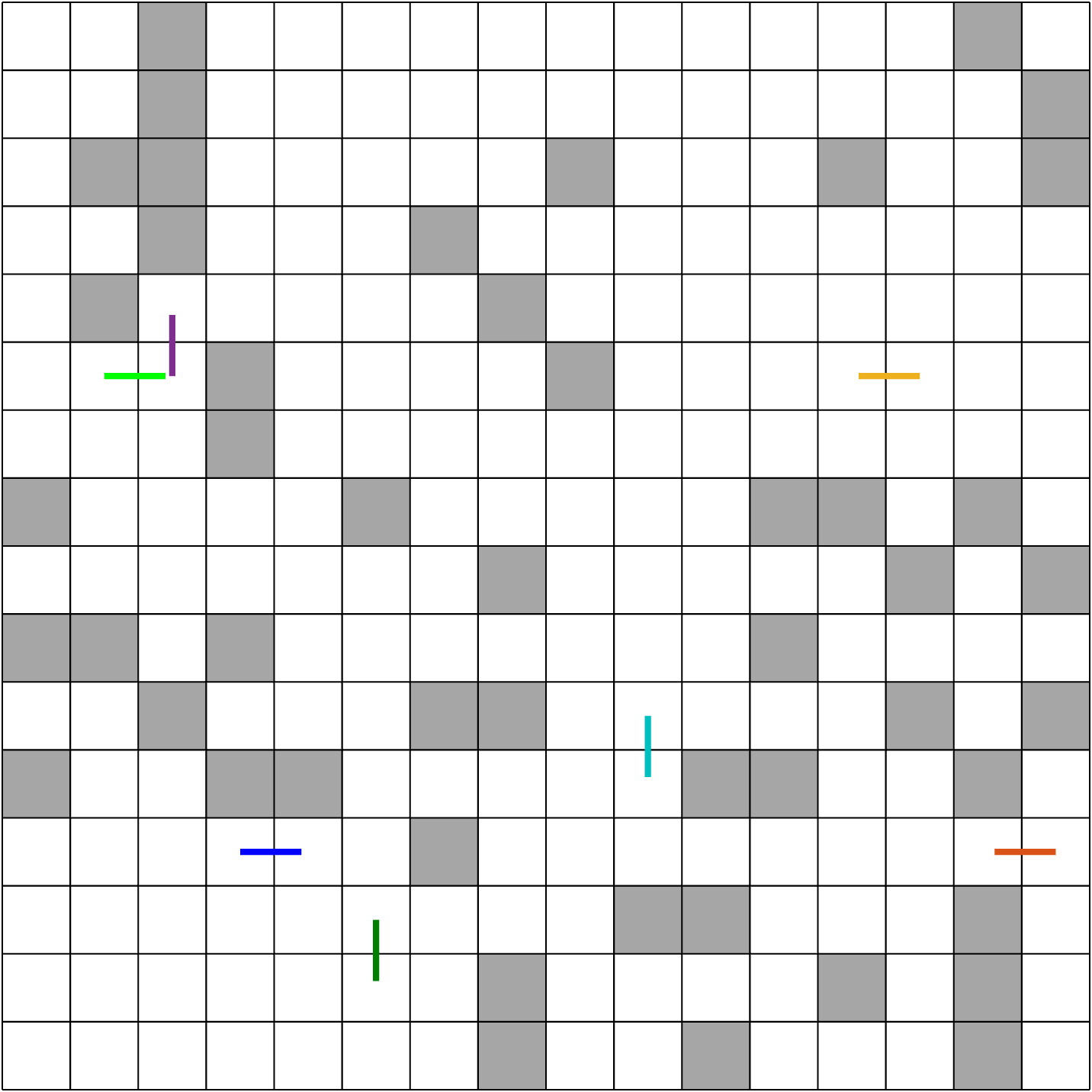}
         \caption{CBS, $\Delta k=[5,6]$}
         \label{fig:ln10_cbs_seg4}
     \end{subfigure}
     \hfill
     \begin{subfigure}{0.3\linewidth}
         \centering
         \includegraphics[scale=0.4]{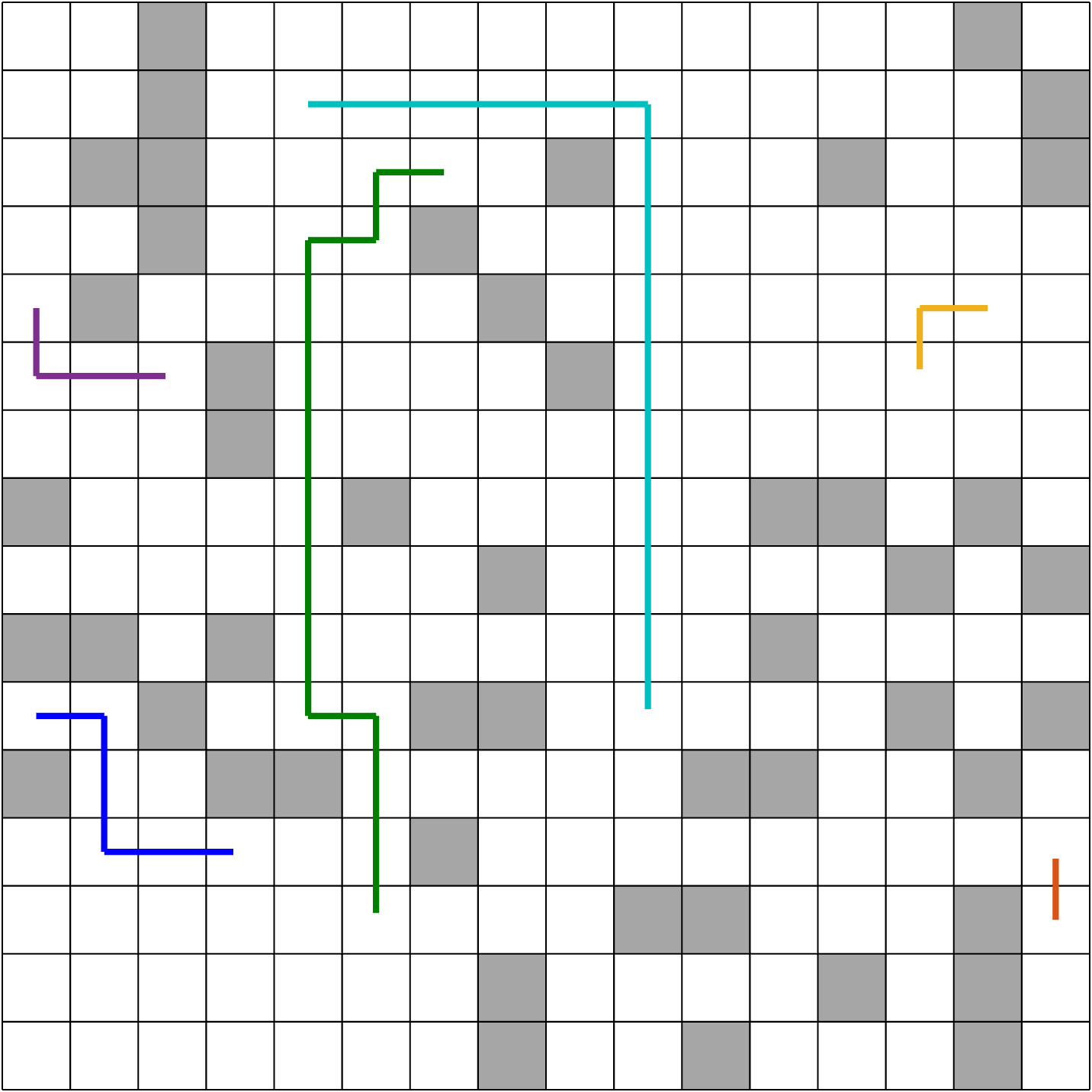}
         \caption{CBS, $\Delta k=[6,20]$}
         \label{fig:ln10_cbs_seg5}
     \end{subfigure}
     \hfill
     \begin{subfigure}{0.3\linewidth}
         \centering
         \includegraphics[scale=0.4]{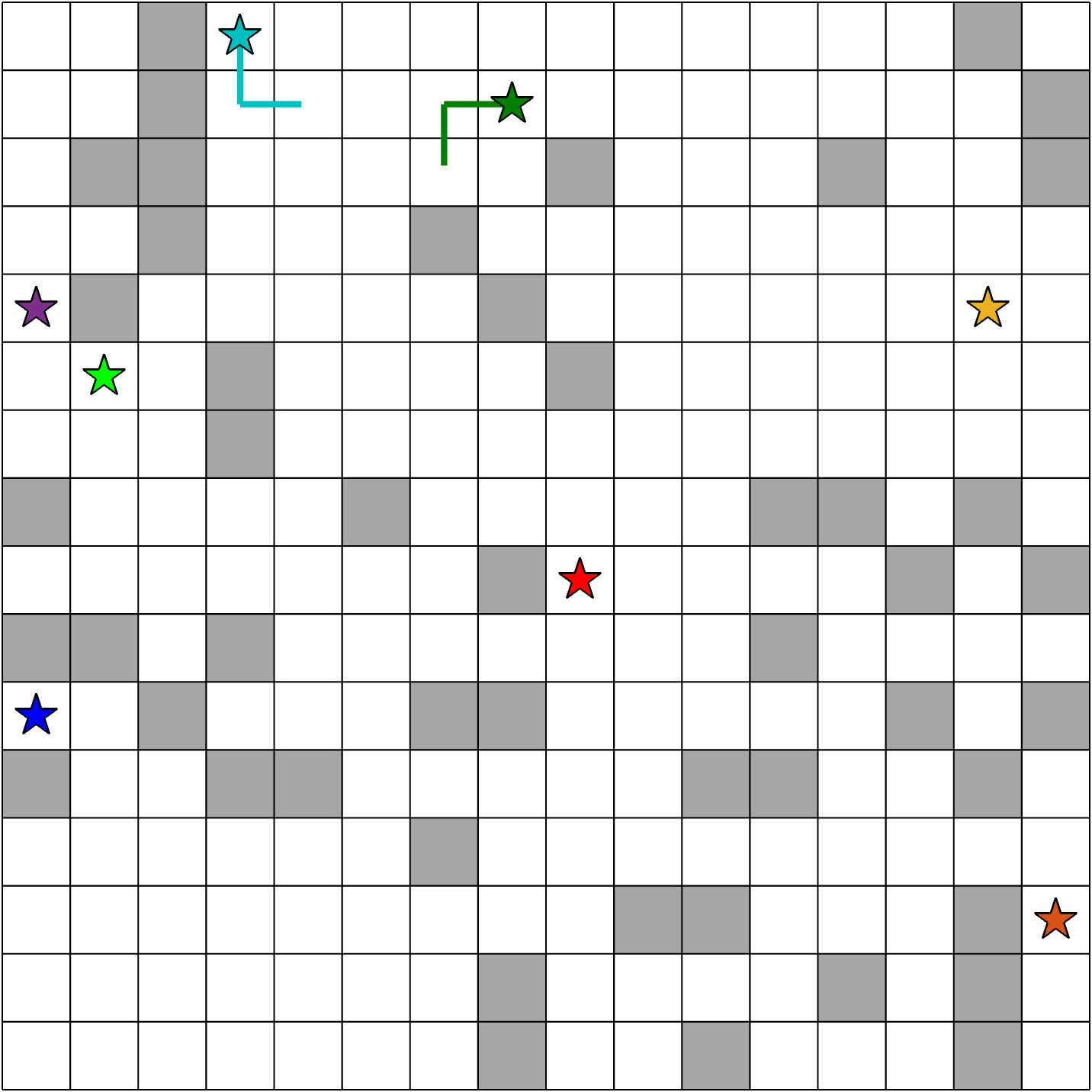}
         \caption{CBS, $\Delta k=[20,22]$}
         \label{fig:ln10_cbs_seg6}
     \end{subfigure}
     \hfill
     \newline
     \begin{subfigure}{0.3\linewidth}
         \centering
         \includegraphics[scale=0.4]{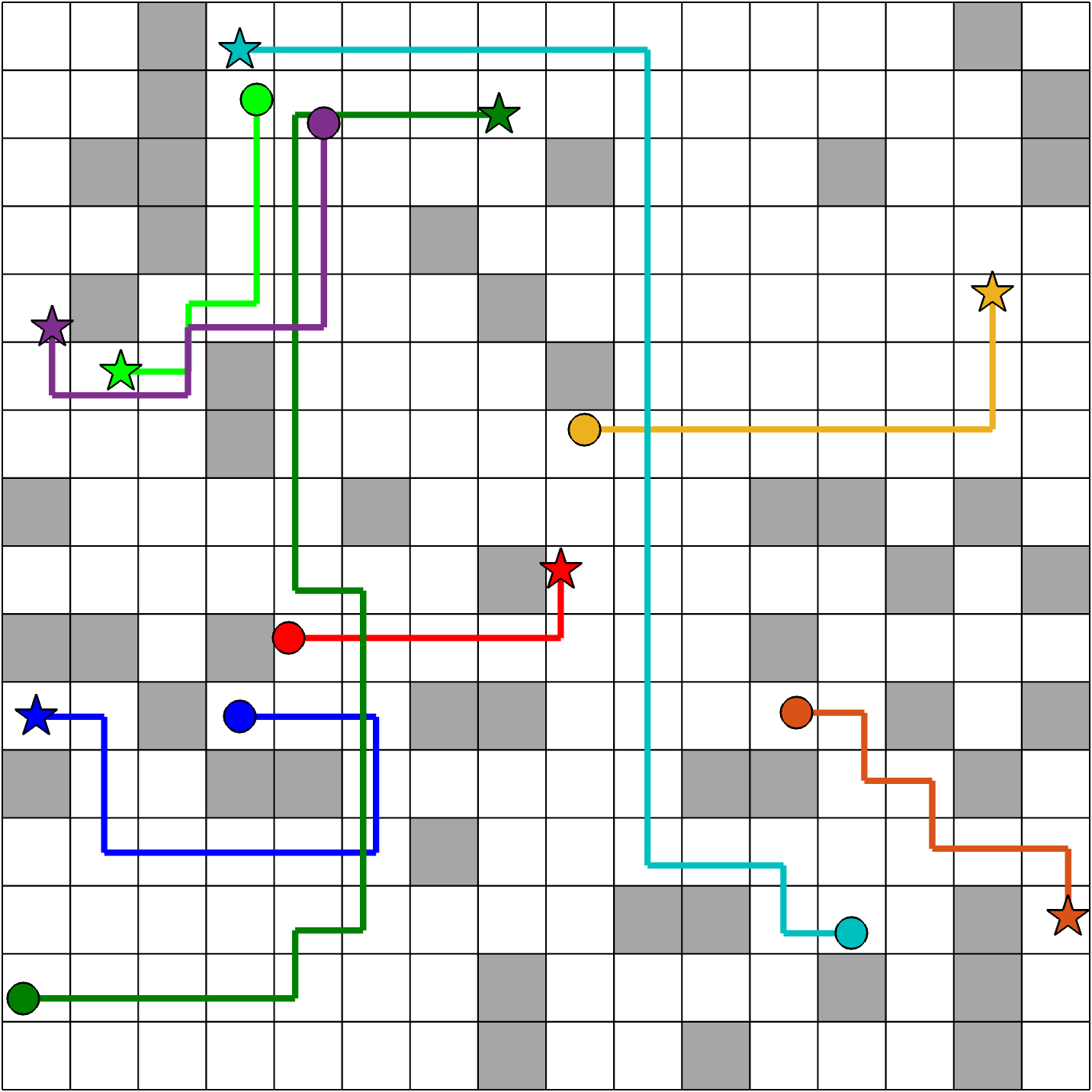}
         \caption{XG-CBS, $r=2$}
         \label{fig:ln10_xgcbs_full}
     \end{subfigure}
     \hfill
     \begin{subfigure}{0.3\linewidth}
         \centering
         \includegraphics[scale=0.4]{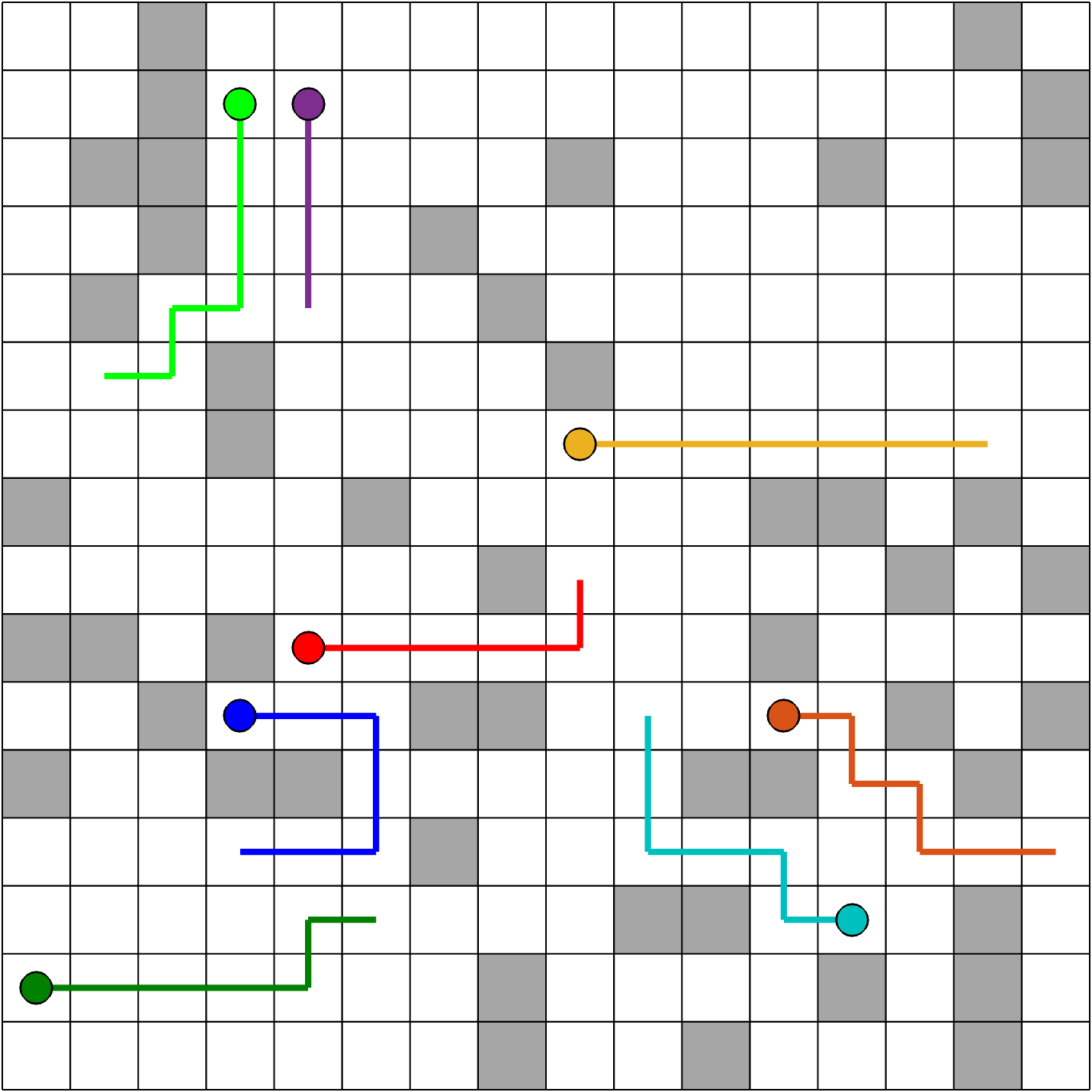}
         \caption{XG-CBS, $\Delta k=[0,6]$}
         \label{fig:ln10_xgcbs_seg1}
     \end{subfigure}
     \hfill
     \begin{subfigure}{0.3\linewidth}
         \centering
         \includegraphics[scale=0.4]{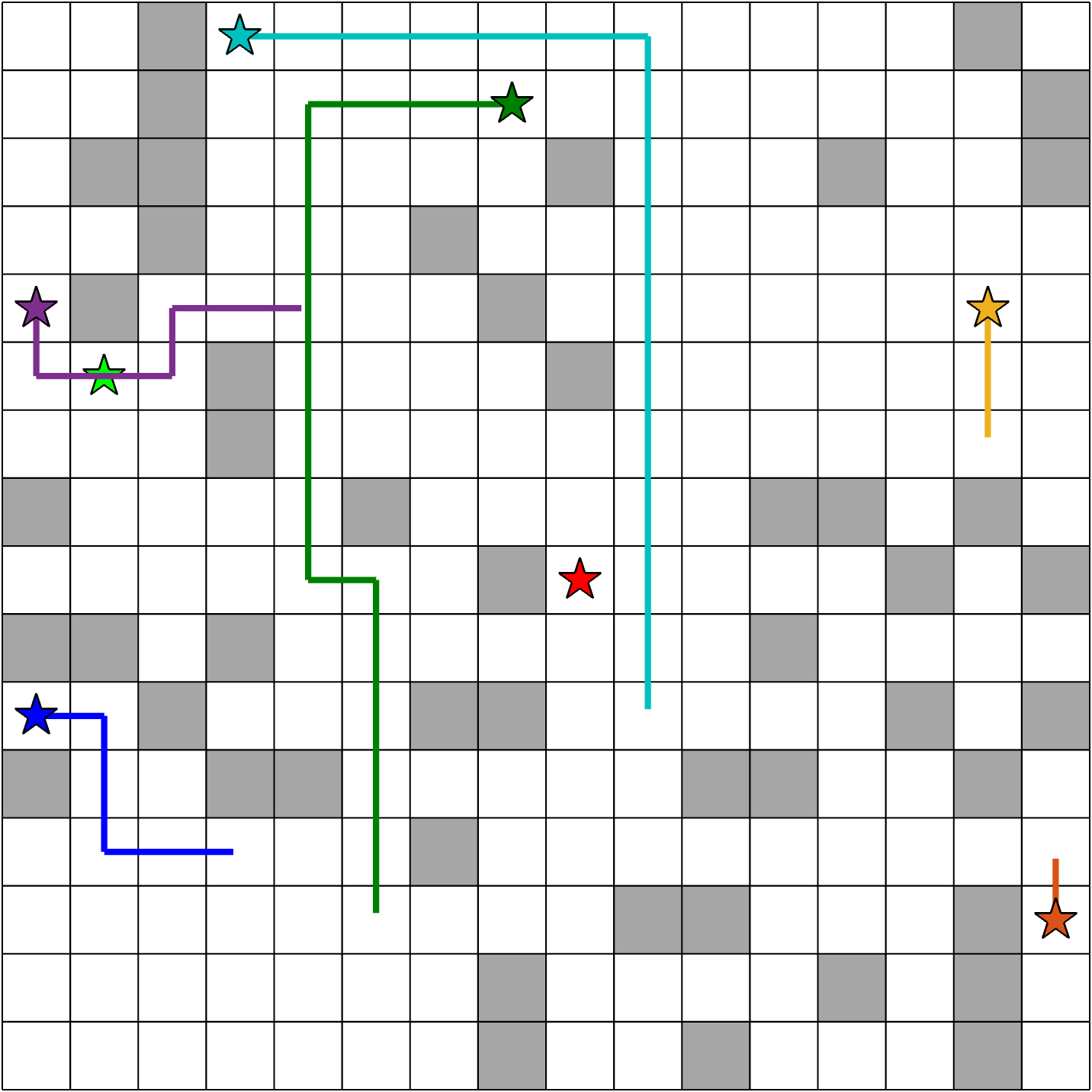}
         \caption{XG-CBS, $\Delta k=[6,22]$}
         \label{fig:ln10_xgcbs_seg2}
     \end{subfigure}
    % \caption{Line $29$ of Table~\ref{tab:final_benchmark}}
    \caption{Example of XG-CBS with 2-segment solution vs 6-segment solution of CBS}
    \label{fig:ln_10} % numbers in labels represent row of Table 1 -- numbers in document should be from Table 2
\end{figure*}

We see one example where XG-CBS with classical $A^*$ performs significantly better than XG-CBS with XG-$A^*$ in Figure~\ref{fig:ln_10}. Here, XG-CBS with XG-$A^*$ does not return a 2-segment solution within 15 minutes. However, XG-CBS with $A^*$ returns the preferred solution in $0.44$ seconds. Notice that the 6-segment plan returned by CBS (Figure~\ref{fig:ln10_cbs_full}) and the 2-segment plan returned by XG-CBS with $A^*$ (Figure~\ref{fig:ln10_xgcbs_full}) look very similar. This is different from earlier examples we have seen thus far, where decreasing segments requires a heavy deviation from the shortest paths for individual agents. Experiments show that in examples like these, where we can greatly simplify the explanation with minor tweaking of the shortest path plan, that XG-CBS with $A^*$ outperforms XG-CBS with XG-$A^*$. We attribute this to the fact that using $A^*$ quickly makes minor changes to the shortest paths while XG-$A^*$ enables large deviations from shortest paths but suffers from computation time as a result. 

\begin{figure*}[p]
     \centering
     \begin{subfigure}{0.3\linewidth}
         \centering
         \includegraphics[scale=0.4]{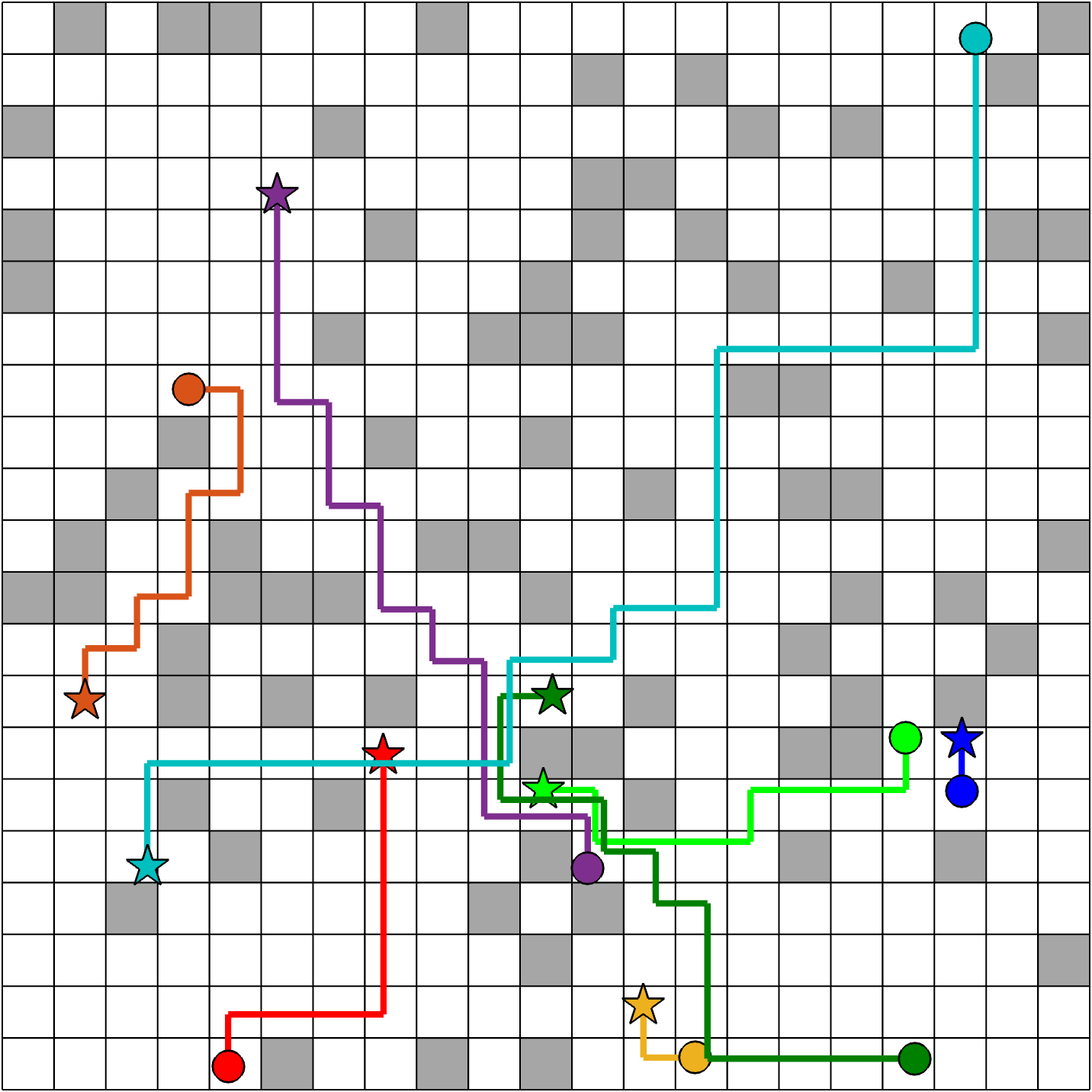}
         \caption{CBS}
         \label{fig:ln11_cbs_full}
     \end{subfigure}
     \hfill
     \begin{subfigure}{0.3\linewidth}
         \centering
         \includegraphics[scale=0.4]{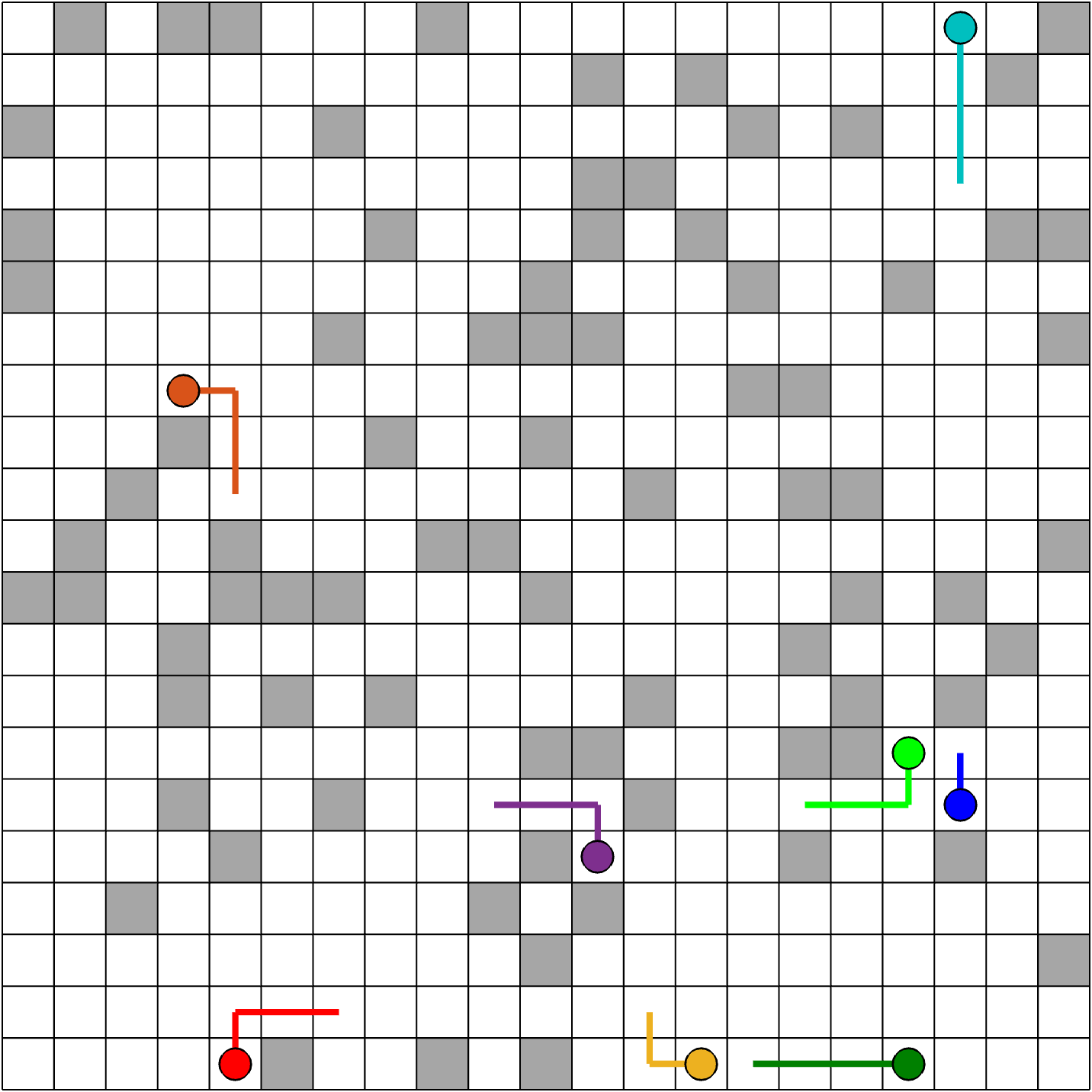}
         \caption{CBS, $\Delta k=[0,3]$}
         \label{fig:ln11_cbs_seg1}
     \end{subfigure}
     \hfill
     \begin{subfigure}{0.3\linewidth}
         \centering
         \includegraphics[scale=0.4]{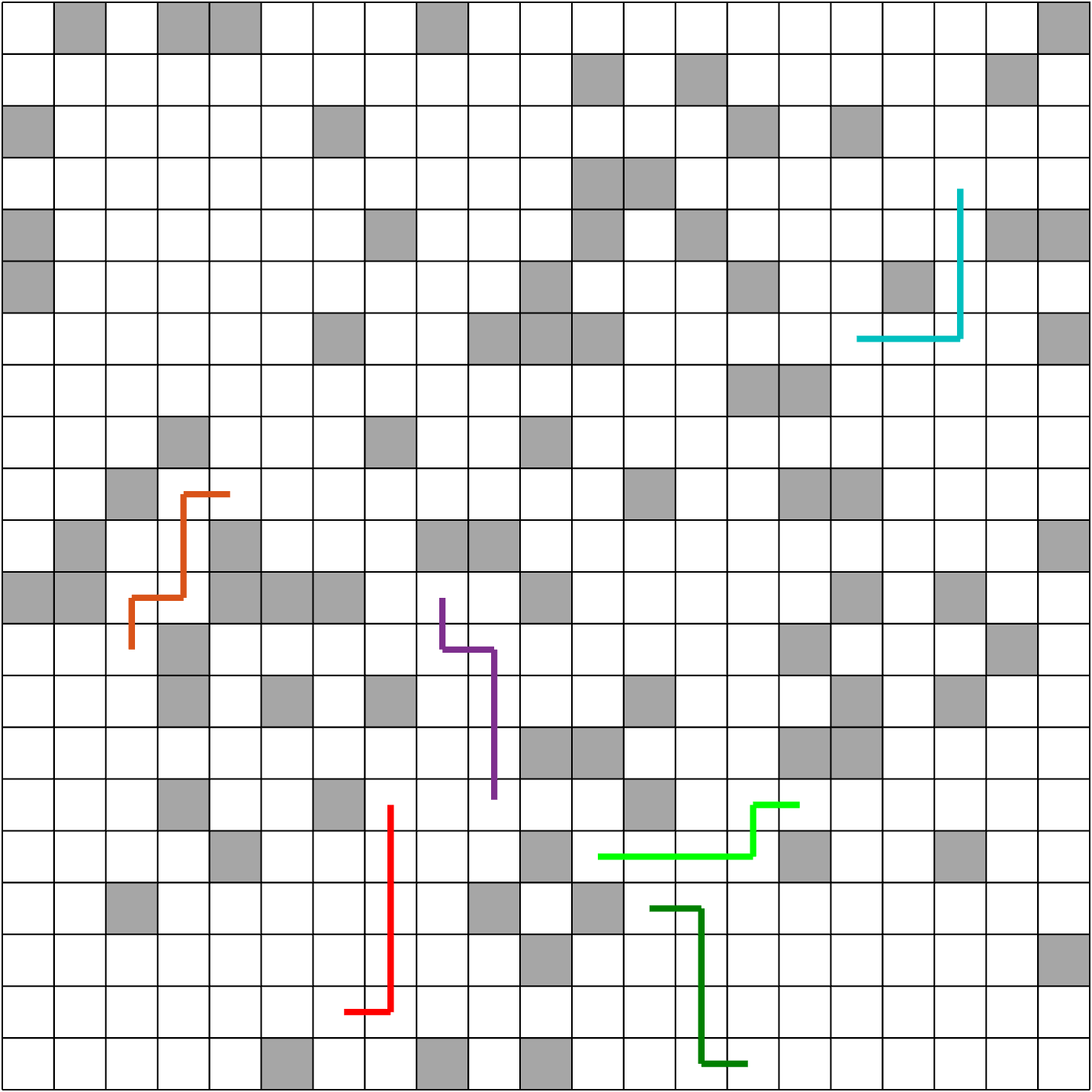}
         \caption{CBS, $\Delta k=[3,8]$}
         \label{fig:ln11_cbs_seg2}
     \end{subfigure}
     \hfill
     \newline
     \begin{subfigure}{0.3\linewidth}
         \centering
         \includegraphics[scale=0.4]{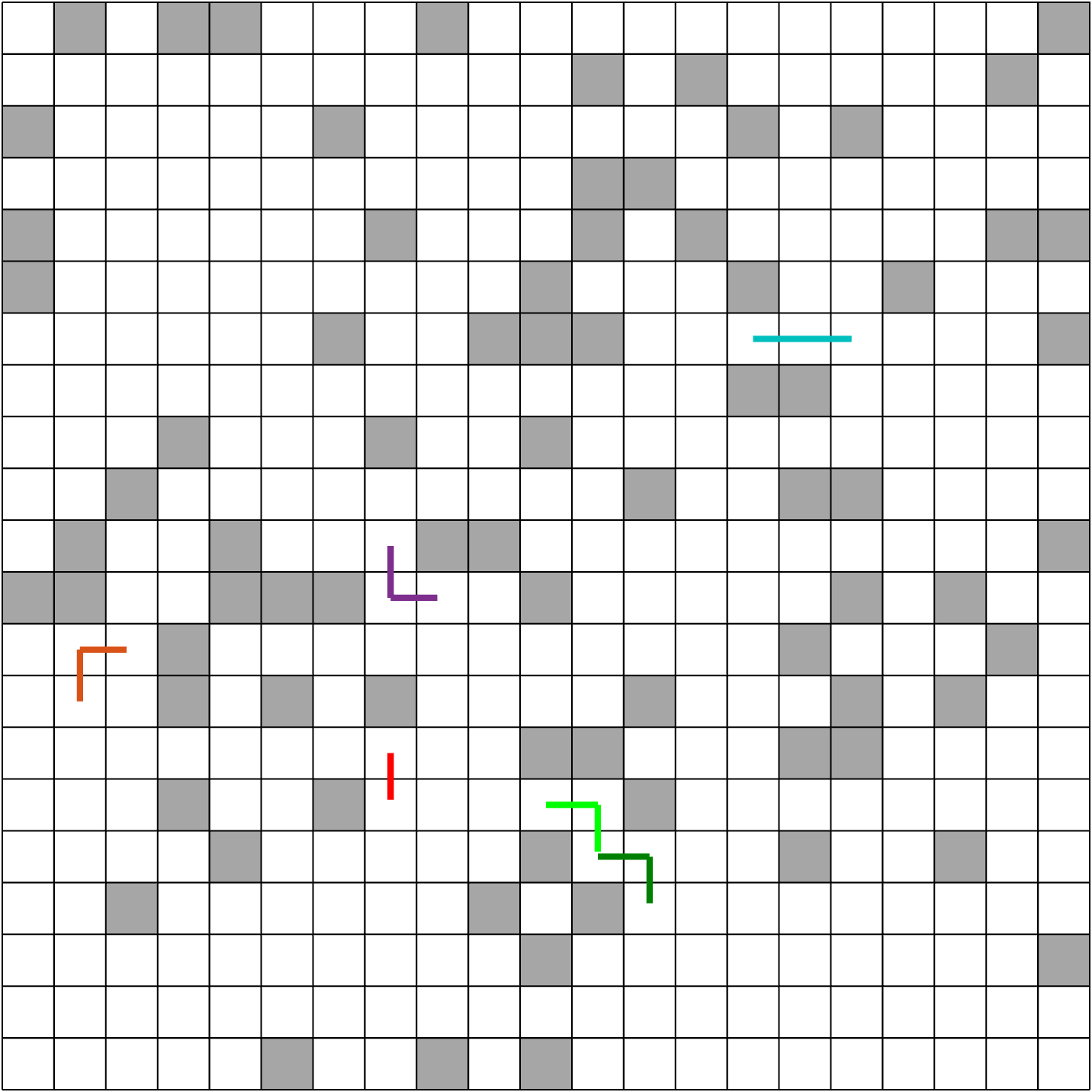}
         \caption{CBS, $\Delta k=[8,10]$}
         \label{fig:ln11_cbs_seg3}
     \end{subfigure}
     \hfill
     \begin{subfigure}{0.3\linewidth}
         \centering
         \includegraphics[scale=0.4]{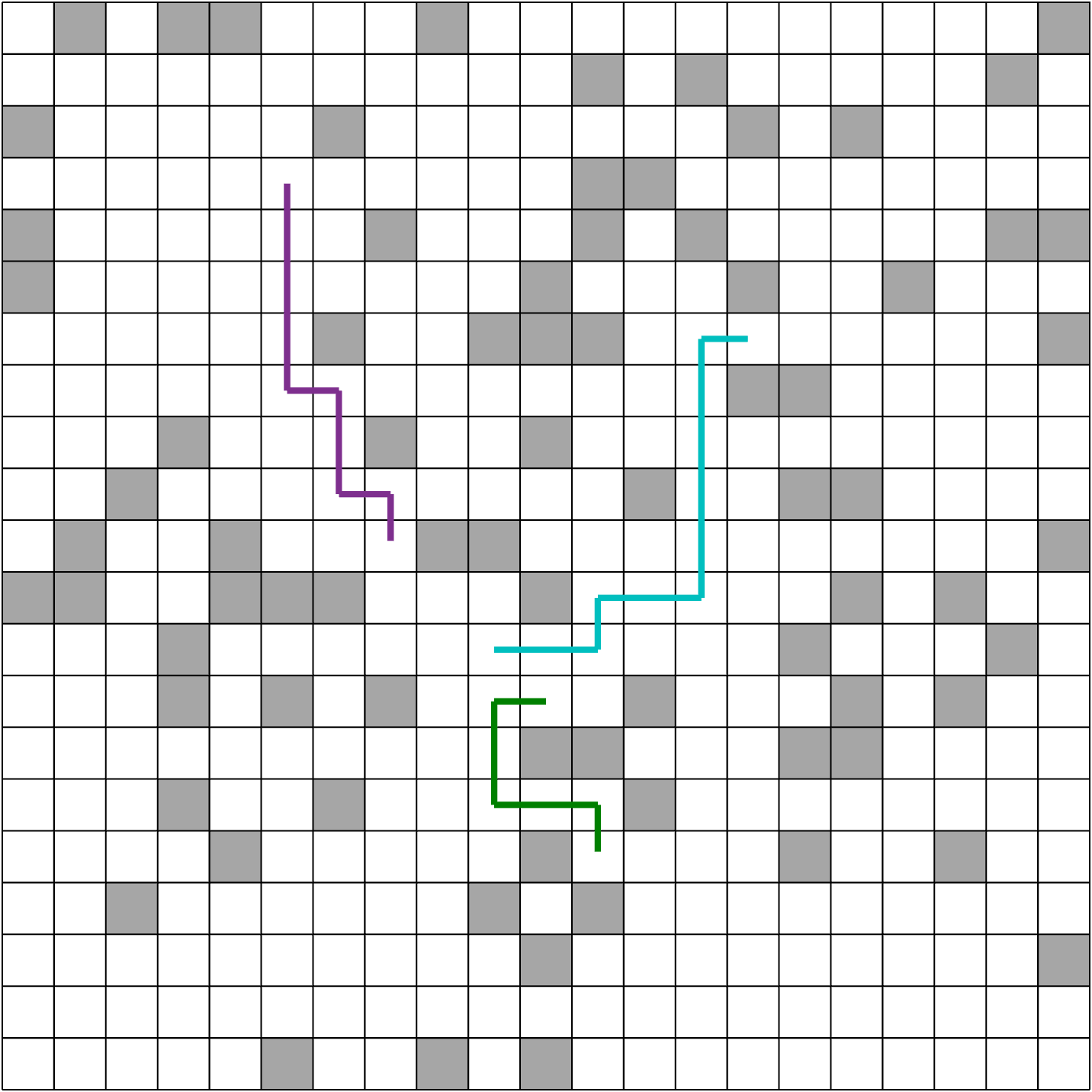}
         \caption{CBS, $\Delta k=[10,19]$}
         \label{fig:ln11_cbs_seg4}
     \end{subfigure}
     \hfill
     \begin{subfigure}{0.3\linewidth}
         \centering
         \includegraphics[scale=0.4]{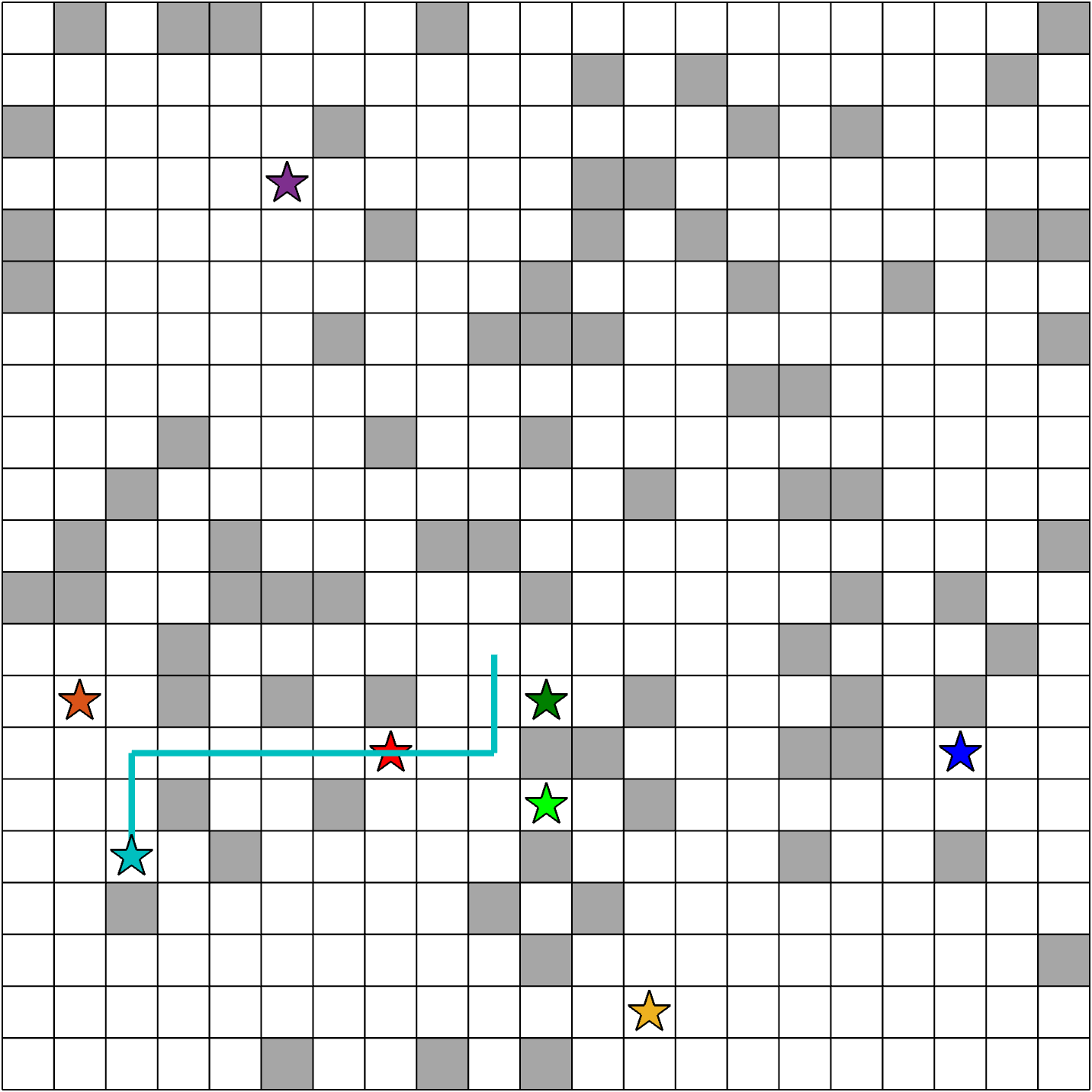}
         \caption{CBS, $\Delta k=[19, 30]$}
         \label{fig:ln11_cbs_seg5}
     \end{subfigure}
     \hfill
     \newline
     \begin{subfigure}{0.3\linewidth}
         \centering
         \includegraphics[scale=0.4]{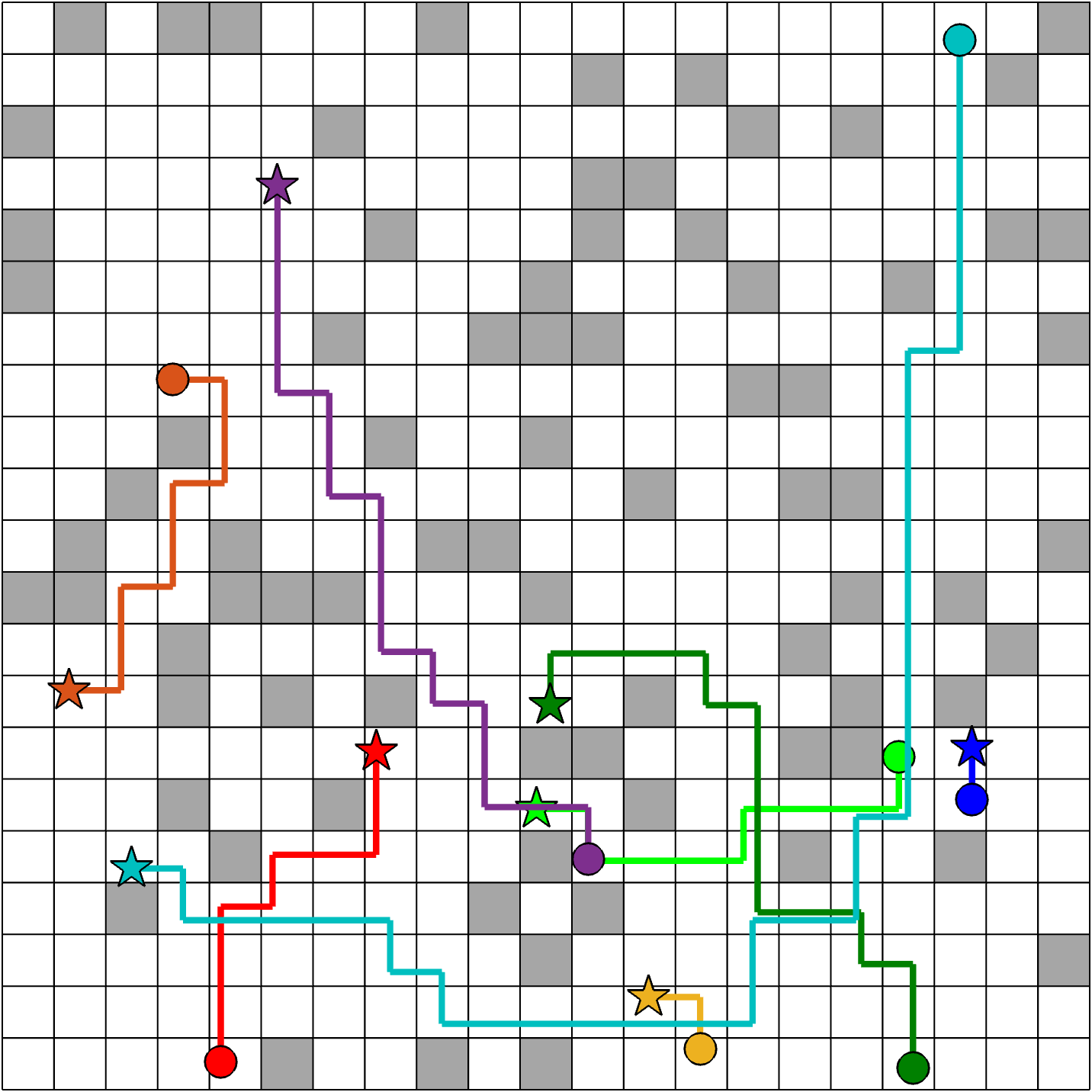}
         \caption{XG-CBS, $r=2$}
         \label{fig:ln11_xgcbs_full}
     \end{subfigure}
     \hfill
     \begin{subfigure}{0.3\linewidth}
         \centering
         \includegraphics[scale=0.4]{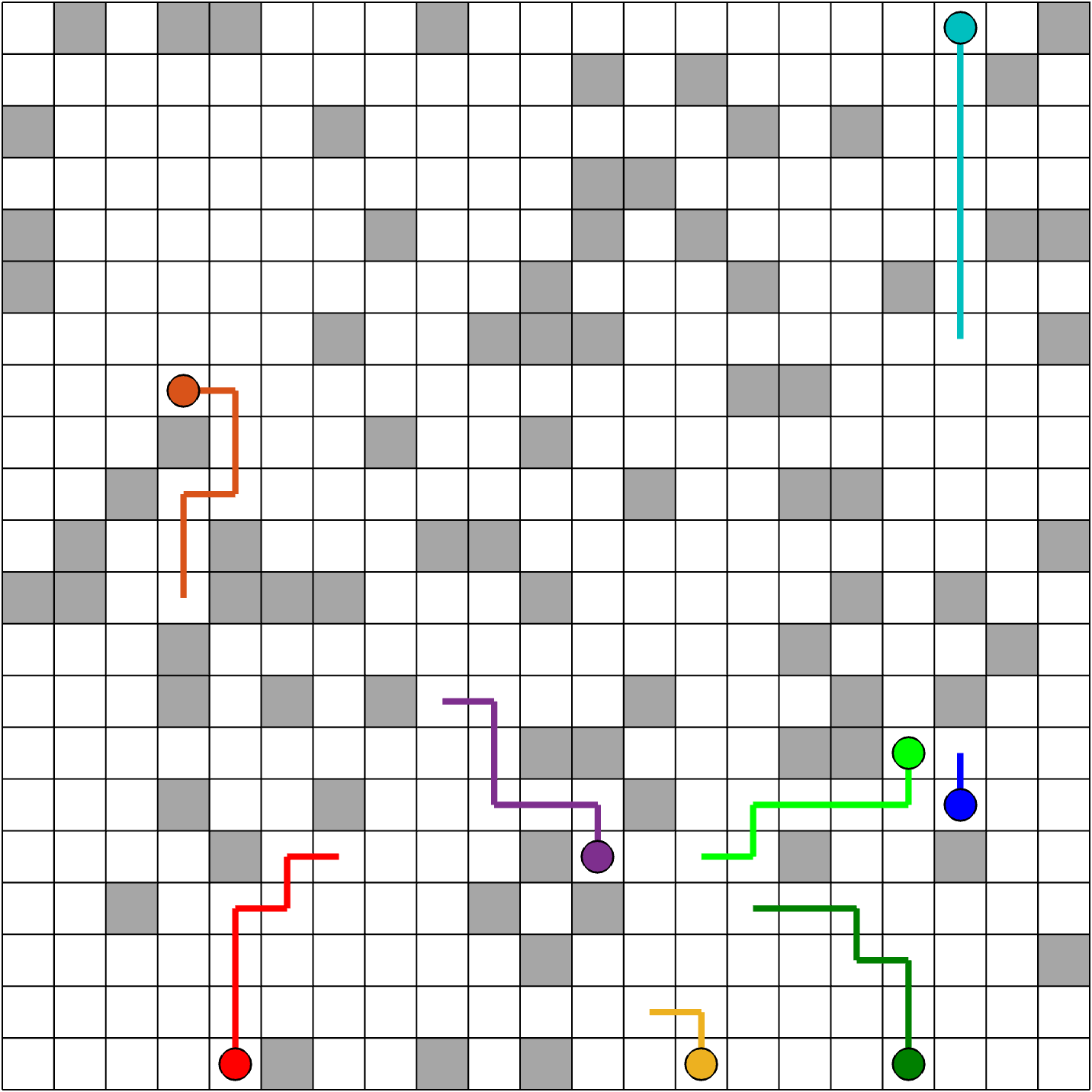}
         \caption{XG-CBS, $\Delta k=[0,6]$}
         \label{fig:ln11_xgcbs_seg1}
     \end{subfigure}
     \hfill
     \begin{subfigure}{0.3\linewidth}
         \centering
         \includegraphics[scale=0.4]{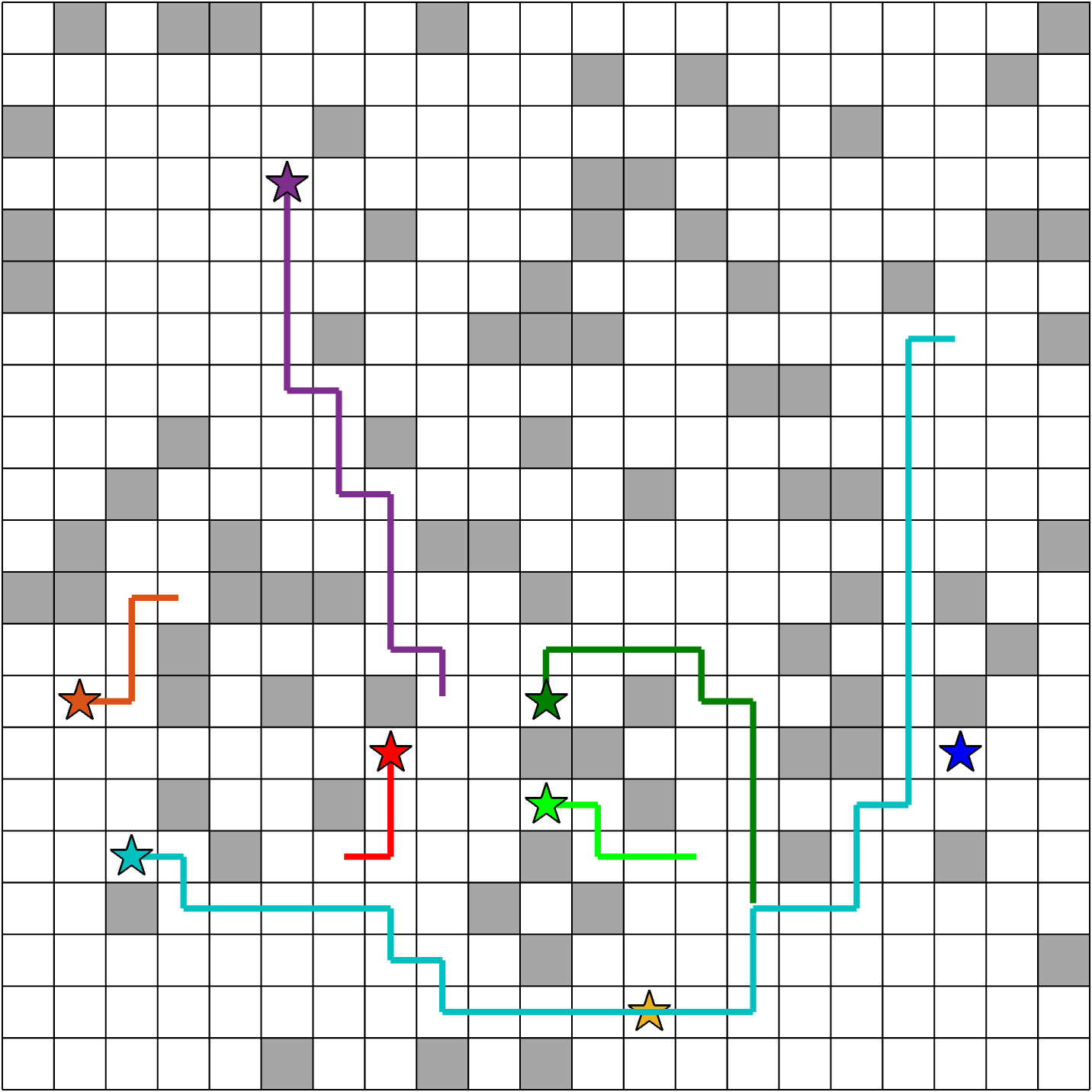}
         \caption{XG-CBS, $\Delta k=[6,38]$}
         \label{fig:ln11_xgcbs_seg2}
     \end{subfigure}
    % \caption{Line $33$ of Table~\ref{tab:final_benchmark}}
    \caption{Example of XG-CBS reducing a 5-segment solution of CBS to a 2-segment plan.}
    \label{fig:ln_11} % numbers in labels represent row of Table 1 -- numbers in document should be from Table 2
\end{figure*}

We witness an opposite behavior in the example shown in Figure~\ref{fig:ln_11}. The shortest path plan generated by CBS in Figure~\ref{fig:ln11_cbs_full} shows many paths overlapping each other, suggesting a high number of segments. Indeed, the plan requires five disjoint segments. Overlapping paths also suggest that large deviations are required by the agents to drastically decrease segmentation. XG-CBS with XG-$A^*$ returns the plan shown in Figure~\ref{fig:ln11_xgcbs_full} in $113.9$ seconds. As expected, XG-CBS with $A^*$ did not find a solution in $15$ minutes of planning. 

\begin{figure*}[p]
     \centering
     \begin{subfigure}{0.24\linewidth}
         \centering
         \includegraphics[scale=0.35]{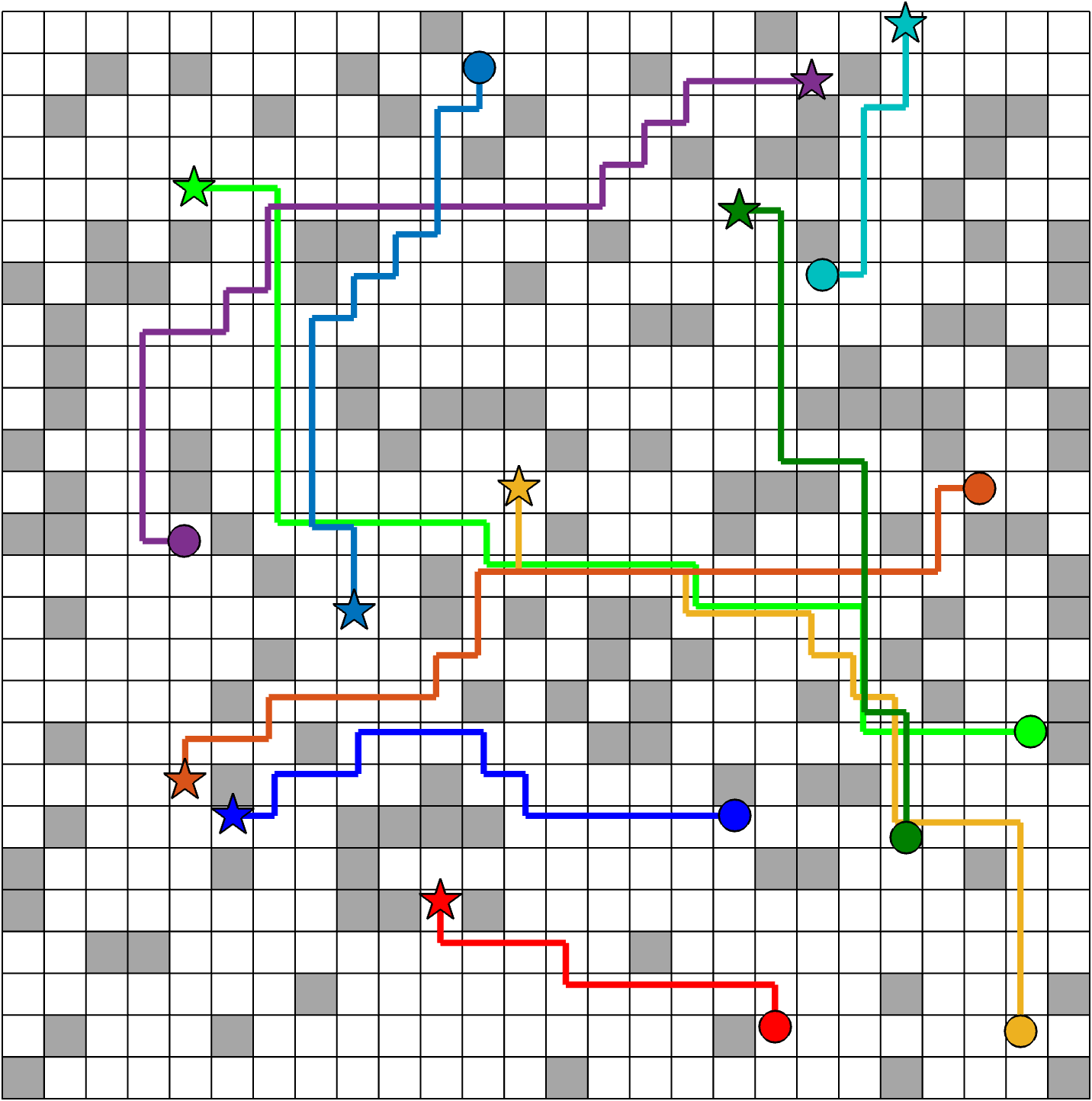}
         \caption{CBS}
         \label{fig:ln15_cbs_full}
     \end{subfigure}
     \hfill
     \begin{subfigure}{0.24\linewidth}
         \centering
         \includegraphics[scale=0.35]{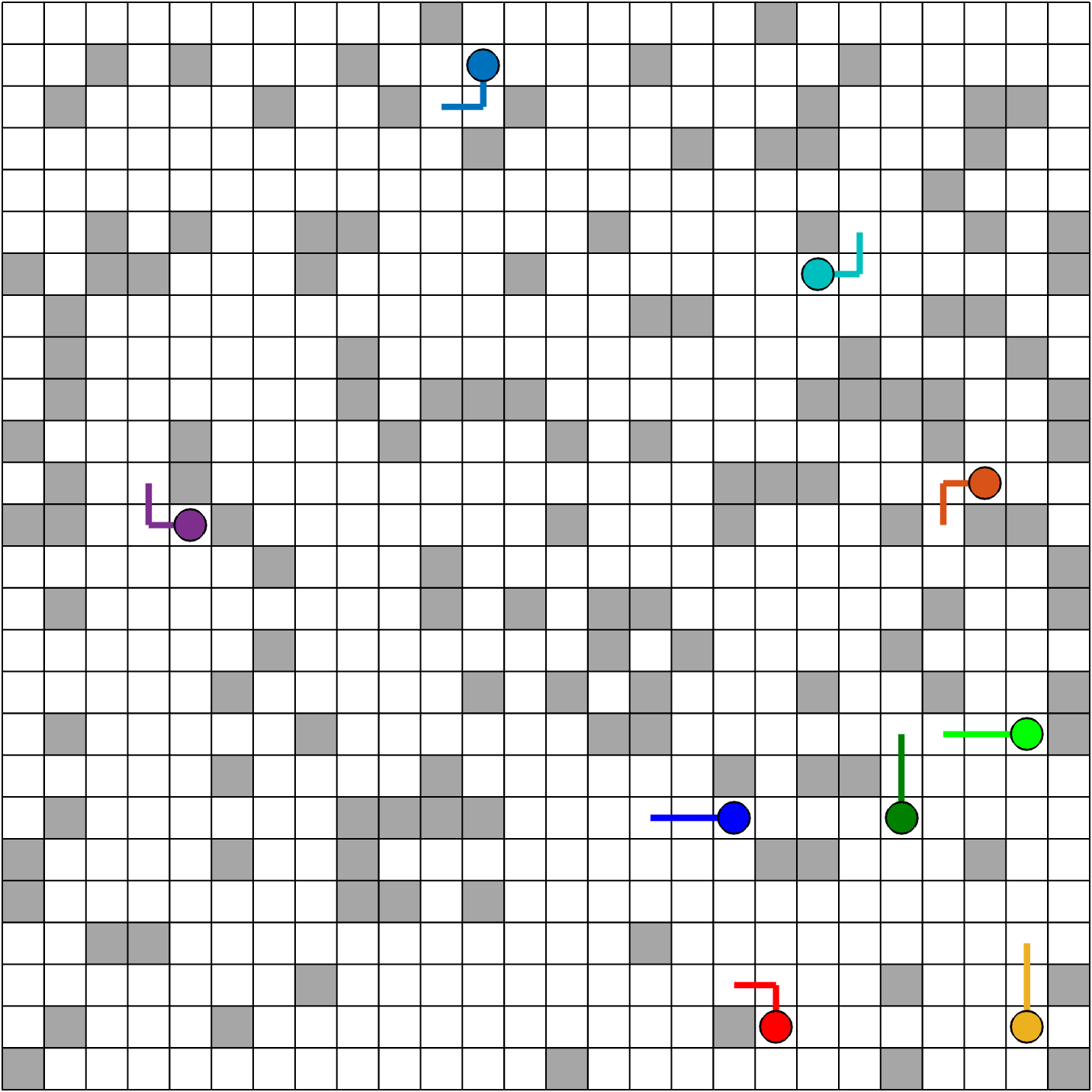}
         \caption{CBS, $\Delta k=[0,2]$}
         \label{fig:ln15_cbs_seg1}
     \end{subfigure}
     \hfill
     \begin{subfigure}{0.24\linewidth}
         \centering
         \includegraphics[scale=0.35]{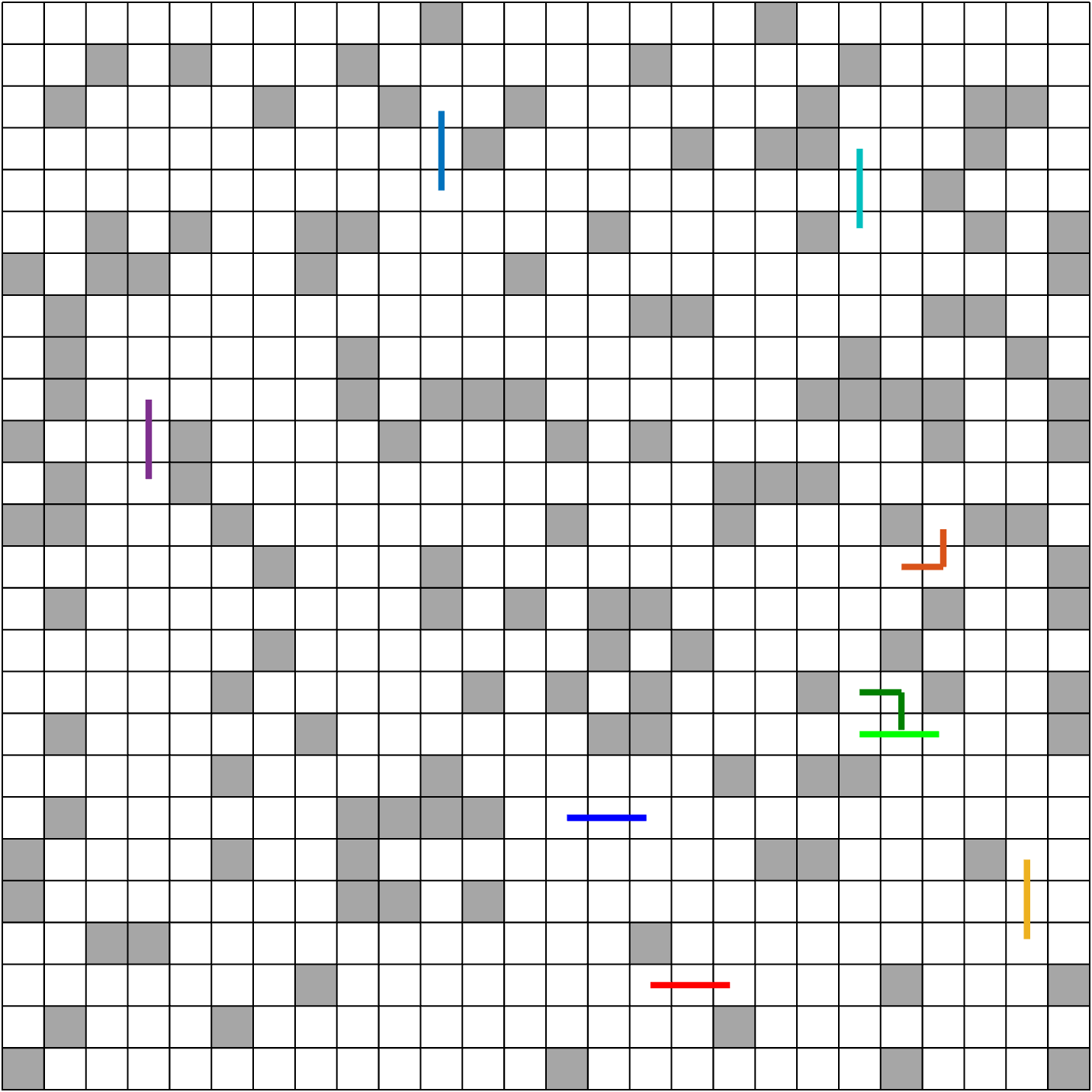}
         \caption{CBS, $\Delta k=[2,4]$}
         \label{fig:ln15_cbs_seg2}
     \end{subfigure}
     \hfill
     \begin{subfigure}{0.24\linewidth}
         \centering
         \includegraphics[scale=0.35]{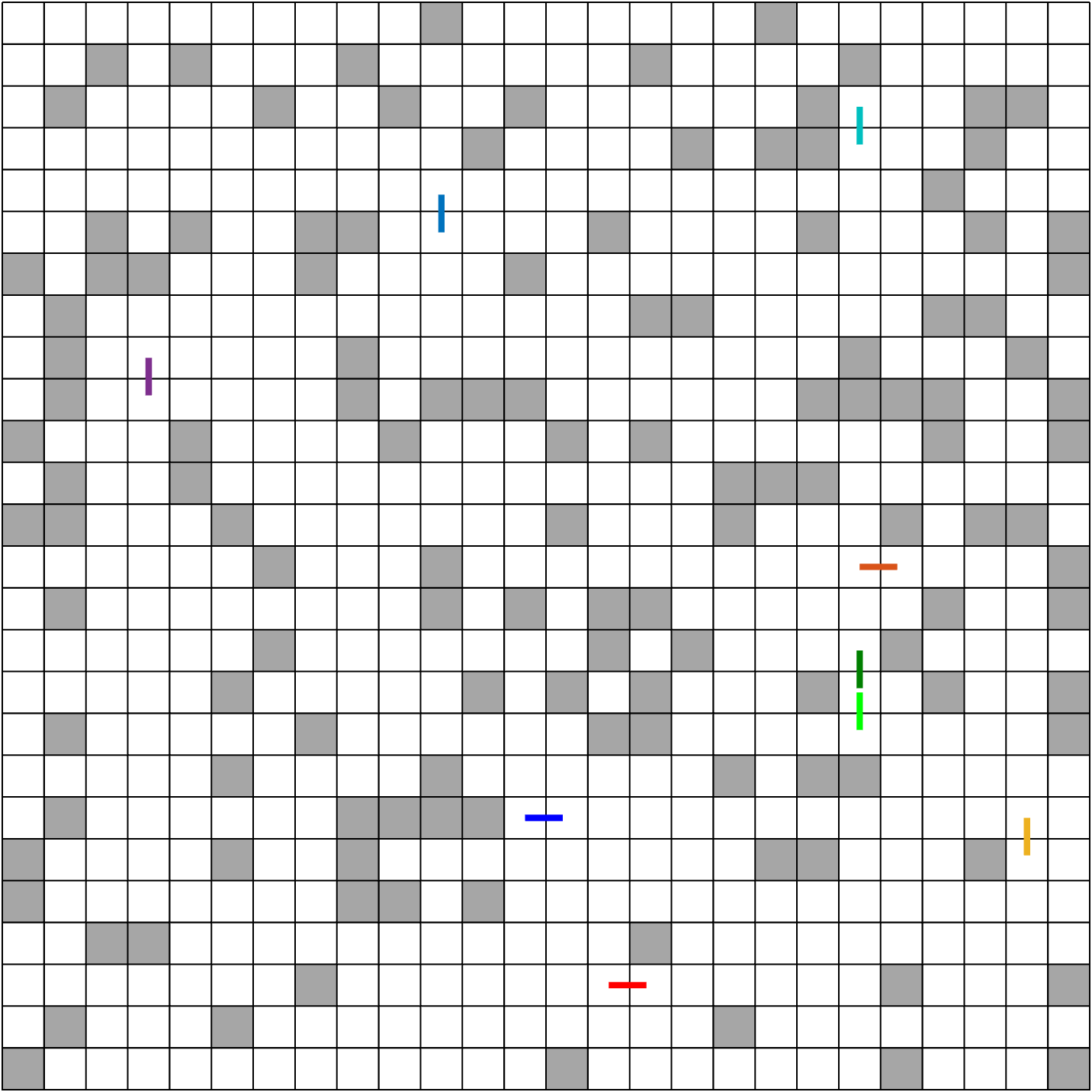}
         \caption{CBS, $\Delta k=[4,5]$}
         \label{fig:ln15_cbs_seg3}
     \end{subfigure}
     \hfill
     \newline
     \begin{subfigure}{0.24\linewidth}
         \centering
         \includegraphics[scale=0.35]{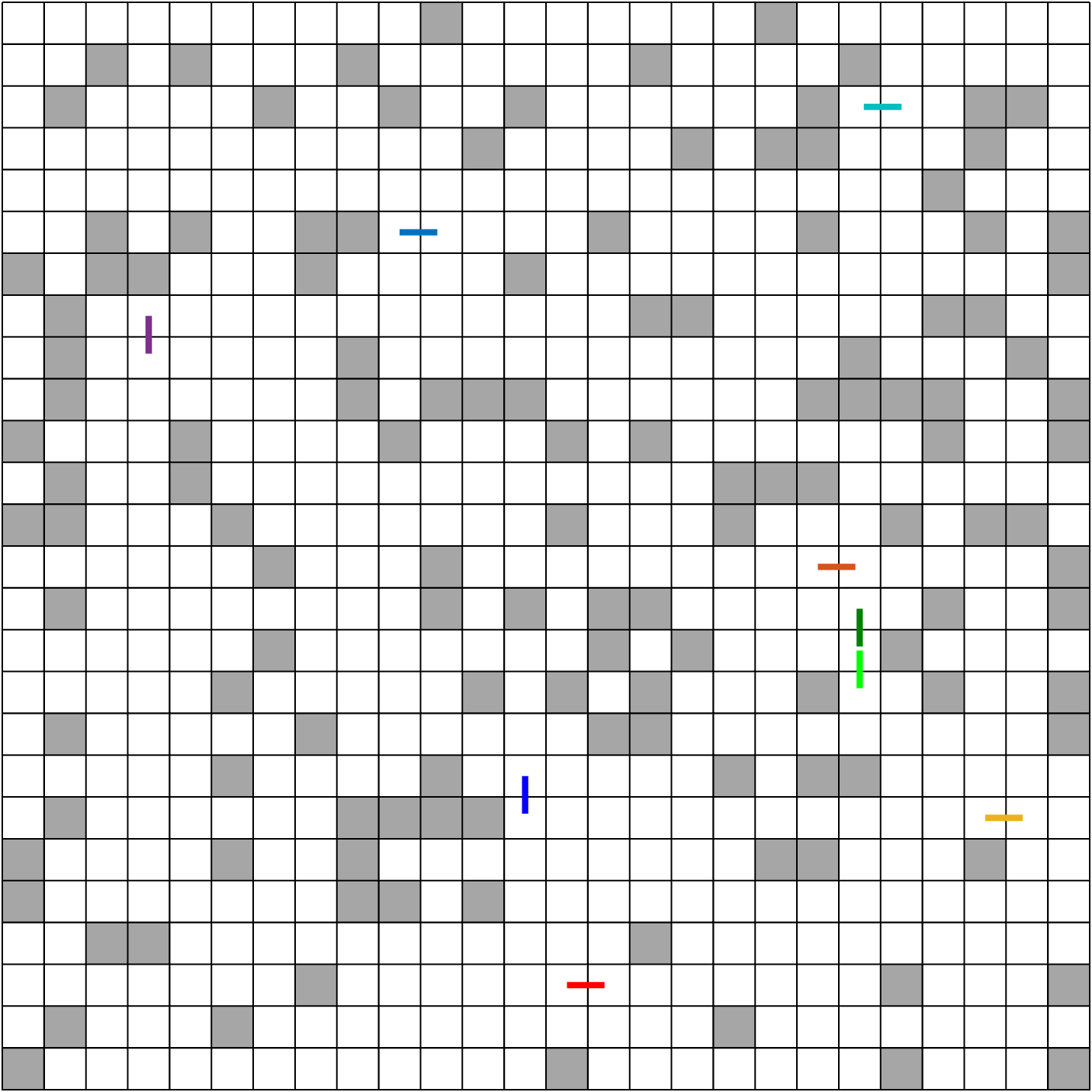}
         \caption{CBS, $\Delta k=[5,6]$}
         \label{fig:ln15_cbs_seg4}
     \end{subfigure}
     \hfill
     \begin{subfigure}{0.24\linewidth}
         \centering
         \includegraphics[scale=0.35]{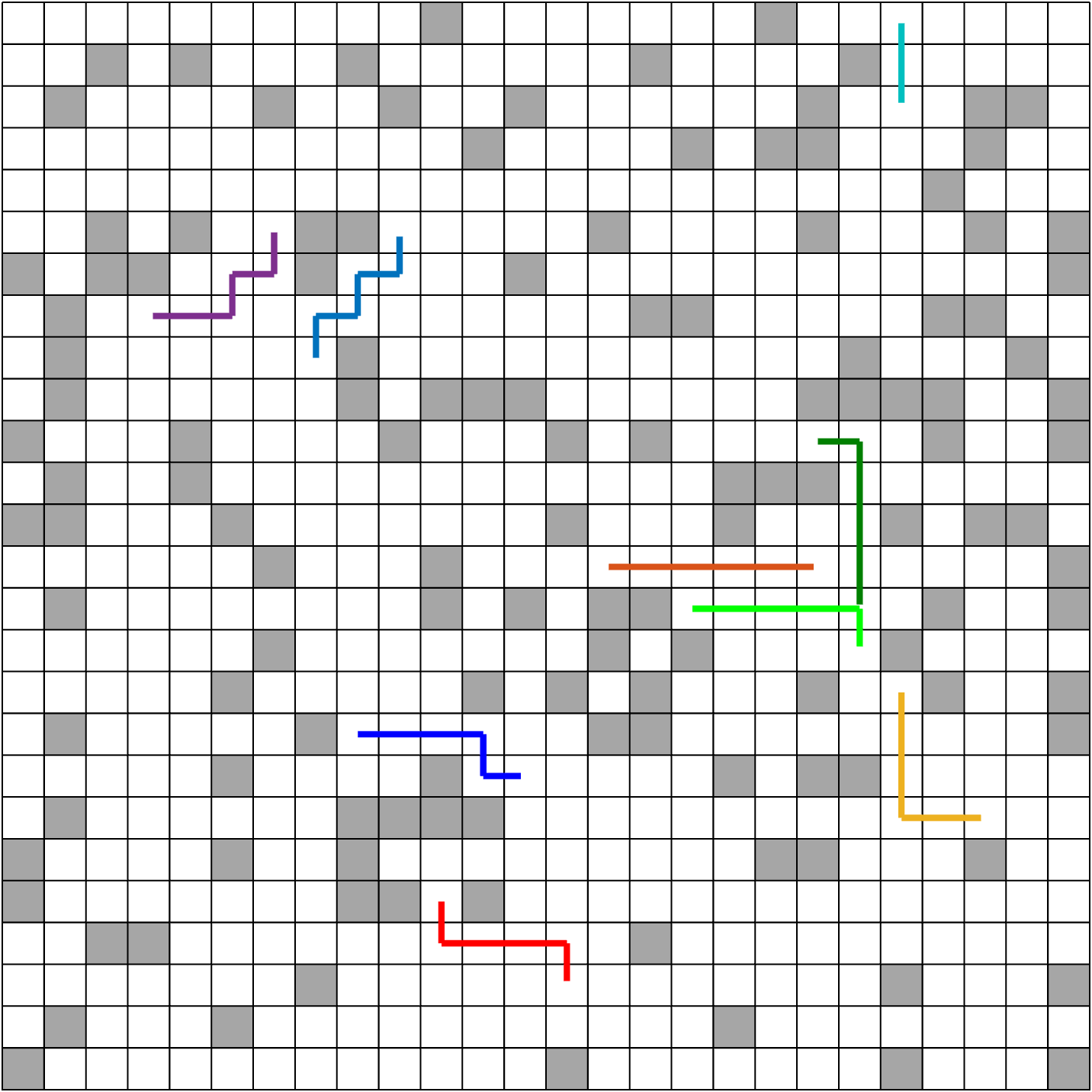}
         \caption{CBS, $\Delta k=[6,11]$}
         \label{fig:ln15_cbs_seg5}
     \end{subfigure}
     \hfill
     \begin{subfigure}{0.24\linewidth}
         \centering
         \includegraphics[scale=0.35]{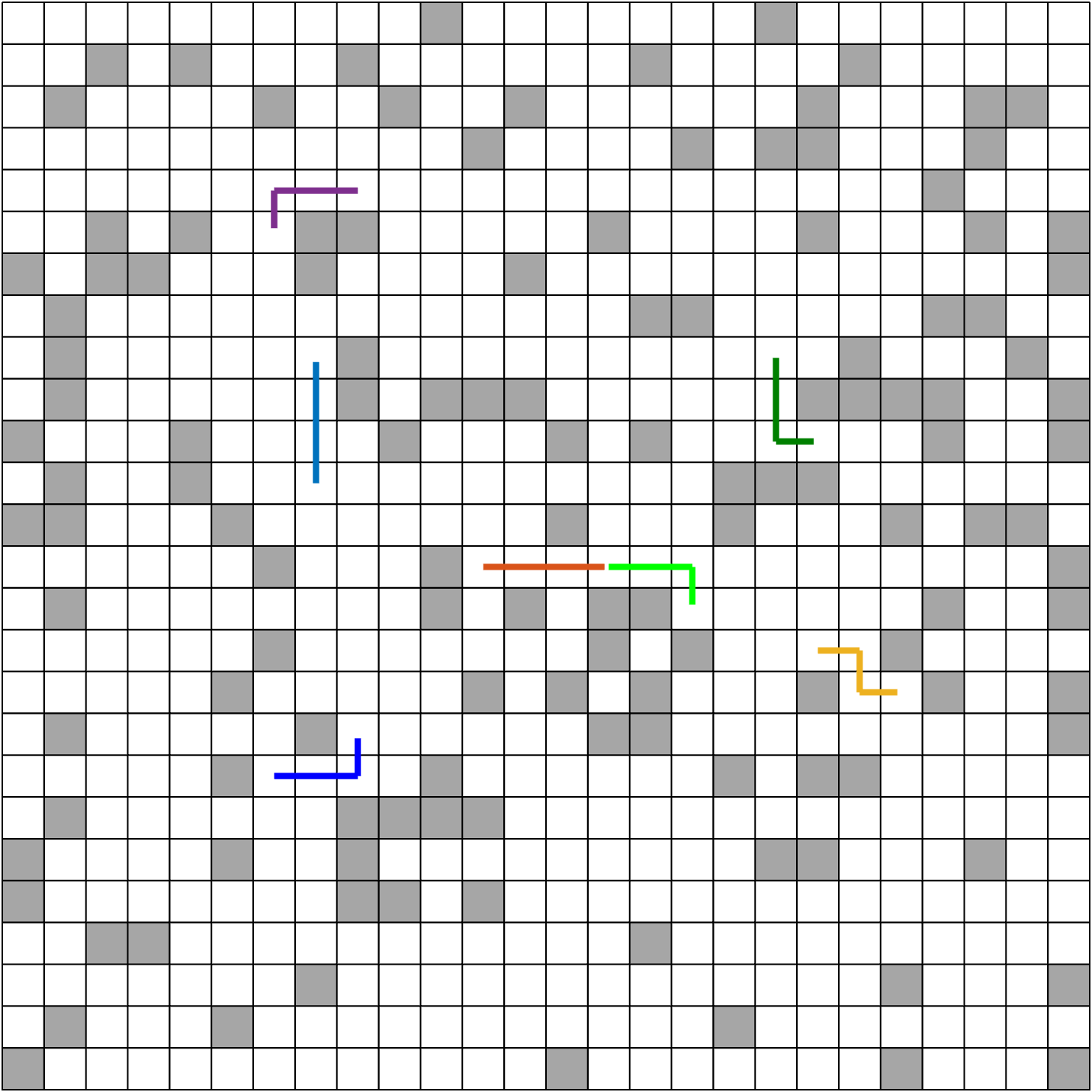}
         \caption{CBS, $\Delta k=[11,14]$}
         \label{fig:ln15_cbs_seg6}
     \end{subfigure}
     \hfill
     \begin{subfigure}{0.24\linewidth}
         \centering
         \includegraphics[scale=0.35]{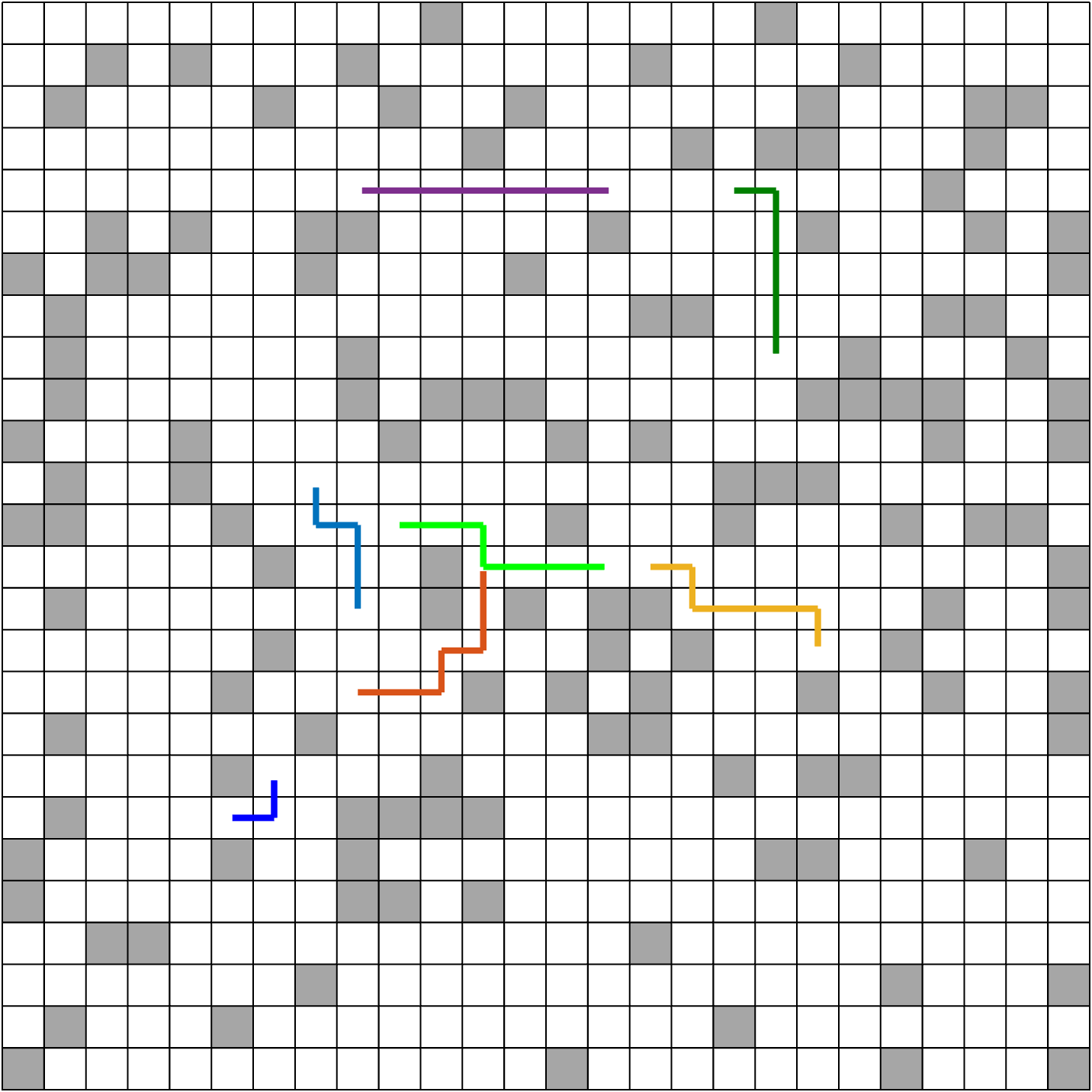}
         \caption{CBS, $\Delta k=[14,20]$}
         \label{fig:ln15_cbs_seg7}
     \end{subfigure}
     \hfill
     \newline
     \begin{subfigure}{0.24\linewidth}
         \centering
         \includegraphics[scale=0.35]{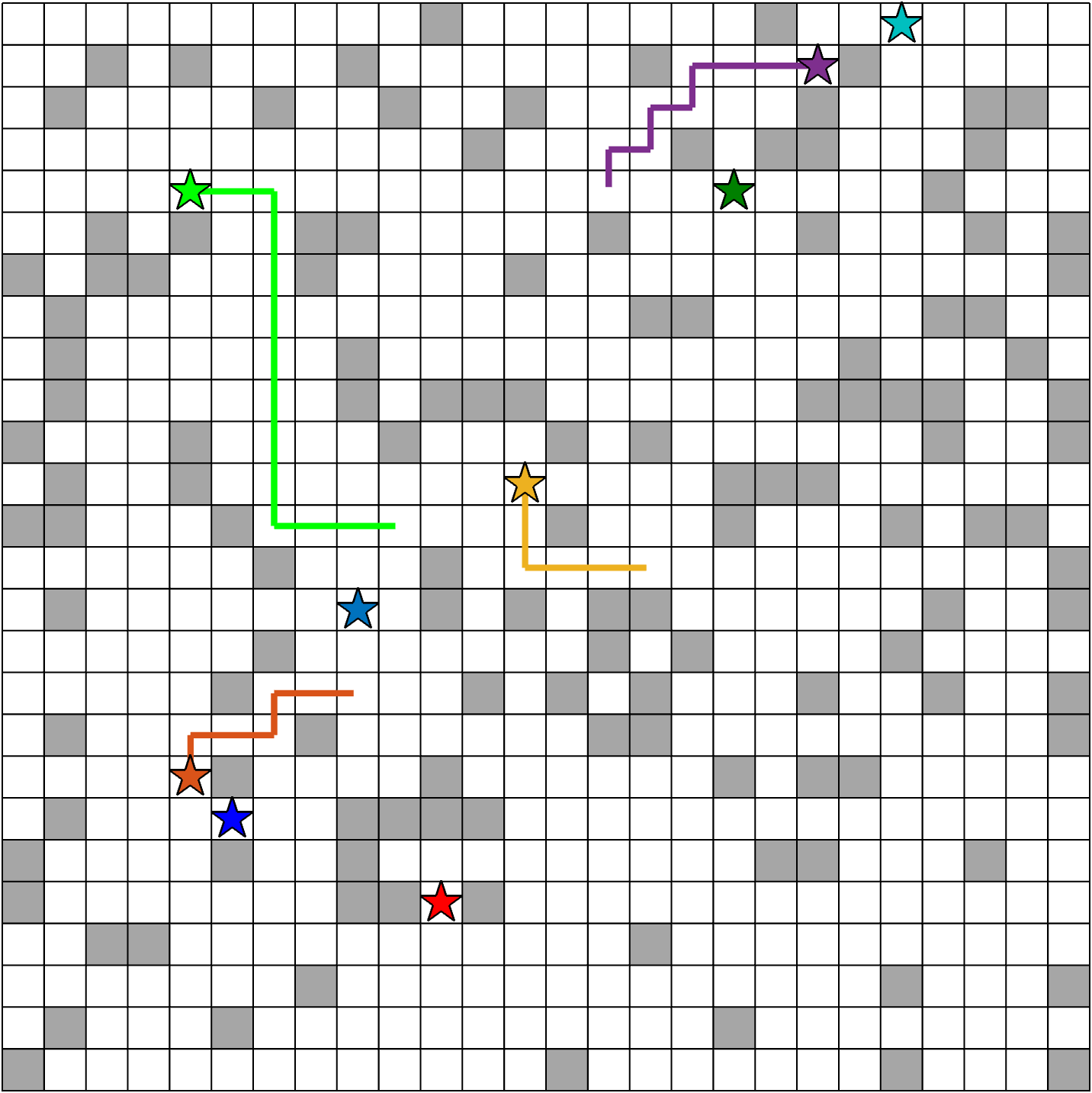}
         \caption{CBS, $\Delta k=[20,33]$}
         \label{fig:ln15_cbs_seg8}
     \end{subfigure}
     \hfill
     \begin{subfigure}{0.24\linewidth}
         \centering
         \includegraphics[scale=0.35]{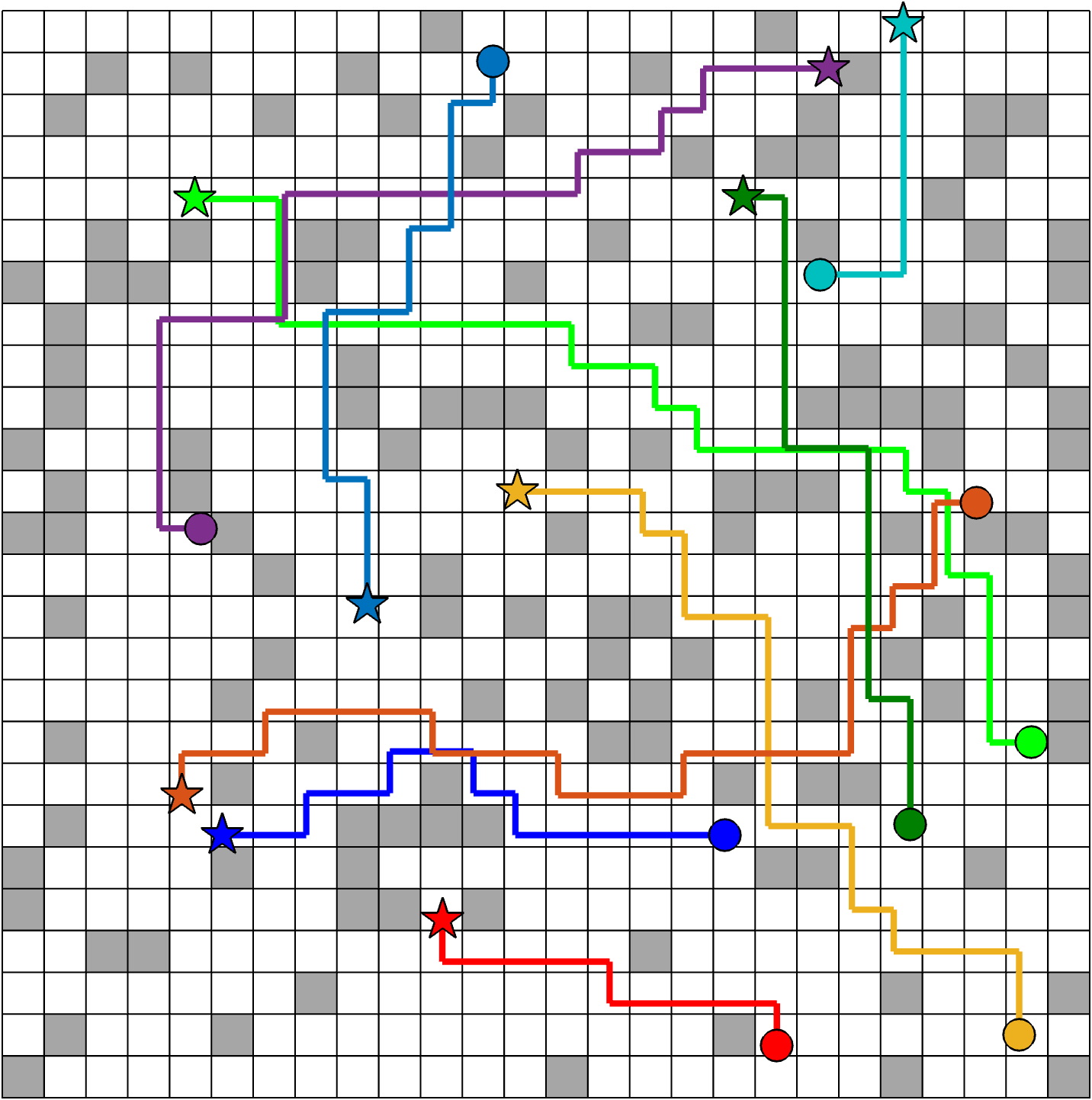}
         \caption{XG-CBS, $r=2$}
         \label{fig:ln15_xgcbs_full}
     \end{subfigure}
     \hfill
     \begin{subfigure}{0.24\linewidth}
         \centering
         \includegraphics[scale=0.35]{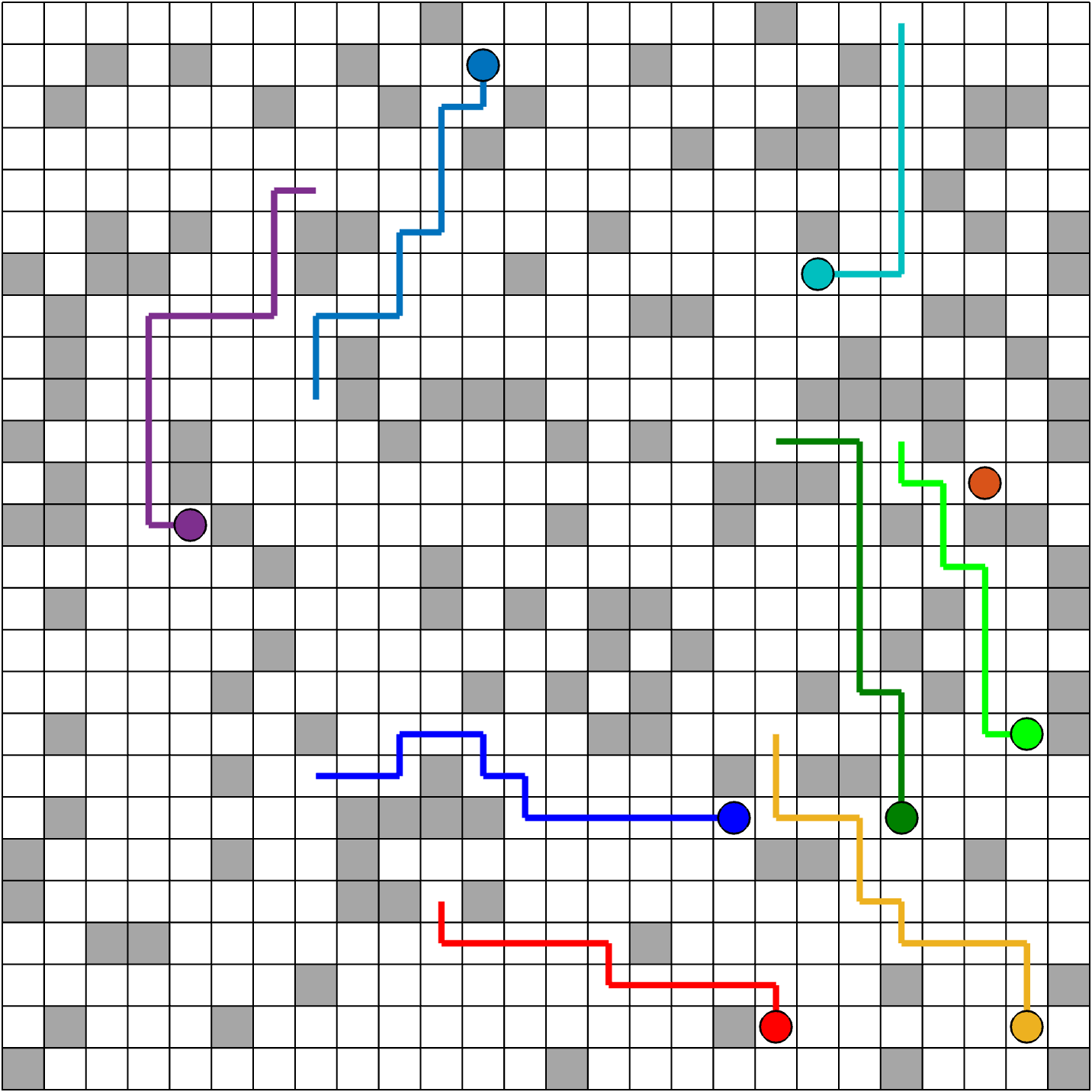}
         \caption{XG-CBS, $\Delta k=[0,13]$}
         \label{fig:ln15_xgcbs_seg1}
     \end{subfigure}
     \hfill
     \begin{subfigure}{0.24\linewidth}
         \centering
         \includegraphics[scale=0.35]{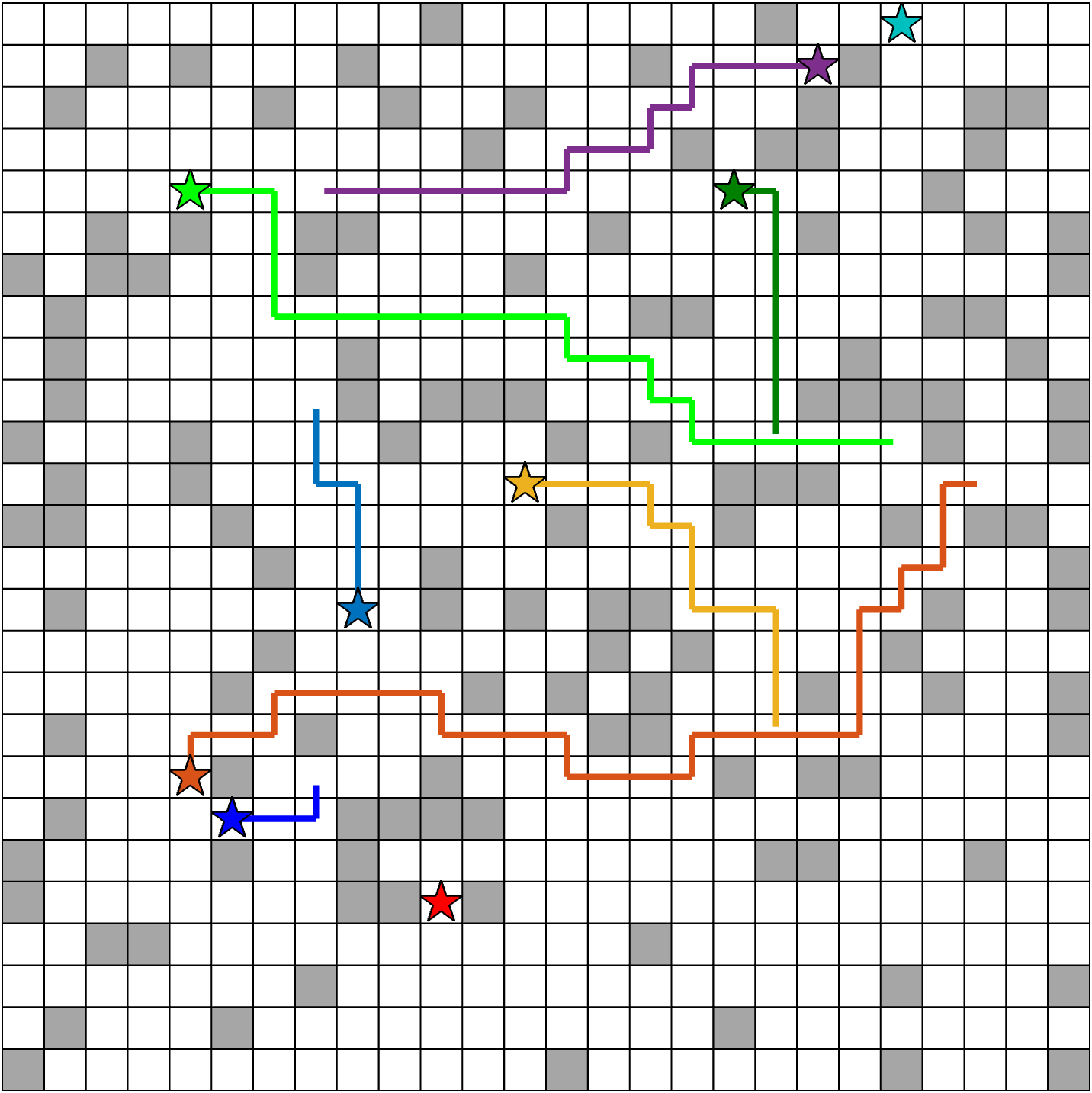}
         \caption{XG-CBS, $\Delta k=[13,43]$}
         \label{fig:ln15_xgcbs_seg2}
     \end{subfigure}
    % \caption{Line $36$ of Table~\ref{tab:final_benchmark}}
    \caption{Example of XG-CBS, reducing 7-segment plan of CBS to a 2-segment plan.}
    \label{fig:ln_15} % numbers in labels represent row of Table 1 -- numbers in document should be from Table 2
\end{figure*}

The decreasing segments becomes increasingly important as the number of agents rises. For example, the CBS solution to a nine agent MAPF problem in a $26\times 26$ grid world (Figure~\ref{fig:ln15_cbs_full}) requires $8$ segments to explain (Figure~\ref{fig:ln15_cbs_seg1}-\ref{fig:ln15_cbs_seg8}). More interestingly, many of the segments show tiny intervals of the plan. XG-CBS with XG-$A^*$ returns an alternative solution, shown in Figure~\ref{fig:ln15_xgcbs_full}, that mitigates the length of the explanation scheme. It presents a plan that only requires two segments (Figure~\ref{fig:ln15_xgcbs_seg1}-\ref{fig:ln15_xgcbs_seg2}). 

\begin{figure*}[p]
     \centering
     \begin{subfigure}{0.19\linewidth}
         \centering
         \includegraphics[scale=0.275]{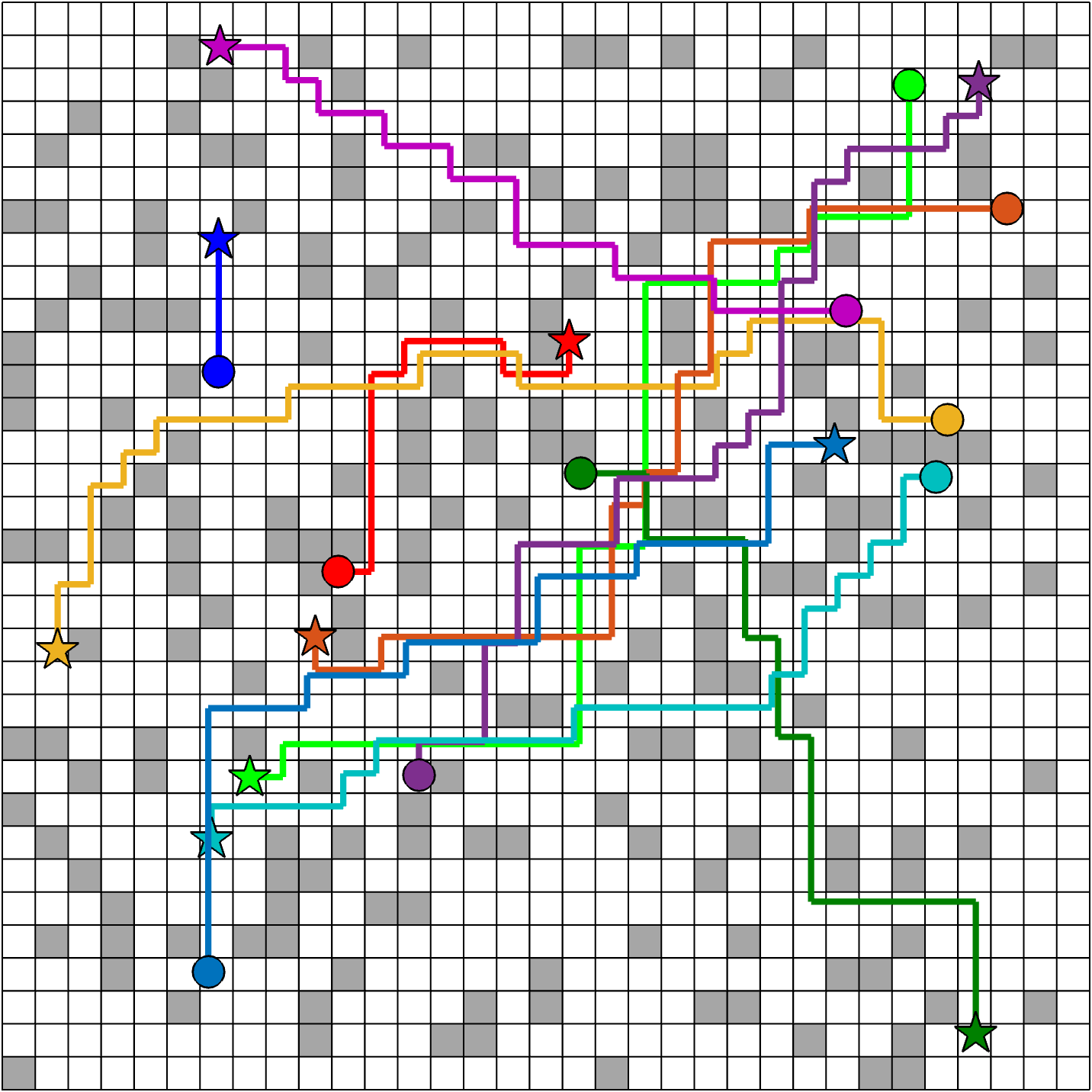}
         \caption{CBS}
         \label{fig:ln18_cbs_full}
     \end{subfigure}
     \hfill
     \begin{subfigure}{0.19\linewidth}
         \centering
         \includegraphics[scale=0.275]{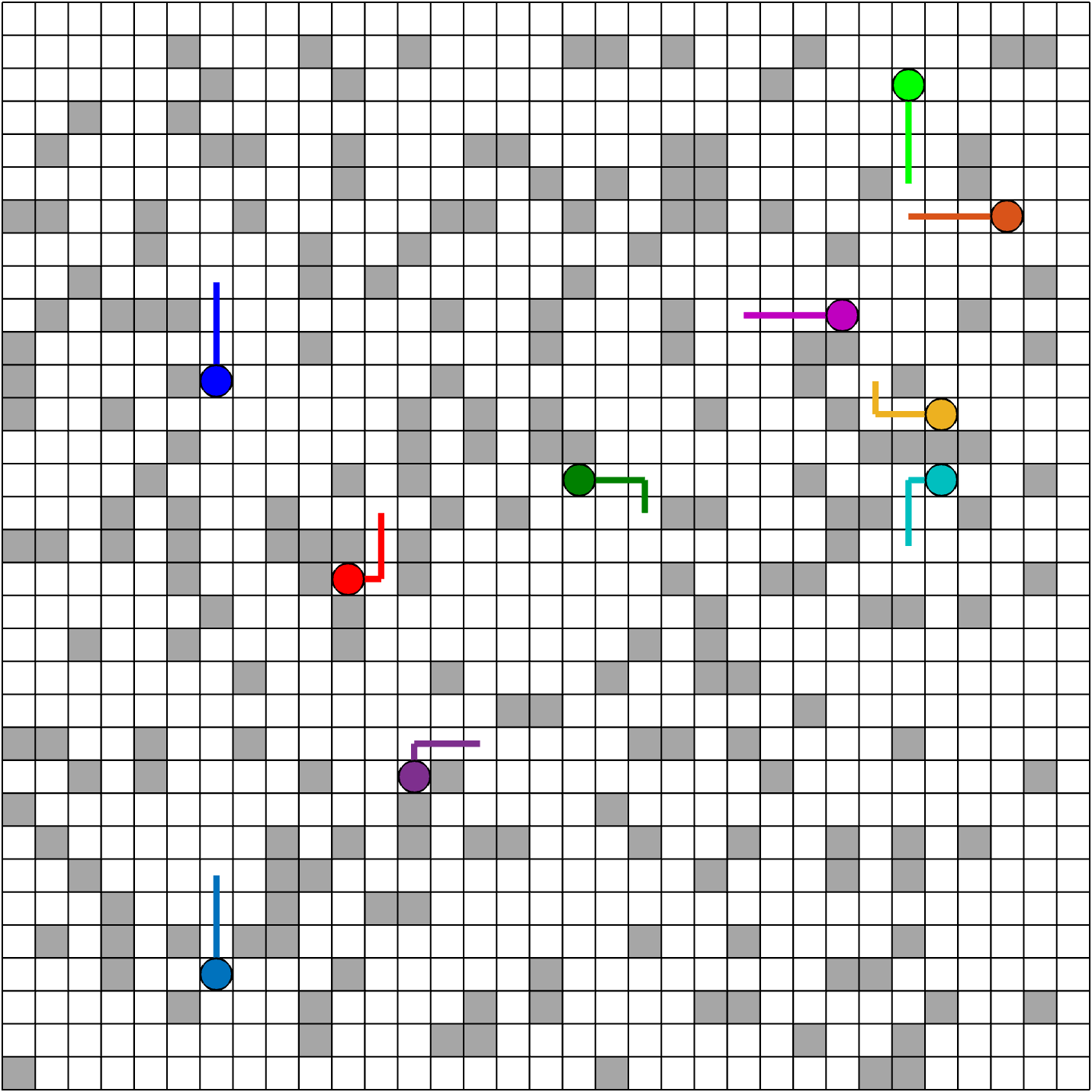}
         \caption{CBS, $\Delta k=[0,3]$}
         \label{fig:ln18_cbs_seg1}
     \end{subfigure}
     \hfill
     \begin{subfigure}{0.19\linewidth}
         \centering
         \includegraphics[scale=0.275]{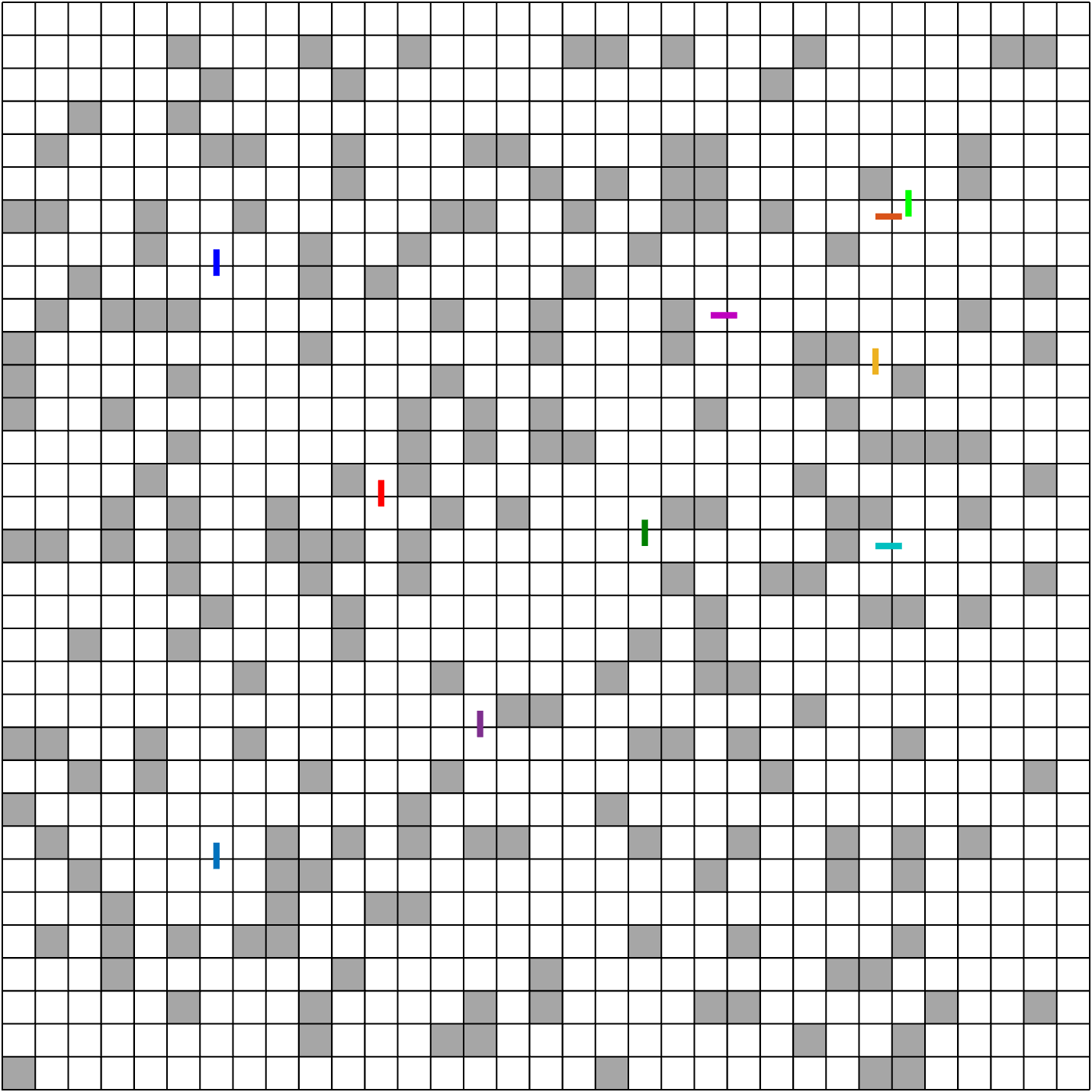}
         \caption{CBS, $\Delta k=[3,4]$}
         \label{fig:ln18_cbs_seg2}
     \end{subfigure}
     \hfill
     \begin{subfigure}{0.19\linewidth}
         \centering
         \includegraphics[scale=0.275]{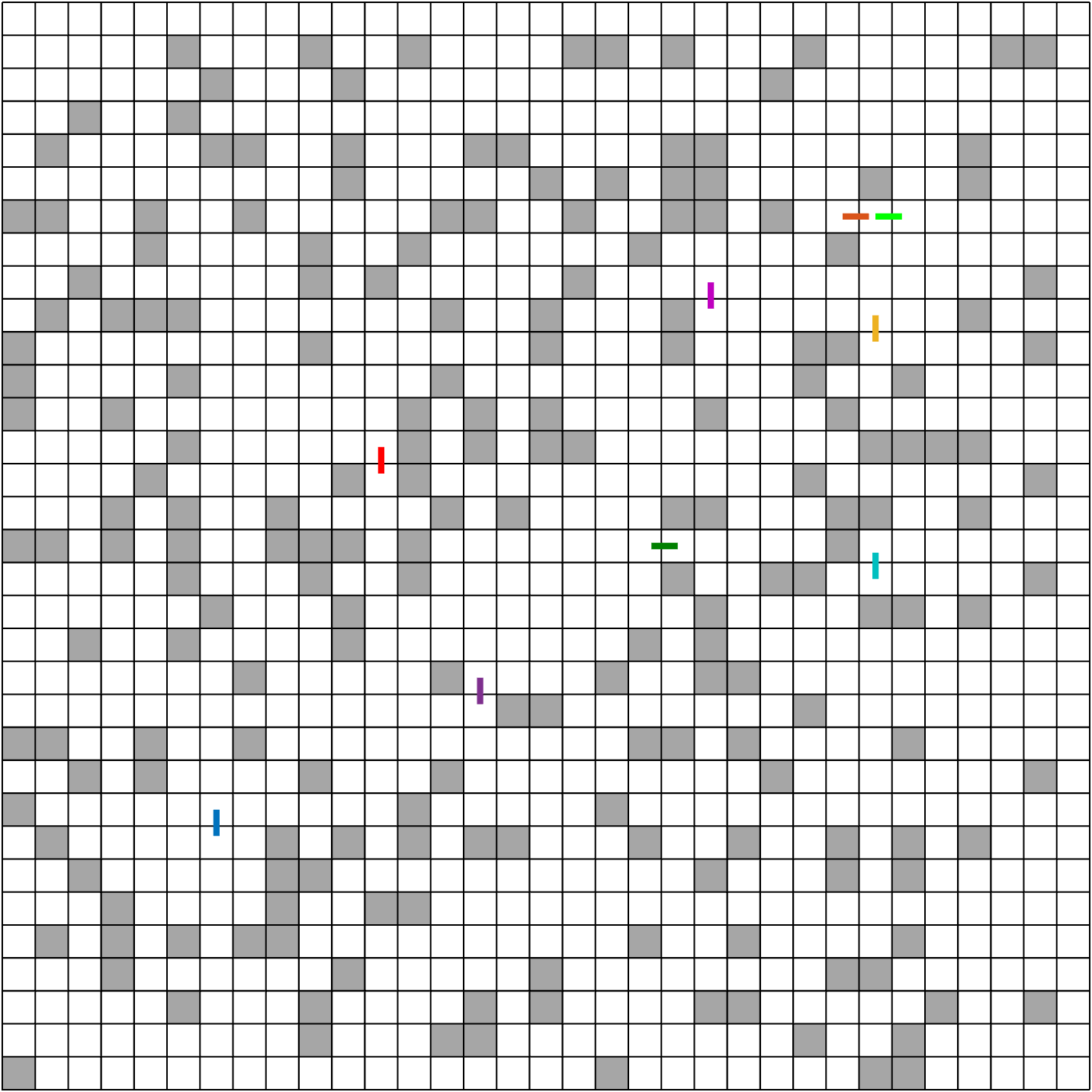}
         \caption{CBS, $\Delta k=[4,5]$}
         \label{fig:ln18_cbs_seg3}
     \end{subfigure}
     \hfill
     \begin{subfigure}{0.19\linewidth}
         \centering
         \includegraphics[scale=0.275]{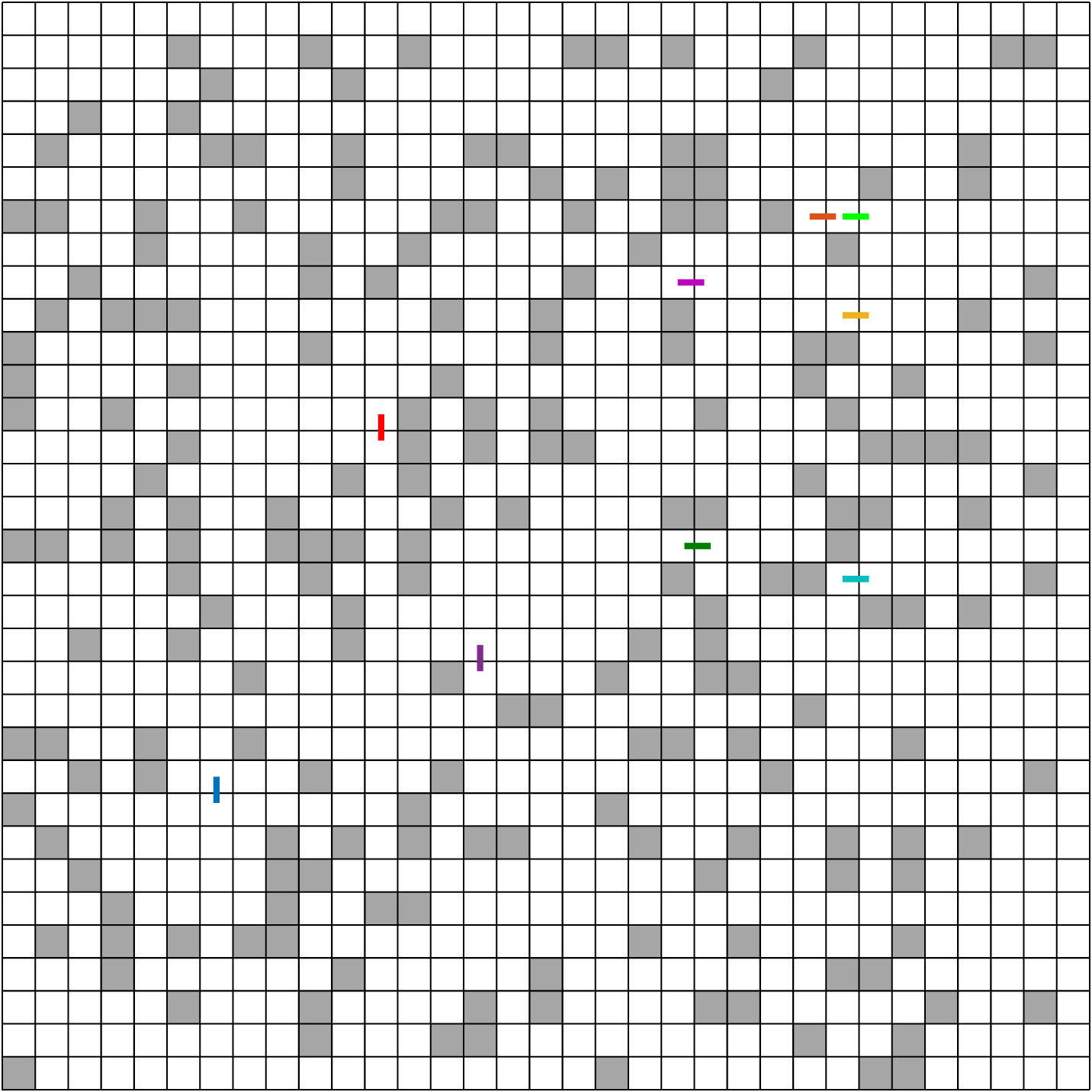}
         \caption{CBS, $\Delta k=[5,6]$}
         \label{fig:ln18_cbs_seg4}
     \end{subfigure}
     \hfill
     \newline
     \begin{subfigure}{0.19\linewidth}
         \centering
         \includegraphics[scale=0.275]{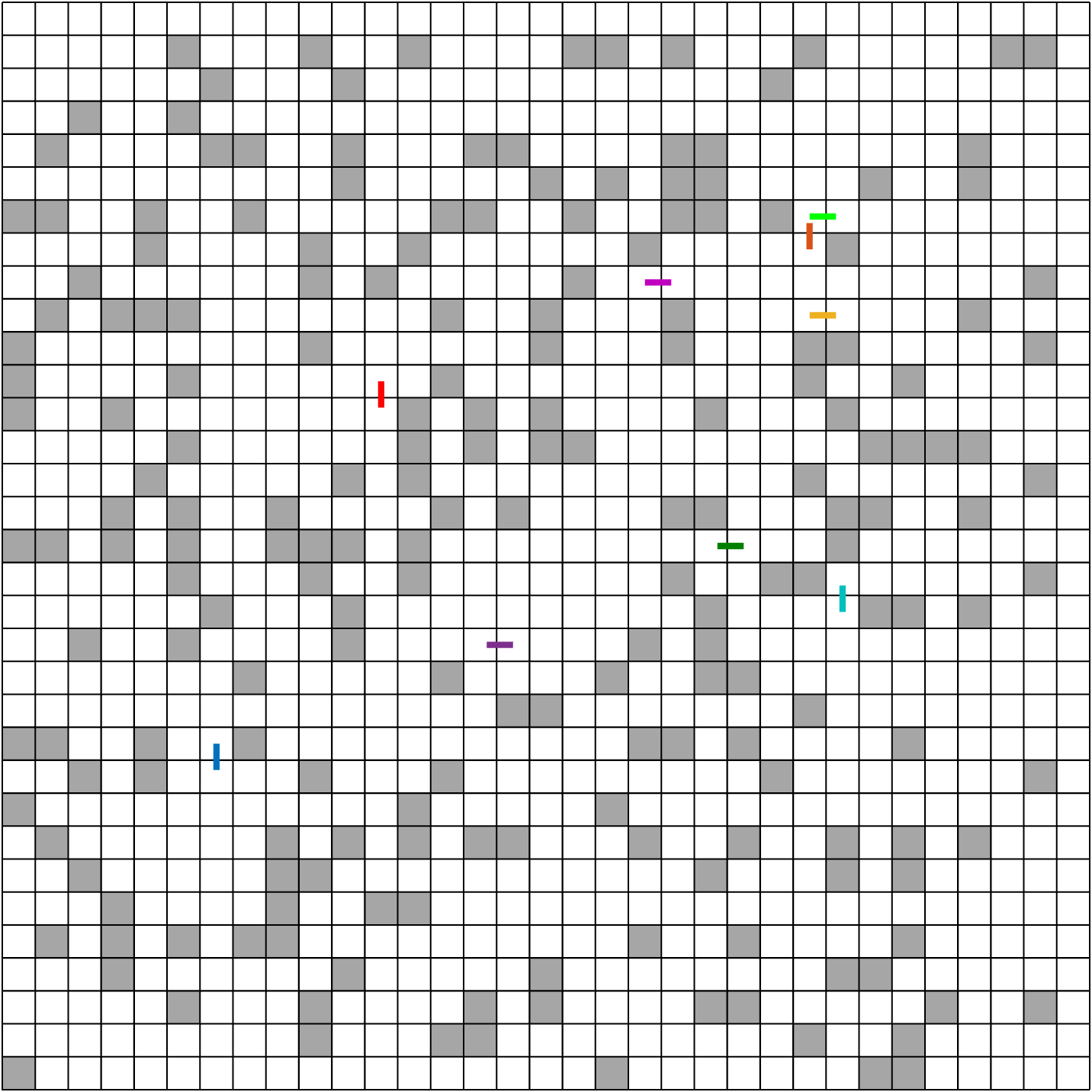}
         \caption{CBS, $\Delta k=[6,7]$}
         \label{fig:ln18_cbs_seg5}
     \end{subfigure}
     \hfill
     \begin{subfigure}{0.19\linewidth}
         \centering
         \includegraphics[scale=0.275]{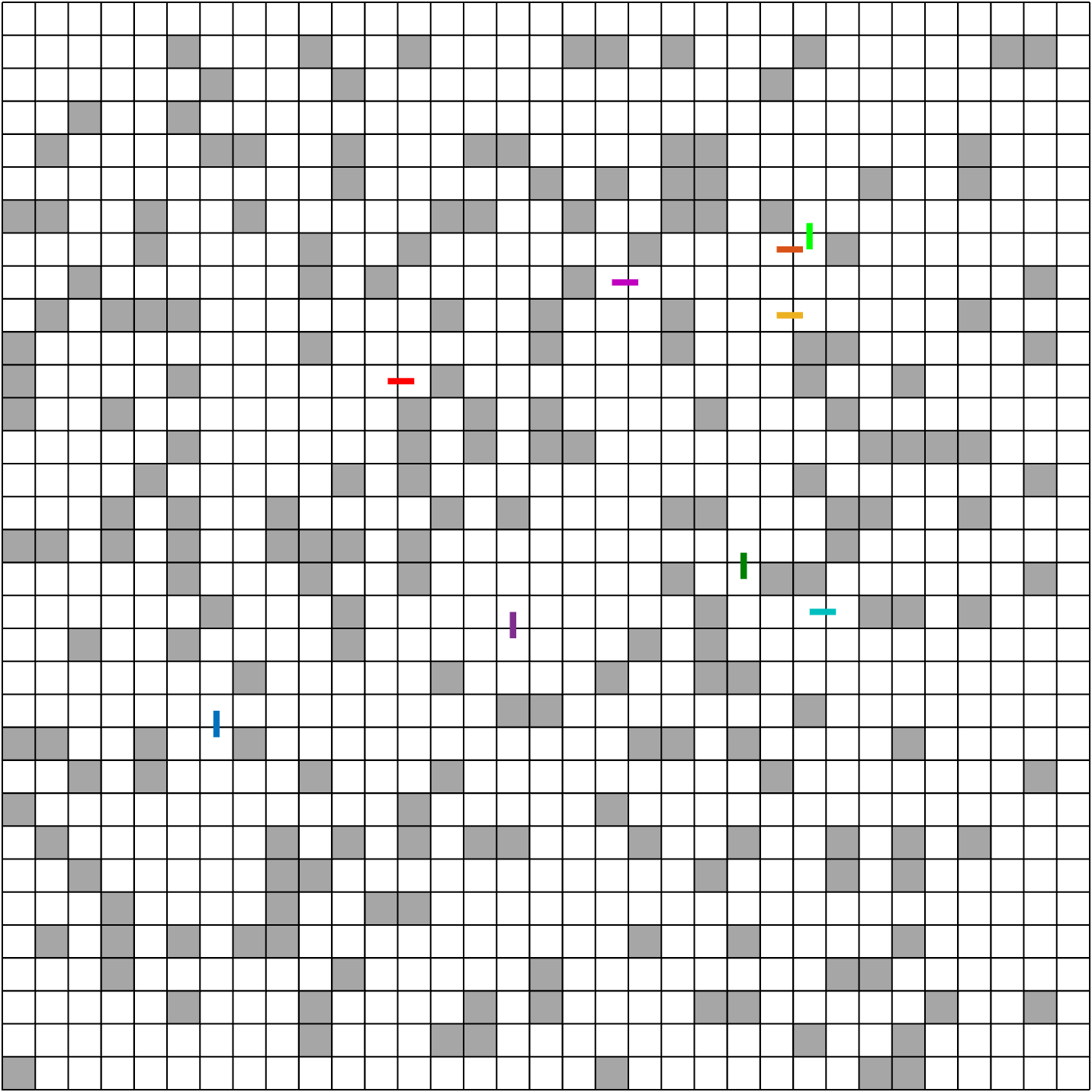}
         \caption{CBS, $\Delta k=[7,8]$}
         \label{fig:ln18_cbs_seg6}
     \end{subfigure}
     \hfill
     \begin{subfigure}{0.19\linewidth}
         \centering
         \includegraphics[scale=0.275]{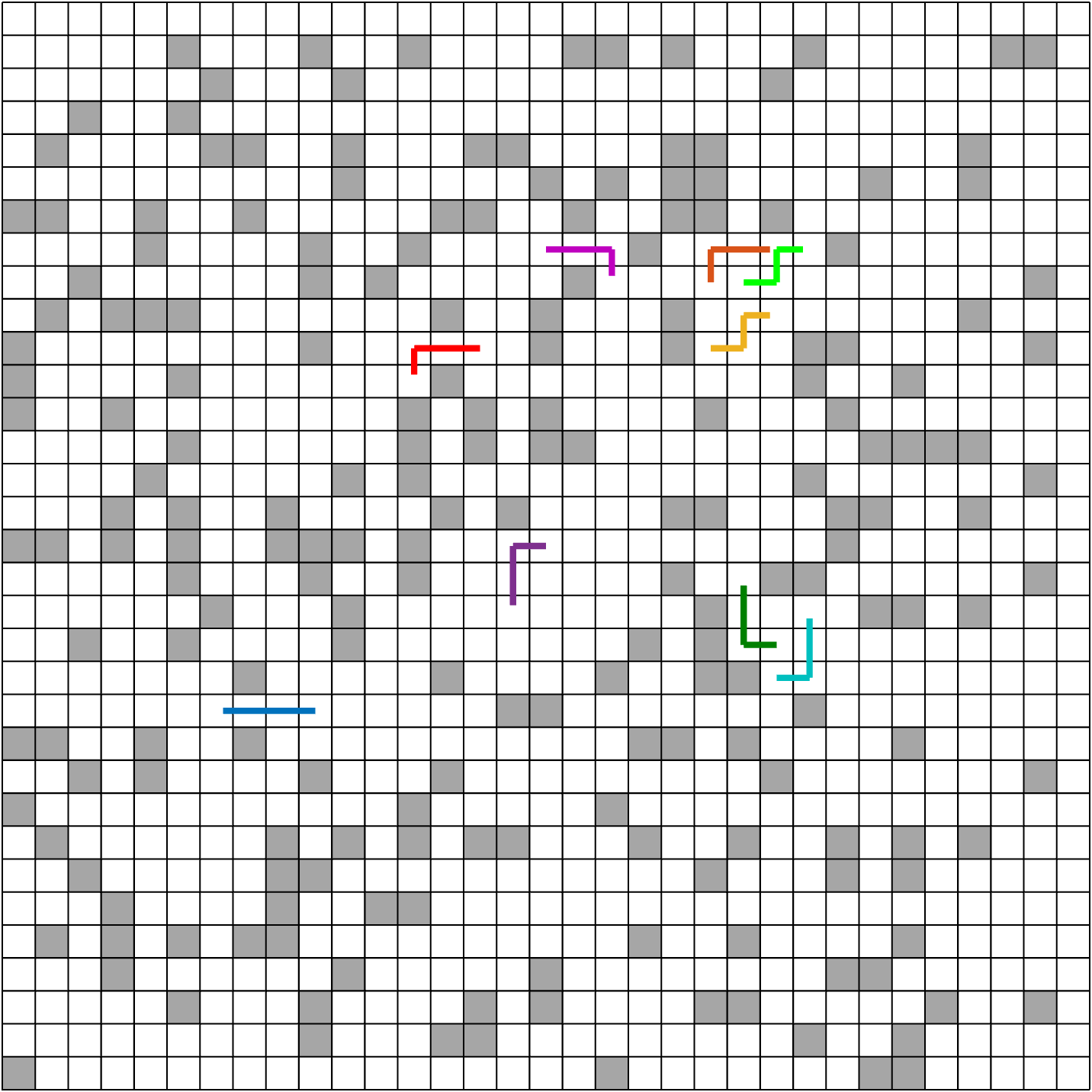}
         \caption{CBS, $\Delta k=[8,11]$}
         \label{fig:ln18_cbs_seg7}
     \end{subfigure}
     \hfill
     \begin{subfigure}{0.19\linewidth}
         \centering
         \includegraphics[scale=0.275]{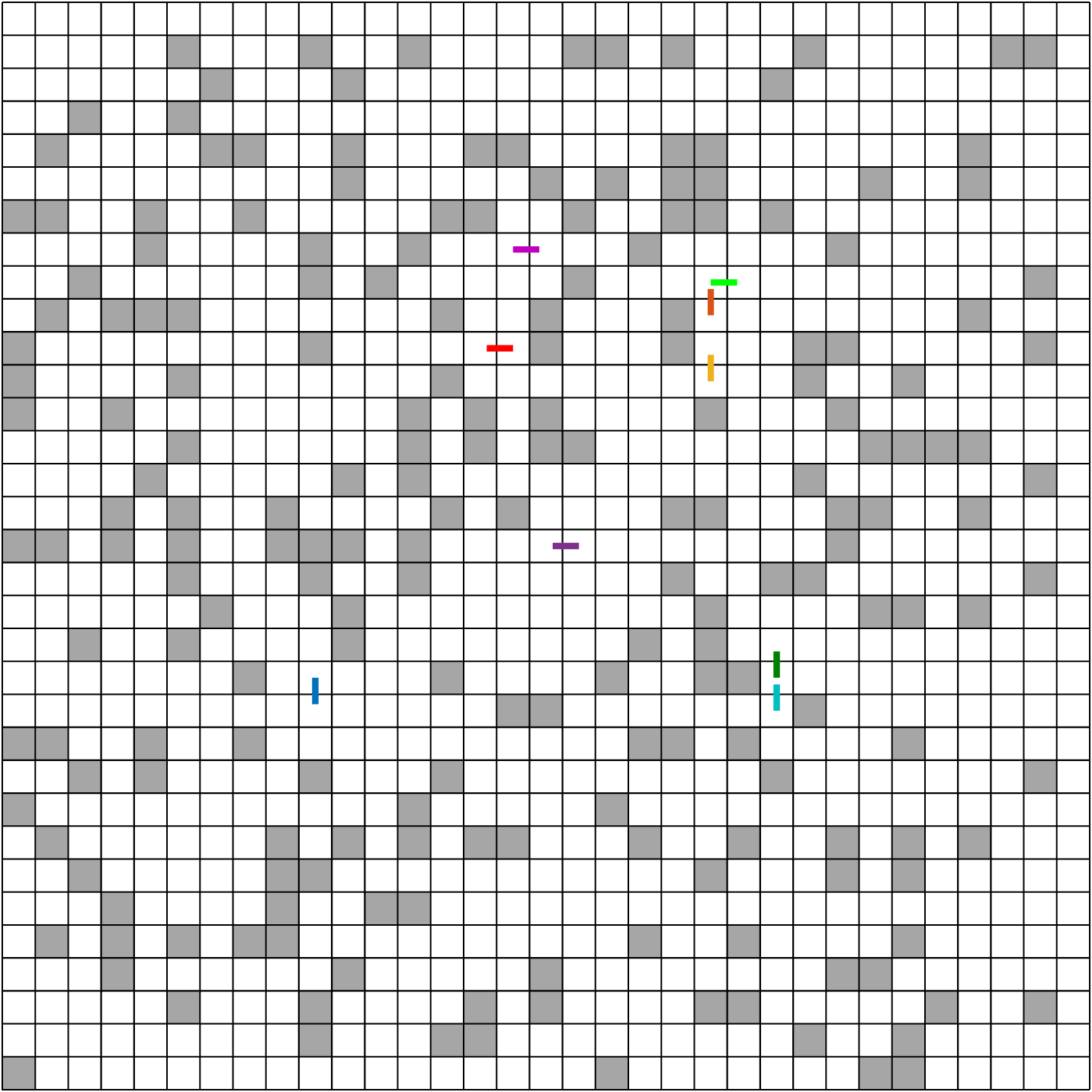}
         \caption{CBS, $\Delta k=[11,12]$}
         \label{fig:ln18_cbs_seg8}
     \end{subfigure}
     \hfill
     \begin{subfigure}{0.19\linewidth}
         \centering
         \includegraphics[scale=0.275]{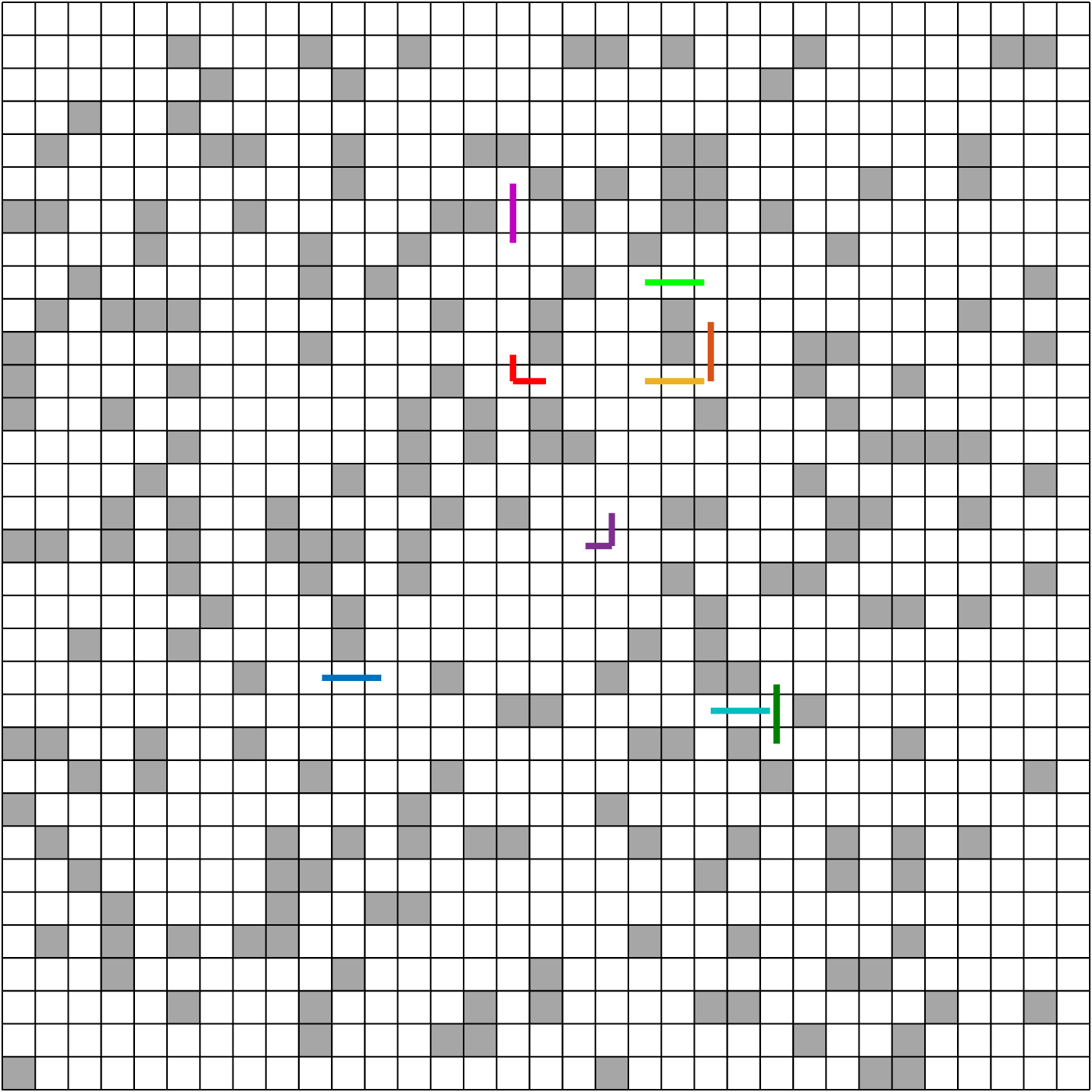}
         \caption{CBS, $\Delta k=[12,14]$}
         \label{fig:ln18_cbs_seg9}
     \end{subfigure}
     \hfill
     \newline
    \begin{subfigure}{0.19\linewidth}
         \centering
         \includegraphics[scale=0.275]{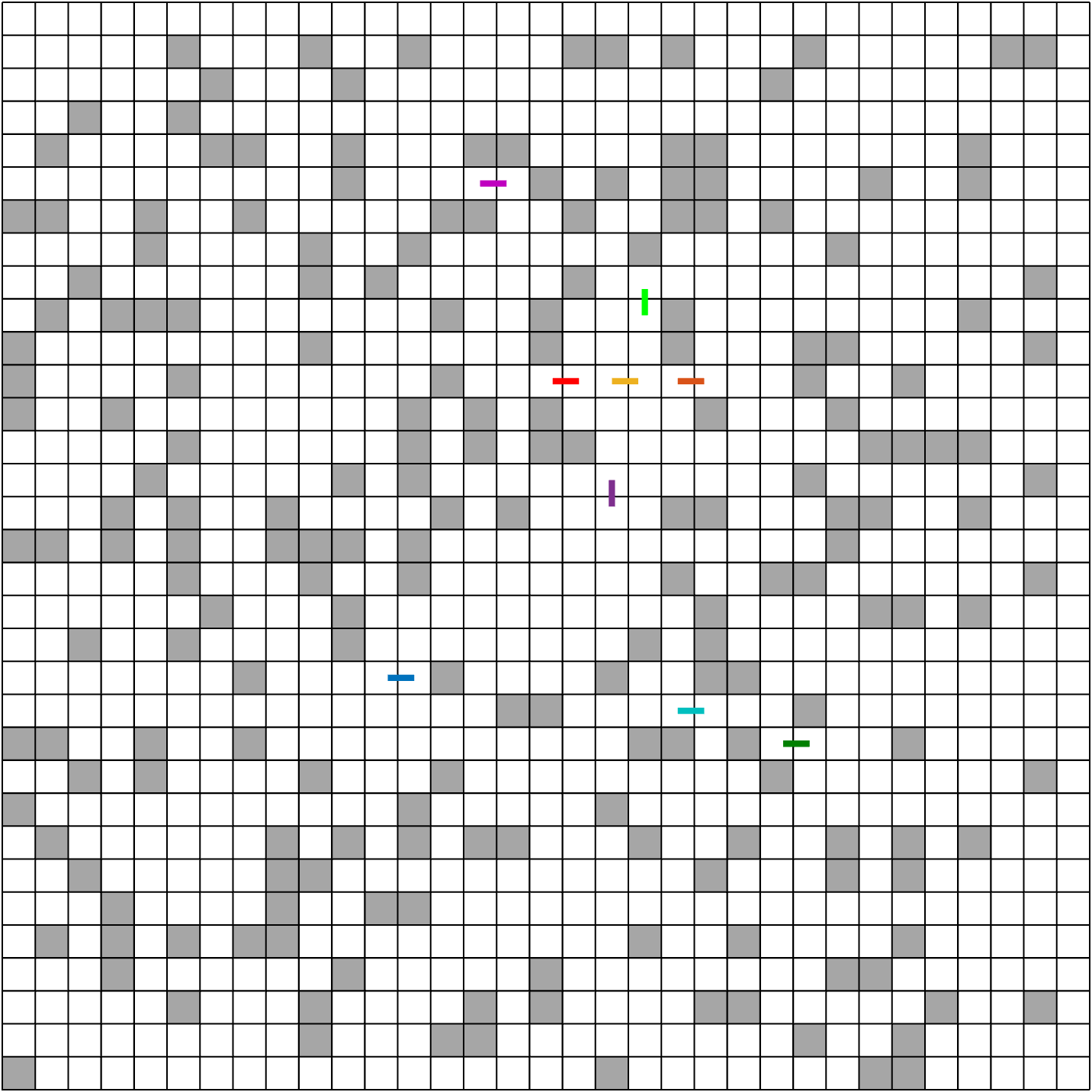}
         \caption{CBS, $\Delta k=[14,15]$}
         \label{fig:ln18_cbs_seg10}
     \end{subfigure}
     \hfill
     \begin{subfigure}{0.19\linewidth}
         \centering
         \includegraphics[scale=0.275]{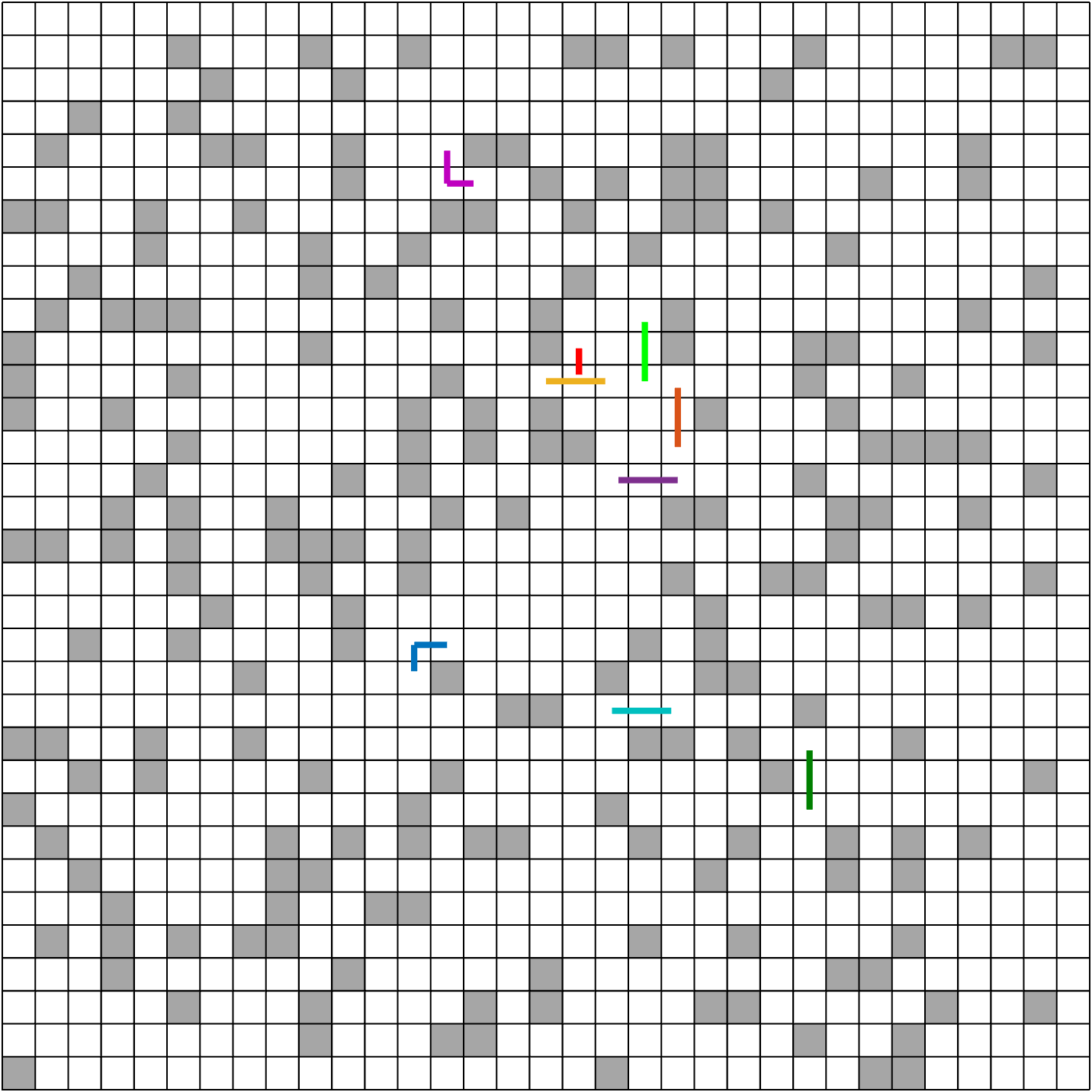}
         \caption{CBS, $\Delta k=[15,17]$}
         \label{fig:ln18_cbs_seg11}
     \end{subfigure}
     \hfill
     \begin{subfigure}{0.19\linewidth}
         \centering
         \includegraphics[scale=0.275]{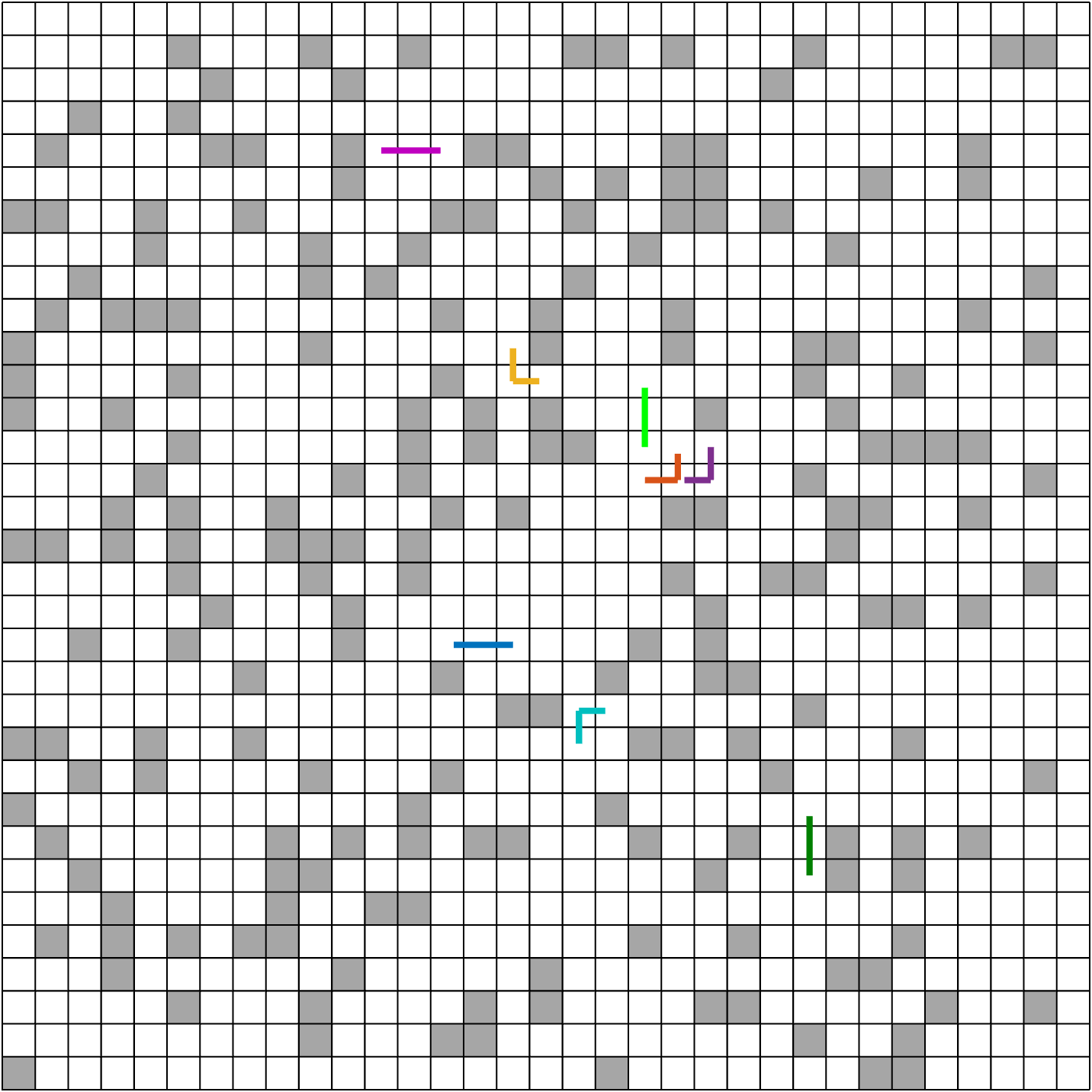}
         \caption{CBS, $\Delta k=[17,19]$}
         \label{fig:ln18_cbs_seg12}
     \end{subfigure}
     \hfill
     \begin{subfigure}{0.19\linewidth}
         \centering
         \includegraphics[scale=0.275]{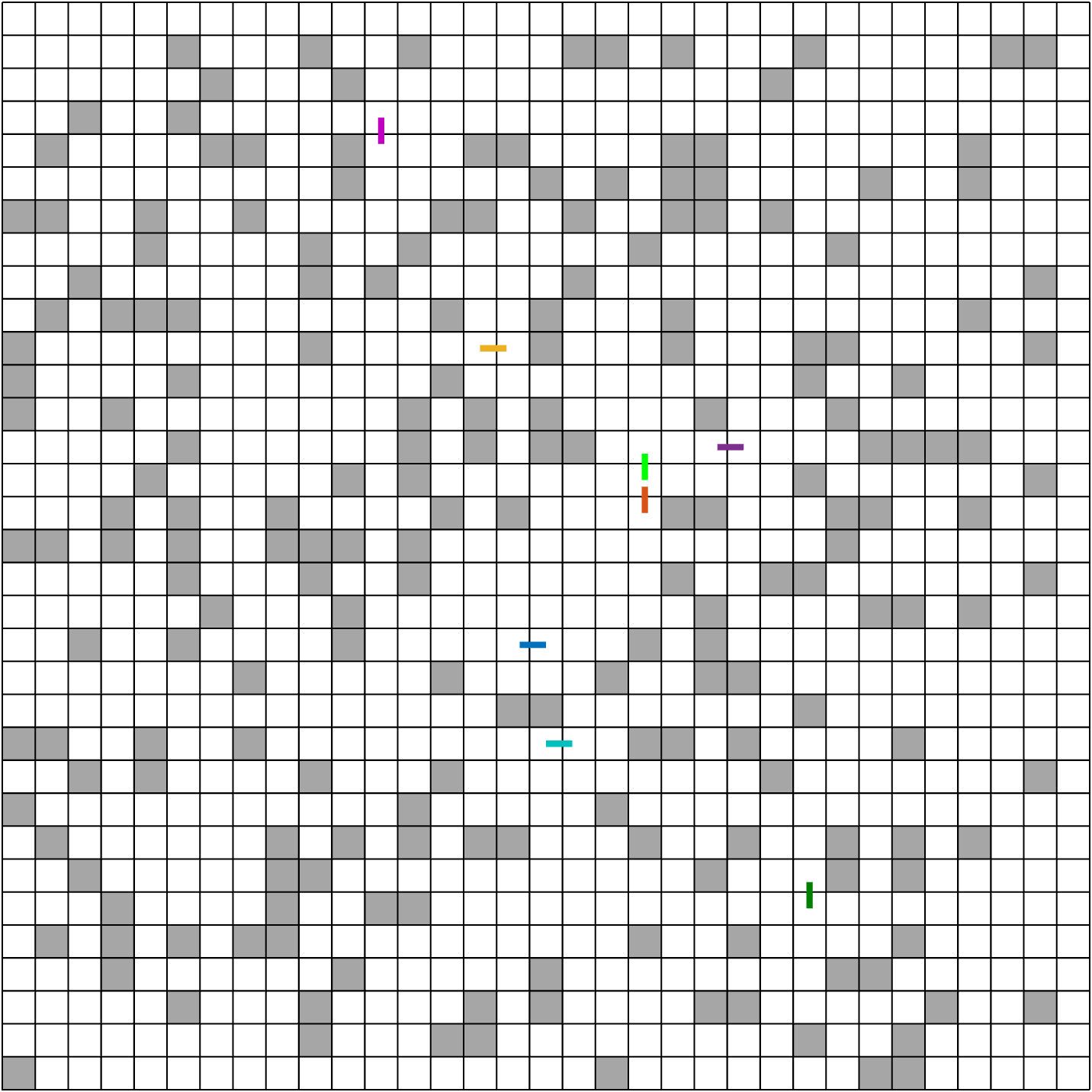}
         \caption{CBS, $\Delta k=[19,20]$}
         \label{fig:ln18_cbs_seg13}
     \end{subfigure}
     \hfill
     \begin{subfigure}{0.19\linewidth}
         \centering
         \includegraphics[scale=0.275]{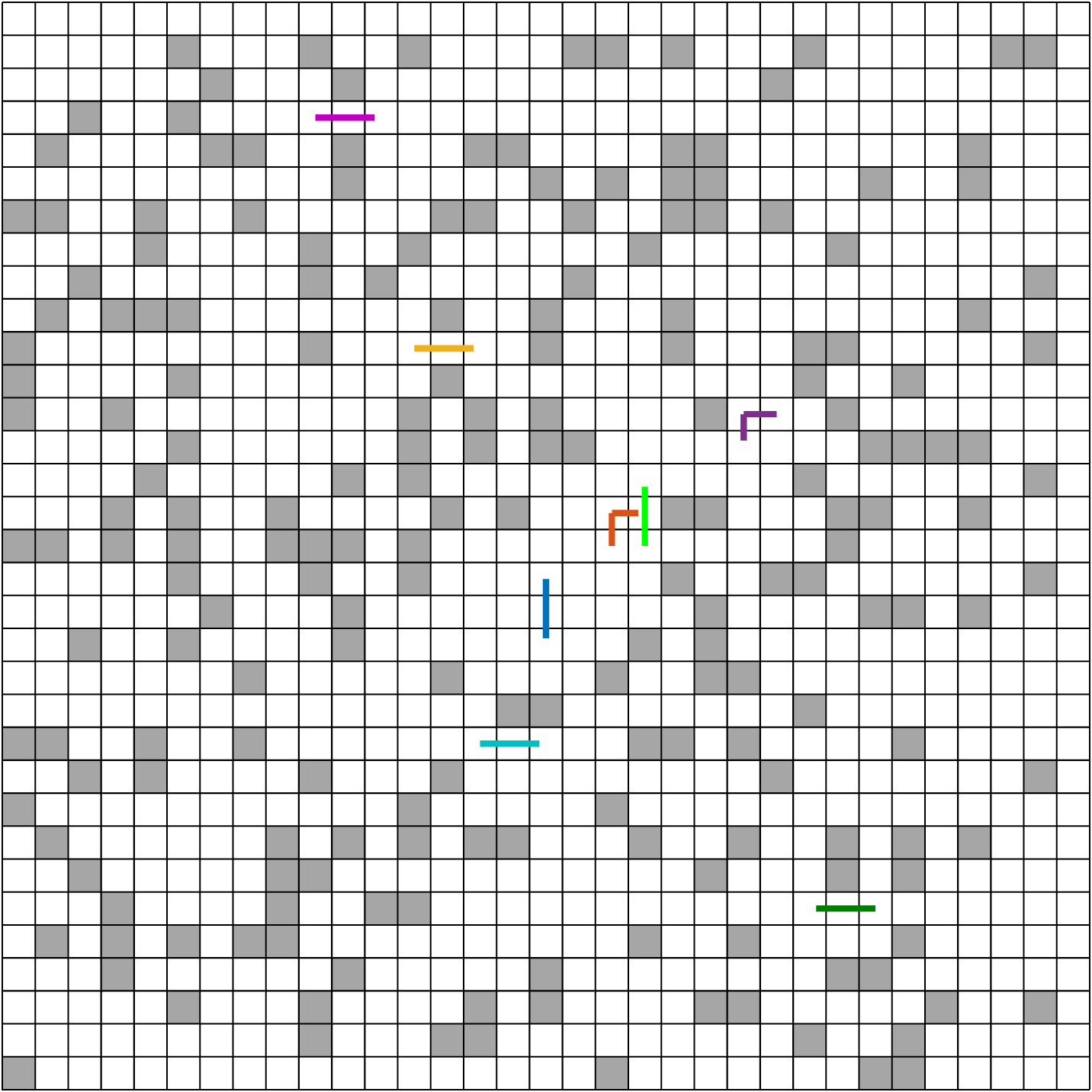}
         \caption{CBS, $\Delta k=[20,22]$}
         \label{fig:ln18_cbs_seg14}
     \end{subfigure}
     \hfill
     \newline
     \begin{subfigure}{0.19\linewidth}
         \centering
         \includegraphics[scale=0.275]{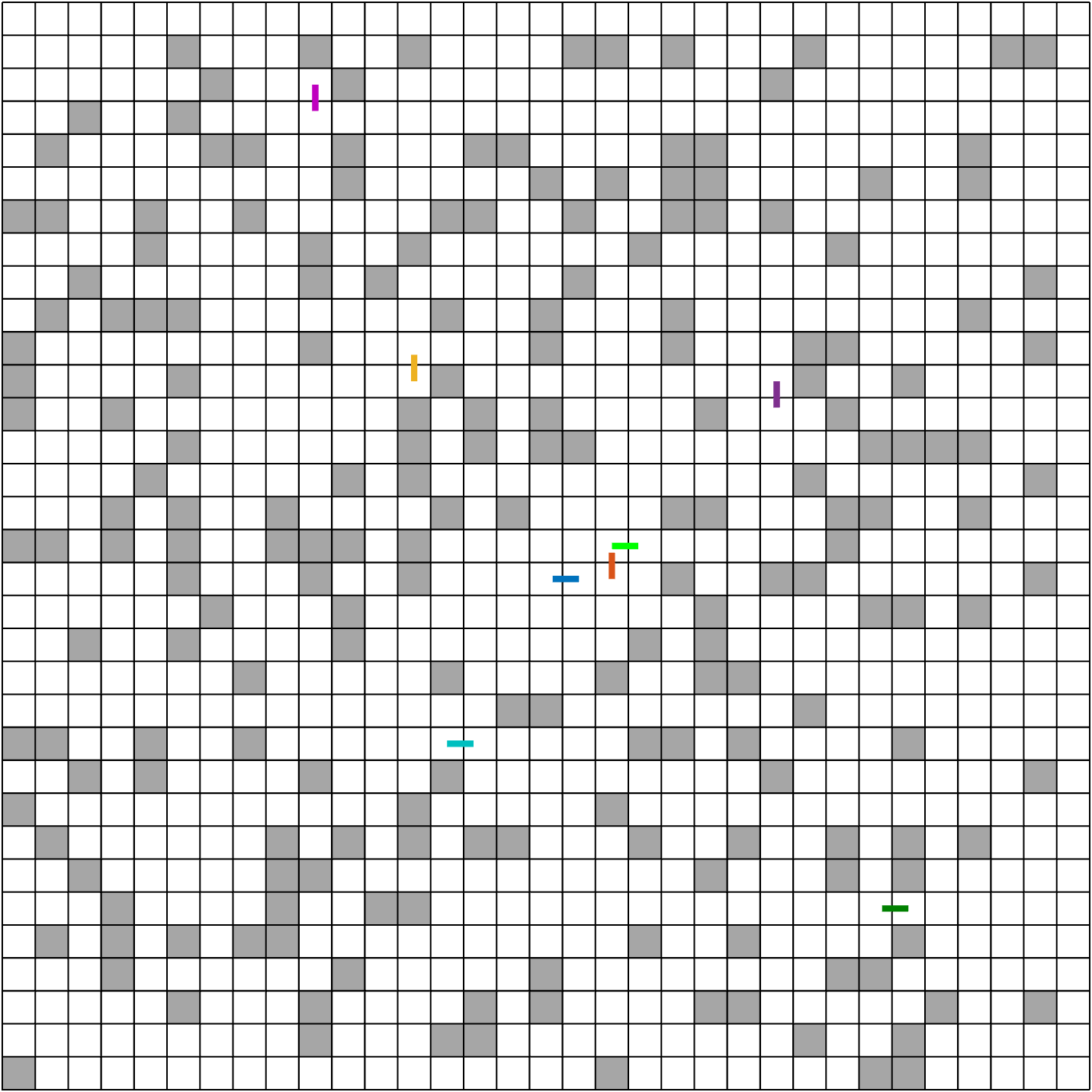}
         \caption{CBS, $\Delta k=[22,23]$}
         \label{fig:ln18_cbs_seg15}
     \end{subfigure}
     \hfill
     \begin{subfigure}{0.19\linewidth}
         \centering
         \includegraphics[scale=0.275]{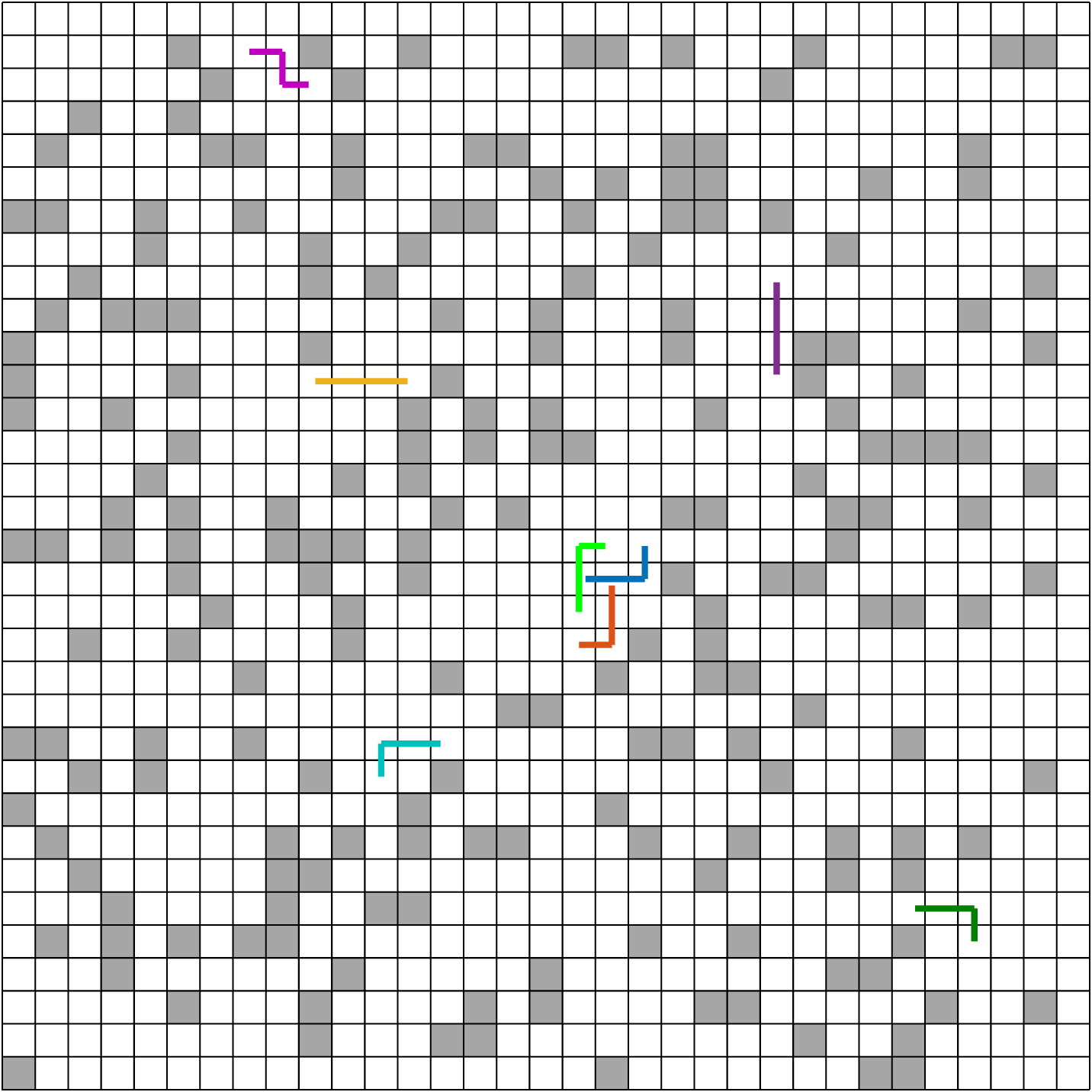}
         \caption{CBS, $\Delta k=[23,26]$}
         \label{fig:ln18_cbs_seg16}
     \end{subfigure}
     \hfill
     \begin{subfigure}{0.19\linewidth}
         \centering
         \includegraphics[scale=0.275]{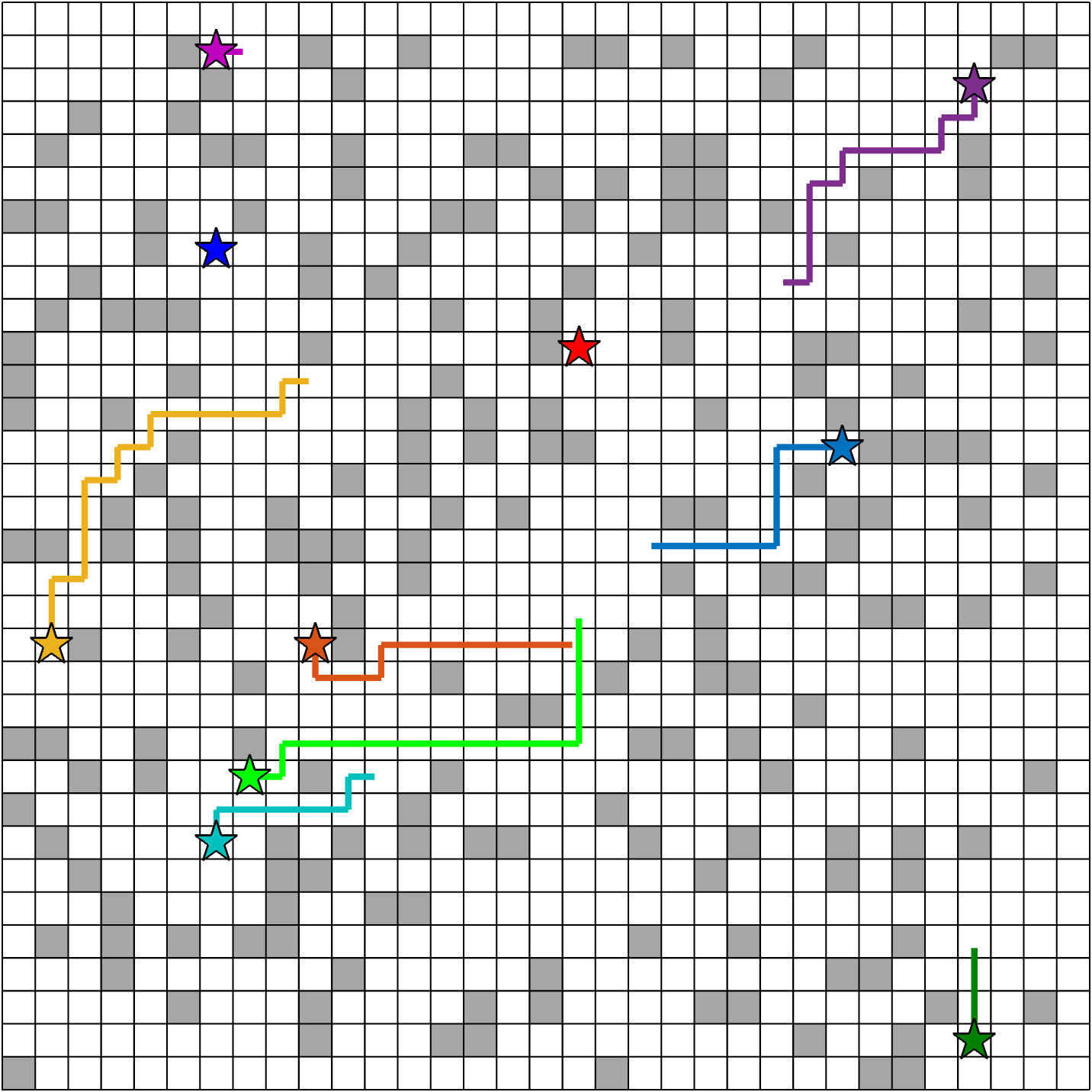}
         \caption{CBS, $\Delta k=[26,42]$}
         \label{fig:ln18_cbs_seg17}
     \end{subfigure}
     \hfill
     \begin{subfigure}{0.19\linewidth}
         \centering
         \includegraphics[scale=0.275]{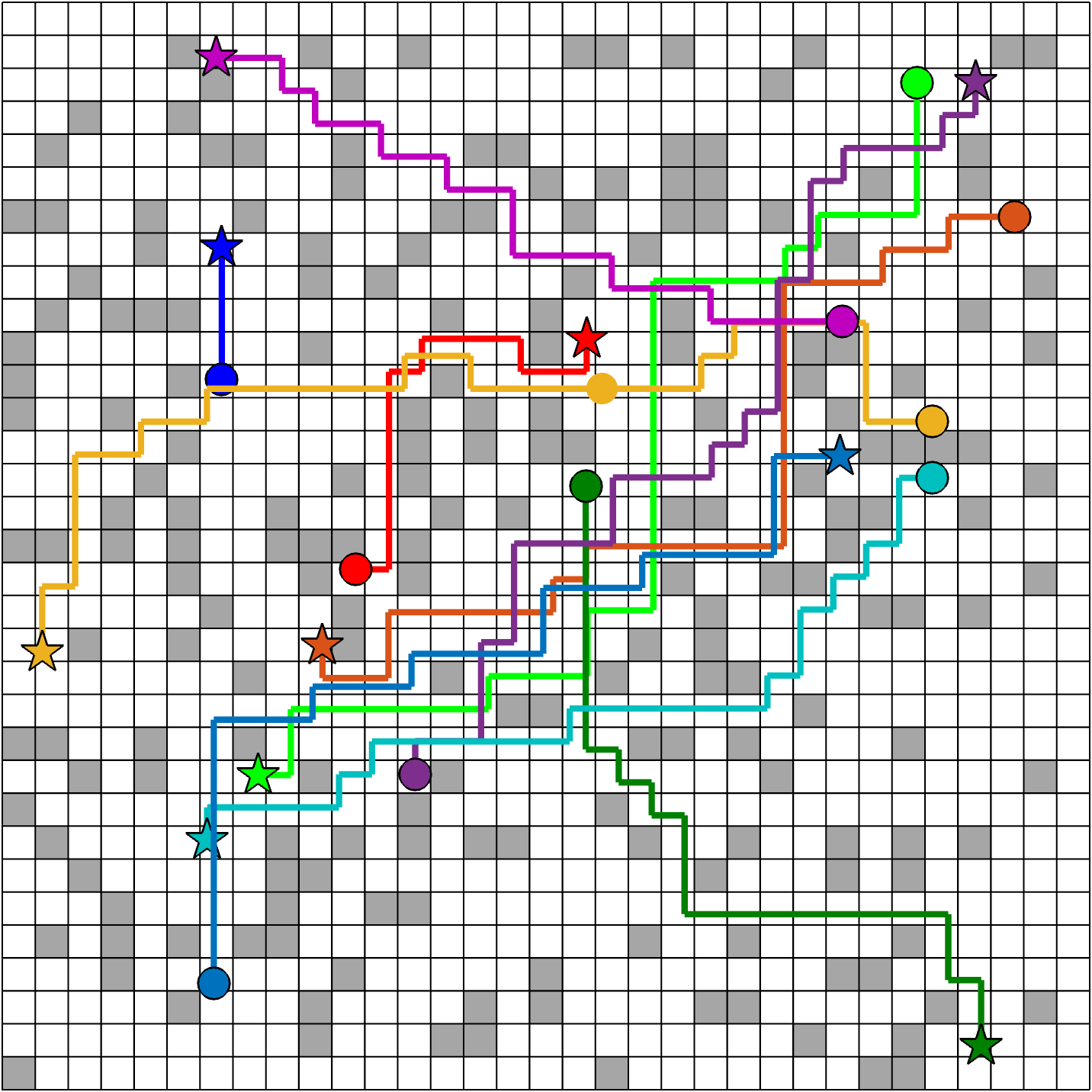}
         \caption{XG-CBS, $r=6$}
         \label{fig:ln18_xgcbs_full}
     \end{subfigure}
     \hfill
     \begin{subfigure}{0.19\linewidth}
         \centering
         \includegraphics[scale=0.275]{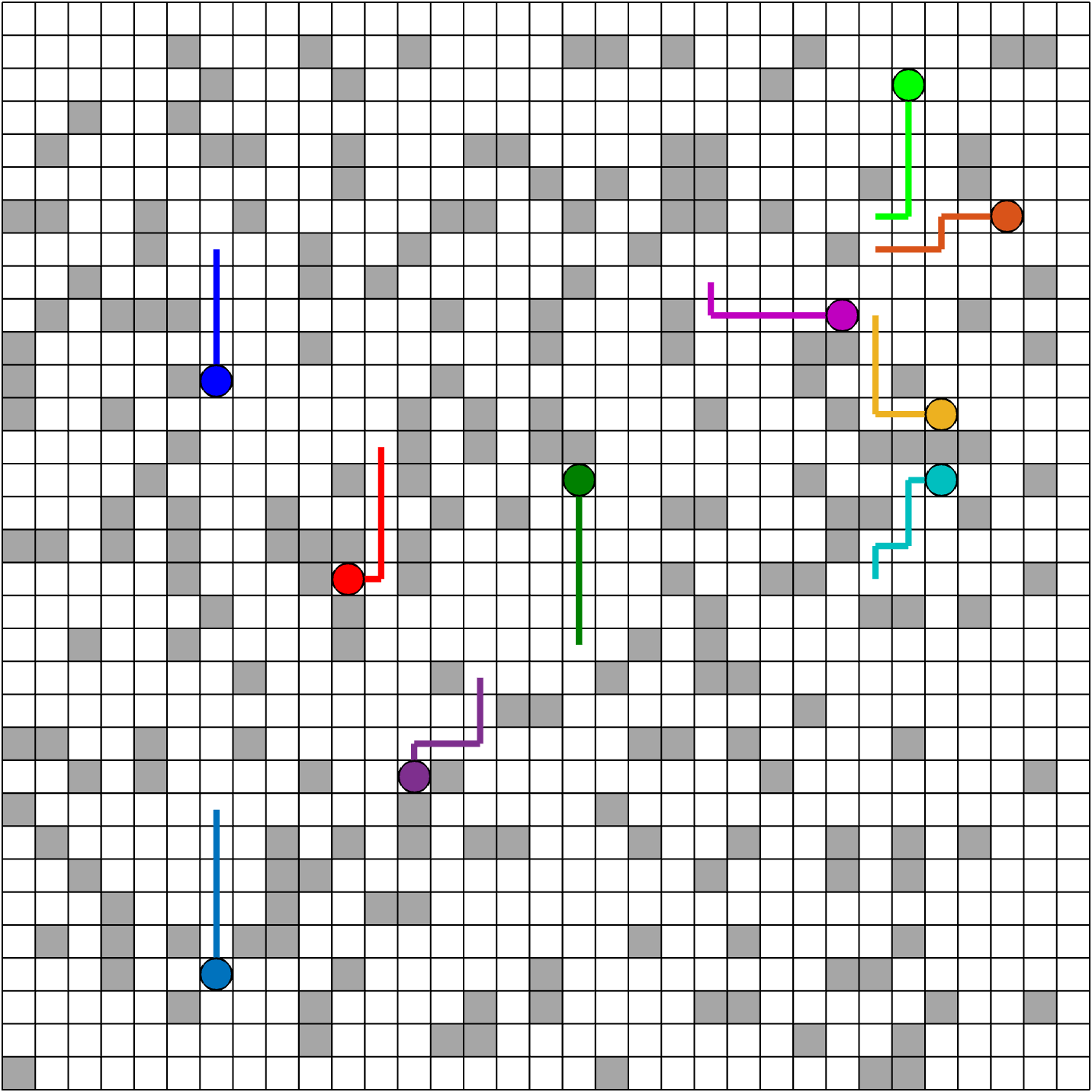}
         \caption{XG-CBS, $\Delta k=[0,5]$}
         \label{fig:ln18_xgcbs_seg1}
     \end{subfigure}
     \hfill
     \newline
     \begin{subfigure}{0.19\linewidth}
         \centering
         \includegraphics[scale=0.275]{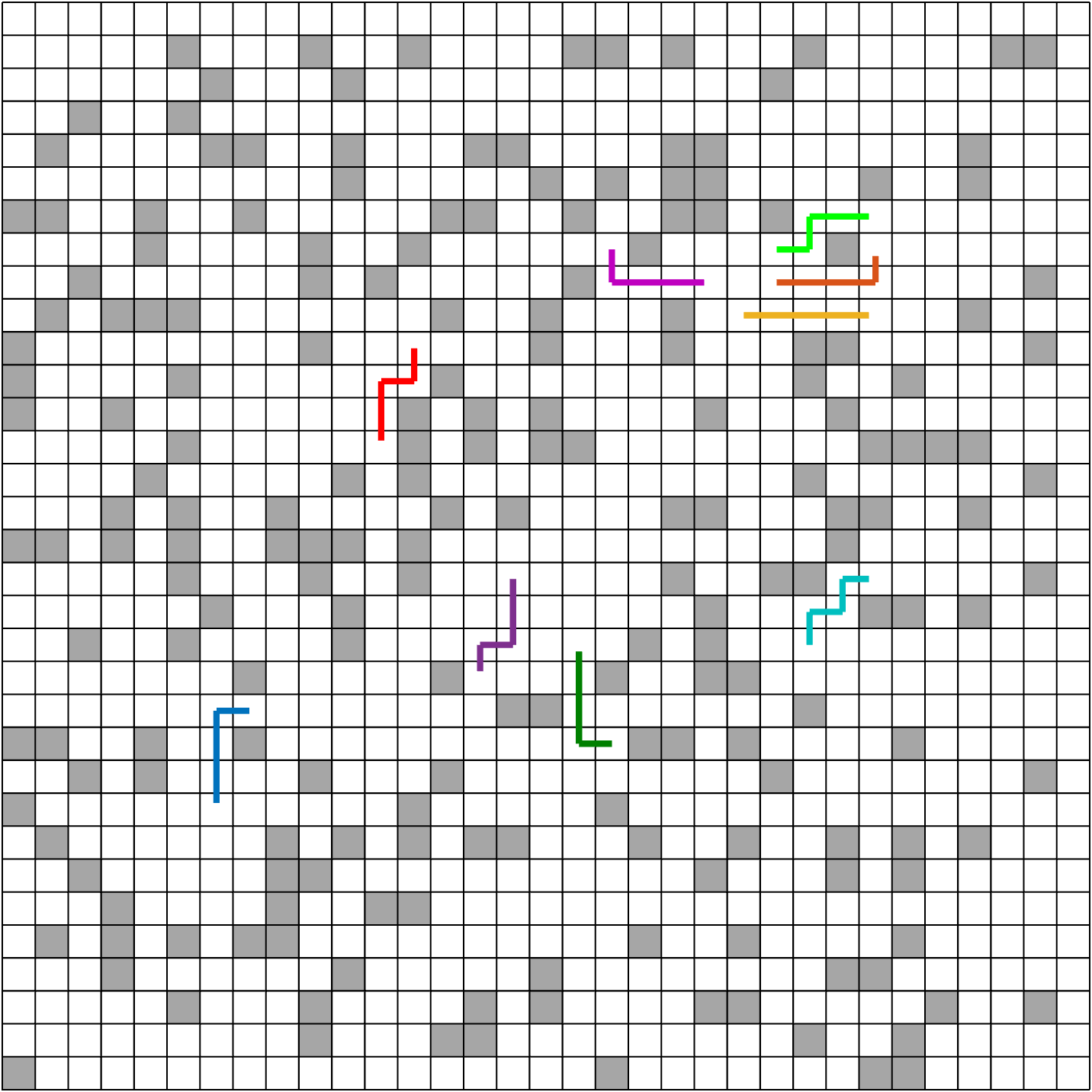}
         \caption{\centering XG-CBS,\hspace{\linewidth}$\Delta k=[5,9]$}
        %  \figuresource{}
         \label{fig:ln18_xgcbs_seg2}
     \end{subfigure}
     \hfill
     \begin{subfigure}{0.19\linewidth}
         \centering
         \includegraphics[scale=0.275]{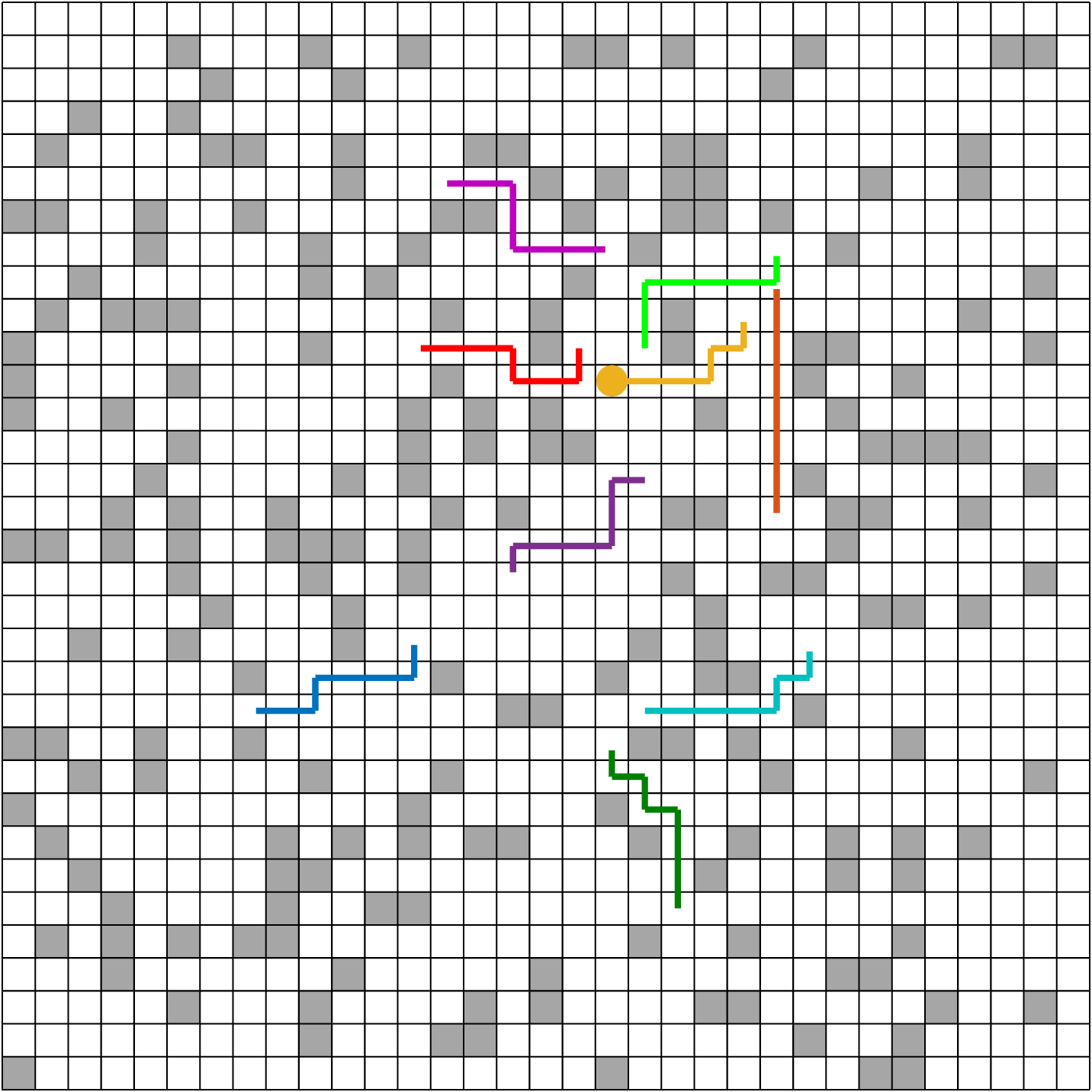}
         \caption{\centering XG-CBS,\hspace{\linewidth}$\Delta k=[9,16]$}
        %  \figuresource{$\Delta k=[9,16]$}
         \label{fig:ln18_xgcbs_seg3}
     \end{subfigure}
     \hfill
     \begin{subfigure}{0.19\linewidth}
         \centering
         \includegraphics[scale=0.275]{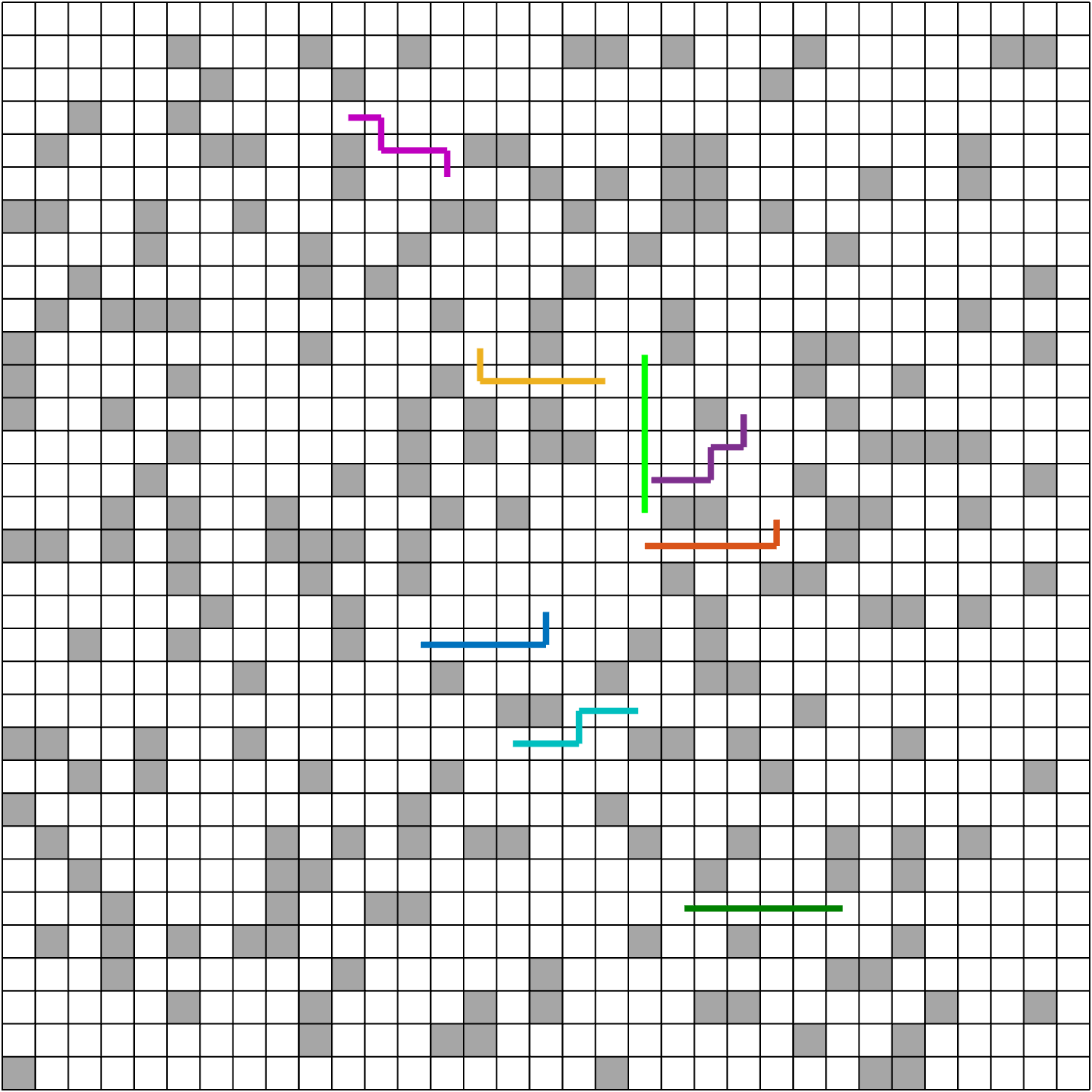}
         \caption{\centering XG-CBS,\hspace{\linewidth}$\Delta k=[16,21]$}
        %  \figuresource{\centering XG-CBS,\hspace{\linewidth}$\Delta k=[16,21]$}
         \label{fig:ln18_xgcbs_seg4}
     \end{subfigure}
     \hfill
     \begin{subfigure}{0.19\linewidth}
         \centering
         \includegraphics[scale=0.275]{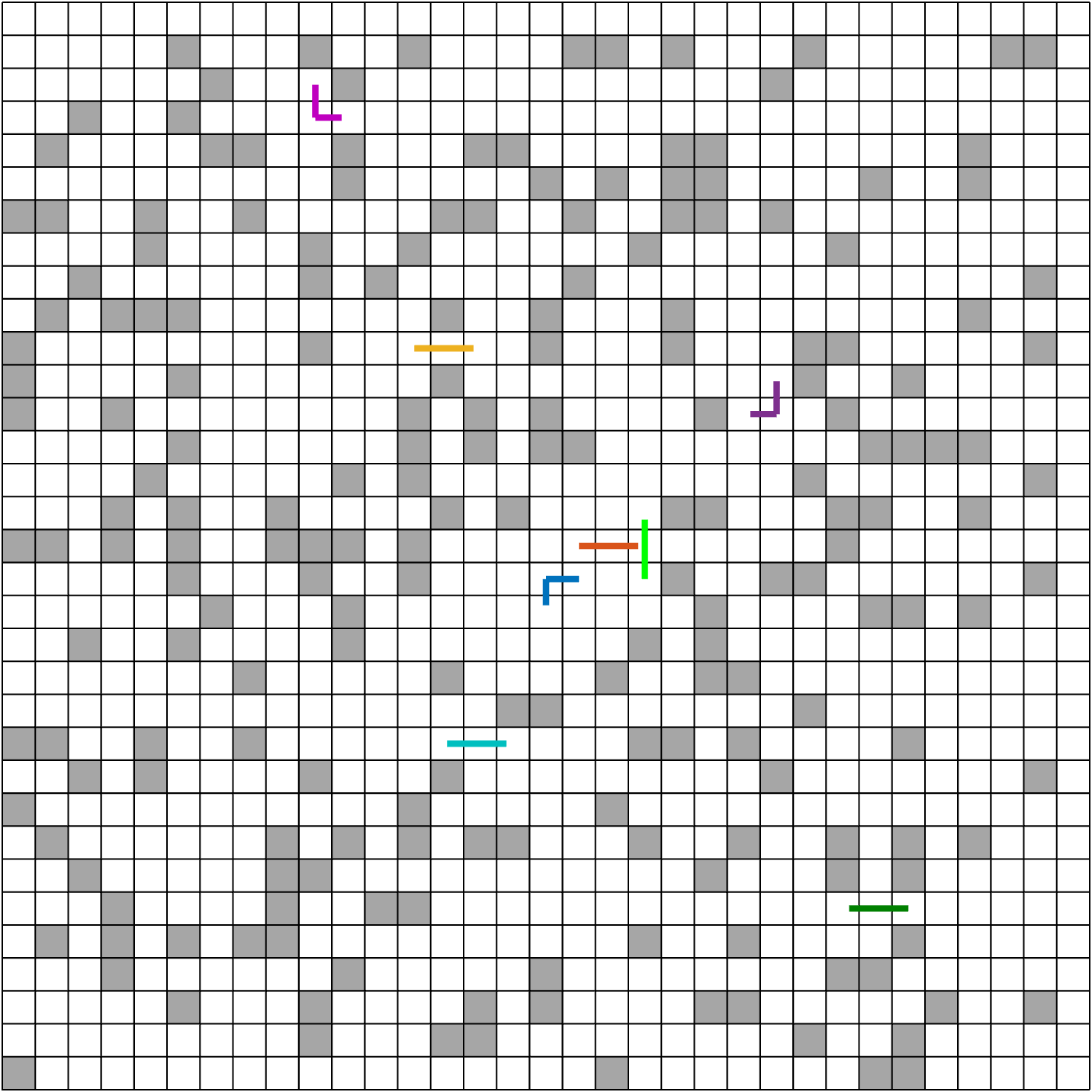}
         \caption{\centering XG-CBS,\hspace{\linewidth}$\Delta k=[21,23]$}
        %  \figuresource{}
         \label{fig:ln18_xgcbs_seg5}
     \end{subfigure}
     \hfill
     \begin{subfigure}{0.19\linewidth}
         \centering
         \includegraphics[scale=0.275]{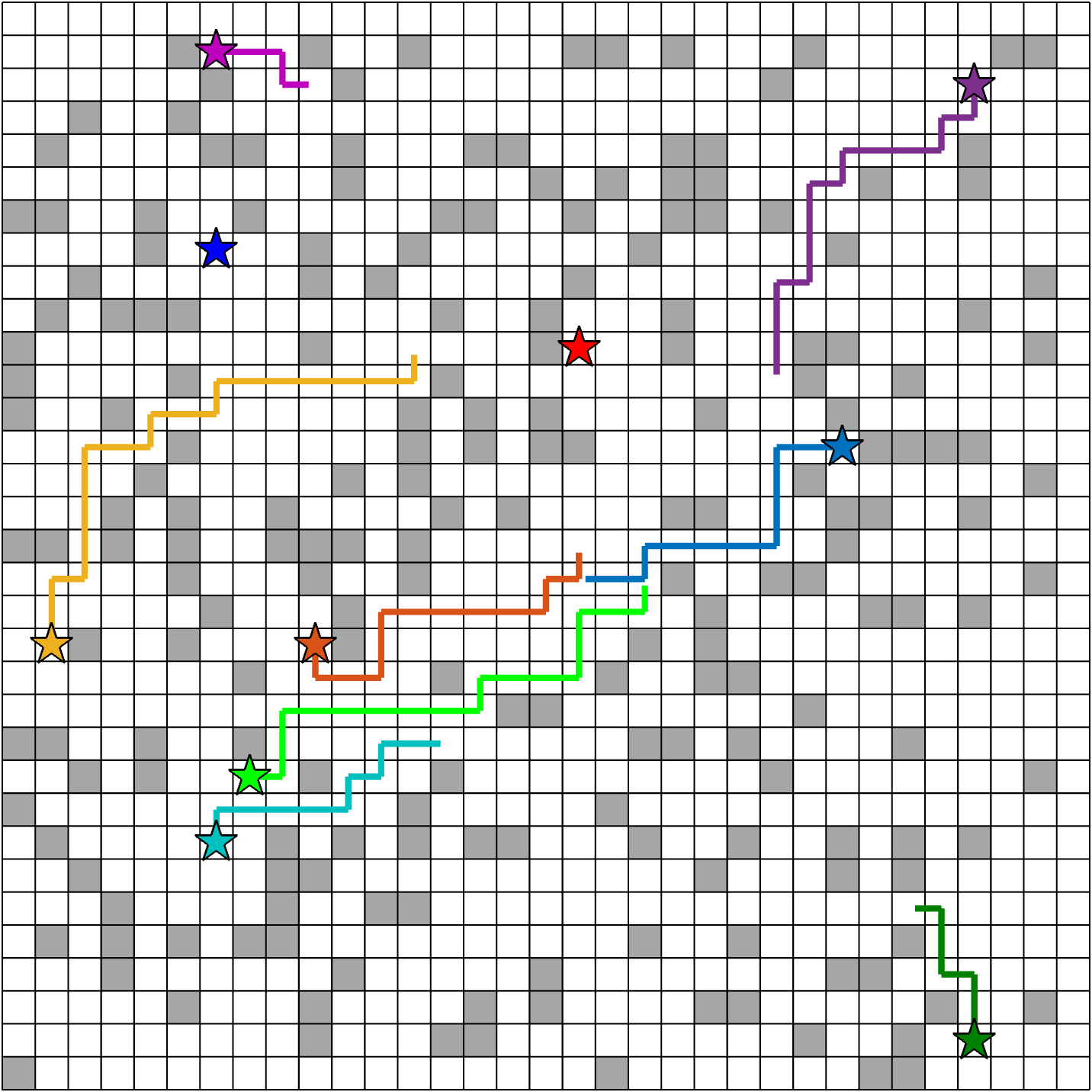}
         \caption{\centering XG-CBS,\hspace{\linewidth}$\Delta k=[23,43]$}
        %  \figuresource{ $\Delta k=[23,43]$}
         \label{fig:ln18_xgcbs_seg6}
     \end{subfigure}
     \hfill
    % \caption{Line $56$ of Table~\ref{tab:final_benchmark}}
    \caption{Example of XG-CBS reducing 18-segment plan of CBS to a 6-segment plan.}
    \label{fig:ln_18} % numbers in labels represent row of Table 1 -- numbers in document should be from Table 2
\end{figure*}

Continuing to scale upwards, we now consider the example in Figure~\ref{fig:ln_18} with ten agents in a $33\times 33$ grid world. In $14$ seconds, XG-CBS with $A^*$ cuts the explanation by approximately $65\%$, making it much simpler for a human user to validate the plan.

\begin{figure*}[p]
     \centering
     \begin{subfigure}{0.24\linewidth}
         \centering
         \includegraphics[scale=0.3]{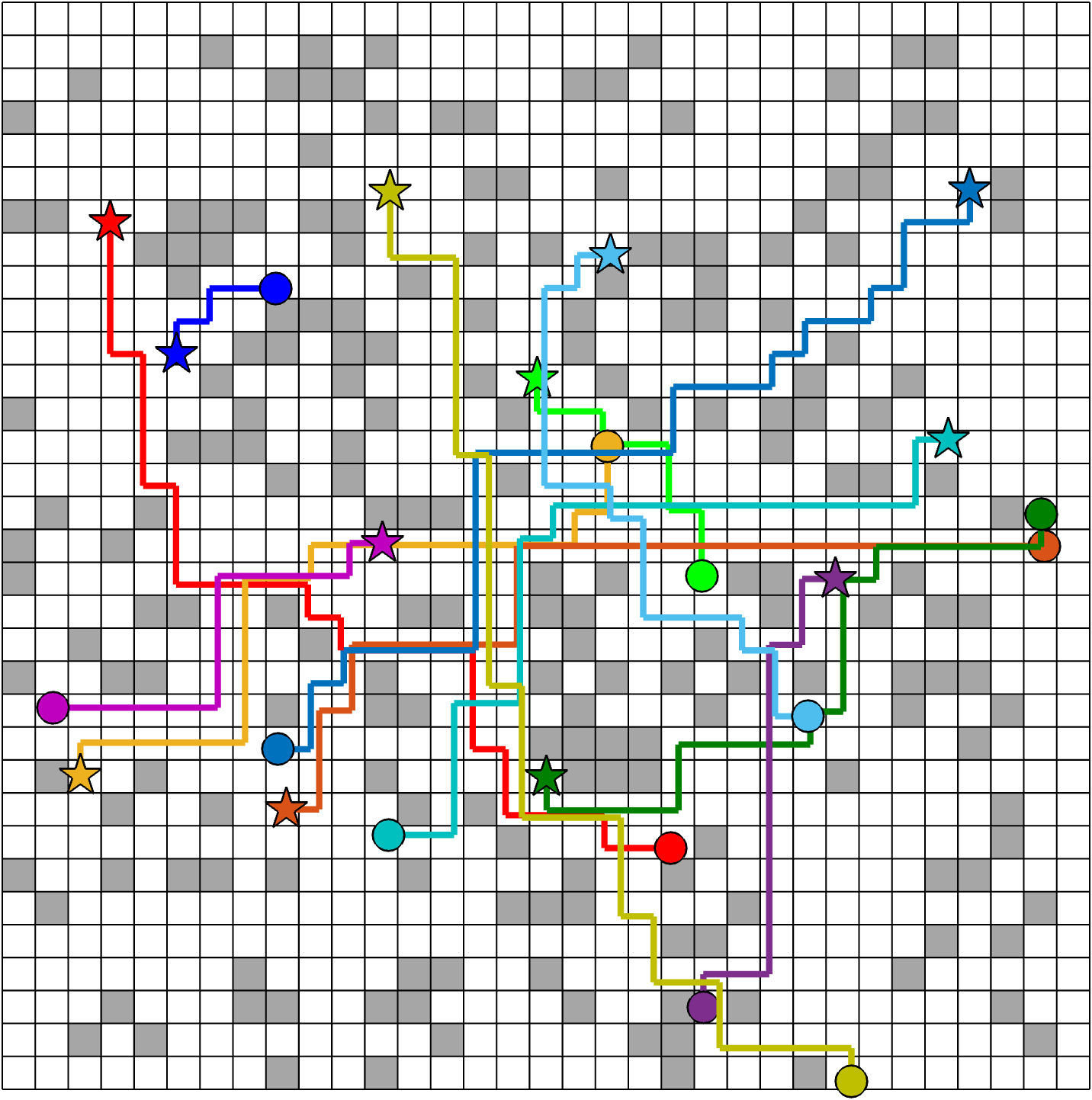}
         \caption{CBS}
         \label{fig:ln23_cbs_full}
     \end{subfigure}
     \hfill
     \begin{subfigure}{0.24\linewidth}
         \centering
         \includegraphics[scale=0.3]{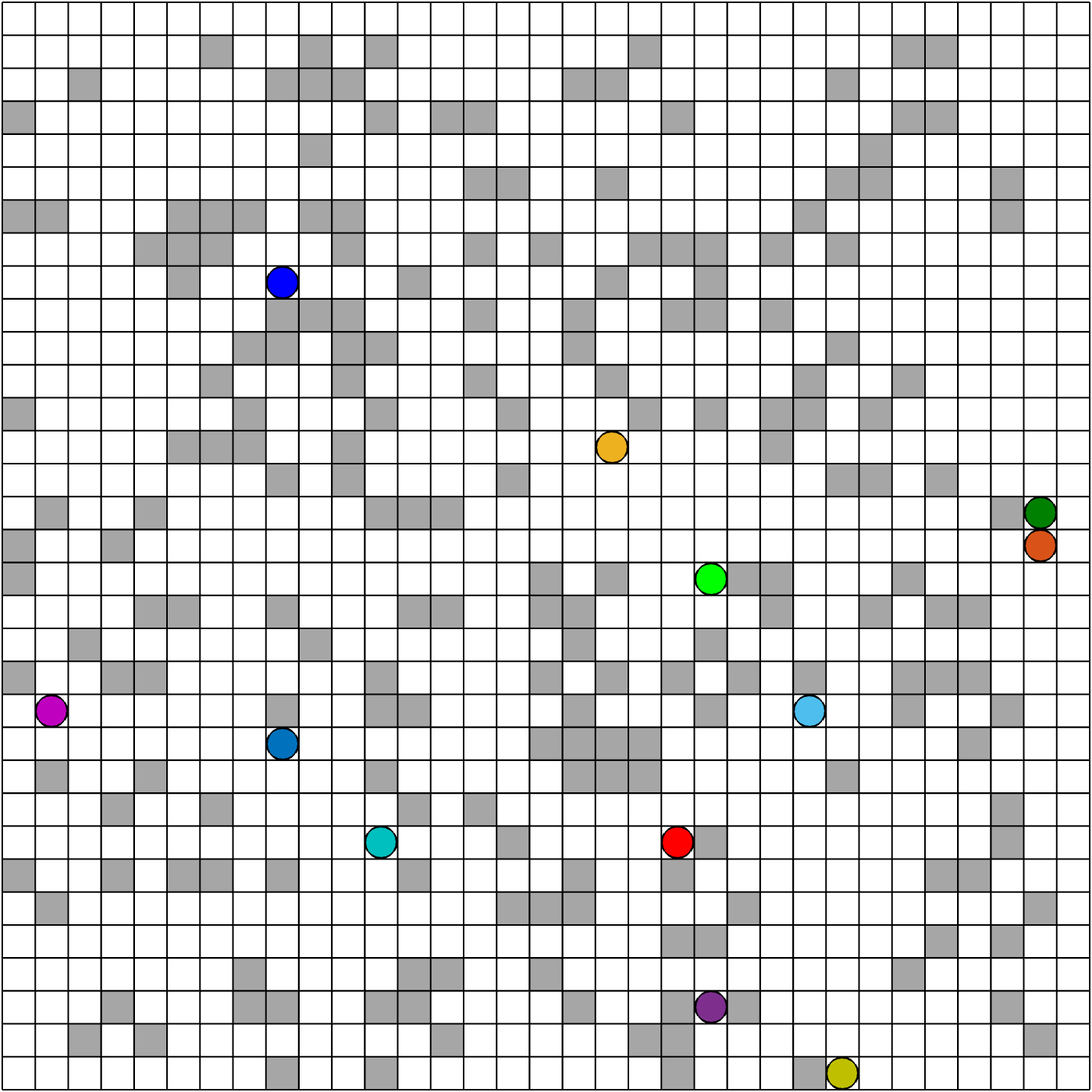}
         \caption{CBS, $\Delta k=[0,0]$}
         \label{fig:ln23_cbs_seg1}
     \end{subfigure}
     \hfill
     \begin{subfigure}{0.24\linewidth}
         \centering
         \includegraphics[scale=0.3]{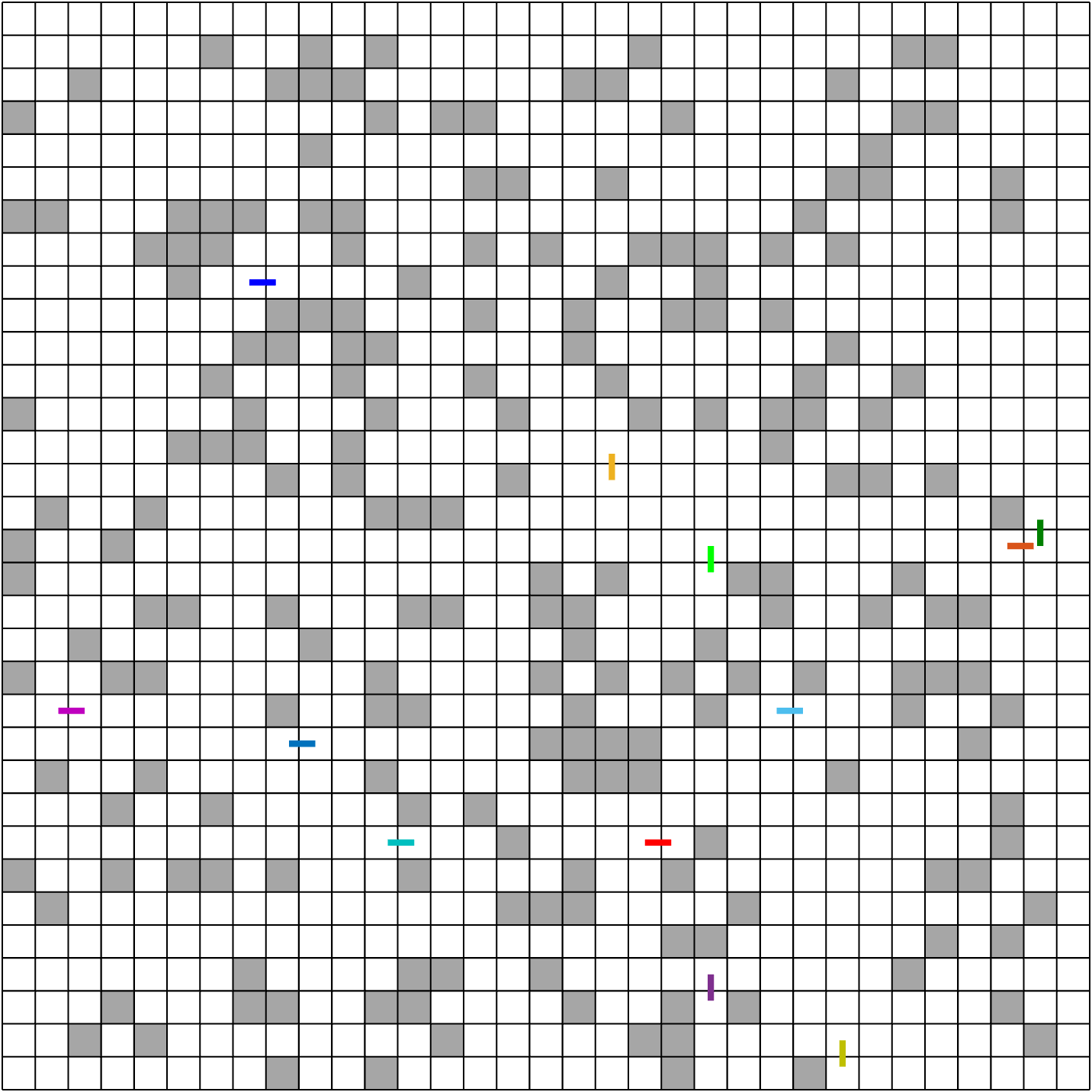}
         \caption{CBS, $\Delta k=[0,1]$}
         \label{fig:ln23_cbs_seg2}
     \end{subfigure}
     \hfill
     \begin{subfigure}{0.24\linewidth}
         \centering
         \includegraphics[scale=0.3]{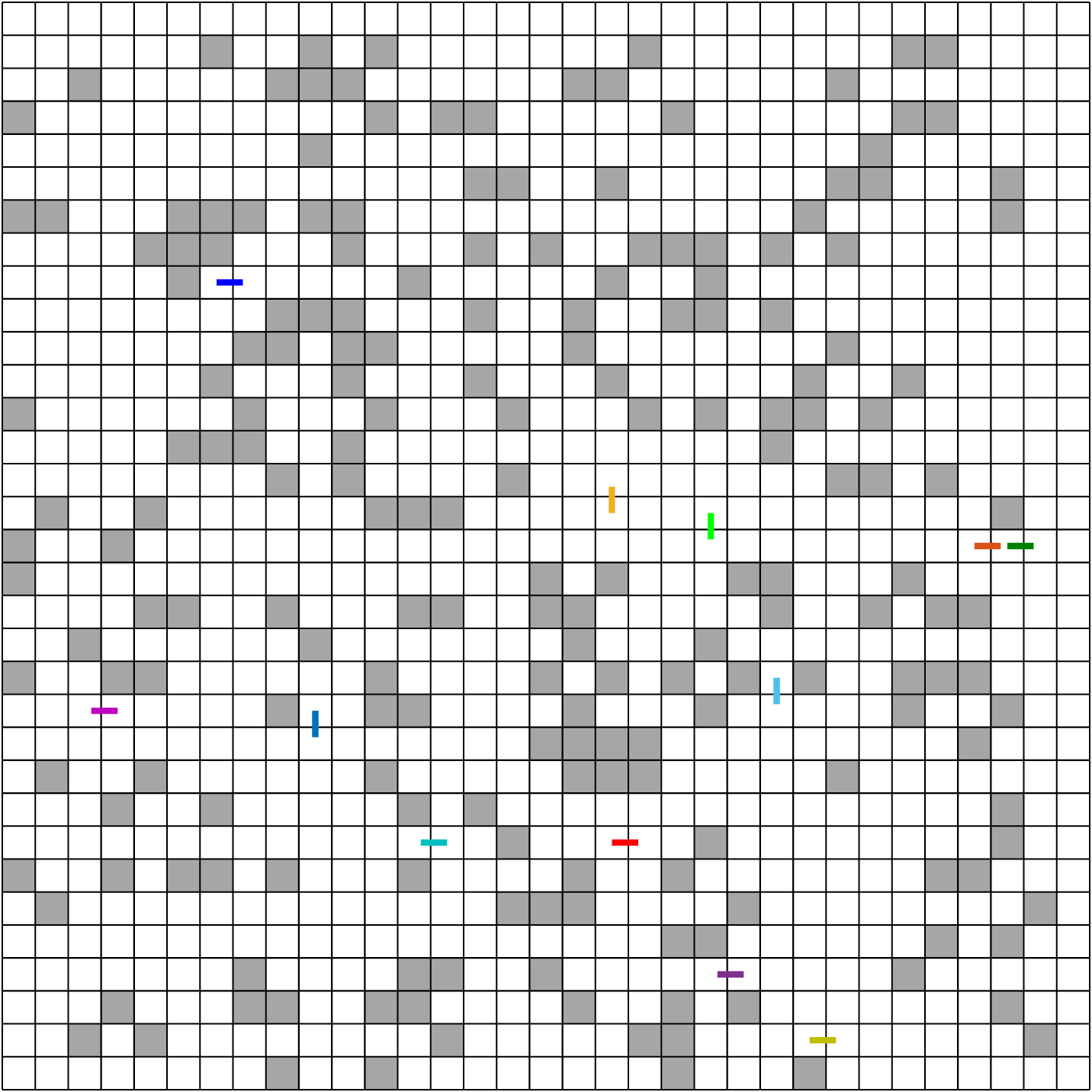}
         \caption{CBS, $\Delta k=[1,2]$}
         \label{fig:ln23_cbs_seg3}
     \end{subfigure}
     \hfill
     \newline
     \begin{subfigure}{0.24\linewidth}
         \centering
         \includegraphics[scale=0.3]{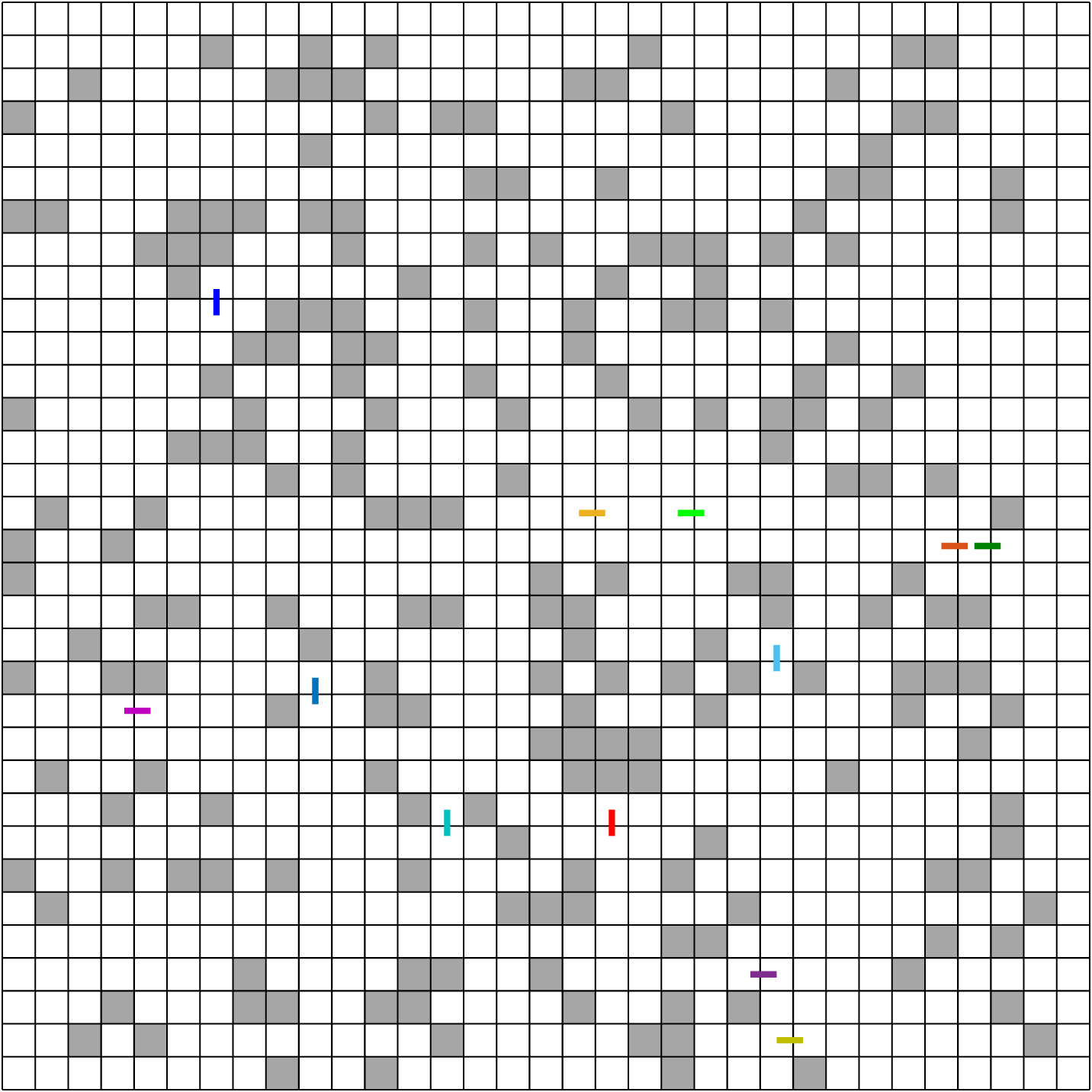}
         \caption{CBS, $\Delta k=[2,3]$}
         \label{fig:ln23_cbs_seg4}
     \end{subfigure}
     \hfill
     \begin{subfigure}{0.24\linewidth}
         \centering
         \includegraphics[scale=0.3]{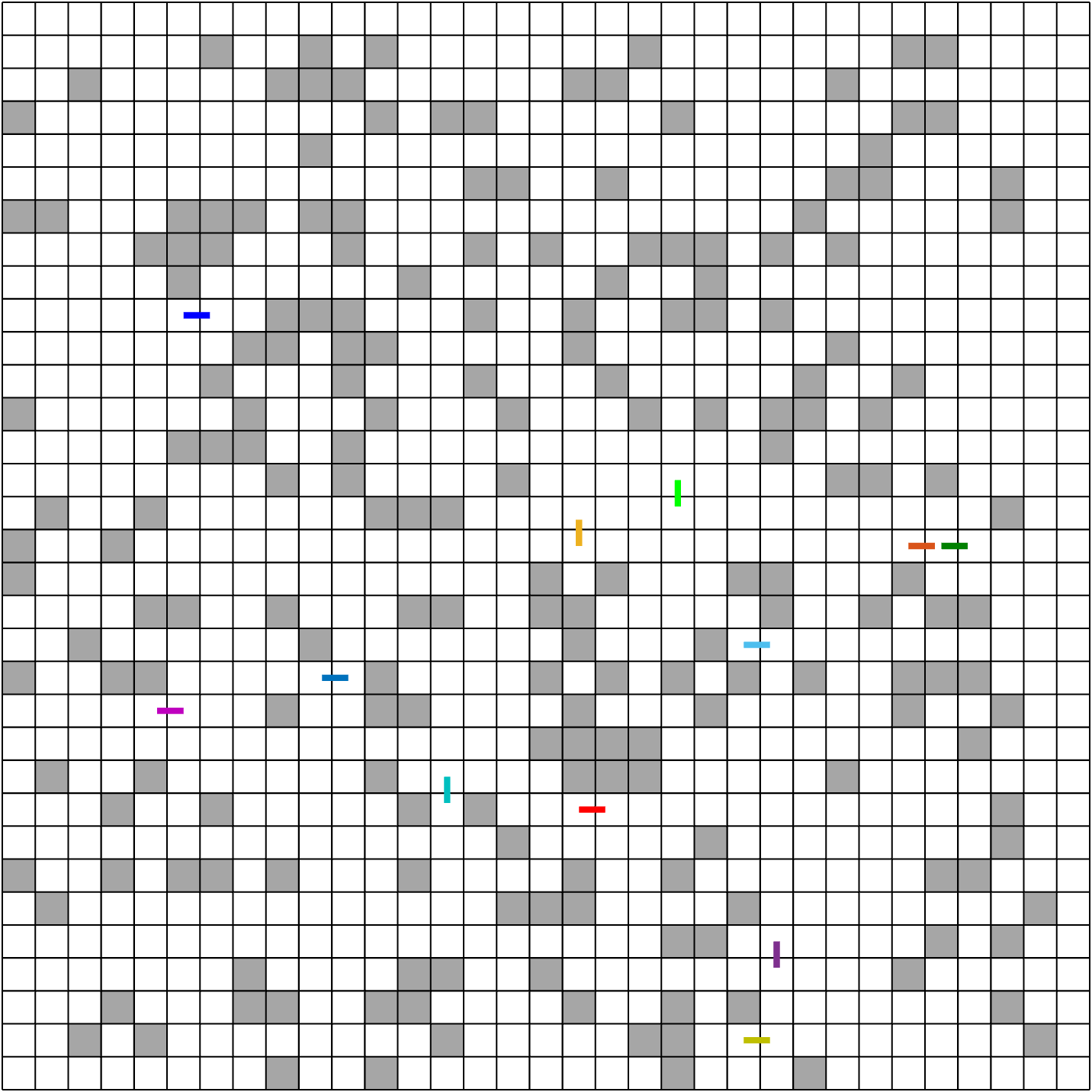}
         \caption{CBS, $\Delta k=[3,4]$}
         \label{fig:ln23_cbs_seg5}
     \end{subfigure}
     \hfill
     \begin{subfigure}{0.24\linewidth}
         \centering
         \includegraphics[scale=0.3]{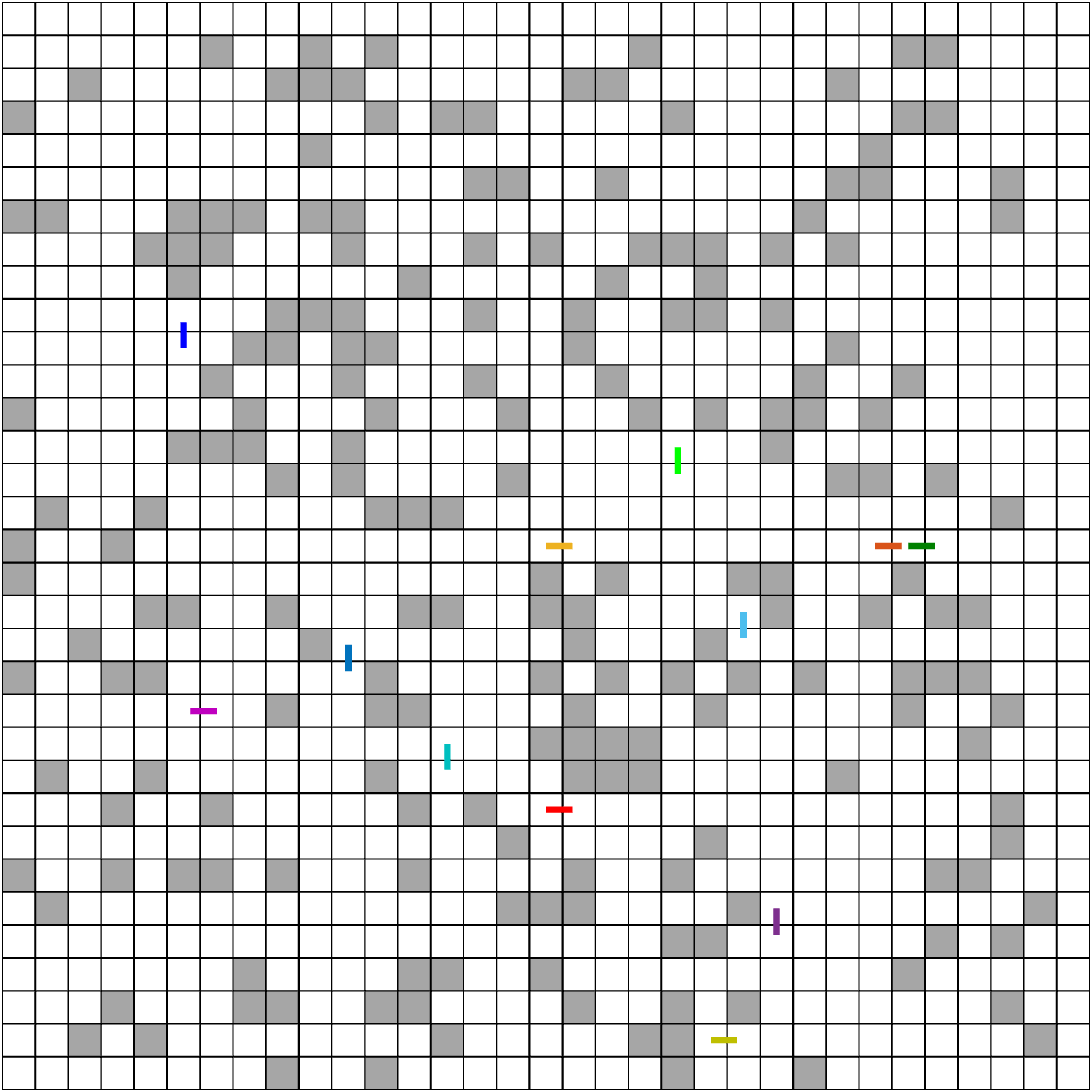}
         \caption{CBS, $\Delta k=[4,5]$}
         \label{fig:ln23_cbs_seg6}
     \end{subfigure}
     \hfill
     \begin{subfigure}{0.24\linewidth}
         \centering
         \includegraphics[scale=0.3]{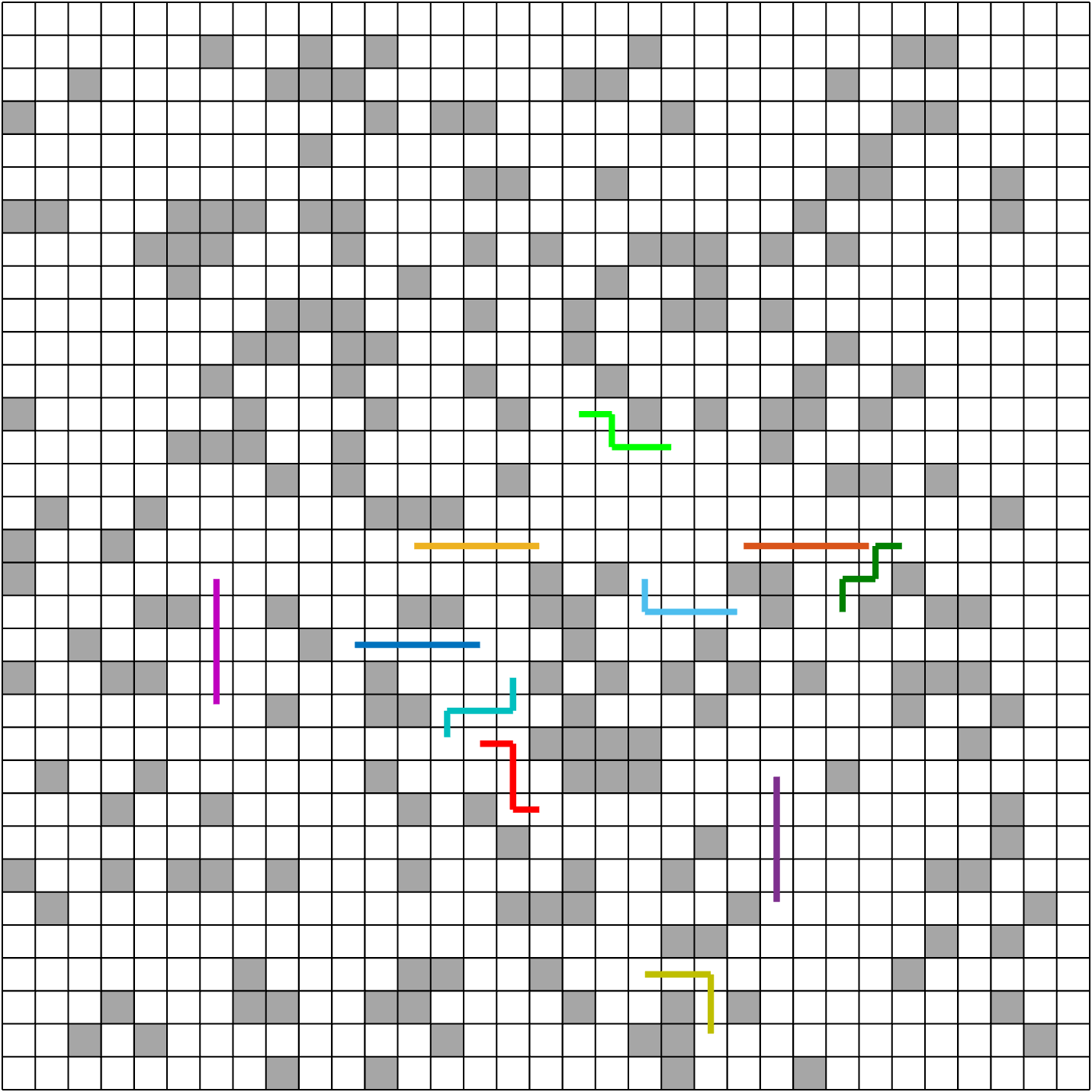}
         \caption{CBS, $\Delta k=[5,9]$}
         \label{fig:ln23_cbs_seg7}
     \end{subfigure}
     \hfill
     \newline
     \begin{subfigure}{0.24\linewidth}
         \centering
         \includegraphics[scale=0.3]{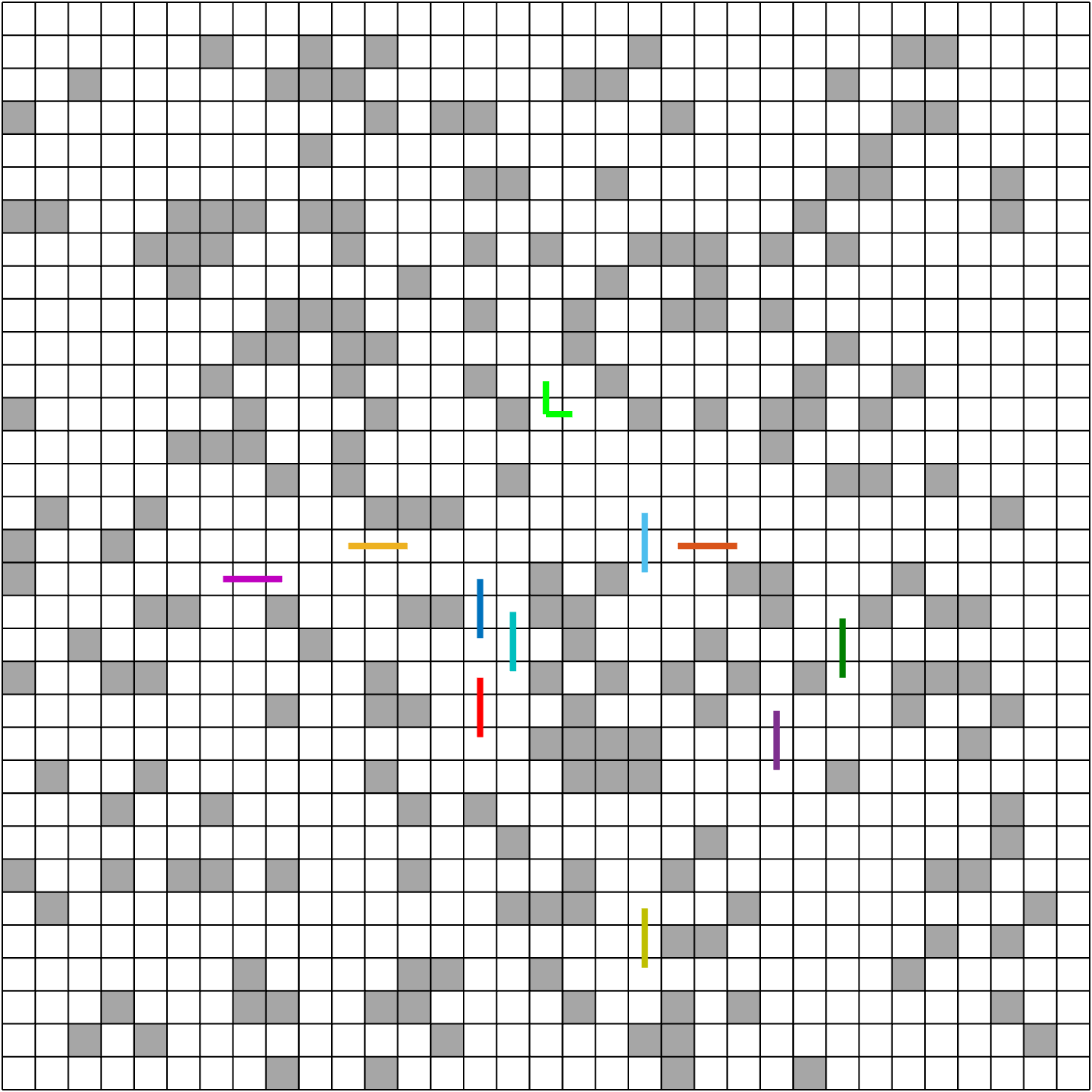}
         \caption{CBS, $\Delta k=[9,11]$}
         \label{fig:ln23_cbs_seg8}
     \end{subfigure}
     \hfill
     \begin{subfigure}{0.24\linewidth}
         \centering
         \includegraphics[scale=0.3]{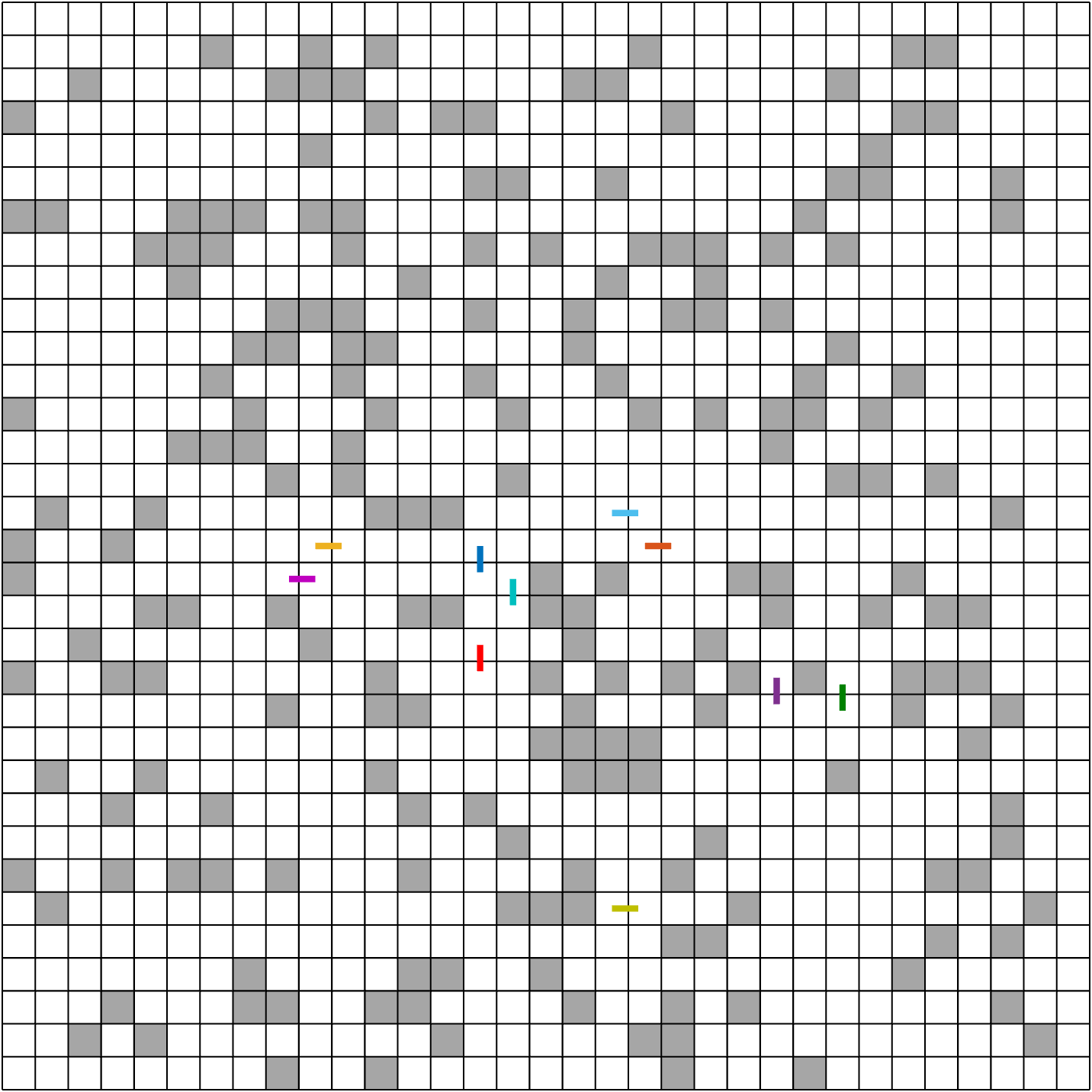}
         \caption{CBS, $\Delta k=[11,12]$}
         \label{fig:ln23_cbs_seg9}
     \end{subfigure}
     \hfill
     \begin{subfigure}{0.24\linewidth}
         \centering
         \includegraphics[scale=0.3]{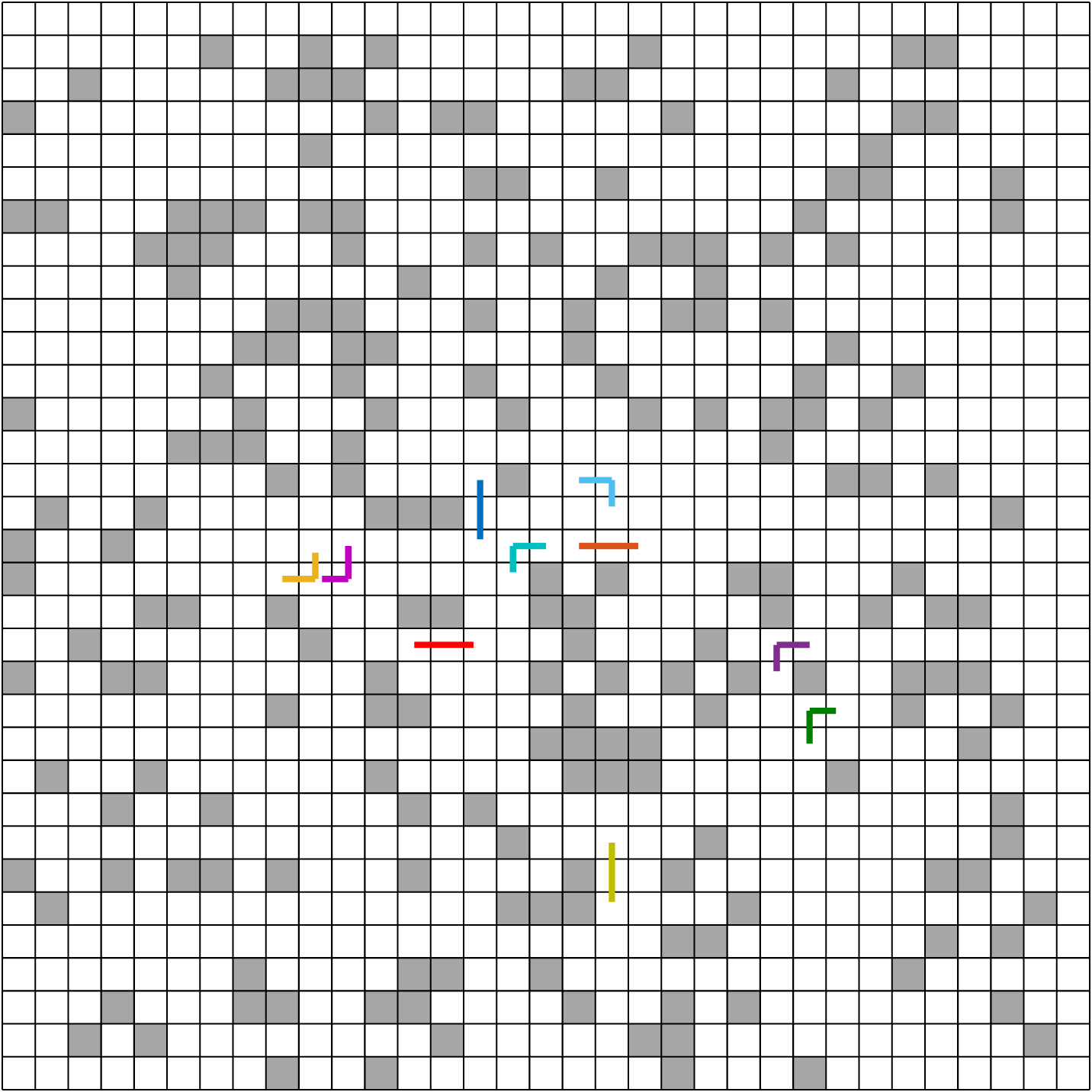}
         \caption{CBS, $\Delta k=[12,14]$}
         \label{fig:ln23_cbs_seg10}
     \end{subfigure}
     \hfill
     \begin{subfigure}{0.24\linewidth}
         \centering
         \includegraphics[scale=0.3]{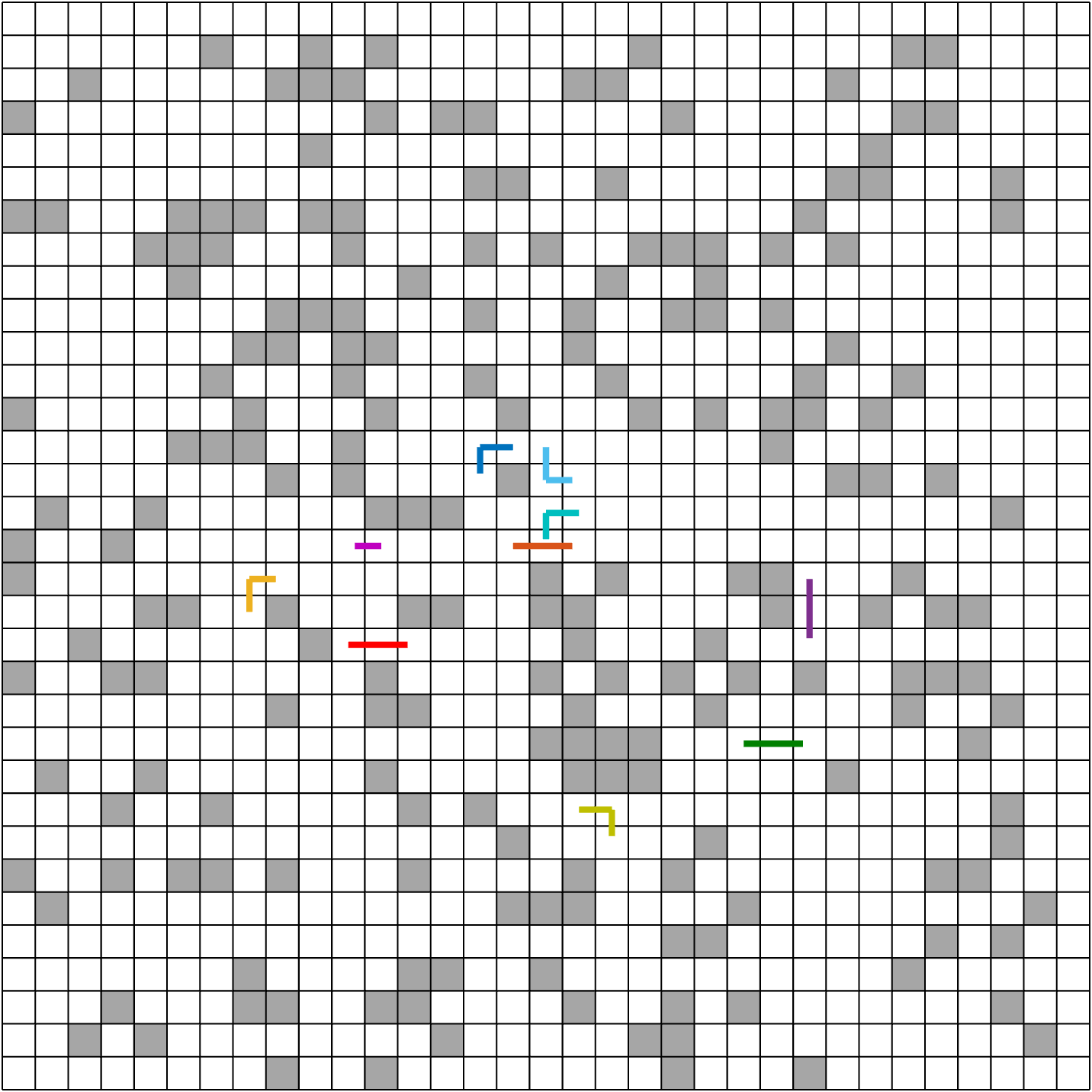}
         \caption{CBS, $\Delta k=[14,16]$}
         \label{fig:ln23_cbs_seg11}
     \end{subfigure}
     \hfill
     \newline
     \begin{subfigure}{0.24\linewidth}
         \centering
         \includegraphics[scale=0.3]{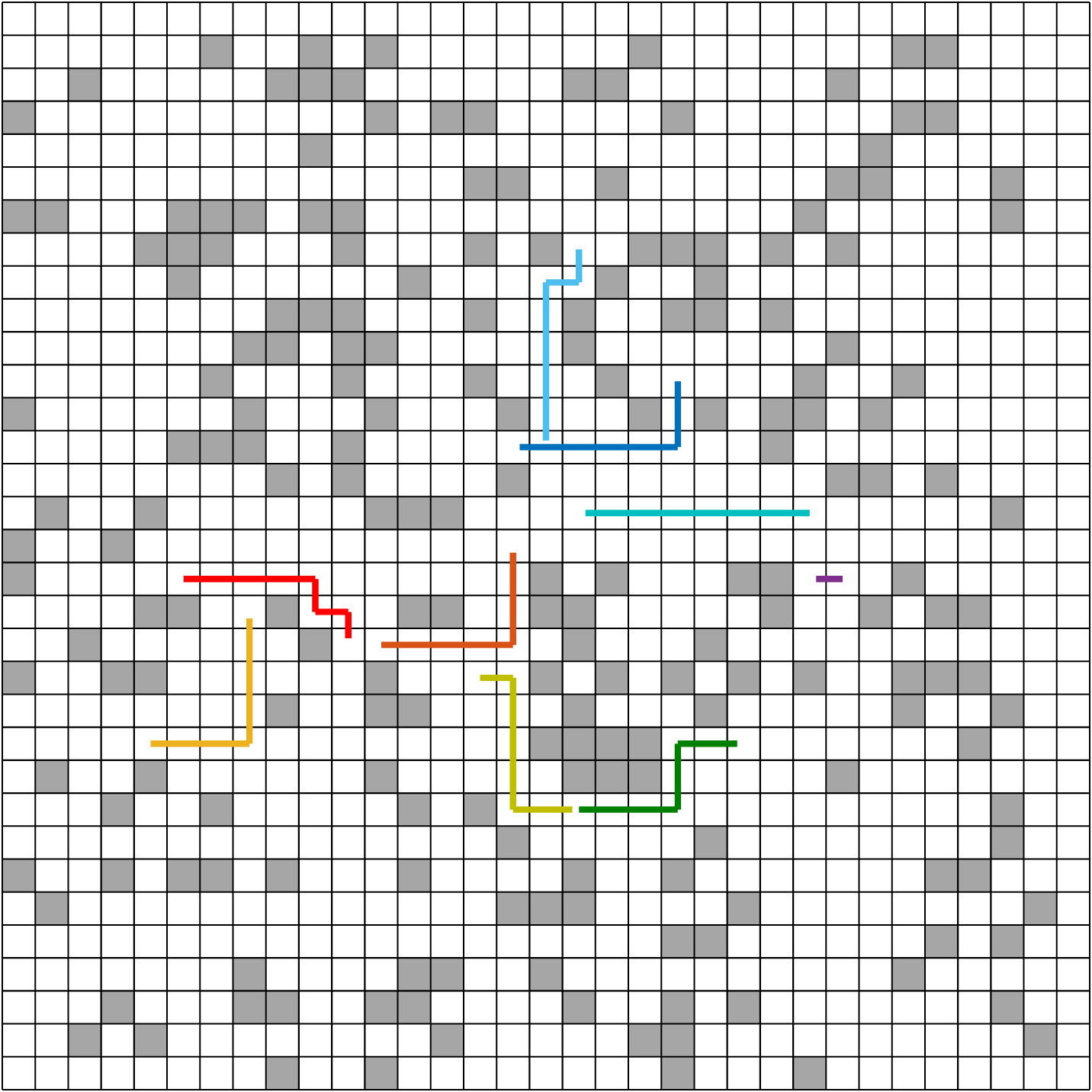}
         \caption{CBS, $\Delta k=[16,23]$}
         \label{fig:ln23_cbs_seg12}
     \end{subfigure}
     \hfill
     \begin{subfigure}{0.24\linewidth}
         \centering
         \includegraphics[scale=0.3]{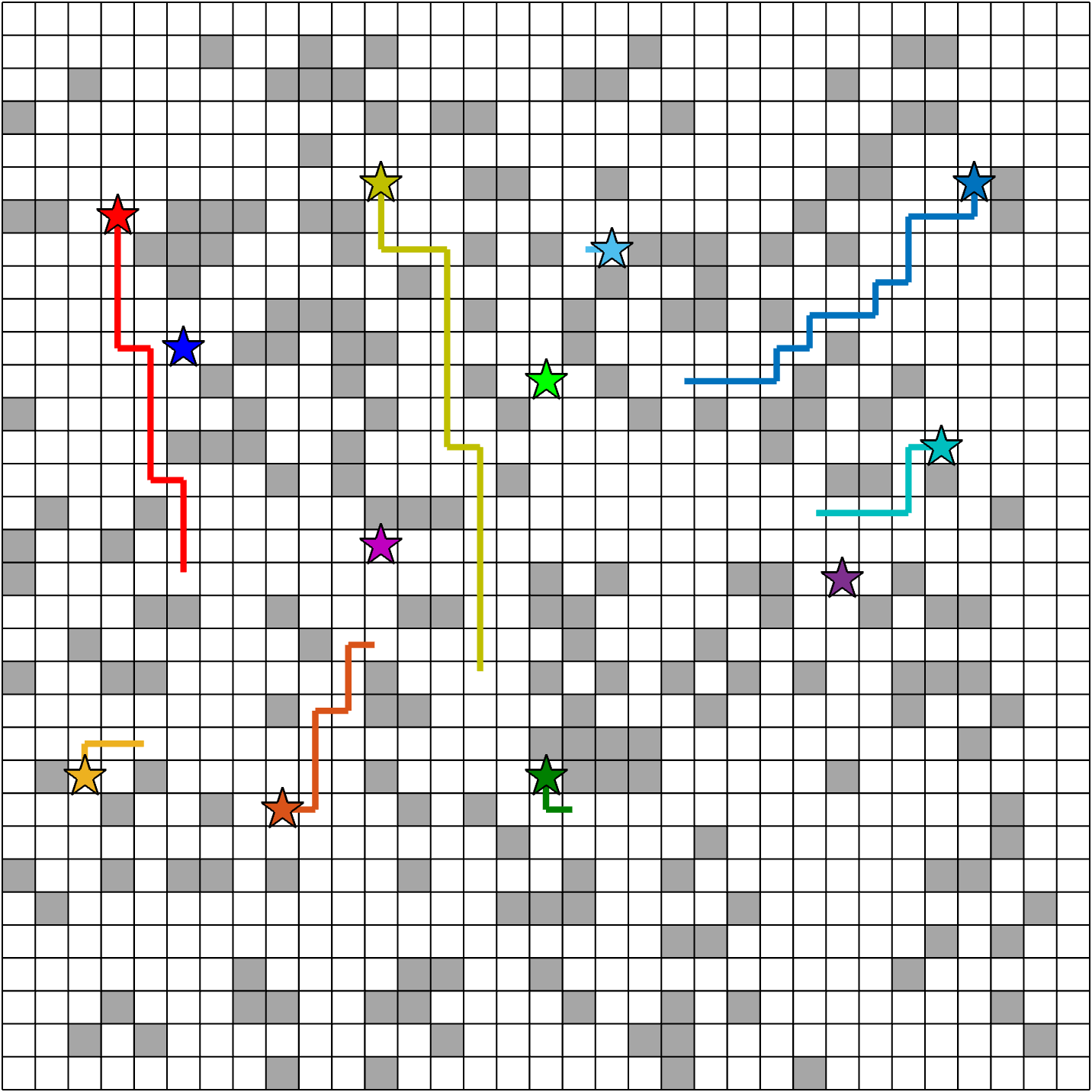}
         \caption{CBS, $\Delta k=[23,41]$}
         \label{fig:ln23_cbs_seg13}
     \end{subfigure}
     \hfill
     \begin{subfigure}{0.24\linewidth}
         \centering
         \includegraphics[scale=0.3]{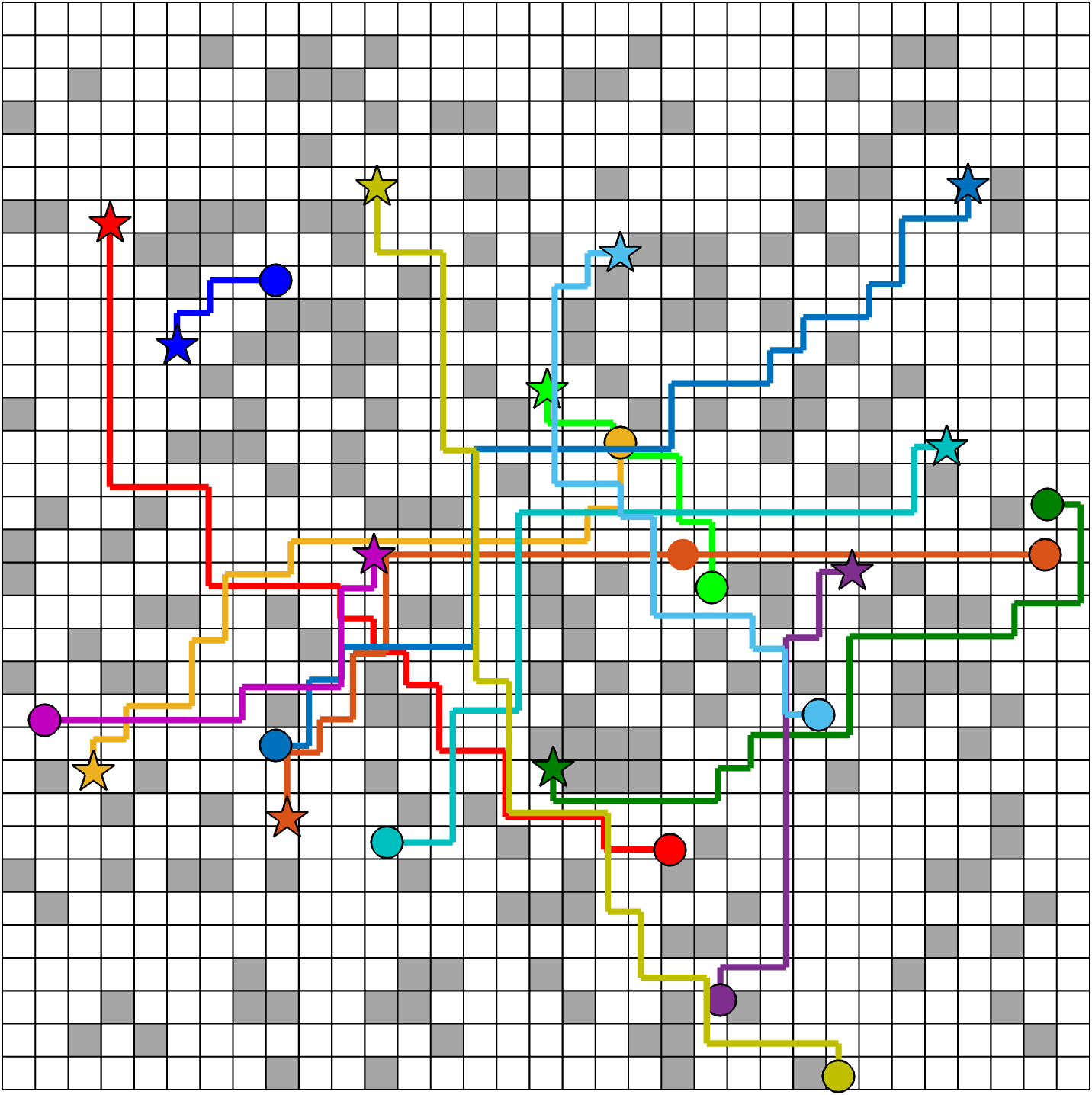}
         \caption{XG-CBS, $r=5$}
         \label{fig:ln23_xgcbs_full}
     \end{subfigure}
     \hfill
     \begin{subfigure}{0.24\linewidth}
         \centering
         \includegraphics[scale=0.3]{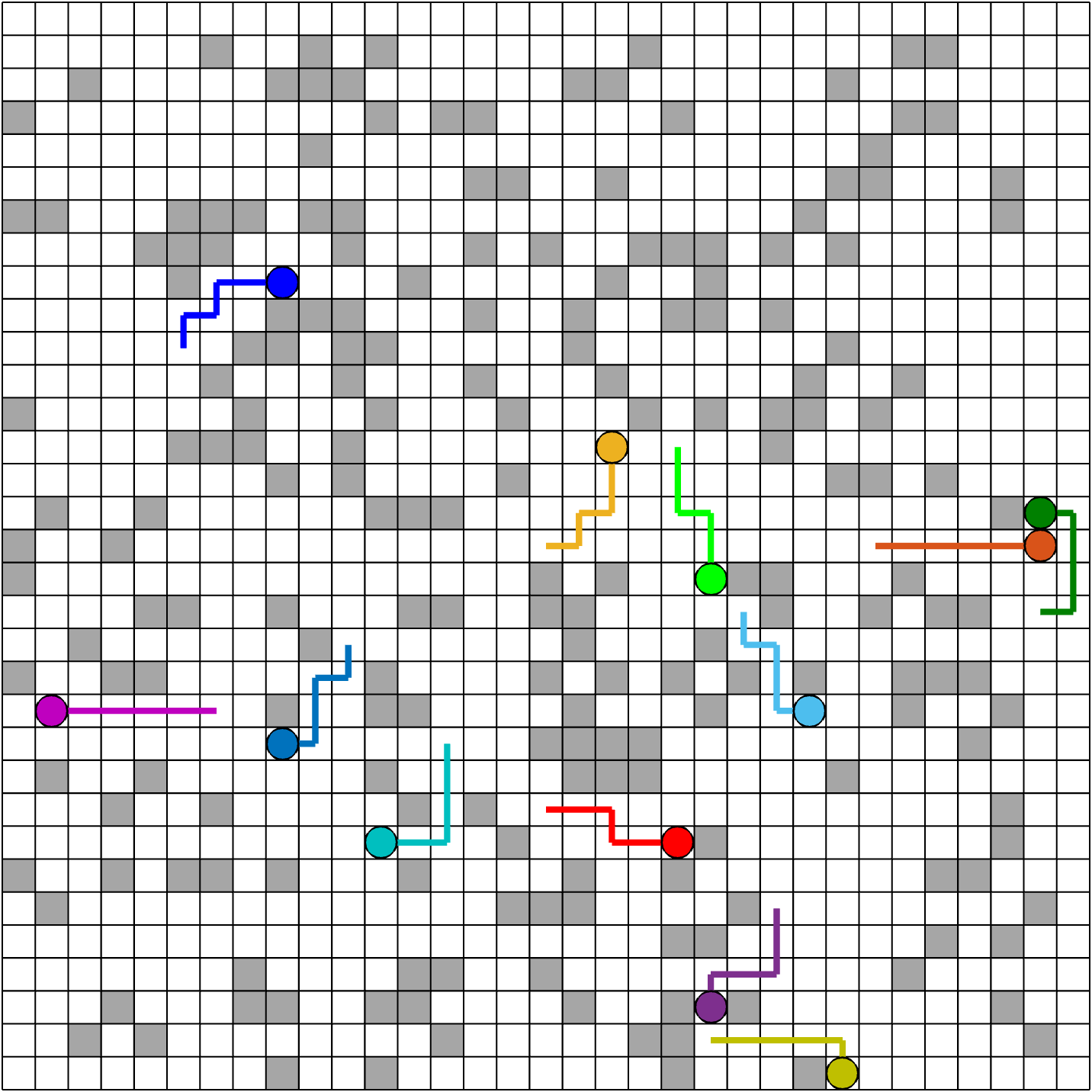}
         \caption{XG-CBS, $\Delta k=[0,5]$}
         \label{fig:ln23_xgcbs_seg1}
     \end{subfigure}
     \newline
     \begin{subfigure}{0.24\linewidth}
         \centering
         \includegraphics[scale=0.3]{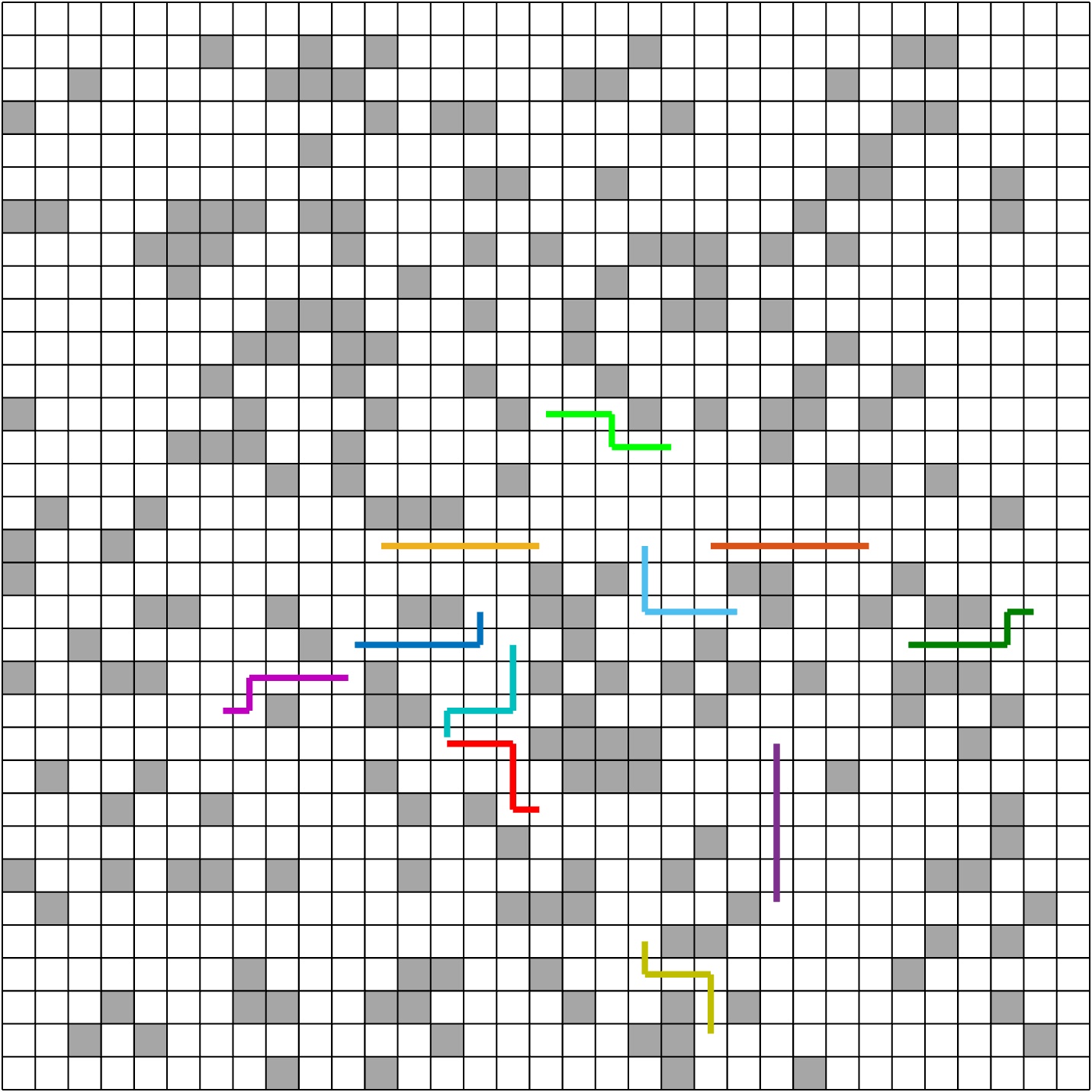}
         \caption{XG-CBS, $\Delta k=[5,10]$}
         \label{fig:ln23_xgcbs_seg2}
     \end{subfigure}
     \hfill
     \begin{subfigure}{0.24\linewidth}
         \centering
         \includegraphics[scale=0.3]{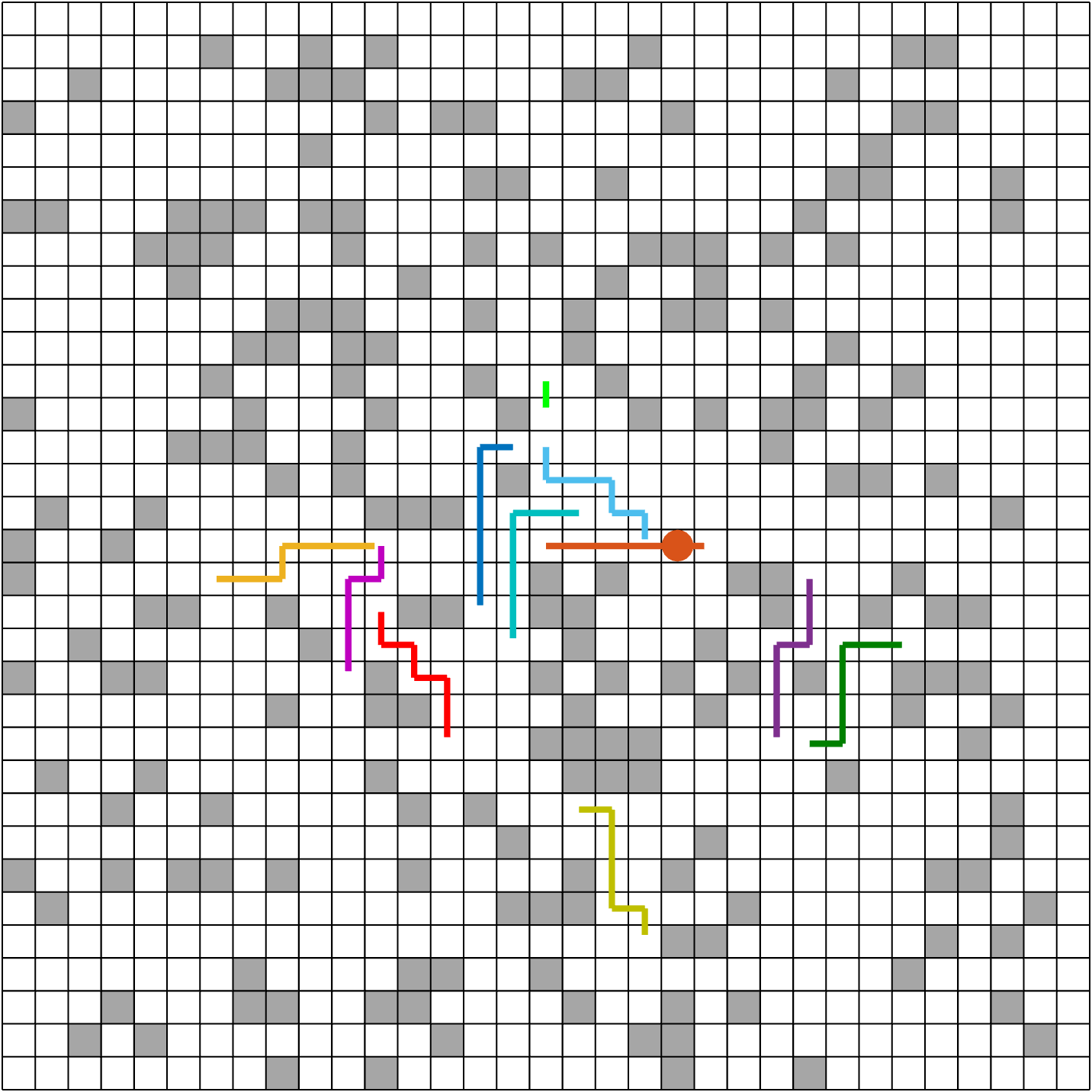}
         \caption{XG-CBS, $\Delta k=[10,16]$}
         \label{fig:ln23_xgcbs_seg3}
     \end{subfigure}
     \hfill
     \begin{subfigure}{0.24\linewidth}
         \centering
         \includegraphics[scale=0.3]{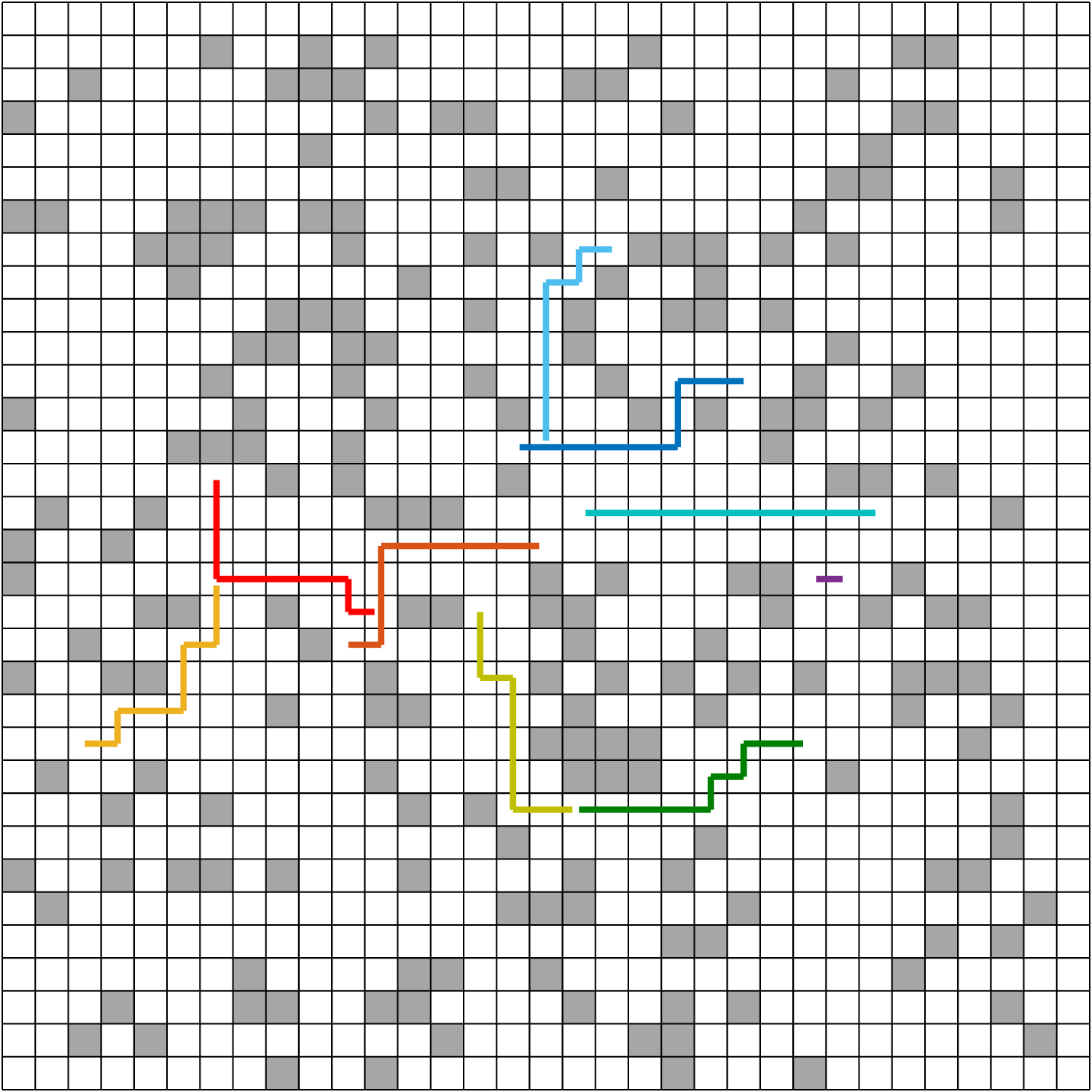}
         \caption{XG-CBS, $\Delta k=[16,25]$}
         \label{fig:ln23_xgcbs_seg4}
     \end{subfigure}
     \hfill
     \begin{subfigure}{0.24\linewidth}
         \centering
         \includegraphics[scale=0.3]{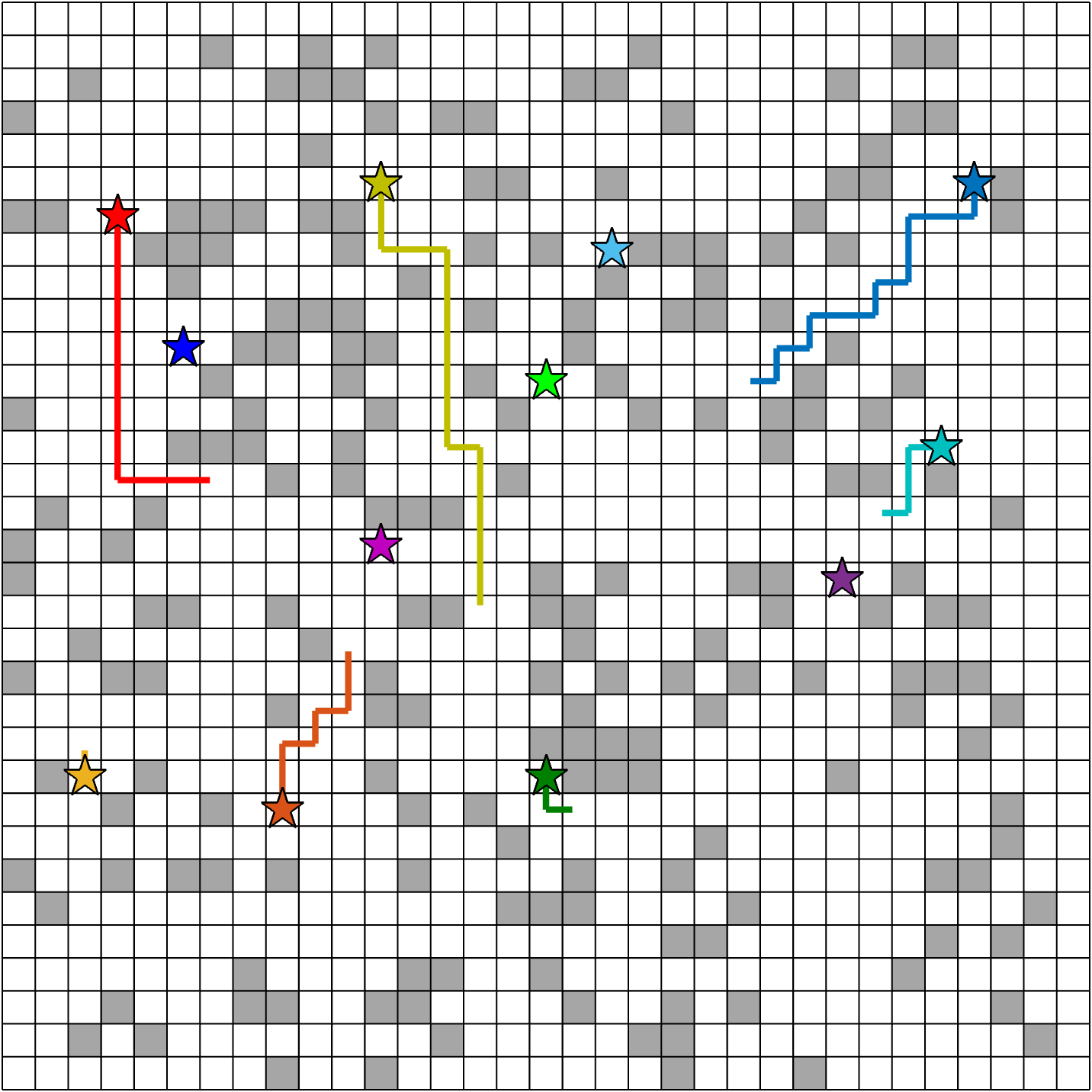}
         \caption{XG-CBS, $\Delta k=[25,41]$}
         \label{fig:ln23_xgcbs_seg5}
     \end{subfigure}
     \hfill
    % \caption{Line $58$ of Table~\ref{tab:final_benchmark}}
    \caption{Example of XG-CBS solution with 5-segments vs 13-segments solution of CBS.}
    \label{fig:ln_23}. % numbers in labels represent row of Table 1 -- numbers in document should be from Table 2
\end{figure*}

We conclude this section by testing our algorithms on a MAPF instance consisting of twelve agents in a $33\times 33$ grid world. The results are shown in Figure~\ref{fig:ln_23}. Notice that the plan returned by CBS (Figure~\ref{fig:ln23_cbs_full}) requires thirteen segments (Figure~\ref{fig:ln23_cbs_seg1}-\ref{fig:ln23_cbs_seg13}). However, planing with XG-CBS using $A^*$, we get the plan shown in Figure~\ref{fig:ln23_xgcbs_full} which only requires five segments, shown in Figure~\ref{fig:ln23_xgcbs_seg1}-\ref{fig:ln23_xgcbs_seg5}.

\subsubsection{Benchmark Evaluation (extended)}
\begin{figure*}[p]
    \centering
    \includegraphics[width=.775\linewidth]{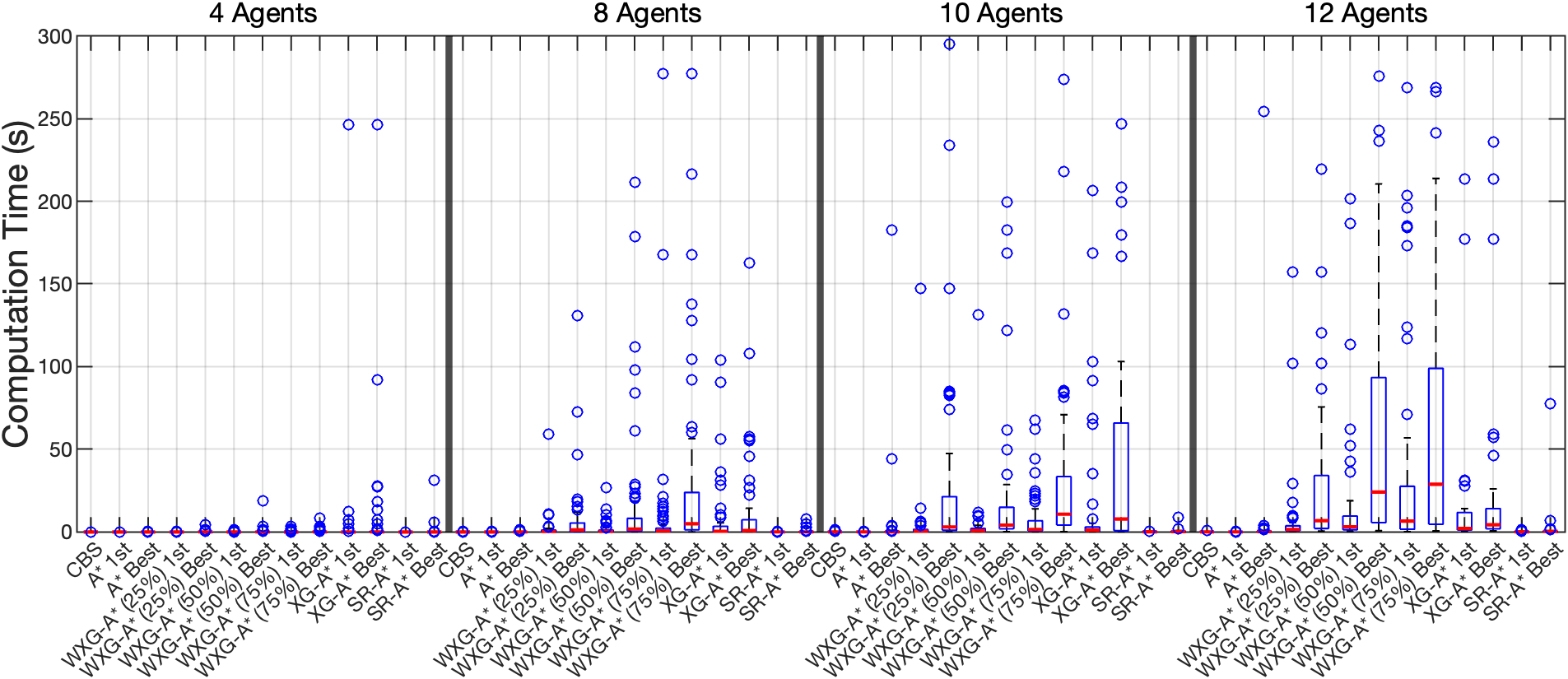}
    \includegraphics[width=.775\linewidth]{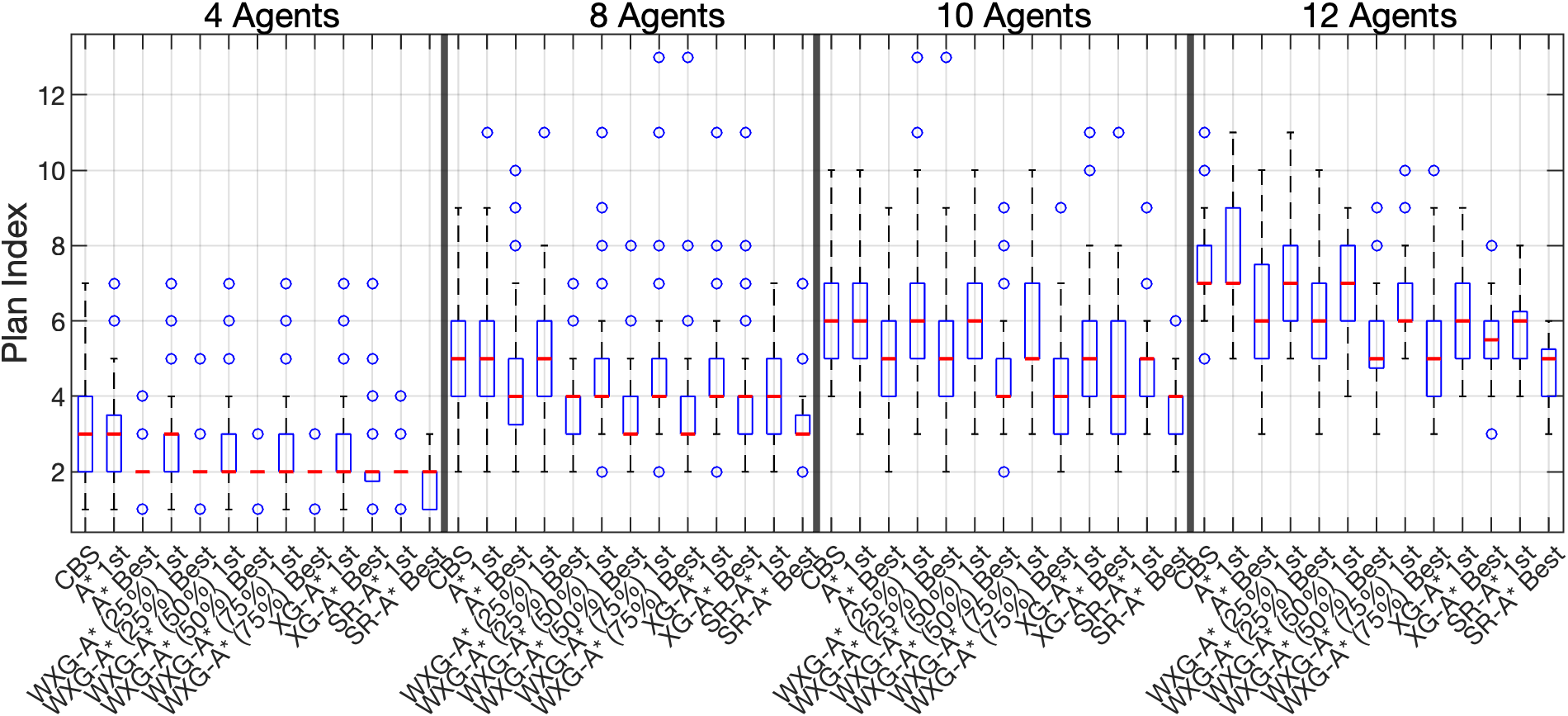}
    \hfill
    \includegraphics[width=.775\linewidth]{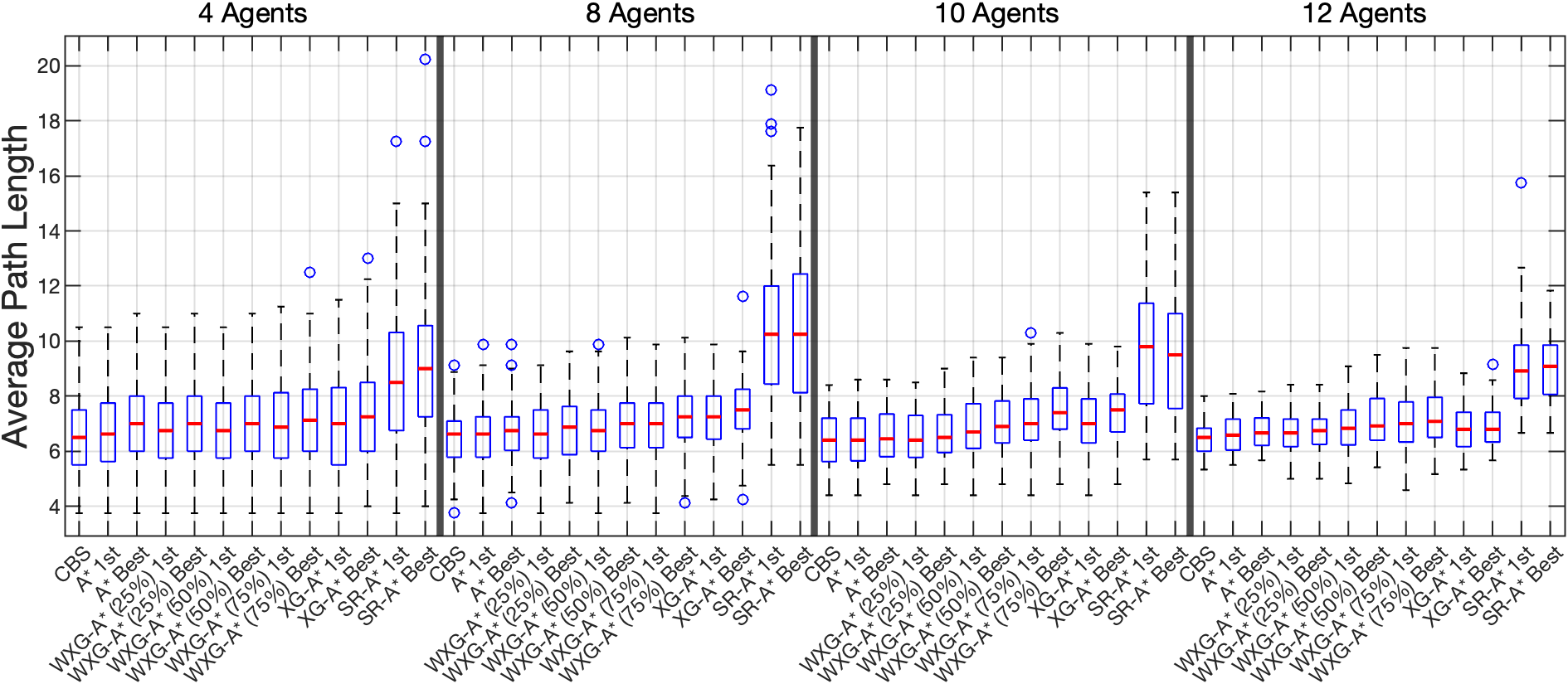}
    \hfill 
    \includegraphics[width=.775\linewidth]{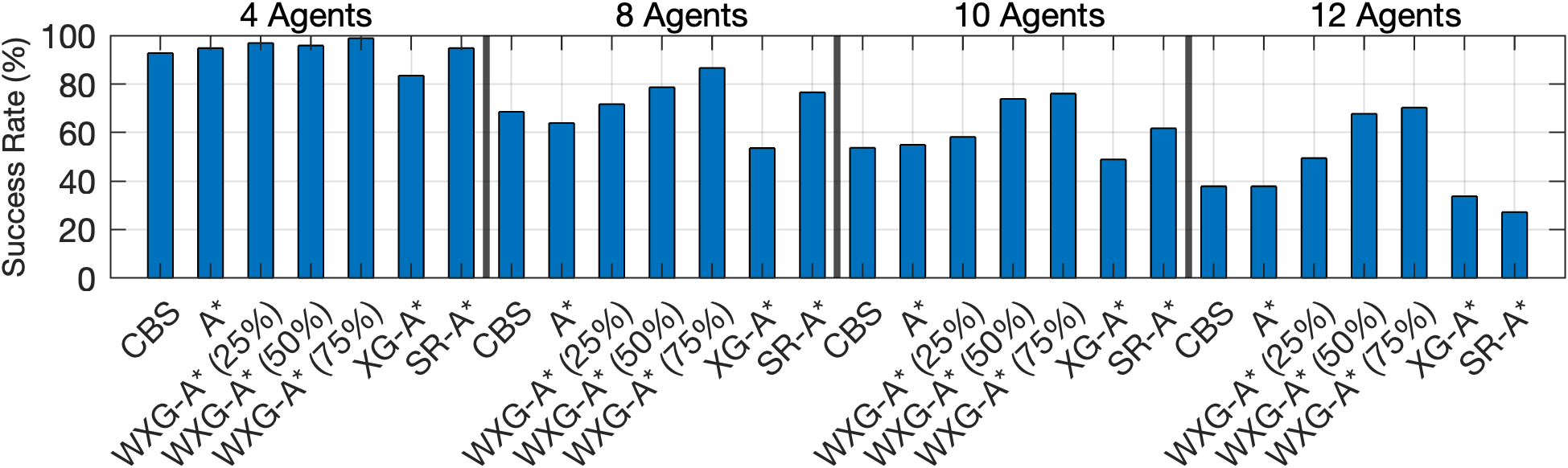}
    \hfill
    \caption{Benchmark results for  $9\times 9$ environments.}
    \label{fig:benchmark_8}
\end{figure*}

\begin{figure*}[p]
    \centering
    \includegraphics[width=.775\linewidth]{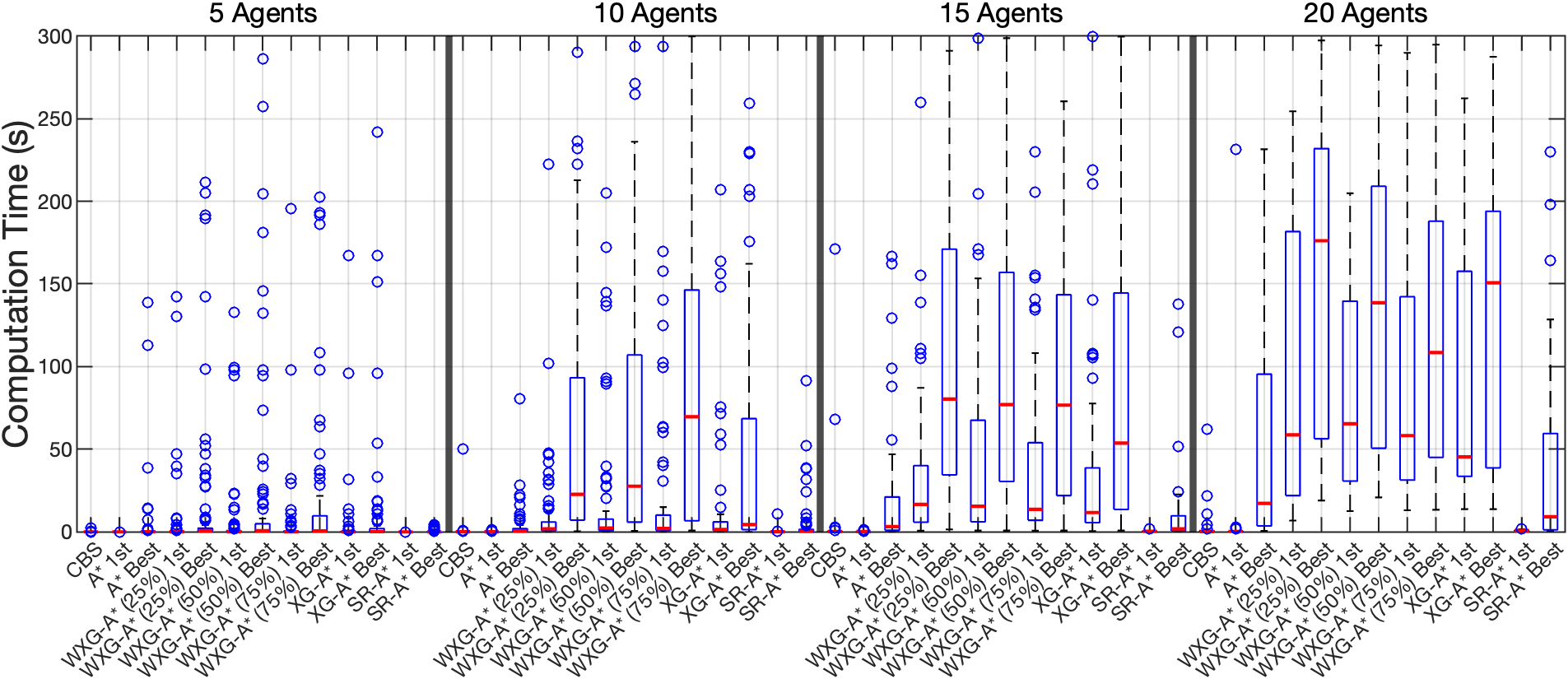}
    \includegraphics[width=.775\linewidth]{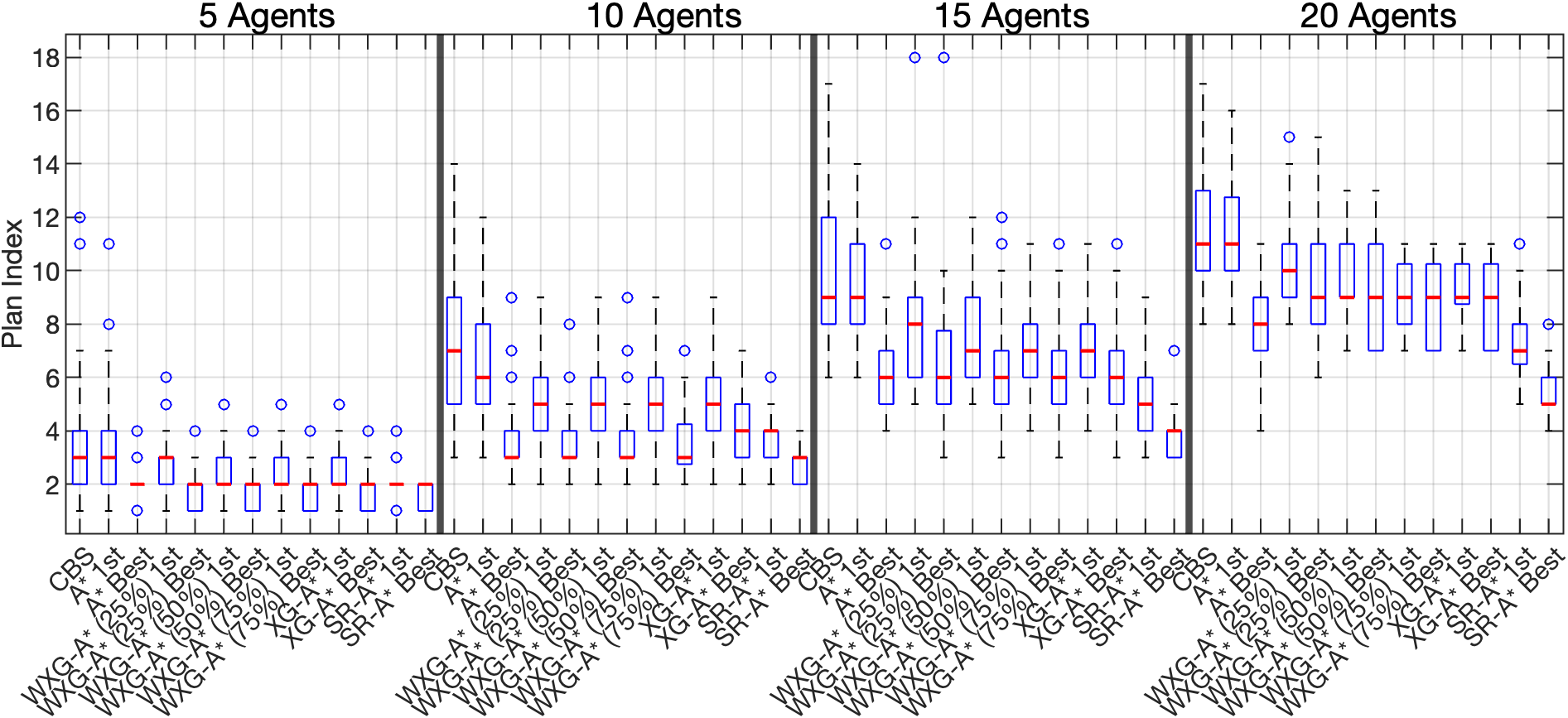}
    \hfill
    \includegraphics[width=.775\linewidth]{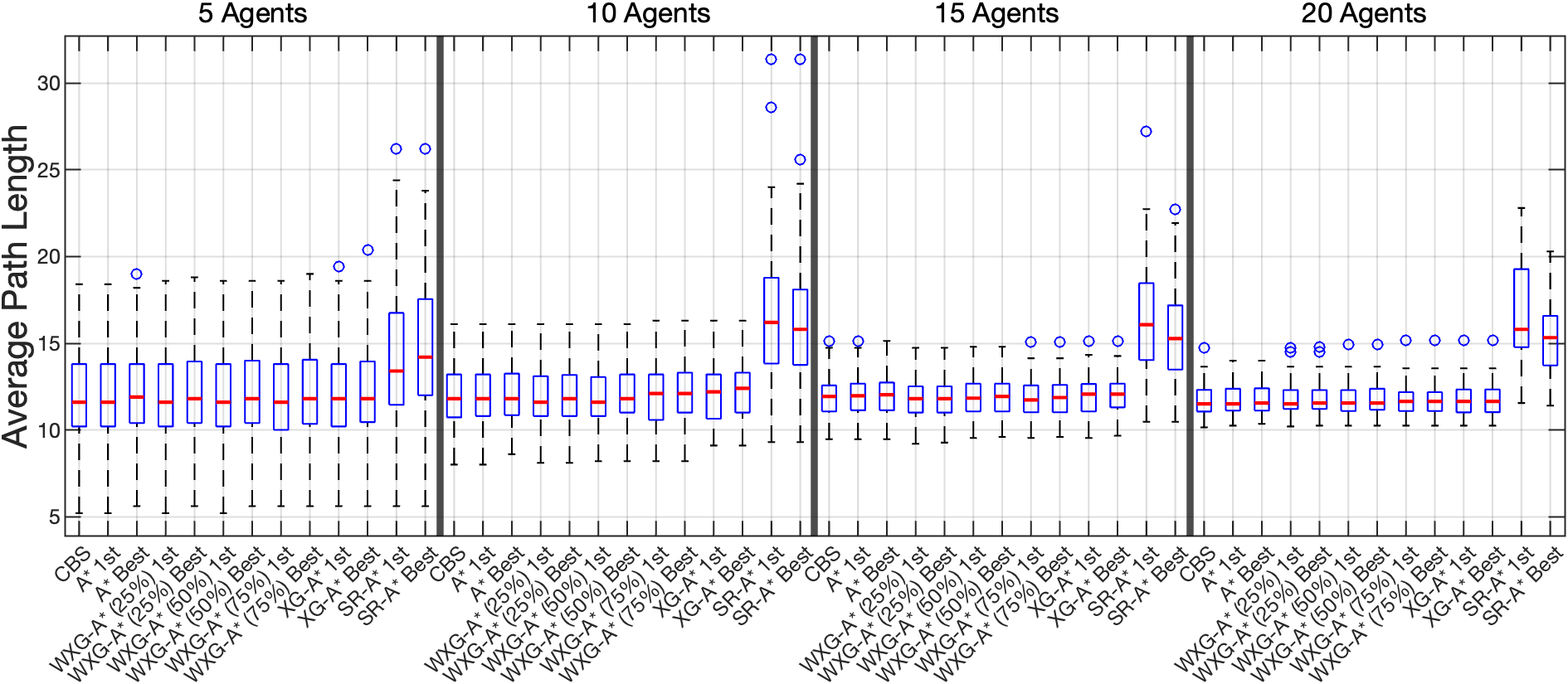}
    \hfill 
    \includegraphics[width=.775\linewidth]{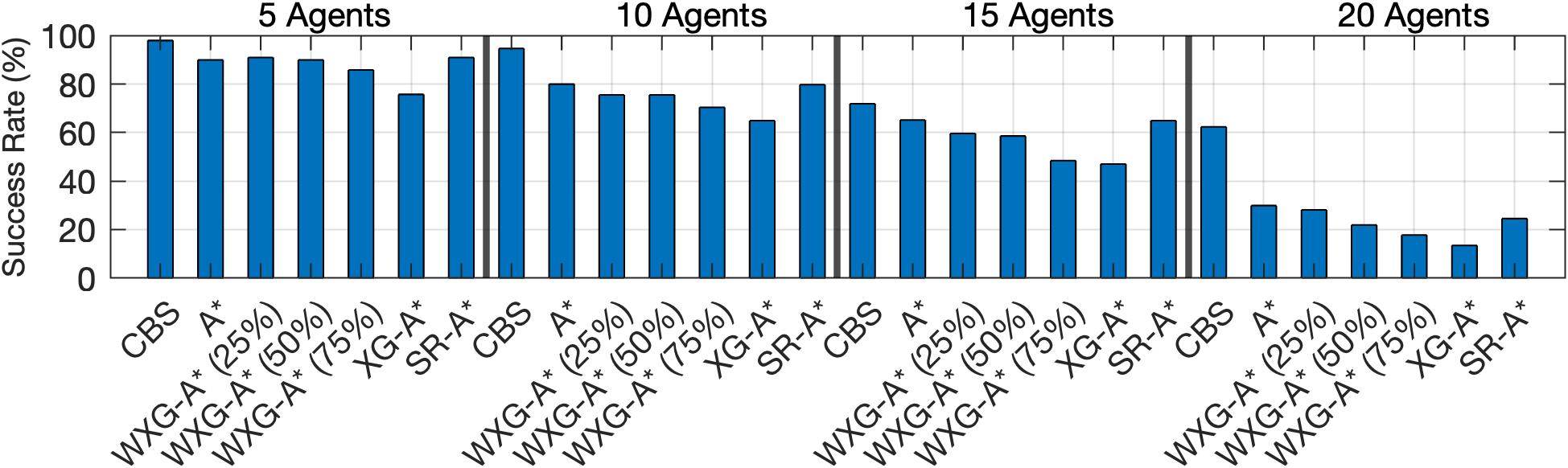}
    \hfill
    \caption{Benchmark results for  $16\times 16$ environments.}
    \label{fig:benchmark_15}
\end{figure*}

% \input{Sections/Full_Table}
% \onecolumn
% \LTXtable{\linewidth}{Sections/Full_Table.tex}
% \newpage
% \twocolumn
% We bring the complete details of the experiments described in Section 5.2. 
% We tested our algorithms on ??? 
% experiments of varying size, planning difficulty, and explanation difficulty. The results are shown in Figures ???????
Our benchmarks were on grid worlds with the following sizes and number of agents:
$9\times 9$ with $4$, $8$, $10$, and $12$ agents, $16\times 16$ with $5$, $10$, $15$, and $20$ agents, and $33\times 33$ with $10$, $20$, and $30$ agents. For each grid size and agent number combination, we ran $100$ unique experiments, where each experiment consisted of CBS, followed by XG-CBS with each of the low-level algorithms. The timeout for a single algorithm on a single benchmark was 5 minutes (while this may seem like a high threshold, recall that Explainable MAPF is computationally harder than MAPF).
The complete results appear in Figures \ref{fig:benchmark_8}, \ref{fig:benchmark_15}, and \ref{fig:benchmark_33} and contain a comparison of computation time, plan length, segmentation index and success rate.

As can be seen from the trend in computation times, for XG-$A^*$ the computation time is already high for $16\times 16$ environments, with a low success rate. This trend carries on to larger environments, rendering XG-$A^*$ with an extremely low success rate. We therefore do not evaluate XG-$A^*$ and its derivative -- WXG-$A^*$ on $33\times 33$ environments.

\begin{figure*}[p]
    \centering
    \includegraphics[width=.8\linewidth]{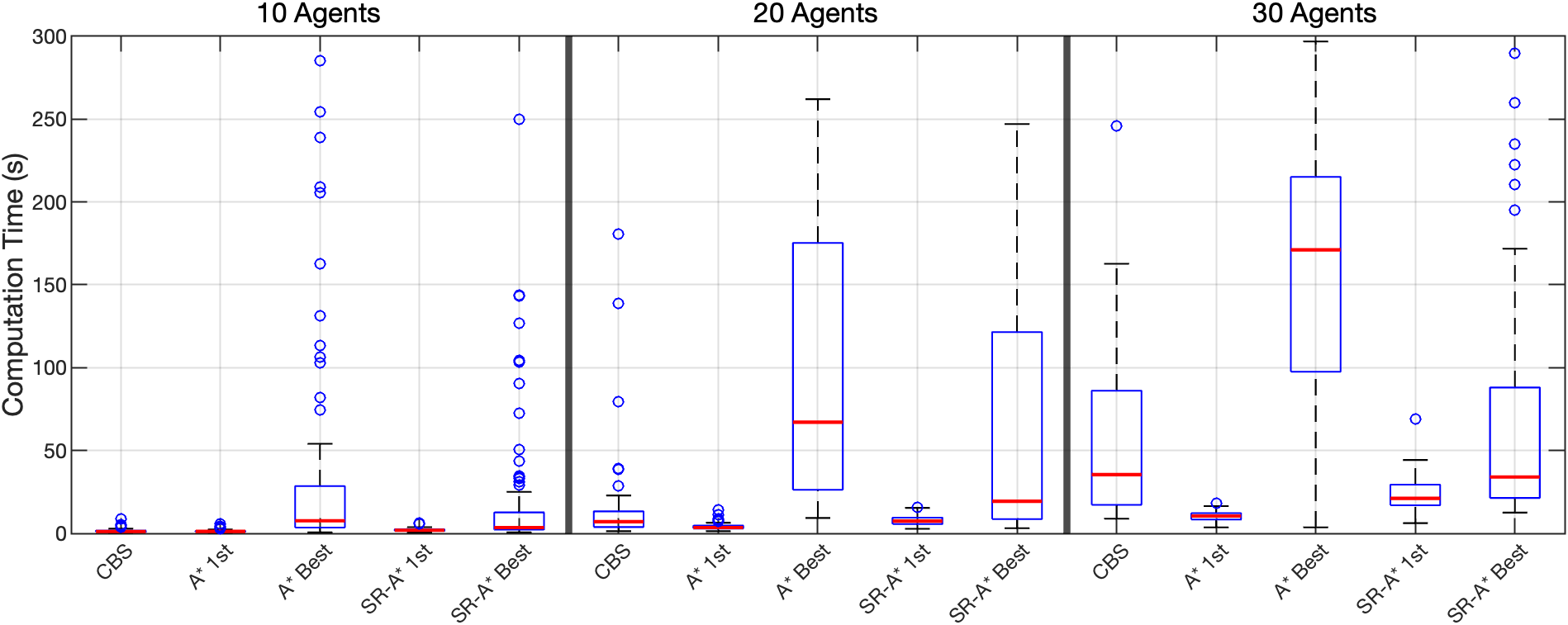}
    \includegraphics[width=.8\linewidth]{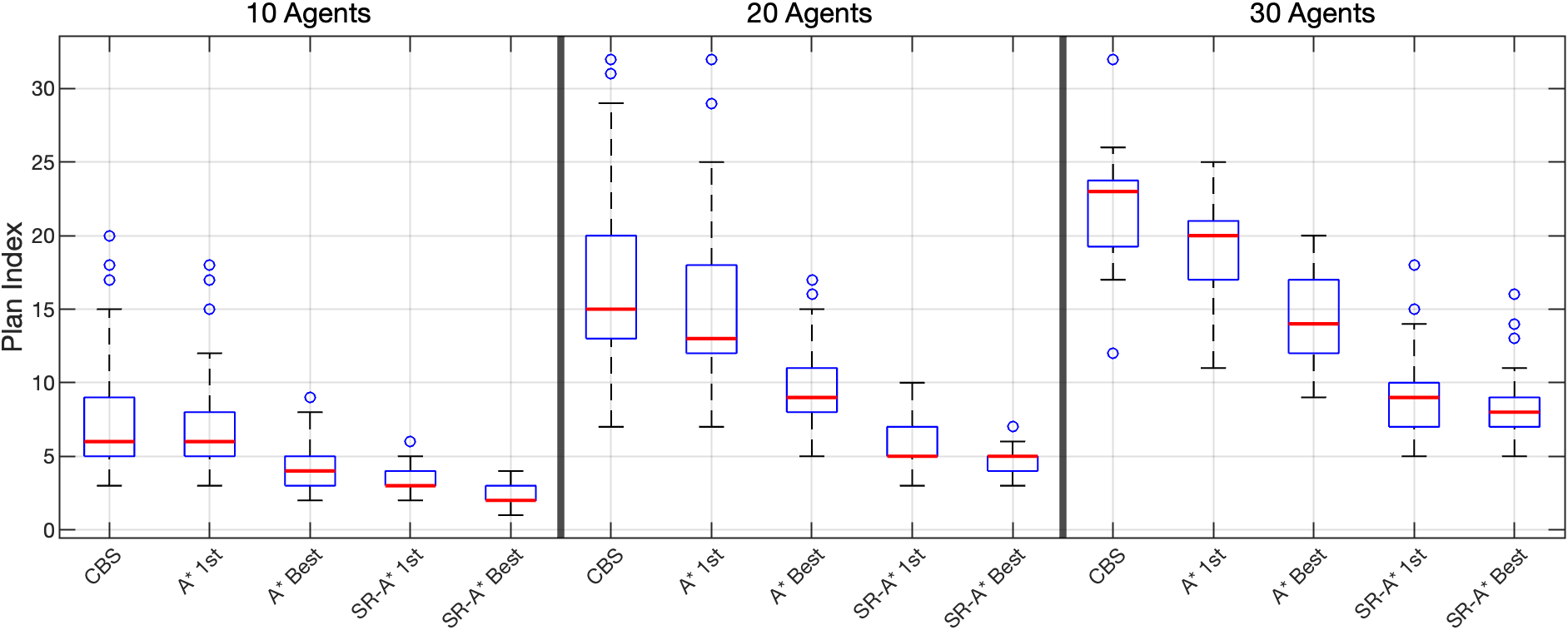}
    \hfill
    \includegraphics[width=.8\linewidth]{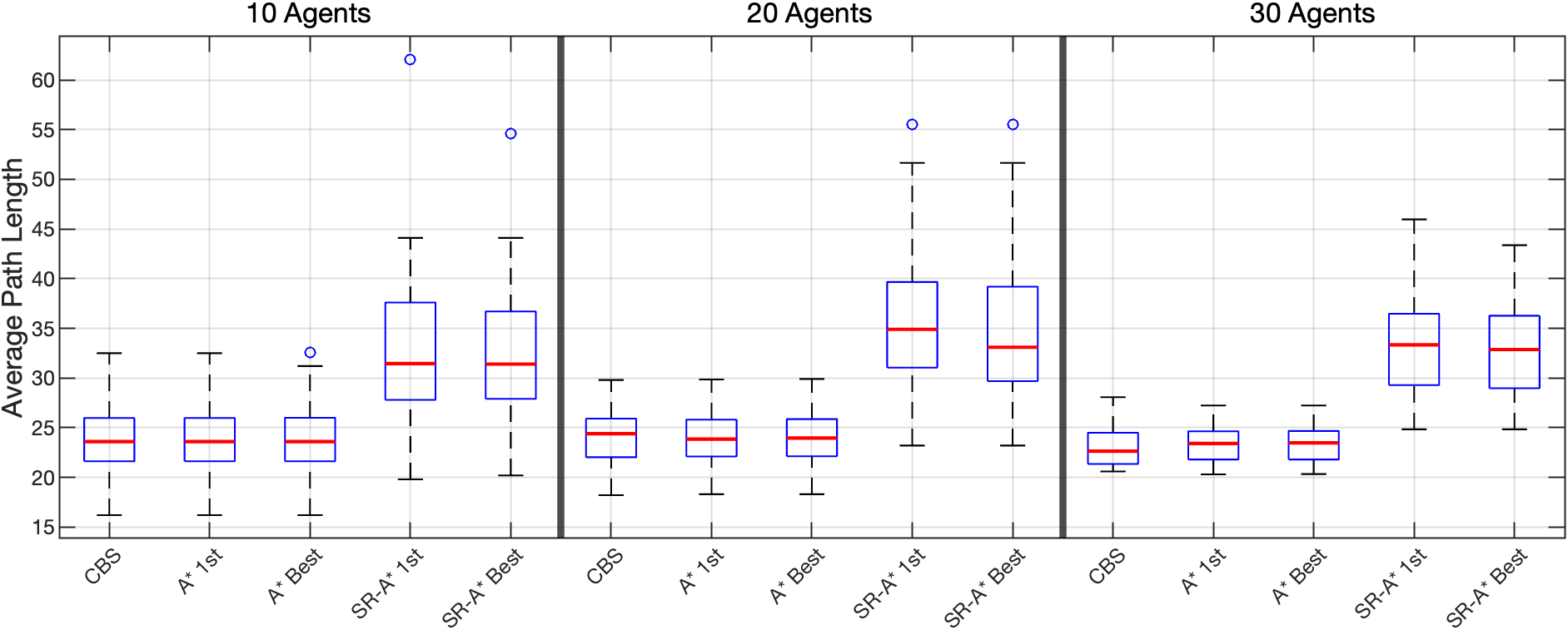}
    \hfill 
    \includegraphics[width=.8\linewidth]{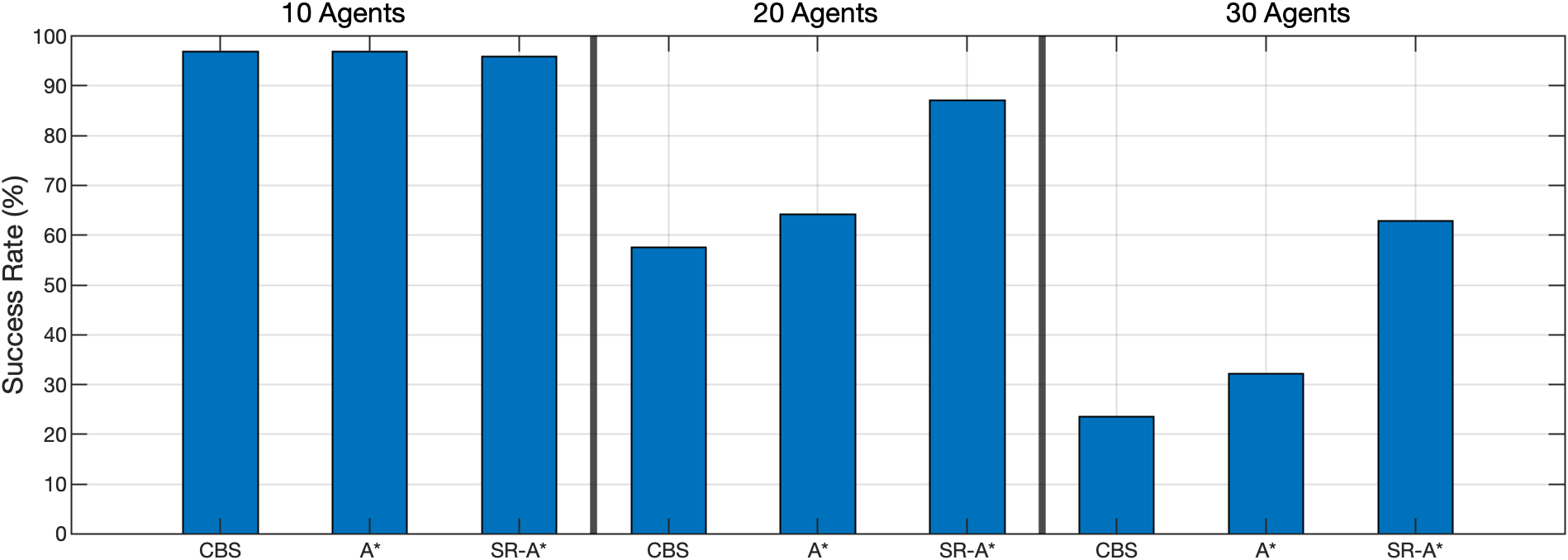}
    \hfill
    \caption{Benchmark results for  $33\times 33$ environments.}
    \label{fig:benchmark_33}
\end{figure*}
\end{document}